\pdfoutput=1

\documentclass[11pt]{article}

\newif\ifreview

\ifreview
\usepackage[review]{acl}
\else
\usepackage[final]{acl}
\fi

\usepackage{times}
\usepackage{latexsym}

\usepackage[T1]{fontenc}

\usepackage[utf8]{inputenc}

\usepackage{microtype}

\usepackage{inconsolata}

\usepackage{graphicx}

\usepackage{amssymb}
\usepackage{amsmath}
\usepackage{comment}
\usepackage{tikz}
\usetikzlibrary{positioning, shapes.multipart, arrows.meta, decorations.pathreplacing}
\usetikzlibrary{fit}
\usepackage{subfig}
\usepackage{booktabs}
\usepackage{enumitem}

\usepackage{soul}
\newcommand{\highlight}[2]{\sethlcolor{#1}\hl{#2}}

\title{Fine-tuning vs.\ In-context Learning in Large Language Models:\\A Formal Language Learning Perspective}

\author{Bishwamittra Ghosh\textsuperscript{1}, Soumi Das\textsuperscript{1}, Till Speicher\textsuperscript{1}, Qinyuan Wu\textsuperscript{1}, Mohammad Aflah Khan\textsuperscript{1},\\{\bf Deepak Garg\textsuperscript{1}, Krishna P. Gummadi\textsuperscript{1}, Evimaria Terzi\textsuperscript{2}} 
\\
\textsuperscript{1}Max Planck Institute for Software Systems, Germany, \textsuperscript{2}Boston University, USA
}

\usepackage[most]{tcolorbox}
\newtcolorbox{rqbox}[1][]{%
  colback=gray!10,
  colframe=black,
  boxrule=0.5pt,
  arc=2pt,
  auto outer arc,
  left=0pt,
  right=0pt,
  top=0pt,     %
  bottom=0pt,  %
  #1
}

\usepackage{amsthm}
\newtheorem{theorem}{Theorem}

\newtheorem{claim}[theorem]{Claim}

\makeatletter
\newtheorem*{rep@theorem}{\rep@title}
\newcommand{\newreptheorem}[2]{%
	\newenvironment{rep#1}[1]{%
		\def\rep@title{{\normalfont \textbf{#2} \ref{##1}}}%
		\begin{rep@theorem}}%
		{\end{rep@theorem}}}
\makeatother
\newreptheorem{claim}{Claim}

\newcommand{\llm}{M}
\newcommand{\dataset}{D}
\newcommand{\example}{s}
\newcommand{\numexample}{n}
\newcommand{\numepoch}{m}
\newcommand{\sep}{\mathtt{[sep]}}
\newcommand{\grammar}{G}
\newcommand{\lang}{L}

\newcommand{\ft}{\ensuremath{\mathtt{FT}}}
\newcommand{\icl}{\ensuremath{\mathtt{ICL}}}

\begin{document}
\maketitle

\begin{abstract}
    Large language models (LLMs) operate in two fundamental learning modes -- \emph{fine-tuning} ({\ft}) and \emph{in-context learning} ({\icl}) -- raising key questions about which mode yields greater language proficiency and whether they differ in their inductive biases. Prior studies comparing {\ft} and {\icl} have yielded mixed and inconclusive results due to inconsistent experimental setups. To enable a rigorous comparison, we propose a \textit{formal language learning} task -- offering precise language boundaries, controlled string sampling, and no data contamination -- and introduce a \textit{discriminative test} for language proficiency, where an LLM succeeds if it assigns higher generation probability to in-language strings than to out-of-language strings.

    Empirically, we find that:
    (a) {\ft} has greater language proficiency than {\icl} on in-distribution generalization, but both perform equally well on out-of-distribution generalization. (b) Their inductive biases, measured by the correlation in string generation probabilities, are similar when both modes partially learn the language but diverge at higher proficiency levels. (c) Unlike {\ft}, {\icl} performance differs substantially across models of varying sizes and families and is sensitive to the token vocabulary of the language. Thus, our work demonstrates the promise of formal languages as a controlled testbed for evaluating LLMs, behaviors that are difficult to isolate in natural language datasets.
    \ifreview
     Our source code is available at an anonymous link.
    \else
    Our source code is available at {\small\url{https://github.com/bishwamittra/formallm}}.
    \fi

\end{abstract}

\section{Introduction}
\label{sec:intro}

Large language models (LLMs) operate in two fundamental learning modes: \textit{fine-tuning} ({\ft}) and \textit{in-context learning} ({\icl}). {\ft} simulates a closed-book exam, where LLMs learn by updating model parameters~\cite{kaplan2020scaling}. {\icl} simulates an open-book exam, where LLMs learn from in-context examples without any parameter update~\cite{brown2020language}. Both modes are widely applied in real-world tasks, including text summarization~\cite{radford2019language}, question-answering~\cite{yang2018hotpotqa}, and conversational agents~\cite{ouyang2022training}. A natural question is, therefore, which learning mode is \textit{more language-proficient}, i.e., which mode recognizes patterns in the language better, and whether their \textit{inductive bias}, i.e., the implicit assumptions about recognizing patterns, is similar or different. Despite its importance, this question remains open due to inconsistent experimental setups in prior studies.

\begin{figure}
    \centering
    \resizebox{\columnwidth}{!}{
    \begin{tikzpicture}[
        model/.style={draw, rectangle, rounded corners, minimum width=2cm, minimum height=0.8cm, fill=blue!25},
        arrow/.style={-{Latex}, thick},
        freeze/.style={draw, rectangle, rounded corners, minimum width=2cm, minimum height=0.8cm, fill=orange!25},
        label/.style={font=\small}
        ]

        \node[draw, rounded corners, fill=green!25] (s1) at (-1,3) {$s^{(1)}$};
        \node[draw, rounded corners, fill=green!25] (s2) at (0,3) {$s^{(2)}$};
        \node (sdot) at (0.75,3) {$ \ldots $};
        \node[draw, rounded corners, fill=green!25] (sn) at (1.5,3) {$s^{(n)}$};

        \node[draw, rounded corners, fit=(s1) (s2) (sdot) (sn), inner sep=2mm] (s-all) {};

        \node[label, above=0cm of s-all, scale=1.2] (annotation-ft-1) {Training Strings};

        \node[draw, rectangle, rounded corners, minimum width=2cm, minimum height=0.8cm, below=0.5cm of s-all] (model-theta) {$ \mathtt{Model} $$($\highlight{orange!25}{$ \theta $} $ \rightarrow $ \highlight{blue!25}{$ \theta^* $}$)$};

        \draw[arrow] (s-all.south) -- (model-theta);

        \node[model, right=0.2cm of model-theta] (model-theta-2) {$ \mathtt{Model} $$(\theta^*)$};

        \node[draw,  minimum height=0.8cm, minimum width=0.8cm, rounded corners, fill=red!25, above=0.5cm of model-theta-2] (st) {$s$};
        
        \node[label, above=0cm of st, scale=1.2, align=center] (annotation-ft-2) {Test\\String};

        \draw[arrow] (st.south) -- (model-theta-2.north);

        \node[label, below=0.5cm of model-theta-2, scale=1.2] (out1) {$\mathtt{Loss}($\highlight{red!25}{$s$}$\mid$\highlight{blue!25}{$\theta^*$}$)$};
        \draw[arrow] (model-theta-2.south) -- (out1.north);

        \node[draw=gray, rounded corners, fit=(out1) (model-theta-2) (annotation-ft-1) (annotation-ft-2) (s-all), inner sep=1mm] (box-ft) {};
        \node[label, below=0.1cm of box-ft, scale=1.2] {\textbf{Fine-Tuning}};

        \node[freeze, right=1.6cm of model-theta-2] (fmodel) {$ \mathtt{Model} $$(\theta)$};

        \node[draw, rounded corners, minimum height=0.8cm, minimum width=4cm, above=0.5cm of fmodel] (prompt) {[\highlight{green!25}{$s^{(1)};\; s^{(2)};\; \ldots;\; s^{(n)};$}\; \highlight{red!25}{$s$}]};

        \node[label, above=0cm of prompt, scale=1.2, align=center] (annotation-icl-1) {Prompt\\ (concatenated input)};

        \draw[arrow] (prompt.south) -- (fmodel.north);

        \node[label, below=0.5cm of fmodel, scale=1.2] (out2) {$\mathtt{Loss}($\highlight{red!25}{$s$}$\mid$\highlight{green!25}{$s^{(1)},\ldots,s^{(n)}$}$;$\highlight{orange!25}{$\theta$}$)$};
        \draw[arrow] (fmodel.south) -- (out2.north);

        \node[draw=gray, rounded corners, fit=(out2) (prompt) (annotation-icl-1) (fmodel), inner sep=1mm] (box-icl) {};
        \node[label, below=0.1cm of box-icl, scale=1.2] {\textbf{In-context Learning}};
    \end{tikzpicture}
    }
    
    \caption{Fine-tuning and in-context learning are two learning modes of an LLM. In formal language learning, the learning task is to generate unseen strings from the language through syntactic pattern recognition (desideratum \textbf{D1}). Under an equal setting (desideratum \textbf{D2}), fine-tuning updates model parameters (\highlight{orange!25}{$ \theta $} $ \rightarrow $ \highlight{blue!25}{$ \theta^* $}) based on the \highlight{green!25}{training strings} and computes the generation loss on a \highlight{red!25}{test string}. In-context learning, however, takes a concatenated input prompt, where \highlight{green!25}{training strings} serve as the prefix for generating the \highlight{red!25}{test string}. Since the two learning modes differ in both input prompt and model parameters, a comparable evaluation metric is needed (desideratum \textbf{D3}).}
    \label{fig:intro}
\end{figure}
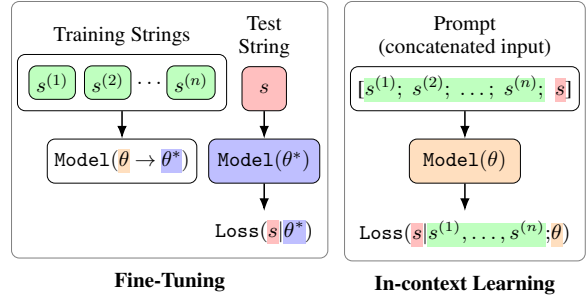

Addressing this gap requires a principled experimental design.

\paragraph{Our Contributions.} Our key contribution is the introduction of the following three-fold desiderata for comparing {\ft} and {\icl}, and a controlled experimental framework that realizes these desiderata (see motivation in Figure~\ref{fig:intro}). Several prior studies attempted to compare {\ft} and {\icl} without satisfying all three desiderata, resulting in mixed and inconclusive results. Specifically, the closest to our work is~\citet{mosbach2023few}, who partially satisfy desiderata \textbf{D1} and \textbf{D2}, but fail to satisfy \textbf{D3}.

\paragraph{D1. {Specification of the Learning Task: Syntax-focused Learning with Zero-prompting.}} We compare {\ft} and {\icl} on learning a \textit{probabilistic formal language}, which is a distribution of strings accepted by a probabilistic formal grammar~\cite{manning2003probabilistic,chater2006probabilistic}.  The learning task is to generate new strings based on recognizing syntactic patterns of the underlying language represented by the training strings (Section~\ref{sec:preliminary}). 

Formal languages offer several advantages for this comparison: (a) they contain syntax only, isolating syntactic pattern recognition from the semantic ambiguity of natural languages -- the main focus of prior studies~\cite{mosbach2023few}; (b) they provide full control over data distribution, enabling precise sampling of training and test strings and differentiation between in-distribution and out-of-distribution languages; (c) they are synthetic, avoiding data contamination and ensuring no model benefits from prior knowledge of the training data~\cite{xu2024benchmark}. These properties are difficult to guarantee with natural language datasets.

A practical challenge in comparing learning modes is \textit{communicating the task via prompt-instructions}, which different LLMs may interpret differently~\cite{wu2025towards,zhao2021calibrate,razavi2025benchmarking,zhuo2024prosa}. Formal language learning sidesteps this subjectivity: we consider a \textit{zero-prompting} setup where the LLM only sees training strings and must generate new strings without any instruction.

\paragraph{D2. {Allocation of Equal Resources.}} A fair comparison requires allocating equal resources to both {\ft} and {\icl}\footnote{Resource fairness includes data and compute fairness. The paper focuses on data fairness. Achieving compute fairness, however, is challenging, since {\ft} and {\icl} incur costs at different stages: {\ft} is costly during training, while {\icl} is costly at inference (Figure~\ref{fig:other_dimension} in the Appendix).}. We provide the same training and test data to both learning modes. Since {\ft} and {\icl} have disjoint hyperparameters -- batch-size, learning rates, and epochs for {\ft}; example repetitions and inference temperature for {\icl} -- we compare their best performance over respective hyperparameter settings, going beyond prior work~\citep{mosbach2023few,yin2024deeper} that addresses this only partially.

\paragraph{D3. {Comparable Evaluation Metric.}} How can we evaluate the language proficiency of a learner in a language? There are two potential tests for language proficiency: generative and discriminative -- the latter introduced in this work. The generative test computes the generation probability of in-language strings, but this is not directly comparable: LLMs vary in their priors, and {\ft} and {\icl} of the same LLM but with different parameters treat the input prompt differently (Figure~\ref{fig:intro}), making a direct numerical comparison of generation probability infeasible. The discriminative test instead checks whether in-language strings are generated with higher probability than \textit{close yet grammatically incorrect} out-of-language strings, and results in a classification score -- a metric that avoids model-specific and prompt-specific biases, making it comparable across both modes. Therefore, we claim that \textit{the discriminative test is the appropriate metric for comparing {\ft} and {\icl}} (Section~\ref{sec:language_proficiency}).

\paragraph{Experimental Results.}  We experiment with $18$ open-source LLMs from $6$ model families and multiple formal languages, and reach the following conclusions: (a) Different LLMs converge to optimal {\ft} performance, while their {\icl} ability varies substantially in formal languages. Model size contributes to improved performance in {\icl} but not in {\ft}.  (b) On in-distribution generalization, where training and test languages are the same, {\ft} dominates {\icl} except in some LLMs where {\icl} is close to {\ft}. On out-of-distribution generalization where training and test languages differ, both learning modes perform equally, and generalize only to out-of-distribution languages that are close to the training language. (c) The inductive bias, measured by the correlation of output generation probability of {\ft} and {\icl}, is similar when both modes partially learn the language but diverges as proficiency improves with a higher number of examples. (d) {\ft} is robust across languages, as assessed by varying the underlying grammar rules or token vocabulary. However, {\icl} performance is affected by the actual tokens used in the language.

Finally, we discuss the pitfalls of testing LLMs with natural language datasets, including imprecise sampling of training and test strings, data contamination, and ill-defined notions of in-distribution vs.\ out-of-distribution tasks, in Appendix~\ref{app_sec:nlp_dataset}. Instead, we propose that synthetic formal languages are necessary for rigorous scientific study of LLMs, and that our work will inspire future research.

\section{Motivation and Related Work}
\label{sec:related_work}

Here, we review related work and motivate why a comprehensive study comparing {\ft} and {\icl} requires satisfying three desiderata: a precise specification of the learning task (\textbf{D1}), equal resource allocation (\textbf{D2}), and a comparable evaluation metric (\textbf{D3}). Prior work on comparing {\ft} and {\icl} has largely overlooked one or more of these desiderata, yielding mixed and inconclusive results.

\paragraph{Independent Studies on {\ft} and {\icl}.} Several works independently investigate {\ft}~\citep{kaplan2020scaling,zhang2024when,hu2024minicpm} and {\icl} in LLMs~\cite{reddy2023mechanistic,pan-etal-2023-context,chen-etal-2025-icleval}, and relate learning performance to model size, training data. However, these studies examine {\ft} and {\icl} in isolation rather than in a controlled head-to-head comparison, making it difficult to draw conclusions about their relative language proficiency. Moreover, these studies rely on natural language benchmarks, where pre-training can disproportionately affect {\ft} and {\icl} performance due to data contamination -- a confounder our synthetic setup avoids.

\paragraph{Benchmarks.} Concerning desideratum \textbf{D1}, natural language datasets~\citep{rajpurkar2016squad,kwiatkwoski2019natq} often provide high-level descriptions of learning tasks, where in-distribution and out-of-distribution tasks are \textit{less precisely defined}.
Even within in-distribution tasks, there is no formal guarantee of coherence between training and test examples -- unlike in a formal language, where all examples belong to the same language. Also, public datasets may result in data contamination, providing an unfair advantage to some LLMs~\cite{dominguez2024training}. For example, we find both issues on the MNLI dataset~\cite{N18-1101}, as previously studied by~\citet{mosbach2023few} on comparing {\ft} and {\icl}. While {\ft} generalizes better out-of-distribution on MNLI -- consistent with~\citet{mosbach2023few} -- we find {\ft} and {\icl} perform equally well out-of-distribution on formal languages (Appendix~\ref{app_sec:nlp_dataset}). This divergence across benchmarks highlights the need for a well-defined learning task
(desideratum \textbf{D1}), with no data contamination.

\paragraph{Comparison of {\ft} and {\icl}.} The comparison between {\ft} and {\icl} yields mixed conclusions, often due to violating desideratum \textbf{D2} on the equality of resources. Several studies conclude that {\ft} outperforms {\icl} due to biased setups -- different model sizes, unequal number of examples, or high variance across runs~\cite{liu2022few,bhatia2023tart,asai2024buffet}.
Other studies find {\icl} better than {\ft}~\cite{yin2024deeper,bertsch2024context,kaneko2025gaps,soudani2024fine,awadalla2022exploring};
some employ suboptimal {\ft} (e.g., $1$ epoch~\cite{yin2024deeper}, or $2$--$5$ epochs~\cite{awadalla2022exploring}), while others evaluate in narrow settings such as bias mitigation~\cite{kaneko2025gaps}, retrieval for low-frequency knowledge~\cite{soudani2024fine}, or the many-shot regime~\cite{bertsch2024context}, where {\icl} retains general language understanding that {\ft} partially forgets.

\paragraph{Evaluation Metrics.} A further gap in prior work is the absence of a comparable evaluation metric (desideratum \textbf{D3}). Existing studies rely on generative metrics such as cross-entropy loss or accuracy~\cite{kallini2024mission,jumelet2023transparency,bhattamishra2020ability,wang2021evaluating,akyurek2024context}, which are not directly comparable across learning modes: in {\ft}, the loss is computed over the updated parameters $\theta^*$, whereas in {\icl}, the loss is conditioned on in-context examples alongside the frozen parameters $\theta$. As a result, a lower generative loss in one mode does not imply greater language proficiency relative to the other. We address this gap by introducing a \textit{discriminative test} that evaluates whether a model assigns higher generation probability (equivalently, lower loss) to in-language strings than to out-of-language strings -- a criterion that is comparable across both learning modes and model families (Section~\ref{sec:language_proficiency}).

\paragraph{Formal Languages in LLM Research.} Owing to their greater controllability, formal languages have been widely used to investigate the linguistic capabilities of LLMs~\cite{jumelet2023transparency}, including their inductive biases in language learning~\citep{papadimitriou2023injecting,white2021examining,hopkins2022towards}. Leveraging formal languages as a testbed, prior studies have compared the representational capacity of LLMs with various sequence-based models~\cite{shi2022learning,chi2023transformer,bhattamishra2020ability,merrill2023formal,strobl2023transformers,hahn2020theoretical}, and analyzed the classes of formal languages that LLMs can learn~\cite{deletang2022neural,hahn2024sensitive,cotterell2018all,mielke2019kind,borenstein2024languages}. Notably, LLMs have been shown to learn hierarchical and probabilistic formal languages that mirror the recursive structure of natural language~\cite{allen2023physics,murty2022characterizing,liu2022transformers}.

To our knowledge, no prior work has employed formal languages to compare the language proficiency and inductive biases of learning modes of LLMs -- this forms our central focus. Extended related work is in Appendix~\ref{sec:related_work_extended}.

\section{Experimental Framework}
\label{sec:preliminary}
We discuss preliminaries on formal languages, how to teach them to LLMs via different learning modes, and the experimental setup -- all of which realize our desiderata.

\begin{figure}
    \scalebox{0.8}
    {
    \small
    \begin{minipage}{0.3\textwidth}
        \centering
        \begin{align*}
            & \textcolor{red}{S}\;\textcolor{black}{\rightarrow}\;\textcolor{red}{A19}\;\textcolor{blue}{[1]}\\
            & \textcolor{red}{A19}\;\textcolor{black}{\rightarrow}\;\textcolor{red}{A18}\;\textcolor{red}{A16}\;\textcolor{blue}{[0.50]}\\
            & \textcolor{red}{A19}\;\textcolor{black}{\rightarrow}\;\textcolor{red}{A16}\;\textcolor{red}{A18}\;\textcolor{red}{A17}\;\textcolor{blue}{[0.50]}\\
            & \textcolor{red}{A18}\;\textcolor{black}{\rightarrow}\;\textcolor{red}{A15}\;\textcolor{red}{A14}\;\textcolor{red}{A13}\;\textcolor{blue}{[0.50]}\\
            & \textcolor{red}{A18}\;\textcolor{black}{\rightarrow}\;\textcolor{red}{A14}\;\textcolor{red}{A15}\;\textcolor{red}{A13}\;\textcolor{blue}{[0.50]}\\
            & \textcolor{red}{A17}\;\textcolor{black}{\rightarrow}\;\textcolor{red}{A14}\;\textcolor{red}{A13}\;\textcolor{red}{A15}\;\textcolor{blue}{[0.50]}\\
            & \textcolor{red}{A17}\;\textcolor{black}{\rightarrow}\;\textcolor{red}{A13}\;\textcolor{red}{A14}\;\textcolor{red}{A15}\;\textcolor{blue}{[0.50]}\\
            & \textcolor{red}{A16}\;\textcolor{black}{\rightarrow}\;\textcolor{red}{A14}\;\textcolor{red}{A15}\;\textcolor{blue}{[0.50]}\\
            & \textcolor{red}{A16}\;\textcolor{black}{\rightarrow}\;\textcolor{red}{A15}\;\textcolor{red}{A14}\;\textcolor{blue}{[0.50]}\\
            & \textcolor{red}{A15}\;\textcolor{black}{\rightarrow}\;\textcolor{red}{A11}\;\textcolor{red}{A12}\;\textcolor{red}{A10}\;\textcolor{blue}{[0.50]}\\
            & \textcolor{red}{A15}\;\textcolor{black}{\rightarrow}\;\textcolor{red}{A12}\;\textcolor{red}{A10}\;\textcolor{red}{A11}\;\textcolor{blue}{[0.50]}\\
        \end{align*}
    \end{minipage}%
    \begin{minipage}{0.33\textwidth}
        \centering
        \begin{align*}
            & \textcolor{red}{A14}\;\textcolor{black}{\rightarrow}\;\textcolor{red}{A11}\;\textcolor{red}{A10}\;\textcolor{red}{A12}\;\textcolor{blue}{[0.50]}\\
            & \textcolor{red}{A14}\;\textcolor{black}{\rightarrow}\;\textcolor{red}{A10}\;\textcolor{red}{A11}\;\textcolor{red}{A12}\;\textcolor{blue}{[0.50]}\\
            & \textcolor{red}{A13}\;\textcolor{black}{\rightarrow}\;\textcolor{red}{A10}\;\textcolor{red}{A12}\;\textcolor{red}{A11}\;\textcolor{blue}{[0.50]}\\
            & \textcolor{red}{A13}\;\textcolor{black}{\rightarrow}\;\textcolor{red}{A12}\;\textcolor{red}{A11}\;\textcolor{red}{A10}\;\textcolor{blue}{[0.50]}\\
            & \textcolor{red}{A12}\;\textcolor{black}{\rightarrow}\;\textcolor{teal}{9}\;\textcolor{teal}{8}\;\textcolor{teal}{7}\;\textcolor{blue}{[0.50]}\\
            & \textcolor{red}{A12}\;\textcolor{black}{\rightarrow}\;\textcolor{teal}{8}\;\textcolor{teal}{7}\;\textcolor{blue}{[0.50]}\\
            & \textcolor{red}{A11}\;\textcolor{black}{\rightarrow}\;\textcolor{teal}{6}\;\textcolor{teal}{5}\;\textcolor{blue}{[0.50]}\\
            & \textcolor{red}{A11}\;\textcolor{black}{\rightarrow}\;\textcolor{teal}{6}\;\textcolor{teal}{4}\;\textcolor{teal}{5}\;\textcolor{blue}{[0.50]}\\
            & \textcolor{red}{A10}\;\textcolor{black}{\rightarrow}\;\textcolor{teal}{3}\;\textcolor{teal}{1}\;\textcolor{blue}{[0.50]}\\
            & \textcolor{red}{A10}\;\textcolor{black}{\rightarrow}\;\textcolor{teal}{1}\;\textcolor{teal}{2}\;\textcolor{teal}{3}\;\textcolor{blue}{[0.50]}\\
        \end{align*}
    \end{minipage}%
    }

    \caption{Inspired by~\citet{allen2023physics}, we illustrate a hierarchical and probabilistic context-free grammar, representing language $ L_1 $. Here, non-terminals are marked in \textcolor{red}{red}, terminal tokens (alphabet) are in \textcolor{teal}{teal}, and rule-probabilities are in \textcolor{blue}{blue}. The grammar contains non-terminal symbols $ S $ and $ A $'s, alphabet $ \mathbf{T} = \{1, 2, \dots, 9\} $, and probabilistic production rules which are applied in a hierarchical way. To generate a string, we start from the non-terminal $ S $ and recursively apply production rules until reaching tokens in $ \mathbf{T} $ only.}
    \label{fig:g1}
    
\end{figure}

\paragraph{Formal Languages.}
We use \emph{probabilistic formal languages}, particularly the class generated by hierarchical probabilistic context free grammars (HPCFGs), as the learning task for LLMs (desideratum \textbf{D1}). HPCFGs capture the recursive structure of natural language. Formally, a probabilistic formal language $L$ is defined on a set of tokens (alphabet)  $\mathbf{T}$, and specifies a probability distribution $P_L$ over strings, $P_L: \mathbf{T}^\ast \rightarrow [0, 1]$, where $\mathbf{T}^\ast$ is the set of all strings. A string $\example$ is \emph{in-language} w.r.t.\ $\lang$ if $P_\lang(\example) > 0$, and \emph{out-of-language} if $P_\lang(\example) = 0$. $\mathbf{T}$ is a proper subset of the vocabulary $\mathbf{V}$ of all tokens of the LLM.

\begin{figure}
    \centering
    \includegraphics[trim={2.5cm 0 2.5cm 2cm},clip,scale=0.7]{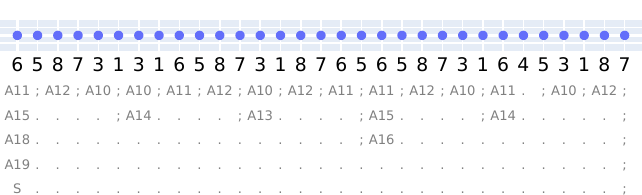}

    \caption{A string $ s $ generated by the grammar in Figure~\ref{fig:g1}. The rule `$ A19 \rightarrow A18\;A16\; [1]$' indicates that non-terminal $ A19 $ is expanded to $ A18 $ followed by $ A16 $ with probability $ 1 $, and so on, until reaching $\mathbf{T}$.  The generation probability of $ s $ is the multiplication of the probabilities of rules applied recursively to generate $ s $, and $ P(s) = (0.5)^{23}  $.}
    \label{fig:representaive_string_main}
\end{figure}

\textbf{Languages.} We consider six languages, denoted by $ \{\lang_i\}_{i=1}^6 $, based on a combination of two distinct HPCFGs and three distinct alphabet sets. For each language, we sample non-overlapping training $(\numexample_{\text{train}} \in \{1, 2, 4, \dots, 1024\})$ and test strings $(\numexample_{\text{test}} = 1024)$, following the distribution in a given language (desideratum \textbf{D2}). Figures~\ref{fig:g1} and~\ref{fig:representaive_string_main} illustrate a representative grammar and a sampled string, respectively. Additional details on formal languages, respective grammars, the sampling process, and length distributions of generated strings are in Appendix~\ref{app_sec:exp_setup}.

\textbf{Construction of Out-of-language Strings.} We quantify the \textit{degree of incorrectness} of an out-of-language string as a \emph{distance} from the language under investigation, which we use in the discriminative test in Section~\ref{sec:language_proficiency} (desideratum \textbf{D3}).
We generate grammatically incorrect strings in two ways: (a) \textit{Incorrect by edit}: We edit in-language strings to create out-of-language strings (through the addition, deletion, and replacement of tokens at random positions), where edit distance is the number of edits made to the in-language string.
(b) \textit{Incorrect by randomization}: We sample random strings over the language's alphabet, matching the length distribution of the language. On average, such random strings have a very high {edit} distance from the language. In both cases, we ensure non-membership of out-of-language strings via a grammar parser.

\paragraph{Teaching the Language to an LLM.} To teach a language $ L $ to an LLM $ \llm $, we sample strings from $ L $ and provide them to $ \llm $ via both learning modes. {\ft} is performed for a fixed number of epochs, denoted by $\numepoch =  50$, where in each epoch the LLM iterates over the strings while minimizing cross-entropy loss. Formally, consider a dataset of $n$ strings $\dataset \triangleq \{ \example^{(j)} \}_{j=1}^\numexample$ sampled from the language, $\dataset \sim \lang$. For a given string $\example$ and its token $\example_i$ at the $i$-th position, let $P_{\llm}(\example_i| \example_{[1,i-1]})$ be the probability that the LLM $\llm$ assigns to the token $\example_i$ given the prefix tokens $\example_{[1,i-1]}$. The cross-entropy loss of the LLM $ \llm $ on the dataset $\dataset$ is the per-token negative log probability at every token position of all strings in $\dataset$,
$
    \small
    \mathtt{loss}_{\llm}(\dataset) \triangleq - \frac{1}{\numexample}\sum_{\example\in \dataset}\frac{1}{|\example|}\sum_{i=1}^{|\example|} \log P_{\llm}(\example_i \mid  \example_{[1,i-1]}).
$

In {\icl}, we provide the same strings in $\dataset $ as in-context examples. Specifically, {\icl} takes a set of ordered examples $\langle s^{(1)},\ldots, s^{(\numexample)} \rangle $  as a prefix for a test string $s$. The {\icl} examples are concatenated using separators, such as semicolons, leading to a prompt $s^{(1)} \sep \ldots \sep s^{(\numexample)} \sep s $. Similar to epochs in {\ft}, we repeat the examples in {\icl} a fixed number of times, $\numepoch \in \{1, 2, 4, 8, 16\}$. In both modes, we compare the language proficiency at the \textit{optimal epoch or repetition} $ m^* $, following desideratum \textbf{D2}.

\paragraph{Models.} We study $18$ open-source LLMs from $6$ model families: Mistral~\cite{jiang2023mistral7b}, Llama~\cite{touvron2023llama,touvron2023llama-2,dubey2024llama}, Qwen~\cite{yang2024qwen2}, Gemma~\cite{gemmateam2024gemmaopenmodelsbased,gemmateam2024gemma2improvingopen}, Pythia~\cite{biderman2023pythia}, and Opt~\cite{zhang2022opt}, ranging from $0.5$B to $13$B parameters. Each experiment is repeated three times by randomly sampling training strings with different seeds. Additional details on hyperparameters are provided in Appendix \ref{app_sec:exp_setup}.

\section{The Test for Language Proficiency}
\label{sec:language_proficiency}

Today, most prior language proficiency tests for LLMs are based on generative measures -- how well an LLM generates strings belonging to the language. These tests, however, do not consider grammatically incorrect strings outside the language. Often, error patterns reveal more about language proficiency -- two non-native speakers may have similar generative performance, but the types of mistakes they make reveal their underlying language prior. Our discriminative test is motivated by this analogy; moreover, it is comparable across learning modes, enabling a direct comparison between {\ft} and {\icl} (desideratum \textbf{D3}).

\paragraph{The Generative Test.} The generative test evaluates how well an LLM generates unseen test strings from the language -- the higher the generation performance, the better the language proficiency.

Formally, consider two LLMs $\llm$ and $\llm'$ and a target language $\lang$. $ \llm $ and $ \llm' $ can also be two learning modes of the same LLM. Using the generative test, $ \llm $ is more language proficient in $ \lang $ than $ \llm' $, if $ \llm  $ generates strings in $ \lang $ with  a lower loss than $ \llm' $, i.e., $ \mathtt{loss}_{\llm}(\lang) < \mathtt{loss}_{\llm'}(\lang) $.

\textbf{Issues with the Generative Test.} Two reasons hinder a direct  comparison between {\ft} and {\icl} using the generative test. (i) Absolute loss (perplexity or probability) is incomparable across LLMs: generation loss is impacted by pre-training setup, vocabulary, model parameters, random initialization, etc. As a result, different LLMs optimally trained on the same language may generate strings with different losses. (ii) {\ft} and {\icl} result in different input prompts and require comparing the same LLM with different parameters (Figure~\ref{fig:intro}). These confounding factors make direct comparison infeasible: if {\ft} and {\icl} generate a string with different loss, we cannot determine whether the difference is due to different input prompts, model parameters, or both. To overcome these issues, we propose a discriminative test, which considers strings outside the language.

\begin{figure}[!t]
    \centering
    \scalebox{0.7}{
        \begin{tikzpicture} 
        
            \draw[fill=red!90, draw=red!90](0,0) circle (2.0);
            \draw[fill=red!60, draw=red!60](0,0) circle (1.67);
            \draw[fill=red!30, draw=red!30](0,0) circle (1.33);
            \draw[fill=green!90, draw=green!90](0,0) circle (1);
    
            \draw [->, bend angle=0, bend left]  (8, 0) to  node[above, xshift=1cm, scale=1.2] {Correct }(0,0);

            \draw [->, bend angle=0, bend left]  (8, - 1.1) to  node[above, xshift=1cm, scale=1.2] {Incorrect (low edit distance)}(0, - 1.1);
            
            \draw [->, bend angle=0, bend left]  (8, - 1.9) to  node[above, xshift=1cm, scale=1.2] {Incorrect (high edit distance)}(0, - 1.9);

        \end{tikzpicture}
    }
    \caption{We visualize the set of all strings
    in a hierarchy, where the inner \textcolor{green}{green} circle denotes {grammatically correct} in-language strings, and the outer \textcolor{red}{red} circle denotes grammatically incorrect out-of-language strings. 
    The generative test focuses on generation performance within the green circle, while the discriminative test focuses on comparative generation performance between green and red (especially at low edit distance) circles.}
    \label{fig:string_hierarchy}
\end{figure}
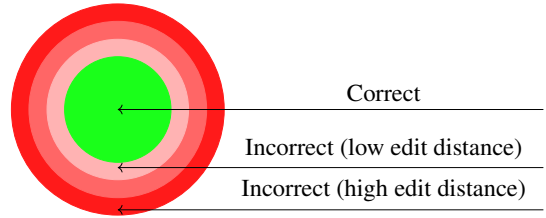

\paragraph{The Discriminative Test.} The key intuition behind the discriminative test is that \textit{if an LLM learned a language, it should generate strings in the language with lower loss than strings outside the language}. Thus, the discriminative test attempts to classify in-language and out-of-language strings based on their generation loss, where the success of classification implies language proficiency. As shown in Figure~\ref{fig:string_hierarchy}, the test can be made stricter by picking out-of-language strings \textit{close} to in-language strings (according to some distance metric such as edit distance) and checking if they can still be identified as out-of-language.

Formally, let $\mathsf{T}(\lang)$ denote out-of-language strings, constructed by editing strings in $\lang$ to ensure they are not in $\lang$. Consider a binary (linear) classifier, where the input is the generation loss assigned by an LLM to strings in $ \lang \cup \mathsf{T}(\lang) $, and the classification task is to determine their membership. Let $ \mathtt{auc}_{\llm}(\lang, \mathsf{T}(\lang)) \in [0, 1] $ be the AUC (area under the receiver operating characteristic curve) of the classifier using model $\llm$; the higher the value, the better. Thus, LLM $\llm$ is more language proficient in $\lang$ than $\llm'$, if $\mathtt{auc}_{\llm}(\lang, \mathsf{T}(\lang)) > \mathtt{auc}_{\llm'}(\lang, \mathsf{T}(\lang))$. We formalize the comparability of the discriminative test in the following claim.

\begin{claim}
    \label{claim:discriminative_test}
    For a given language, the discriminative test yields a numerically comparable score between two learning modes of an LLM and across LLMs, unlike the generative test.
\end{claim}

To support our claim, the discriminative test asks the same LLM or learning mode (i.e., equal parameters) to generate in-language and out-of-language strings, where all strings use the same input prompt format. Thus, the derived classification score is comparable across learning modes and LLMs (details in Appendix~\ref{app_sec:discriminative_test}).

\begin{figure}[!t]
    \captionsetup[subfigure]{justification=centering}
    \centering
    
    \subfloat[Generative Test, {\ft}]{
    \includegraphics[scale=0.35]
        {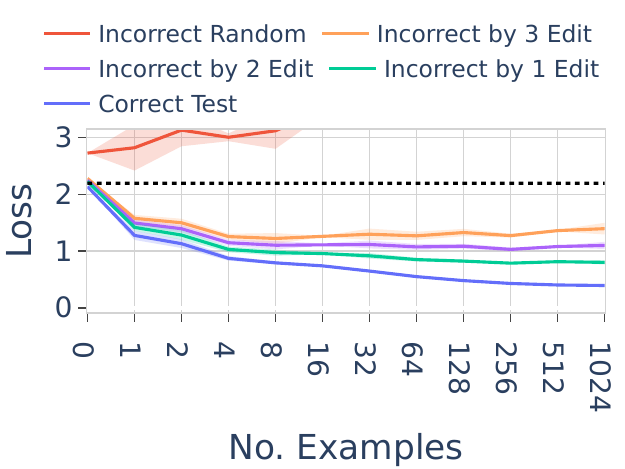}
        \label{fig:language_proficiency_gen_ft}
    }
    \subfloat[Generative Test, {\icl}]{
    \includegraphics[scale=0.35]{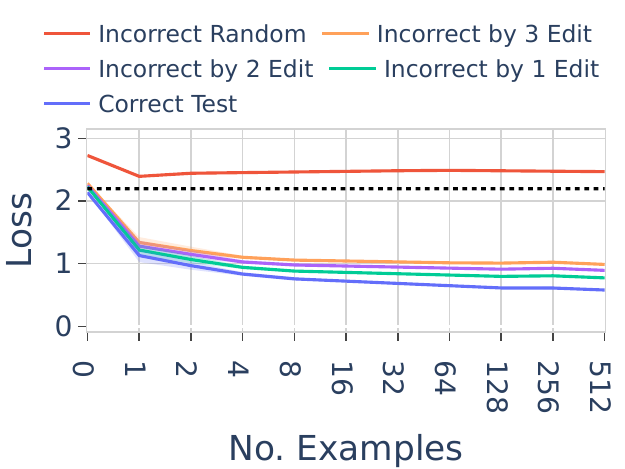}
    \label{fig:language_proficiency_gen_icl}
    }

    \subfloat[Discriminative Test, {\ft}]{
    \includegraphics[scale=0.35]{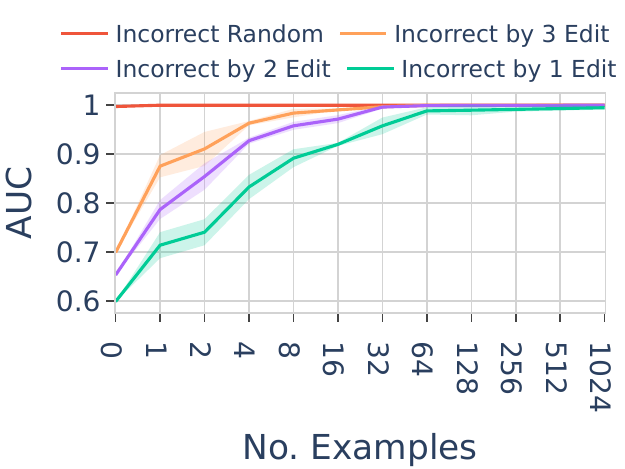}
    \label{fig:language_proficiency_discr_ft}
    }
	\subfloat[Discriminative Test, {\icl}]{
    \includegraphics[scale=0.35]{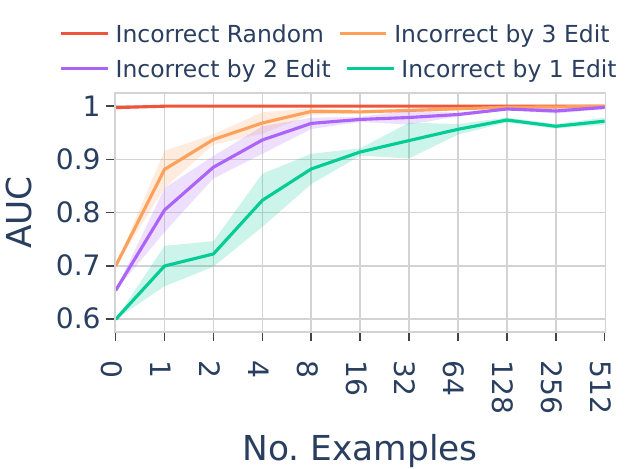}
    \label{fig:language_proficiency_discr_icl}
    }

    \caption{Language proficiency of Mistral-$7$B on language $\lang_1$, while varying the number of examples in both learning modes. 
    }

    \label{fig:language_proficiency}
        
\end{figure}

\paragraph{Demonstration of Language Proficiency Tests.} We now illustrate the behavior of both tests empirically. Figure~\ref{fig:language_proficiency} shows the language proficiency of an LLM w.r.t.\ the generative test (loss) in the top row and the discriminative test (AUC) in the bottom row, for both {\ft} and {\icl}.

\textbf{Observation 1. Generative test alone is misleading.} In Figures~\ref{fig:language_proficiency_gen_ft} and~\ref{fig:language_proficiency_gen_icl}, with increasing examples, the loss decreases on in-language test strings, shown in the \textcolor{blue}{blue} line. From this observation alone, we cannot determine whether language proficiency is achieved. Because, loss also decreases on out-of-language strings that are close, and there is often a loss-overlap between in-language and out-of-language strings, especially when the number of examples is low.  Therefore, \textit{the generative test alone is insufficient in determining whether language proficiency is achieved in the target language or in nearby languages}.

\textbf{Observation 2. Discriminative test score is correlated with training size and edit distance of out-of-language strings.} In Figures~\ref{fig:language_proficiency_discr_ft} and~\ref{fig:language_proficiency_discr_icl}, the AUC of the discriminator increases with the number of examples. Hence, the LLM becomes increasingly proficient in the language, by not only generating strings from the language with lower loss, but also distinguishing them from strings outside the language. Moreover, AUC is correlated with the edit distance of out-of-language strings; the higher the edit distance, the higher the AUC. \textit{Importantly, the AUC scores of {\ft} and {\icl} are comparable when both modes use the same number of examples and degree of grammatical incorrectness.}

In the next section, we apply the discriminative test to compare {\ft} and {\icl}, and report the AUC for discriminating in-language test strings from out-of-language strings at edit distance $1$ -- the most stringent setting of the discriminative test.

\section{Fine-tuning vs. In-context Learning}
\label{sec:comparison}

We study the language proficiency of {\ft} and {\icl} in LLMs by learning syntactic patterns from formal languages. Specifically, we address the following research questions.\footnote{Additional results including evaluation on natural language datasets, capability of utilizing full {\icl} context by different LLMs, and the detailed implications of research questions are in the Appendix.}

\begin{rqbox}
    \textbf{RQ1.} When evaluating {\ft} and {\icl} independently on a given language, how language proficient are different LLMs of varying sizes and families?\\
    \textbf{RQ2.} Which learning mode is more language proficient when evaluated jointly on in-distribution and out-of-distribution generalization?\\
    \textbf{RQ3.} Do {\ft} and {\icl} result in similar inductive bias while learning a formal language?\\
    \textbf{RQ4.} How robust is the performance of {\ft} and {\icl} to changes in languages?
\end{rqbox}

\begin{figure}[!t]
	\centering

    \subfloat[{\ft}]{
	\includegraphics[scale=0.35]{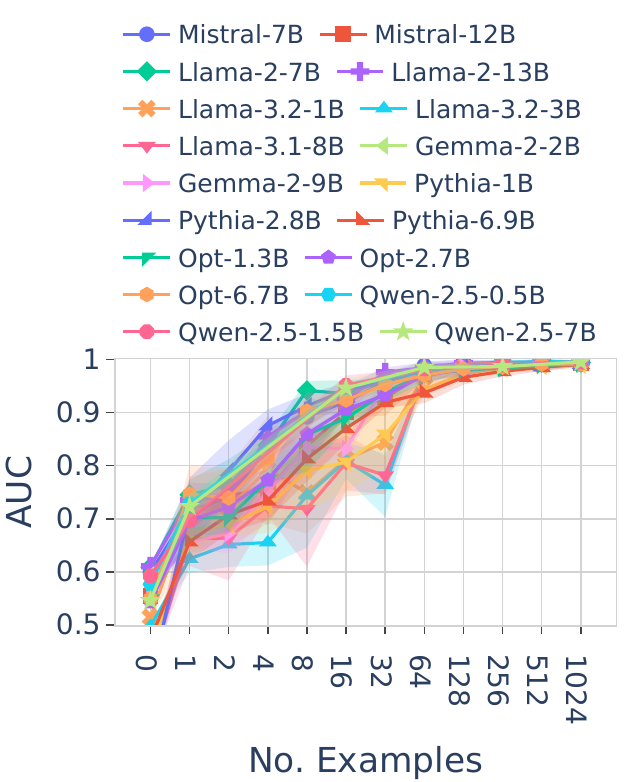}
	}\subfloat[{\icl}]{
	\includegraphics[scale=0.35]{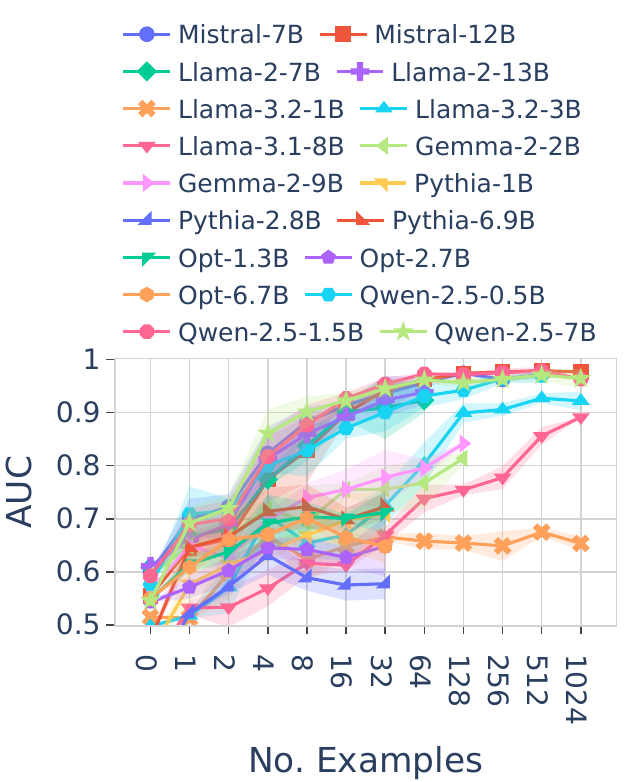}
	}

	\caption{{\ft} and {\icl} across different LLMs while learning language $\lang_1$. Different LLMs demonstrate similar {\ft} performance, but their {\icl} ability varies.}

    \label{fig:fine_tuning_vs_few_shot_all_models}
\end{figure}

\paragraph{Answer to RQ1: Different LLMs attain a similar and near-optimal language proficiency under {\ft}, but their {\icl} ability varies substantially.} In Figure~\ref{fig:fine_tuning_vs_few_shot_all_models}, we report the AUC of {\ft} and {\icl} across LLMs and example sizes while learning language $ \lang_1 $. In both modes, AUC increases with examples, indicating better learning.

\textbf{Fine-tuning.} During {\ft}, all models across families and parameter sizes eventually converge to the optimal AUC ($> 0.99$) after sufficient training examples, such as $ 512 $. Across example sizes $ \{1, 16, 64, 256, 1024\} $, the average AUC of {\ft} is similar across models: Llama-$2$ $(0.93)$ $>$ Qwen ($ 0.92 $) $>$  Mistral $(0.91)$ $>$ Opt $ (0.91)$ $>$ Gemma $(0.90)$ $\ge$ Pythia $(0.90)$ $>$  Llama-$3$ $(0.88)$, where the respective AUC is inside the parentheses.  Only in a few families (e.g., Opt) does the largest model achieve the highest AUC. In addition, a more proficient family often achieves its best language proficiency in an earlier epoch. For example, the median epoch is $7.5$ for Llama-$2$, $12$ for Opt, and $37$ for Llama-$3$. \textit{Therefore, different LLMs, regardless of sizes and families, can achieve similar language proficiency under {\ft} on a tailored task like formal language learning.}

\begin{table}
    \centering
	
        \resizebox{\columnwidth}{!}{
            \begin{tabular}{l p{0.90\columnwidth}}
                \toprule
                  {{\icl} ability (AUC range)} & Model \\
        
                \midrule
                Good ($\ge 0.75$) & 
                    Qwen-$ 2.5 $-$ 7 $B,  
                    Mistral-$7$B, 
                    Qwen-$ 2.5 $-$ 1.5 $B, 
                    Llama-$2$-$13$B,
                    Qwen-$ 2.5 $-$ 0.5 $B, 
                    Llama-$2$-$7$B,
                    Mistral-$12$B
                    \\

                  Moderate ($ \ge 0.6 $) & 
                    Gemma-$2$-$2$B,
                    Gemma-$2$-$9$B,
                    Pythia-$6.9$B,
                    Opt-$1.3$B,
                    Opt-$6.7$B,
                    Pythia-$1$B,
                    Llama-$3.2$-$3$B,
                    Opt-$2.7$B,
                    Llama-$3.2$-$1$B                    
                  \\

                  Poor ($ < 0.6 $) & 
                    Llama-$3.1$-8B,
                    Pythia-$2.8$B
                  \\
                  
                \bottomrule
            \end{tabular}
        }
    \caption{{\icl} ability of LLMs on language $ L_1 $ with up to $ 32 $  examples, based on discriminative AUC. In each group, LLMs are sorted in descending  {\icl} ability.}
    \label{tab:categorizatoin_few_shot}
\end{table}

\textbf{In-context Learning.} In {\icl}, the AUC varies substantially within a model family and across model families. First, we observe that different LLMs have variable context length, restricting each model to a different maximum number of {\icl} examples. To compare all models fairly, we limit our analysis to $32$ {\icl} examples, which all models can fit in their context. We find the following order of {\icl} ability of LLM families: Qwen $(0.78)$ $\ge$ Mistral $(0.78)$ $>$ Llama-$2$ $(0.77)$ $>$  Gemma $(0.69)$ $>$ Opt $(0.64)$ $>$   Pythia $(0.61)$ $>$ Llama-$3$ $(0.59)$. Due to variable performance, we propose a ranking of {\icl} ability of LLMs in Table~\ref{tab:categorizatoin_few_shot}. Importantly, within a family, {\icl} ability does not always correlate with model size (Mistral $ 7 $B $ > $ Mistral-$ 12 $B) or model generations
(Llama-$ 2 $-$ 7 $B $ > $ Llama-$ 3.1 $-$ 8 $B). Only in some families, such as Qwen, Pythia, and Llama-$ 2 $, is the largest model better in {\icl}.  Unlike {\ft}, repeating {\icl} examples more than once worsens {\icl} performance: repeating examples takes up context space, and it is thus better to sample examples from the language distribution without repetition. \textit{To conclude, {\icl} ability is more variable across LLMs, compared to {\ft}.}

\begin{figure}[!t]
	\centering
    \captionsetup[subfigure]{justification=centering}

	\subfloat[Qwen-$2.5$-$7$B]{
		\includegraphics[scale=0.35]{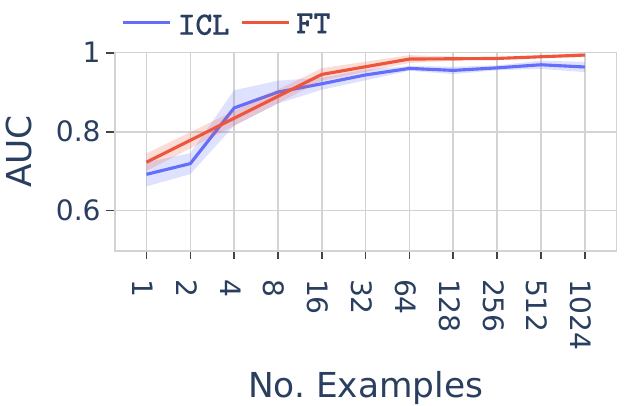}
	}
    \subfloat[Mistral-$7$B]{
	   \includegraphics[scale=0.35]{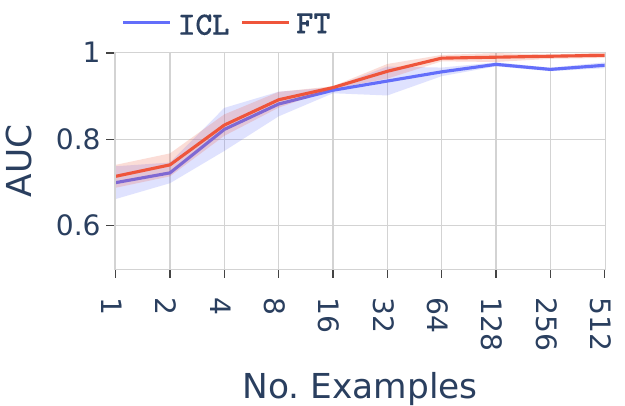}
	}

    \subfloat[Llama-$2$-$7$B]{
        \includegraphics[scale=0.35]{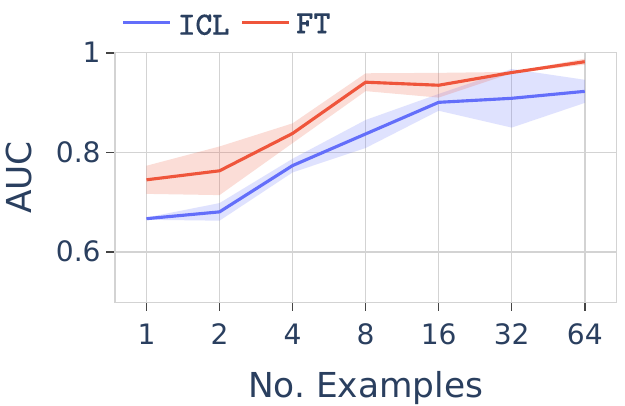}
    }
    \subfloat[Gemma-$2$-$9$B]{
		\includegraphics[scale=0.35]{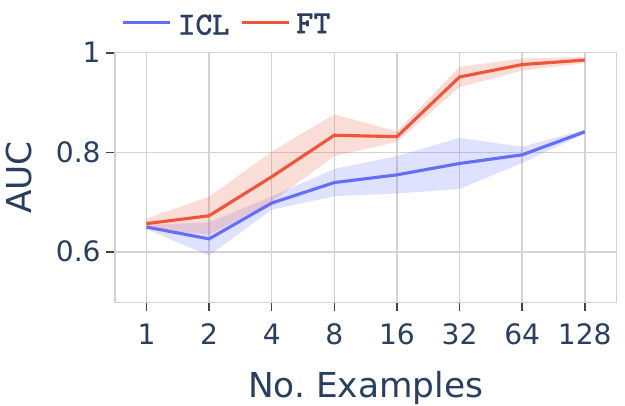}
	}

    \subfloat[Pythia-$6.9$B]{
        \includegraphics[scale=0.35]{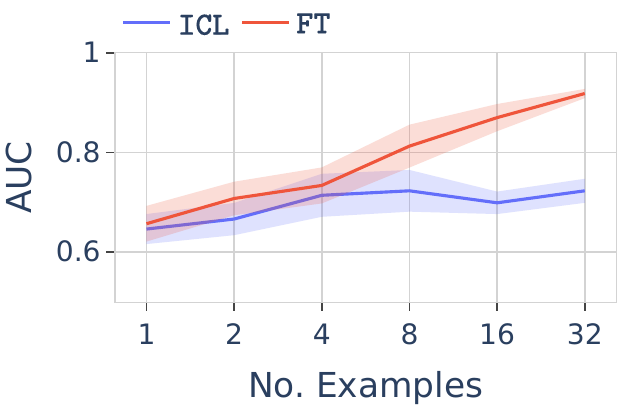}
	}
	\subfloat[Opt-$6.7$B]{
        \includegraphics[scale=0.35]{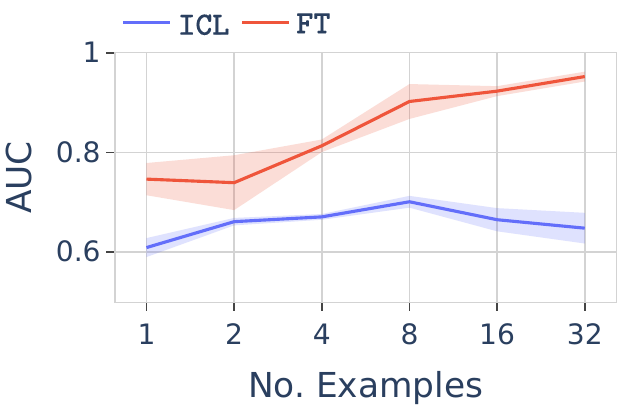}
	}

	\caption{In-distribution generalization of {\ft} vs.\ {\icl} on $ \lang_1 $ in comparable $\approx$ $7$B parameter size LLMs. {\ft} usually dominates {\icl}, except in Qwen-$2.5$-$7$B, Mistral-$7$B, and Llama-$2$-$7$B, where {\icl} is close to {\ft}.
    }

    \label{fig:fine_tuning_vs_few_shot_direct_comparison}
    
\end{figure}

\begin{figure}[!t]
    \captionsetup[subfigure]{justification=centering}
    \centering

    \subfloat[{\ft},  Mistral-$7$B]{
        \includegraphics[scale=0.35]{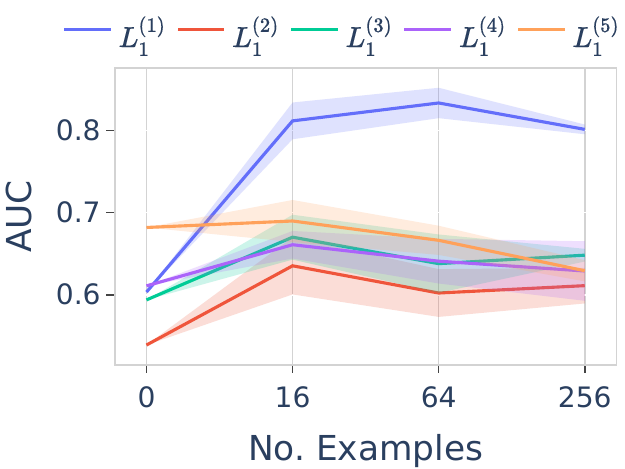
        }
    }
    \subfloat[{\icl},  Mistral-$7$B]{
        \includegraphics[scale=0.35]{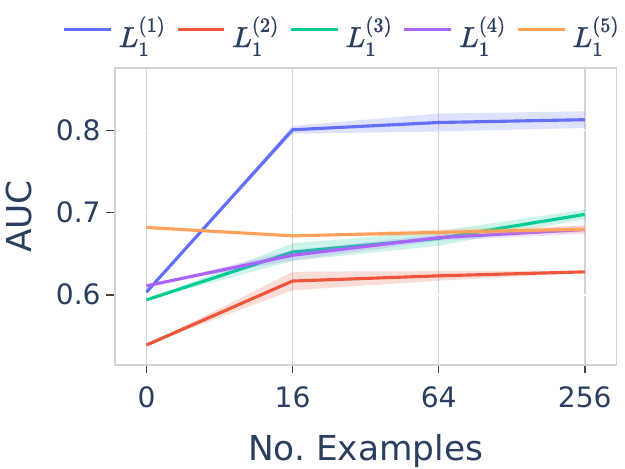
        }
    }

    \subfloat[{\ft},  Llama-$2$-$7$B]{
        \includegraphics[scale=0.35]{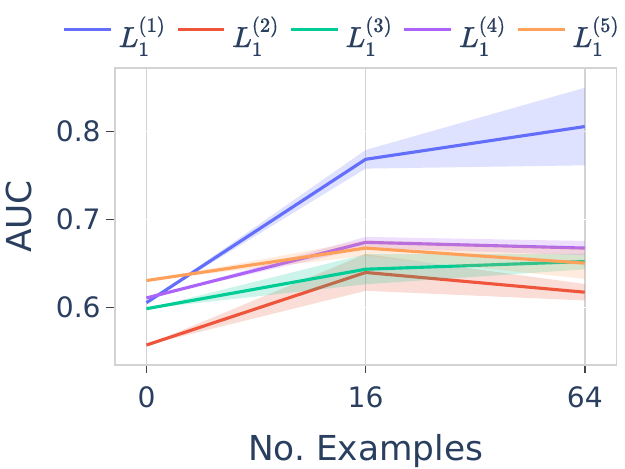
        }
    }
    \subfloat[{\icl},  Llama-$2$-$7$B]{
        \includegraphics[scale=0.35]{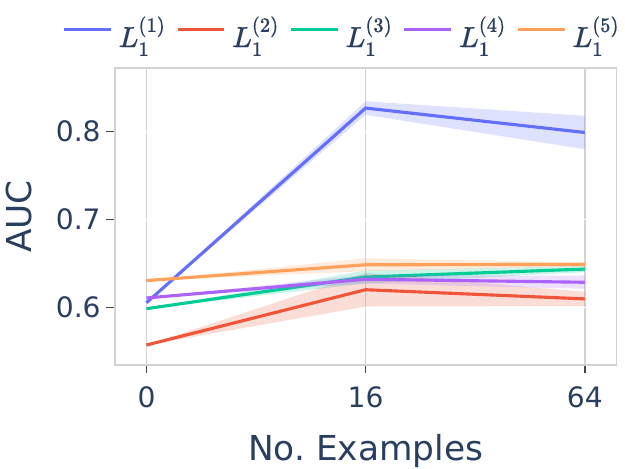
        }
    }

    \caption{Out-of-distribution generalization of {\ft} and {\icl} on increasingly distant languages, where both modes perform almost equally. $\lang_1$ is the base learned language, and generalization is performed on  $\lang_1^{(\ell)}$, by changing $\ell$ rules in the {grammar} of $\lang_1$. $\lang_1^{(\ell)}$ contains all changed rules in $\lang_1^{(\ell-1)}$. Therefore, $\mathtt{dist}(\lang_1, \lang_1^{(\ell-1)}) \leq \mathtt{dist}(\lang_1, \lang_1^{(\ell)})$, where $ 2 \le \ell \le 5$ (see Eq.~\eqref{eq:language_distance}).}

    \label{fig:ood_generalization}
\end{figure}

\paragraph{Answer to RQ2: On in-distribution language generalization, {\ft} dominates {\icl} in most LLMs; only in a subset of LLMs, {\icl} is close to {\ft}. On out-of-distribution generalization, both {\ft} and {\icl} perform similarly, and generalize well to the nearest language only.} In Figure~\ref{fig:fine_tuning_vs_few_shot_direct_comparison}, we compare {\ft} and {\icl} of an LLM on in-distribution language generalization, where evaluation is performed on the same teaching language. In most LLMs, {\ft} dominates {\icl}, and the performance difference becomes more pronounced with more examples. However, in a subset of models, such as Mistral-$7$B, Qwen-$2.5$-$7$B, and Llama-$2$-$7$B, {\icl} is close to {\ft} -- these models are usually ranked as having good {\icl} ability in Table~\ref{tab:categorizatoin_few_shot}. \textit{Therefore, {\ft} is more language proficient than {\icl} on in-distribution language generalization.}

For the comparison of {\ft} and {\icl} on out-of-distribution generalization, the LLM first learns the language $\lang_1$, and then we evaluate the LLM on five other languages $\{\lang_1^{(1)}, \dots, \lang_1^{(5)}\}$ of increasing distances from $\lang_1$ (Figure~\ref{fig:ood_generalization}). We emphasize that formal languages offer a systematic distance computation between two languages (i.e., out-of-distribution tasks), unlike natural language datasets (Appendix~\ref{app_sec:nlp_dataset}). Surprisingly,  both modes perform similarly on out-of-distribution languages, and only perform well on the nearest language $ \lang_1^{(1)}$. \textit{Therefore, the superiority of {\ft} over {\icl} on in-distribution generalization does not extend to out-of-distribution generalization.}

\begin{figure}
    \centering
    \captionsetup[subfigure]{justification=centering}

    \subfloat[Qwen-$2.5$-$7$B]{
        \includegraphics[scale=0.35]{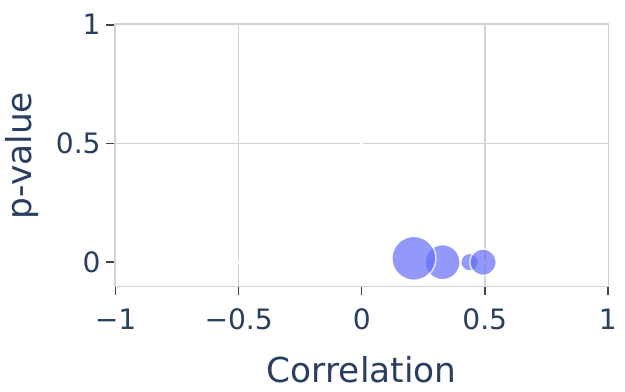}
    }
    \subfloat[Mistral-$7$B]{
        \includegraphics[scale=0.35]{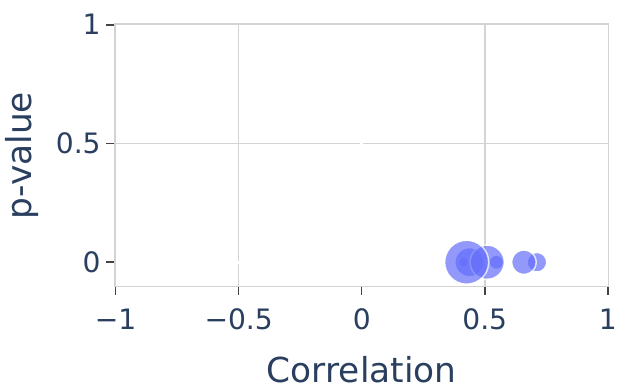}
    }

    \subfloat[Llama-$2$-$7$B]{
        \includegraphics[scale=0.35]{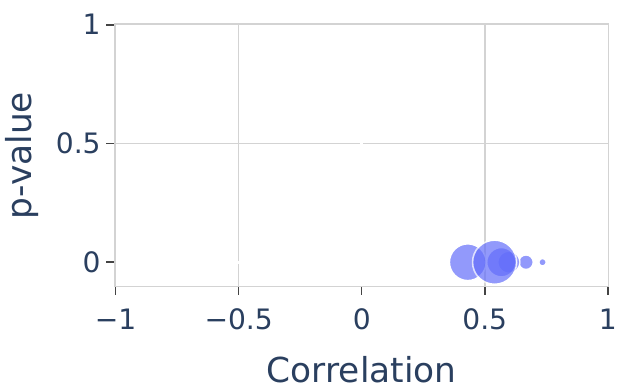}
    }
    \subfloat[Gemma-$2$-$9$B]{
        \includegraphics[scale=0.35]{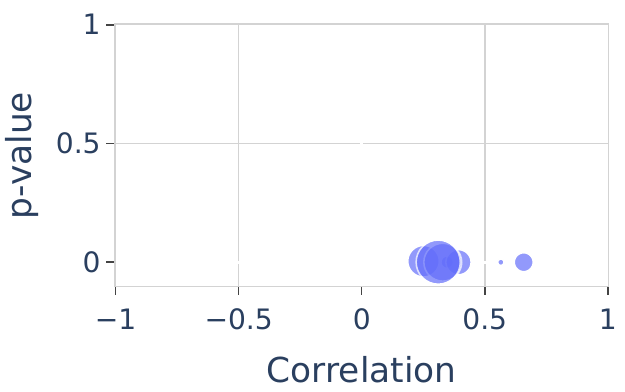}
    }
    \caption{Inductive bias of {\icl} and {\ft}, computed as the Pearson correlation of generation loss of {\ft} and {\icl} on identical test strings. Correlation, despite being positive, tends to decrease with more examples (larger markers).}
    \label{fig:inductive_bias}
\end{figure}

\paragraph{Answer to RQ3: The inductive bias of {\ft} and {\icl} is often similar, but not equal. Similarity decreases with training examples.}

To compare the inductive bias of {\ft} and {\icl}, we do not focus on how each mode operates internally, but on the correlation between their generation losses when evaluated on the same set of strings. Thus, if correlation is high, inductive bias is similar, since both modes find the language similarly easy or difficult to generate.  In Figure~\ref{fig:inductive_bias}, the Pearson correlation is positive ($ < 0.8 $). However, correlation tends to decrease with more training examples, implying that as each mode learns the language better, they do so differently. \textit{To summarize, the inductive bias of {\ft} and {\icl} is often similar, but similarity decreases as each mode learns the language better with more training examples.}

\begin{figure}[!t]
	\centering
    \captionsetup[subfigure]{justification=centering}

    \subfloat[Language $ \lang_1 $\\($ G_\alpha^{\text{Numerical}} $)]{
		\includegraphics[trim={0 1cm 1.2cm 0},clip, scale=0.4]{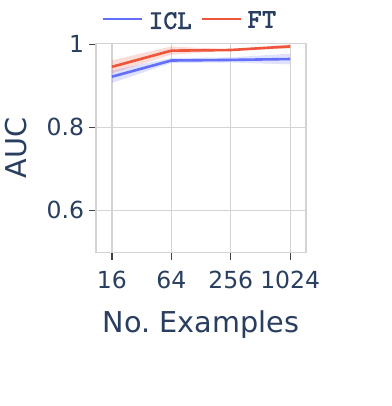}
	}
    \subfloat[Language $ \lang_2 $\\($ G_\alpha^{\text{Latin}} $)]{
		\includegraphics[trim={0 1cm 1.2cm 0},clip,scale=0.4]{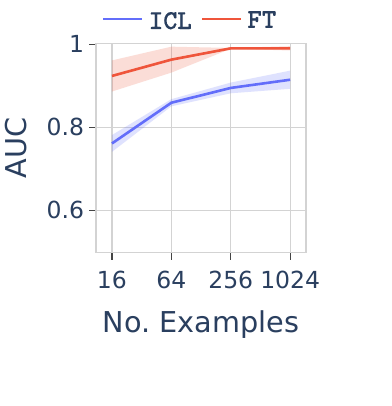}
	}\hfill
    \subfloat[Language $ \lang_3 $\\($ G_\alpha^{\text{Under-trained}} $)]{
		\includegraphics[trim={0 1cm 0 0},clip,scale=0.4]{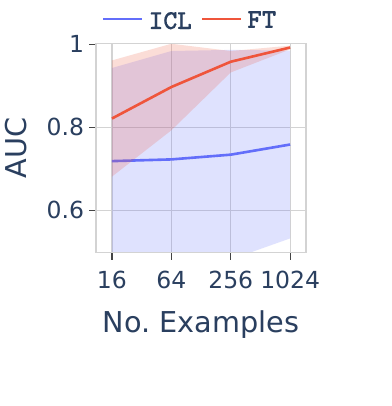}
	}

    \subfloat[Language $ \lang_4 $\\($ G_\beta^{\text{Numerical}} $)]{
		\includegraphics[trim={0 1cm 1.2cm 0},clip, scale=0.4]{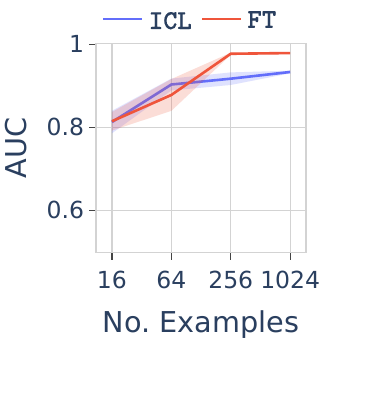}
	}\hfil
    \subfloat[Language $ \lang_5 $\\($ G_\beta^{\text{Latin}} $)]{
		\includegraphics[trim={0 1cm 1.2cm 0},clip, scale=0.4]{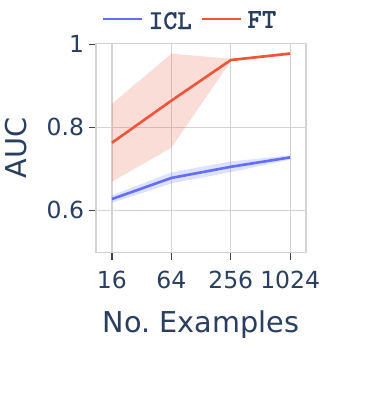}
	}\hfil
    \subfloat[Language $ \lang_6 $\\($ G_\beta^{\text{Under-trained}} $)]{
		\includegraphics[trim={0 1cm 0 0},clip, scale=0.4]{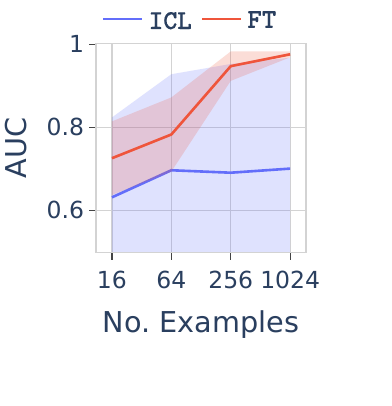}
	}
    
    \caption{Robustness of language proficiency of {\ft} and {\icl} in Qwen-$2.5$-$7$B while varying languages in two ways: changing the grammar rules (rows) and changing the alphabet tokens (columns). The underlying grammar for a language is inside the parentheses. Compared to {\ft}, {\icl} is sensitive to the tokens used in the language, despite having the same underlying grammar.}
    \label{fig:robustness}
    
\end{figure}

\paragraph{Answer to RQ4: {\ft} is more robust to changes in languages than {\icl}.} In Figure~\ref{fig:robustness}, we study the robustness of {\ft} and {\icl} on different languages, by changing the underlying grammar rules: $ G_\alpha $ and $ G_\beta $, and the alphabet: numerical, Latin, and under-trained tokens. {\ft} is better than {\icl} in all languages, consistent with results in in-distribution generalization (Figure~\ref{fig:fine_tuning_vs_few_shot_direct_comparison}). Importantly, changing only the tokens (across columns) introduces more variability than changing the grammar rules (across rows), and the variability is more pronounced in {\icl} than {\ft}. For example, when considering under-trained tokens, i.e., tokens barely seen in pre-training~\cite{land2024fishing}, {\icl} performance is the worst. \textit{Therefore, for robust performance, {\ft} is preferred over {\icl}}.

We further validate our findings \textbf{beyond formal languages to natural languages}, as studied by~\citet{mosbach2023few} (Appendix~\ref{app_sec:nlp_dataset}). These findings hold for natural language on in-distribution generalization, but not for out-of-distribution generalization. In doing so, we identify several issues in natural language datasets such as data contamination and a poor differentiation between in-distribution and out-of-distribution tasks, factors that we carefully avoid in formal languages.

\subsection*{Key Implications of the Study} We draw the following implications from our study (details in Appendix~\ref{app_sec:implications}):

\begin{itemize}[leftmargin=*]
    \item {\ft} is better than {\icl} if the test and training languages are the same. {\icl} is, however, preferred on out-of-distribution languages, where general language understanding of the model is retained as parameters are not updated.

    \item {\ft} and {\icl} are likely to recognize patterns similarly when few examples are given (i.e., before the language is well learned). With more examples, the inductive bias of {\ft} and {\icl} usually differs.
    
    \item Within a family, a larger model size can lead to better {\icl}, but not necessarily better {\ft}. Among model families, Qwen, Mistral, and Llama-$ 2 $ are better in both modes. 
    
    \item Unlike {\ft}, {\icl} is more token-sensitive, despite following the same grammar rules. Since models in the same family may have different pre-training recipes impacting the same tokens differently, we may expect variability in {\icl} within a family (Mistral-$ 7 $B $ > $ Mistral-$ 12 $B, Llama-$ 2 $-$ 7 $B $ > $ Llama-$ 3.1 $-$ 8 $B). However, if the target language contains less-common (under-trained) tokens, {\ft} is preferred.

    \item The discriminative test considers strings outside the language in determining language proficiency. If LLM $ \llm $ assigns higher generation loss to strings in $ L $ than LLM $ \llm' $, but the discriminative AUC of $\llm$ is higher than that of $\llm'$, we still expect $ \llm $ to be more proficient in $ L $, contrary to what generation loss alone would suggest. Here, $ \llm $ may have numerically high generation loss possibly due to model-specific priors, but its higher ability to differentiate strings inside and outside the language makes it more proficient.

\end{itemize}

\section{Conclusion}
We study the language proficiency and inductive bias of {\ft} and {\icl} -- two fundamental learning modes in LLMs. We propose three desiderata for a fair comparison between learning modes, which prior studies overlooked. To satisfy these desiderata, we consider the task of %
formal language learning by an LLM, and propose a comparable discriminative test for evaluating language proficiency.

Our controlled experimental framework leads to important findings: {\ft} is better than {\icl} on in-distribution language generalization, but both perform equally on out-of-distribution generalization. Their inductive bias is similar, but this similarity decreases as both modes learn the language better with more training examples. Unlike {\ft}, {\icl} performance is more sensitive to the tokens used in the language, even with the same grammar rules.

Many of our results on synthetic formal languages are difficult to achieve with poorly controlled natural language datasets. More broadly, formal language learning opens up the possibility of evaluating LLMs in a controlled testbed, enabling precise study of their capabilities beyond what natural language datasets afford.

\section*{Limitations}
Despite the precise controllability of our formal language setup and the utility of our discriminative test as a comparative metric between {\ft} and {\icl}, this work has limitations that warrant further investigation.

\textbf{Formal languages are limited to context-free languages.} The paper focuses on hierarchical context-free languages, which mimic the recursive structure of natural languages. However, we highlight the need for further study to confirm our findings in other classes of formal languages, such as regular and context-sensitive languages.

\textbf{Scope of LLMs.} Our goal is to compare {\ft} and {\icl} on LLMs of equal parameter size. Since {\ft} is more compute-intensive, we limit our experiments to a maximum of $ 13 $B parameter size models. Moreover, we do not perform an extensive hyperparameter search in {\ft}, such as batch size and learning rate. Rather, we find the optimal epoch for each {\ft} run and compare it with the optimal repetition of examples in {\icl}. Furthermore, we restrict experiments to full fine-tuning, while acknowledging that several parameter-efficient fine-tuning methods exist and may lead to different conclusions. We experiment with the non-instruction-tuned models, since a formal language learning task with only syntactic pattern recognition does not require instructions -- comparing {\ft} and {\icl} on instruction-tuned models is left for future work.

\textbf{Larger models ($ > 13 $B) may have better in-context learning performance. Does it invalidate our results?} Since {\icl} is inferior to {\ft} on in-distribution performance, a natural question is whether considering larger models would further improve {\icl}. While we expect {\icl} to improve with model-size, so does {\ft}, and our finding that {\ft} is better than {\icl} on in-distribution languages remains unchanged.

\textbf{We find variable {\icl} performance across LLMs. How can we explain this?} To explain the variability of {\icl} performance, we have conducted two studies: (a) determining whether existing LLMs utilize their full {\icl} context (see Appendix~\ref{app_sec:icl_limit}), and (b) identifying the sensitivity of {\icl} to tokens used in our experiments (see \textbf{RQ4} in Section~\ref{sec:comparison}). The former result identifies which LLMs fully utilize their {\icl} context and which do not. The latter result shows that the tokens used for experimentation have a large impact on {\icl} performance, and the same set of tokens may have been pre-trained to varying degrees across LLMs. While these results are important, we leave a more informed explanation of model-specific {\icl} performance for future work.

\textbf{Inductive bias comparison is based only on the generative test.} We measure inductive bias via generation loss on individual strings. Extending this to a discriminative test requires per-string discrimination: we define a string as \textit{learned} if the LLM assigns it lower loss than all its out-of-language neighbors -- making it a local minimum. Inductive bias then reduces to the correlation of per-string discrimination, which we leave for future work.

\section*{Ethics Statement}
The paper investigates how different learning modes of large language models (LLMs), namely fine-tuning ({\ft}) and in-context learning ({\icl}), compare in their language proficiency and inductive bias. Our experiments involve controlled and synthetically generated formal languages with no human subject involvement or use of private data. As such, the research study does not present immediate ethical risks from the data collection or model training processes. Our scientific results have profound implications for choosing the right mode of learning for LLMs in various applications.

\bibliography{main}

\begin{thebibliography}{77}
\providecommand{\natexlab}[1]{#1}

\bibitem[{Aky{\"u}rek et~al.(2024)Aky{\"u}rek, Wang, Kim, and Andreas}]{akyurek2024context}
Ekin Aky{\"u}rek, Bailin Wang, Yoon Kim, and Jacob Andreas. 2024.
\newblock In-context language learning: Architectures and algorithms.
\newblock In \emph{Proceedings of the 41st International Conference on Machine Learning}, ICML'24. JMLR.org.

\bibitem[{Allen-Zhu and Li(2023)}]{allen2023physics}
Zeyuan Allen-Zhu and Yuanzhi Li. 2023.
\newblock Physics of language models: Part 1, learning hierarchical language structures.
\newblock \emph{arXiv preprint arXiv:2305.13673}.

\bibitem[{Asai et~al.(2024)Asai, Kudugunta, Yu, Blevins, Gonen, Reid, Tsvetkov, Ruder, and Hajishirzi}]{asai2024buffet}
Akari Asai, Sneha Kudugunta, Xinyan Yu, Terra Blevins, Hila Gonen, Machel Reid, Yulia Tsvetkov, Sebastian Ruder, and Hannaneh Hajishirzi. 2024.
\newblock \href {https://doi.org/10.18653/v1/2024.naacl-long.100} {{BUFFET}: Benchmarking large language models for few-shot cross-lingual transfer}.
\newblock In \emph{Proceedings of the 2024 Conference of the North American Chapter of the Association for Computational Linguistics: Human Language Technologies (Volume 1: Long Papers)}, pages 1771--1800, Mexico City, Mexico. Association for Computational Linguistics.

\bibitem[{Awadalla et~al.(2022)Awadalla, Wortsman, Ilharco, Min, Magnusson, Hajishirzi, and Schmidt}]{awadalla2022exploring}
Anas Awadalla, Mitchell Wortsman, Gabriel Ilharco, Sewon Min, Ian Magnusson, Hannaneh Hajishirzi, and Ludwig Schmidt. 2022.
\newblock Exploring the landscape of distributional robustness for question answering models.
\newblock In \emph{Findings of the Association for Computational Linguistics: EMNLP 2022}, Abu Dhabi, United Arab Emirates. Association for Computational Linguistics.

\bibitem[{Bertsch et~al.(2024)Bertsch, Ivgi, Alon, Berant, Gormley, and Neubig}]{bertsch2024context}
Amanda Bertsch, Maor Ivgi, Uri Alon, Jonathan Berant, Matthew~R. Gormley, and Graham Neubig. 2024.
\newblock \href {https://openreview.net/forum?id=4KAmc7vUbq} {In-context learning with long-context models: An in-depth exploration}.
\newblock In \emph{First Workshop on Long-Context Foundation Models @ ICML 2024}.

\bibitem[{Bhatia et~al.(2023)Bhatia, Narayan, De~Sa, and R{\'e}}]{bhatia2023tart}
Kush Bhatia, Avanika Narayan, Christopher~M De~Sa, and Christopher R{\'e}. 2023.
\newblock {TART}: A plug-and-play transformer module for task-agnostic reasoning.
\newblock \emph{Advances in Neural Information Processing Systems}, 36:9751--9788.

\bibitem[{Bhattamishra et~al.(2020)Bhattamishra, Ahuja, and Goyal}]{bhattamishra2020ability}
Satwik Bhattamishra, Kabir Ahuja, and Navin Goyal. 2020.
\newblock On the ability and limitations of transformers to recognize formal languages.
\newblock In \emph{Proceedings of the 2020 Conference on Empirical Methods in Natural Language Processing}, Online. Association for Computational Linguistics.

\bibitem[{Biderman et~al.(2023)Biderman, Schoelkopf, Anthony, Bradley, O’Brien, Hallahan, Khan, Purohit, Prashanth, Raff et~al.}]{biderman2023pythia}
Stella Biderman, Hailey Schoelkopf, Quentin~Gregory Anthony, Herbie Bradley, Kyle O’Brien, Eric Hallahan, Mohammad~Aflah Khan, Shivanshu Purohit, USVSN~Sai Prashanth, Edward Raff, and 1 others. 2023.
\newblock Pythia: A suite for analyzing large language models across training and scaling.
\newblock In \emph{International Conference on Machine Learning}, pages 2397--2430. PMLR.

\bibitem[{Borenstein et~al.(2024)Borenstein, Svete, Chan, Valvoda, Nowak, Augenstein, Chodroff, and Cotterell}]{borenstein2024languages}
Nadav Borenstein, Anej Svete, Robin Chan, Josef Valvoda, Franz Nowak, Isabelle Augenstein, Eleanor Chodroff, and Ryan Cotterell. 2024.
\newblock What languages are easy to language-model? a perspective from learning probabilistic regular languages.
\newblock In \emph{Proceedings of the 62nd Annual Meeting of the Association for Computational Linguistics (Volume 1: Long Papers)}.

\bibitem[{Brown et~al.(2020)Brown, Mann, Ryder, Subbiah, Kaplan, Dhariwal, Neelakantan, Shyam, Sastry, Askell et~al.}]{brown2020language}
Tom Brown, Benjamin Mann, Nick Ryder, Melanie Subbiah, Jared~D Kaplan, Prafulla Dhariwal, Arvind Neelakantan, Pranav Shyam, Girish Sastry, Amanda Askell, and 1 others. 2020.
\newblock Language models are few-shot learners.
\newblock \emph{Advances in neural information processing systems}, 33:1877--1901.

\bibitem[{Chater and Manning(2006)}]{chater2006probabilistic}
Nick Chater and Christopher~D Manning. 2006.
\newblock Probabilistic models of language processing and acquisition.
\newblock \emph{Trends in cognitive sciences}, 10(7):335--344.

\bibitem[{Chen et~al.(2025)Chen, Lin, Zhou, Huang, Jia, Cao, and Wen}]{chen-etal-2025-icleval}
Wentong Chen, Yankai Lin, ZhenHao Zhou, HongYun Huang, YanTao Jia, Zhao Cao, and Ji-Rong Wen. 2025.
\newblock \href {https://aclanthology.org/2025.coling-main.693/} {{ICLE}val: Evaluating in-context learning ability of large language models}.
\newblock In \emph{Proceedings of the 31st International Conference on Computational Linguistics}, pages 10398--10422, Abu Dhabi, UAE. Association for Computational Linguistics.

\bibitem[{Chi et~al.(2023)Chi, Fan, Rudnicky, and Ramadge}]{chi2023transformer}
Ta-Chung Chi, Ting-Han Fan, Alexander~I Rudnicky, and Peter~J Ramadge. 2023.
\newblock Transformer working memory enables regular language reasoning and natural language length extrapolation.
\newblock In \emph{Findings of the Association for Computational Linguistics: EMNLP 2023}, Singapore. Association for Computational Linguistics.

\bibitem[{Chomsky(1956)}]{chomsky1956three}
Noam Chomsky. 1956.
\newblock Three models for the description of language.
\newblock \emph{IRE Transactions on information theory}, 2(3):113--124.

\bibitem[{Collins(2013)}]{collins2013probabilistic}
Michael Collins. 2013.
\newblock Probabilistic context-free grammars ({PCFGs}).

\bibitem[{Cotterell et~al.(2018)Cotterell, Mielke, Eisner, and Roark}]{cotterell2018all}
Ryan Cotterell, Sabrina~J Mielke, Jason Eisner, and Brian Roark. 2018.
\newblock Are all languages equally hard to language-model?
\newblock In \emph{Proceedings of the 2018 Conference of the North American Chapter of the Association for Computational Linguistics: Human Language Technologies, Volume 2 (Short Papers)}, New Orleans, Louisiana. Association for Computational Linguistics.

\bibitem[{de~la Higuera et~al.(2014)de~la Higuera, Scicluna, and Nederhof}]{de2014computation}
Colin de~la Higuera, James Scicluna, and Mark-Jan Nederhof. 2014.
\newblock On the computation of distances for probabilistic context-free grammars.
\newblock \emph{arXiv preprint arXiv:1407.1513}.

\bibitem[{Del{\'e}tang et~al.(2023)Del{\'e}tang, Ruoss, Grau-Moya, Genewein, Wenliang, Catt, Cundy, Hutter, Legg, Veness et~al.}]{deletang2022neural}
Gr{\'e}goire Del{\'e}tang, Anian Ruoss, Jordi Grau-Moya, Tim Genewein, Li~Kevin Wenliang, Elliot Catt, Chris Cundy, Marcus Hutter, Shane Legg, Joel Veness, and 1 others. 2023.
\newblock Neural networks and the chomsky hierarchy.
\newblock In \emph{The Eleventh International Conference on Learning Representations}.

\bibitem[{Dominguez-Olmedo et~al.(2025)Dominguez-Olmedo, Dorner, and Hardt}]{dominguez2024training}
Ricardo Dominguez-Olmedo, Florian~E Dorner, and Moritz Hardt. 2025.
\newblock Training on the test task confounds evaluation and emergence.
\newblock In \emph{The Thirteenth International Conference on Learning Representations}.

\bibitem[{Dubey et~al.(2024)Dubey, Jauhri, Pandey, Kadian, Al-Dahle, Letman, Mathur, Schelten, Yang, Fan et~al.}]{dubey2024llama}
Abhimanyu Dubey, Abhinav Jauhri, Abhinav Pandey, Abhishek Kadian, Ahmad Al-Dahle, Aiesha Letman, Akhil Mathur, Alan Schelten, Amy Yang, Angela Fan, and 1 others. 2024.
\newblock The {Llama} 3 herd of models.
\newblock \emph{arXiv preprint arXiv:2407.21783}.

\bibitem[{Gupta et~al.(2023)Gupta, Sawant, Mishra, Nakamura, Mitra, Mashetty, and Baral}]{gupta2023instruction}
Himanshu Gupta, Saurabh~Arjun Sawant, Swaroop Mishra, Mutsumi Nakamura, Arindam Mitra, Santosh Mashetty, and Chitta Baral. 2023.
\newblock Instruction tuned models are quick learners.
\newblock \emph{arXiv preprint arXiv:2306.05539}.

\bibitem[{Hahn(2020)}]{hahn2020theoretical}
Michael Hahn. 2020.
\newblock Theoretical limitations of self-attention in neural sequence models.
\newblock \emph{Transactions of the Association for Computational Linguistics}, 8:156--171.

\bibitem[{Hahn and Rofin(2024)}]{hahn2024sensitive}
Michael Hahn and Mark Rofin. 2024.
\newblock Why are sensitive functions hard for transformers?
\newblock In \emph{Proceedings of the 62nd Annual Meeting of the Association for Computational Linguistics (Volume 1: Long Papers)}, Bangkok, Thailand. Association for Computational Linguistics.

\bibitem[{Hopkins(2022)}]{hopkins2022towards}
Mark Hopkins. 2022.
\newblock Towards more natural artificial languages.
\newblock In \emph{Proceedings of the 26th Conference on Computational Natural Language Learning (CoNLL)}, pages 85--94.

\bibitem[{Hu et~al.(2022)Hu, yelong shen, Wallis, Allen-Zhu, Li, Wang, Wang, and Chen}]{hu2022lora}
Edward~J Hu, yelong shen, Phillip Wallis, Zeyuan Allen-Zhu, Yuanzhi Li, Shean Wang, Lu~Wang, and Weizhu Chen. 2022.
\newblock \href {https://openreview.net/forum?id=nZeVKeeFYf9} {Lo{RA}: Low-rank adaptation of large language models}.
\newblock In \emph{International Conference on Learning Representations}.

\bibitem[{Hu et~al.(2024)Hu, Tu, Han, Cui, He, Zhao, Long, Zheng, Fang, Huang, Zhang, Thai, Wang, Yao, Zhao, Zhou, Cai, Zhai, Ding, Jia, Zeng, dahai li, Liu, and Sun}]{hu2024minicpm}
Shengding Hu, Yuge Tu, Xu~Han, Ganqu Cui, Chaoqun He, Weilin Zhao, Xiang Long, Zhi Zheng, Yewei Fang, Yuxiang Huang, Xinrong Zhang, Zhen~Leng Thai, Chongyi Wang, Yuan Yao, Chenyang Zhao, Jie Zhou, Jie Cai, Zhongwu Zhai, Ning Ding, and 5 others. 2024.
\newblock \href {https://openreview.net/forum?id=3X2L2TFr0f} {Mini{CPM}: Unveiling the potential of small language models with scalable training strategies}.
\newblock In \emph{First Conference on Language Modeling}.

\bibitem[{Icard(2020)}]{icard2020calibrating}
Thomas~F Icard. 2020.
\newblock Calibrating generative models: The probabilistic {Chomsky--Sch{\"u}tzenberger} hierarchy.
\newblock \emph{Journal of Mathematical Psychology}, 95:102308.

\bibitem[{Jiang et~al.(2023)Jiang, Sablayrolles, Mensch, Bamford, Chaplot, de~las Casas, Bressand, Lengyel, Lample, Saulnier, Lavaud, Lachaux, Stock, Scao, Lavril, Wang, Lacroix, and Sayed}]{jiang2023mistral7b}
Albert~Q. Jiang, Alexandre Sablayrolles, Arthur Mensch, Chris Bamford, Devendra~Singh Chaplot, Diego de~las Casas, Florian Bressand, Gianna Lengyel, Guillaume Lample, Lucile Saulnier, Lélio~Renard Lavaud, Marie-Anne Lachaux, Pierre Stock, Teven~Le Scao, Thibaut Lavril, Thomas Wang, Timothée Lacroix, and William~El Sayed. 2023.
\newblock \href {https://arxiv.org/abs/2310.06825} {Mistral 7b}.

\bibitem[{Jumelet and Zuidema(2023)}]{jumelet2023transparency}
Jaap Jumelet and Willem Zuidema. 2023.
\newblock Transparency at the source: Evaluating and interpreting language models with access to the true distribution.
\newblock In \emph{Findings of the Association for Computational Linguistics: EMNLP 2023}, Singapore. Association for Computational Linguistics.

\bibitem[{Kallini et~al.(2024)Kallini, Papadimitriou, Futrell, Mahowald, and Potts}]{kallini2024mission}
Julie Kallini, Isabel Papadimitriou, Richard Futrell, Kyle Mahowald, and Christopher Potts. 2024.
\newblock Mission: Impossible language models.
\newblock In \emph{Proceedings of the 62nd Annual Meeting of the Association for Computational Linguistics (Volume 1: Long Papers)}, Bangkok, Thailand. Association for Computational Linguistics.

\bibitem[{Kaneko et~al.(2025)Kaneko, Bollegala, and Baldwin}]{kaneko2025gaps}
Masahiro Kaneko, Danushka Bollegala, and Timothy Baldwin. 2025.
\newblock The gaps between fine tuning and in-context learning in bias evaluation and debiasing.
\newblock In \emph{Proceedings of the 31st International Conference on Computational Linguistics}, pages 2758--2764.

\bibitem[{Kaplan et~al.(2020)Kaplan, McCandlish, Henighan, Brown, Chess, Child, Gray, Radford, Wu, and Amodei}]{kaplan2020scaling}
Jared Kaplan, Sam McCandlish, Tom Henighan, Tom~B Brown, Benjamin Chess, Rewon Child, Scott Gray, Alec Radford, Jeffrey Wu, and Dario Amodei. 2020.
\newblock Scaling laws for neural language models.
\newblock \emph{arXiv preprint arXiv:2001.08361}.

\bibitem[{Kwiatkowski et~al.(2019)Kwiatkowski, Palomaki, Redfield, Collins, Parikh, Alberti, Epstein, Polosukhin, Kelcey, Devlin, Lee, Toutanova, Jones, Chang, Dai, Uszkoreit, Le, and Petrov}]{kwiatkwoski2019natq}
Tom Kwiatkowski, Jennimaria Palomaki, Olivia Redfield, Michael Collins, Ankur Parikh, Chris Alberti, Danielle Epstein, Illia Polosukhin, Matthew Kelcey, Jacob Devlin, Kenton Lee, Kristina~N. Toutanova, Llion Jones, Ming-Wei Chang, Andrew Dai, Jakob Uszkoreit, Quoc Le, and Slav Petrov. 2019.
\newblock Natural questions: a benchmark for question answering research.
\newblock \emph{Transactions of the Association of Computational Linguistics}.

\bibitem[{Land and Bartolo(2024)}]{land2024fishing}
Sander Land and Max Bartolo. 2024.
\newblock Fishing for {Magikarp}: Automatically detecting under-trained tokens in large language models.
\newblock In \emph{Proceedings of the 2024 Conference on Empirical Methods in Natural Language Processing}, Miami, Florida, USA. Association for Computational Linguistics.

\bibitem[{Le~Scao and Rush(2021)}]{le2021many}
Teven Le~Scao and Alexander~M Rush. 2021.
\newblock How many data points is a prompt worth?
\newblock In \emph{Proceedings of the 2021 Conference of the North American Chapter of the Association for Computational Linguistics: Human Language Technologies}, pages 2627--2636.

\bibitem[{Lehman et~al.(2023)Lehman, Hernandez, Mahajan, Wulff, Smith, Ziegler, Nadler, Szolovits, Johnson, and Alsentzer}]{lehman2023we}
Eric Lehman, Evan Hernandez, Diwakar Mahajan, Jonas Wulff, Micah~J Smith, Zachary Ziegler, Daniel Nadler, Peter Szolovits, Alistair Johnson, and Emily Alsentzer. 2023.
\newblock Do we still need clinical language models?
\newblock In \emph{Conference on health, inference, and learning}, pages 578--597. PMLR.

\bibitem[{Lin and Lee(2024)}]{lin2024dual}
Ziqian Lin and Kangwook Lee. 2024.
\newblock \href {https://openreview.net/forum?id=5H4nJIGqmK} {Dual operating modes of in-context learning}.
\newblock In \emph{ICLR 2024 Workshop on Mathematical and Empirical Understanding of Foundation Models}.

\bibitem[{Liu et~al.(2023)Liu, Ash, Goel, Krishnamurthy, and Zhang}]{liu2022transformers}
Bingbin Liu, Jordan~T Ash, Surbhi Goel, Akshay Krishnamurthy, and Cyril Zhang. 2023.
\newblock Transformers learn shortcuts to automata.
\newblock In \emph{The Eleventh International Conference on Learning Representations}.

\bibitem[{Liu et~al.(2022)Liu, Tam, Muqeeth, Mohta, Huang, Bansal, and Raffel}]{liu2022few}
Haokun Liu, Derek Tam, Mohammed Muqeeth, Jay Mohta, Tenghao Huang, Mohit Bansal, and Colin~A Raffel. 2022.
\newblock Few-shot parameter-efficient fine-tuning is better and cheaper than in-context learning.
\newblock \emph{Advances in Neural Information Processing Systems}, 35:1950--1965.

\bibitem[{Manning(2003)}]{manning2003probabilistic}
Christopher~D Manning. 2003.
\newblock Probabilistic syntax.
\newblock \emph{Probabilistic linguistics}, 289341.

\bibitem[{Merrill(2023)}]{merrill2023formal}
William Merrill. 2023.
\newblock Formal languages and the {NLP} black box.
\newblock In \emph{International Conference on Developments in Language Theory}, pages 1--8. Springer.

\bibitem[{Mesnard et~al.(2024)Mesnard, Hardin, Dadashi, Bhupatiraju, Pathak, Sifre, Rivière, Kale, Love, Tafti, Hussenot, Sessa, Chowdhery, Roberts, Barua, Botev, Castro-Ros, Slone, Héliou, Tacchetti, Bulanova, Paterson, Tsai, Shahriari, Lan, Choquette-Choo, Crepy, Cer, Ippolito, Reid, Buchatskaya, Ni, Noland, Yan, Tucker, Muraru, Rozhdestvenskiy, Michalewski, Tenney, Grishchenko, Austin, Keeling, Labanowski, Lespiau, Stanway, Brennan, Chen, Ferret, Chiu, Mao-Jones, Lee, Yu, Millican, Sjoesund, Lee, Dixon, Reid, Mikuła, Wirth, Sharman, Chinaev, Thain, Bachem, Chang, Wahltinez, Bailey, Michel, Yotov, Chaabouni, Comanescu, Jana, Anil, McIlroy, Liu, Mullins, Smith, Borgeaud, Girgin, Douglas, Pandya, Shakeri, De, Klimenko, Hennigan, Feinberg, Stokowiec, hui Chen, Ahmed, Gong, Warkentin, Peran, Giang, Farabet, Vinyals, Dean, Kavukcuoglu, Hassabis, Ghahramani, Eck, Barral, Pereira, Collins, Joulin, Fiedel, Senter, Andreev, and Kenealy}]{gemmateam2024gemmaopenmodelsbased}
Thomas Mesnard, Cassidy Hardin, Robert Dadashi, Surya Bhupatiraju, Shreya Pathak, Laurent Sifre, Morgane Rivière, Mihir~Sanjay Kale, Juliette Love, Pouya Tafti, Léonard Hussenot, Pier~Giuseppe Sessa, Aakanksha Chowdhery, Adam Roberts, Aditya Barua, Alex Botev, Alex Castro-Ros, Ambrose Slone, Amélie Héliou, and 88 others. 2024.
\newblock Gemma: Open models based on {Gemini} research and technology.
\newblock \emph{arXiv preprint arXiv:2403.08295}.

\bibitem[{Mielke et~al.(2019)Mielke, Cotterell, Gorman, Roark, and Eisner}]{mielke2019kind}
Sabrina~J Mielke, Ryan Cotterell, Kyle Gorman, Brian Roark, and Jason Eisner. 2019.
\newblock What kind of language is hard to language-model?
\newblock In \emph{Proceedings of the 57th Annual Meeting of the Association for Computational Linguistics}, Florence, Italy. Association for Computational Linguistics.

\bibitem[{Mosbach et~al.(2023)Mosbach, Pimentel, Ravfogel, Klakow, and Elazar}]{mosbach2023few}
Marius Mosbach, Tiago Pimentel, Shauli Ravfogel, Dietrich Klakow, and Yanai Elazar. 2023.
\newblock \href {https://doi.org/10.18653/v1/2023.findings-acl.779} {Few-shot fine-tuning vs. in-context learning: A fair comparison and evaluation}.
\newblock In \emph{Findings of the Association for Computational Linguistics: ACL 2023}, pages 12284--12314, Toronto, Canada. Association for Computational Linguistics.

\bibitem[{Murty et~al.(2023)Murty, Sharma, Andreas, and Manning}]{murty2022characterizing}
Shikhar Murty, Pratyusha Sharma, Jacob Andreas, and Christopher~D Manning. 2023.
\newblock Characterizing intrinsic compositionality in transformers with tree projections.
\newblock In \emph{The Eleventh International Conference on Learning Representations}.

\bibitem[{Oliver and Wang(2024)}]{oliver2024crafting}
Michael Oliver and Guan Wang. 2024.
\newblock Crafting efficient fine-tuning strategies for large language models.
\newblock \emph{arXiv preprint arXiv:2407.13906}.

\bibitem[{Ouyang et~al.(2022)Ouyang, Wu, Jiang, Almeida, Wainwright, Mishkin, Zhang, Agarwal, Slama, Ray et~al.}]{ouyang2022training}
Long Ouyang, Jeffrey Wu, Xu~Jiang, Diogo Almeida, Carroll Wainwright, Pamela Mishkin, Chong Zhang, Sandhini Agarwal, Katarina Slama, Alex Ray, and 1 others. 2022.
\newblock Training language models to follow instructions with human feedback.
\newblock \emph{Advances in neural information processing systems}, 35:27730--27744.

\bibitem[{Pan et~al.(2023)Pan, Gao, Chen, and Chen}]{pan-etal-2023-context}
Jane Pan, Tianyu Gao, Howard Chen, and Danqi Chen. 2023.
\newblock \href {https://doi.org/10.18653/v1/2023.findings-acl.527} {What in-context learning {\textquotedblleft}learns{\textquotedblright} in-context: Disentangling task recognition and task learning}.
\newblock In \emph{Findings of the Association for Computational Linguistics: ACL 2023}, pages 8298--8319, Toronto, Canada. Association for Computational Linguistics.

\bibitem[{Papadimitriou and Jurafsky(2023)}]{papadimitriou2023injecting}
Isabel Papadimitriou and Dan Jurafsky. 2023.
\newblock Injecting structural hints: Using language models to study inductive biases in language learning.
\newblock In \emph{Findings of the Association for Computational Linguistics: EMNLP 2023}, Singapore. Association for Computational Linguistics.

\bibitem[{Pecher et~al.(2025)Pecher, Srba, and Bielikova}]{pecher2025comparing}
Branislav Pecher, Ivan Srba, and Maria Bielikova. 2025.
\newblock Comparing specialised small and general large language models on text classification: 100 labelled samples to achieve break-even performance.
\newblock In \emph{Proceedings of the 2025 Conference on Empirical Methods in Natural Language Processing}, pages 165--184.

\bibitem[{Radford et~al.(2019)Radford, Wu, Child, Luan, Amodei, and Sutskever}]{radford2019language}
Alec Radford, Jeff Wu, Rewon Child, David Luan, Dario Amodei, and Ilya Sutskever. 2019.
\newblock Language models are unsupervised multitask learners.

\bibitem[{Rajpurkar et~al.(2016)Rajpurkar, Zhang, Lopyrev, and Liang}]{rajpurkar2016squad}
Pranav Rajpurkar, Jian Zhang, Konstantin Lopyrev, and Percy Liang. 2016.
\newblock \href {https://doi.org/10.18653/v1/D16-1264} {{SQ}u{AD}: 100,000+ questions for machine comprehension of text}.
\newblock In \emph{Proceedings of the 2016 Conference on Empirical Methods in Natural Language Processing}, pages 2383--2392, Austin, Texas. Association for Computational Linguistics.

\bibitem[{Ravfogel et~al.(2019)Ravfogel, Goldberg, and Linzen}]{ravfogel2019studying}
Shauli Ravfogel, Yoav Goldberg, and Tal Linzen. 2019.
\newblock Studying the inductive biases of {RNNs} with synthetic variations of natural languages.
\newblock In \emph{Proceedings of the 2019 Conference of the North American Chapter of the Association for Computational Linguistics: Human Language Technologies, Volume 1 (Long and Short Papers)}, Minneapolis, Minnesota. Association for Computational Linguistics.

\bibitem[{Razavi et~al.(2025)Razavi, Soltangheis, Arabzadeh, Salamat, Zihayat, and Bagheri}]{razavi2025benchmarking}
Amirhossein Razavi, Mina Soltangheis, Negar Arabzadeh, Sara Salamat, Morteza Zihayat, and Ebrahim Bagheri. 2025.
\newblock Benchmarking prompt sensitivity in large language models.
\newblock In \emph{European Conference on Information Retrieval}, pages 303--313. Springer.

\bibitem[{Reddy(2024)}]{reddy2023mechanistic}
Gautam Reddy. 2024.
\newblock The mechanistic basis of data dependence and abrupt learning in an in-context classification task.
\newblock In \emph{The Twelfth International Conference on Learning Representations}.

\bibitem[{Riviere et~al.(2024)Riviere, Pathak, Sessa, Hardin, Bhupatiraju, Hussenot, Mesnard, Shahriari, Ramé, Ferret, Liu, Tafti, Friesen, Casbon, Ramos, Kumar, Lan, Jerome, Tsitsulin, Vieillard, Stanczyk, Girgin, Momchev, Hoffman, Thakoor, Grill, Neyshabur, Bachem, Walton, Severyn, Parrish, Ahmad, Hutchison, Abdagic, Carl, Shen, Brock, Coenen, Laforge, Paterson, Bastian, Piot, Wu, Royal, Chen, Kumar, Perry, Welty, Choquette-Choo, Sinopalnikov, Weinberger, Vijaykumar, Rogozińska, Herbison, Bandy, Wang, Noland, Moreira, Senter, Eltyshev, Visin, Rasskin, Wei, Cameron, Martins, Hashemi, Klimczak-Plucińska, Batra, Dhand, Nardini, Mein, Zhou, Svensson, Stanway, Chan, Zhou, Carrasqueira, Iljazi, Becker, Fernandez, van Amersfoort, Gordon, Lipschultz, Newlan, yeong Ji, Mohamed, Badola, Black, Millican, McDonell, Nguyen, Sodhia, Greene, Sjoesund, Usui, Sifre, Heuermann, Lago, McNealus, Soares, Kilpatrick, Dixon, Martins, Reid, Singh, Iverson, Görner, Velloso, Wirth, Davidow, Miller, Rahtz, Watson, Risdal, Kazemi, Moynihan, Zhang, Kahng, Park, Rahman, Khatwani, Dao, Bardoliwalla, Devanathan, Dumai, Chauhan, Wahltinez, Botarda, Barnes, Barham, Michel, Jin, Georgiev, Culliton, Kuppala, Comanescu, Merhej, Jana, Rokni, Agarwal, Mullins, Saadat, Carthy, Cogan, Perrin, Arnold, Krause, Dai, Garg, Sheth, Ronstrom, Chan, Jordan, Yu, Eccles, Hennigan, Kocisky, Doshi, Jain, Yadav, Meshram, Dharmadhikari, Barkley, Wei, Ye, Han, Kwon, Xu, Shen, Gong, Wei, Cotruta, Kirk, Rao, Giang, Peran, Warkentin, Collins, Barral, Ghahramani, Hadsell, Sculley, Banks, Dragan, Petrov, Vinyals, Dean, Hassabis, Kavukcuoglu, Farabet, Buchatskaya, Borgeaud, Fiedel, Joulin, Kenealy, Dadashi, and Andreev}]{gemmateam2024gemma2improvingopen}
Morgane Riviere, Shreya Pathak, Pier~Giuseppe Sessa, Cassidy Hardin, Surya Bhupatiraju, Léonard Hussenot, Thomas Mesnard, Bobak Shahriari, Alexandre Ramé, Johan Ferret, Peter Liu, Pouya Tafti, Abe Friesen, Michelle Casbon, Sabela Ramos, Ravin Kumar, Charline~Le Lan, Sammy Jerome, Anton Tsitsulin, and 178 others. 2024.
\newblock Gemma 2: Improving open language models at a practical size.
\newblock \emph{arXiv preprint arXiv:2408.00118}.

\bibitem[{Shen et~al.(2023)Shen, Mishra, and Khashabi}]{shen2023pretrained}
Lingfeng Shen, Aayush Mishra, and Daniel Khashabi. 2023.
\newblock Do pretrained transformers really learn in-context by gradient descent?
\newblock \emph{arXiv preprint arXiv:2310.08540}.

\bibitem[{Shi et~al.(2022)Shi, Gao, Tian, Chen, and Zhao}]{shi2022learning}
Hui Shi, Sicun Gao, Yuandong Tian, Xinyun Chen, and Jishen Zhao. 2022.
\newblock Learning bounded context-free-grammar via {LSTM} and the transformer: difference and the explanations.
\newblock In \emph{Proceedings of the AAAI conference on artificial intelligence}, volume~36, pages 8267--8276.

\bibitem[{Soudani et~al.(2024)Soudani, Kanoulas, and Hasibi}]{soudani2024fine}
Heydar Soudani, Evangelos Kanoulas, and Faegheh Hasibi. 2024.
\newblock Fine tuning vs. retrieval augmented generation for less popular knowledge.
\newblock In \emph{Proceedings of the 2024 Annual International ACM SIGIR Conference on Research and Development in Information Retrieval in the Asia Pacific Region}, pages 12--22.

\bibitem[{Srinivasan et~al.(2024)Srinivasan, Gumpena, Yattapu, and Brahmbhatt}]{srinivasan2024comparative}
Krishna Prasad~Varadarajan Srinivasan, Prasanth Gumpena, Madhusudhana Yattapu, and Vishal~H Brahmbhatt. 2024.
\newblock Comparative analysis of different efficient fine tuning methods of large language models ({LLMs}) in low-resource setting.
\newblock \emph{arXiv preprint arXiv:2405.13181}.

\bibitem[{Strobl et~al.(2023)Strobl, Merrill, Weiss, Chiang, and Angluin}]{strobl2023transformers}
Lena Strobl, William Merrill, Gail Weiss, David Chiang, and Dana Angluin. 2023.
\newblock Transformers as recognizers of formal languages: A survey on expressivity.
\newblock \emph{arXiv preprint arXiv:2311.00208}.

\bibitem[{Su et~al.(2023)Su, Kasai, Wu, Shi, Wang, Xin, Zhang, Ostendorf, Zettlemoyer, Smith et~al.}]{su2022selective}
Hongjin Su, Jungo Kasai, Chen~Henry Wu, Weijia Shi, Tianlu Wang, Jiayi Xin, Rui Zhang, Mari Ostendorf, Luke Zettlemoyer, Noah~A Smith, and 1 others. 2023.
\newblock Selective annotation makes language models better few-shot learners.
\newblock In \emph{The Eleventh International Conference on Learning Representations}.

\bibitem[{Touvron et~al.(2023{\natexlab{a}})Touvron, Lavril, Izacard, Martinet, Lachaux, Lacroix, Rozi{\`e}re, Goyal, Hambro, Azhar et~al.}]{touvron2023llama}
Hugo Touvron, Thibaut Lavril, Gautier Izacard, Xavier Martinet, Marie-Anne Lachaux, Timoth{\'e}e Lacroix, Baptiste Rozi{\`e}re, Naman Goyal, Eric Hambro, Faisal Azhar, and 1 others. 2023{\natexlab{a}}.
\newblock Llama: Open and efficient foundation language models.
\newblock \emph{arXiv preprint arXiv:2302.13971}.

\bibitem[{Touvron et~al.(2023{\natexlab{b}})Touvron, Martin, Stone, Albert, Almahairi, Babaei, Bashlykov, Batra, Bhargava, Bhosale et~al.}]{touvron2023llama-2}
Hugo Touvron, Louis Martin, Kevin Stone, Peter Albert, Amjad Almahairi, Yasmine Babaei, Nikolay Bashlykov, Soumya Batra, Prajjwal Bhargava, Shruti Bhosale, and 1 others. 2023{\natexlab{b}}.
\newblock Llama 2: Open foundation and fine-tuned chat models.
\newblock \emph{arXiv preprint arXiv:2307.09288}.

\bibitem[{Wang(2021)}]{wang2021evaluating}
Shunjie Wang. 2021.
\newblock \emph{Evaluating transformer’s ability to learn mildly context-sensitive languages}.
\newblock University of Washington.

\bibitem[{Wei et~al.(2023)Wei, Wei, Tay, Tran, Webson, Lu, Chen, Liu, Huang, Zhou et~al.}]{wei2023larger}
Jerry Wei, Jason Wei, Yi~Tay, Dustin Tran, Albert Webson, Yifeng Lu, Xinyun Chen, Hanxiao Liu, Da~Huang, Denny Zhou, and 1 others. 2023.
\newblock Larger language models do in-context learning differently.
\newblock \emph{arXiv preprint arXiv:2303.03846}.

\bibitem[{White and Cotterell(2021)}]{white2021examining}
Jennifer~C White and Ryan Cotterell. 2021.
\newblock Examining the inductive bias of neural language models with artificial languages.
\newblock In \emph{Proceedings of the 59th Annual Meeting of the Association for Computational Linguistics and the 11th International Joint Conference on Natural Language Processing (Volume 1: Long Papers)}, Online. Association for Computational Linguistics.

\bibitem[{Williams et~al.(2018)Williams, Nangia, and Bowman}]{N18-1101}
Adina Williams, Nikita Nangia, and Samuel Bowman. 2018.
\newblock \href {http://aclweb.org/anthology/N18-1101} {A broad-coverage challenge corpus for sentence understanding through inference}.
\newblock In \emph{Proceedings of the 2018 Conference of the North American Chapter of the Association for Computational Linguistics: Human Language Technologies, Volume 1 (Long Papers)}, pages 1112--1122. Association for Computational Linguistics.

\bibitem[{Wu et~al.(2025)Wu, Khan, Das, Nanda, Ghosh, Kolling, Speicher, Bindschaedler, Gummadi, and Terzi}]{wu2025towards}
Qinyuan Wu, Mohammad~Aflah Khan, Soumi Das, Vedant Nanda, Bishwamittra Ghosh, Camila Kolling, Till Speicher, Laurent Bindschaedler, Krishna Gummadi, and Evimaria Terzi. 2025.
\newblock Towards reliable latent knowledge estimation in llms: Zero-prompt many-shot based factual knowledge extraction.
\newblock In \emph{Proceedings of the Eighteenth ACM International Conference on Web Search and Data Mining}, pages 754--763.

\bibitem[{Xu et~al.(2024)Xu, Guan, Greene, Kechadi et~al.}]{xu2024benchmark}
Cheng Xu, Shuhao Guan, Derek Greene, M~Kechadi, and 1 others. 2024.
\newblock Benchmark data contamination of large language models: A survey.
\newblock \emph{arXiv preprint arXiv:2406.04244}.

\bibitem[{Yang et~al.(2024)Yang, Yang, Zhang, Hui, Zheng, Yu, Li, Liu, Huang, Wei et~al.}]{yang2024qwen2}
An~Yang, Baosong Yang, Beichen Zhang, Binyuan Hui, Bo~Zheng, Bowen Yu, Chengyuan Li, Dayiheng Liu, Fei Huang, Haoran Wei, and 1 others. 2024.
\newblock Qwen2.5 technical report.
\newblock \emph{arXiv preprint arXiv:2412.15115}.

\bibitem[{Yang et~al.(2018)Yang, Qi, Zhang, Bengio, Cohen, Salakhutdinov, and Manning}]{yang2018hotpotqa}
Zhilin Yang, Peng Qi, Saizheng Zhang, Yoshua Bengio, William Cohen, Ruslan Salakhutdinov, and Christopher~D. Manning. 2018.
\newblock \href {https://doi.org/10.18653/v1/D18-1259} {{H}otpot{QA}: A dataset for diverse, explainable multi-hop question answering}.
\newblock In \emph{Proceedings of the 2018 Conference on Empirical Methods in Natural Language Processing}, pages 2369--2380, Brussels, Belgium. Association for Computational Linguistics.

\bibitem[{Yin et~al.(2024)Yin, He, Leong, Wang, Yan, Shen, and Zhang}]{yin2024deeper}
Qingyu Yin, Xuzheng He, Chak~Tou Leong, Fan Wang, Yanzhao Yan, Xiaoyu Shen, and Qiang Zhang. 2024.
\newblock \href {https://doi.org/10.18653/v1/2024.findings-emnlp.239} {Deeper insights without updates: The power of in-context learning over fine-tuning}.
\newblock In \emph{Findings of the Association for Computational Linguistics: EMNLP 2024}, pages 4138--4151, Miami, Florida, USA. Association for Computational Linguistics.

\bibitem[{Zhang et~al.(2024)Zhang, Liu, Cherry, and Firat}]{zhang2024when}
Biao Zhang, Zhongtao Liu, Colin Cherry, and Orhan Firat. 2024.
\newblock \href {https://openreview.net/forum?id=5HCnKDeTws} {When scaling meets {LLM} finetuning: The effect of data, model and finetuning method}.
\newblock In \emph{The Twelfth International Conference on Learning Representations}.

\bibitem[{Zhang et~al.(2022)Zhang, Roller, Goyal, Artetxe, Chen, Chen, Dewan, Diab, Li, Lin et~al.}]{zhang2022opt}
Susan Zhang, Stephen Roller, Naman Goyal, Mikel Artetxe, Moya Chen, Shuohui Chen, Christopher Dewan, Mona Diab, Xian Li, Xi~Victoria Lin, and 1 others. 2022.
\newblock Opt: Open pre-trained transformer language models.
\newblock \emph{arXiv preprint arXiv:2205.01068}.

\bibitem[{Zhao et~al.(2021)Zhao, Wallace, Feng, Klein, and Singh}]{zhao2021calibrate}
Zihao Zhao, Eric Wallace, Shi Feng, Dan Klein, and Sameer Singh. 2021.
\newblock Calibrate before use: Improving few-shot performance of language models.
\newblock In \emph{International conference on machine learning}, pages 12697--12706. PMLR.

\bibitem[{Zhuo et~al.(2024)Zhuo, Zhang, Fang, Duan, Lin, and Chen}]{zhuo2024prosa}
Jingming Zhuo, Songyang Zhang, Xinyu Fang, Haodong Duan, Dahua Lin, and Kai Chen. 2024.
\newblock {ProSA}: Assessing and understanding the prompt sensitivity of llms.
\newblock In \emph{Findings of the Association for Computational Linguistics: EMNLP 2024}, pages 1950--1976.

\end{thebibliography}

\appendix

\clearpage
\section{Extended Related Work }
\label{sec:related_work_extended}
This section reviews two bodies of related work: studies on {\ft} and {\icl} as learning modes, and prior use of formal languages in the context of LLMs.
\subsection{Learning Modes in LLM: Fine-tuning and In-context Learning}

We discuss existing studies that independently investigate fine-tuning and in-context learning, followed by their direct comparison.

\textbf{Fine-tuning:} A number of works including \citet{kaplan2020scaling,zhang2024when,hu2024minicpm,srinivasan2024comparative,oliver2024crafting,hu2022lora} study the effects of fine-tuning or its variants with respect to model scaling, where larger fine-tuned models with less data outperform smaller models with the same amount of data, leading to compute-efficient training. Our experiments on synthetic formal languages do not demonstrate such a pattern, possibly because we allow all models of different sizes to reach their optimal fine-tuning performance on formal languages, where there is no tangible benefit of being large. 

\textbf{In-context learning:} Given a set of examples as demonstrations, {\icl} allows LLMs to extract patterns without updating model parameters. Several studies attempt to explain how learning is achieved in {\icl}, by comparing it to gradient descent~\cite{shen2023pretrained}, in-weights learning \cite{reddy2023mechanistic}, and Boolean function learning in a controlled setting~\cite{bhattamishra2020ability}. Recently,~\citet{pan-etal-2023-context,lin2024dual} explore the dual characteristics of {\icl}: (i) task learning, where the test examples are unseen during pre-training, and (ii) task recognition/retrieval, where test examples are seen during the pre-training, and LLMs are asked to retrieve them using a different prompt. 

A related question is how {\icl} performance scales with model size.~\citet{wei2023larger} study the relationship between {\icl} and model scale, where overriding semantic priors like flipping labels improves with larger models. In contrast,~\citet{chen-etal-2025-icleval} observe that {\icl} ability does not linearly correlate with model size. Our study finds that in the majority of model families, model size improves {\icl}, while in a few families, a medium-sized model performs better at {\icl}.

\textbf{Fine-tuning versus In-context learning.} Several works compare {\ft} and {\icl}, with inconclusive results.~\citet{mosbach2023few,liu2022few,bhatia2023tart,asai2024buffet} agree that {\ft} is better than {\icl}. However, these conclusions are not fully comparable: \citet{mosbach2023few} uses generative evaluation metrics that are not comparable across learning modes (violating desideratum \textbf{D3}), while others operate under unequal conditions (violating desideratum \textbf{D2}): using incomparable models or unequal numbers of examples~\cite{liu2022few,bhatia2023tart}, or observing high variance across different choices of examples~\cite{asai2024buffet}. 

Another group of works, including~\citet{yin2024deeper,bertsch2024context,kaneko2025gaps,soudani2024fine,awadalla2022exploring}, finds that {\icl} is better than {\ft}. Some employ suboptimal {\ft}: e.g.,~\citet{yin2024deeper} fine-tune for only $ 1 $ epoch, and~\citet{awadalla2022exploring} fine-tune for $ 2 $--$ 5 $ epochs. Others evaluate in settings where {\icl} benefits from retaining general language understanding while {\ft} suffers from parameter updates that cause forgetting, such as bias mitigation~\cite{kaneko2025gaps} and retrieval for low-frequency knowledge~\cite{soudani2024fine}, or in the many-shot regime~\cite{bertsch2024context}. These confounding factors reinforce the need for a neutral testbed -- such as formal language learning -- where the influence of prior training on {\ft} and {\icl} can be maximally controlled, satisfying desideratum \textbf{D2}.

Several works further compound this issue by comparing {\ft} and {\icl} across models of different sizes, occasionally using different datasets. For example, ~\citet{pecher2025comparing,lehman2023we} investigate whether smaller {\ft} models are better than larger general-purpose models adapted via {\icl}.~\citet{su2022selective} showcase the usefulness of selective annotation, but when comparing between {\ft} and {\icl}, they use different model sizes: smaller models for {\ft} vs larger models for {\icl}. ~\citet{gupta2023instruction} compare among {\icl}, instruction-tuning, and {\ft}, where {\icl} and instruction-tuning are conducted on the same model, but {\ft} is performed on a different larger model. Furthermore, the data used for instruction-tuning are not given to {\icl}. Finally,~\citet{le2021many} compare {\ft} and {\icl} on an identical masked language model -- this model is fundamentally different from autoregressive LLMs in terms of generating tokens, which is our focus. In all the papers, there is still a chance of data contamination, which we have carefully avoided using formal language learning.

\subsection{Formal Languages and LLMs}

To address the data contamination and experimental inconsistencies inherent in natural language testbeds, we ground our comparison in formal languages. Many prior works have studied formal languages in the context of LLMs, focusing primarily on the expressive power of language models and the learnability of formal grammars — neither of which directly addresses our goal of comparing {\ft} and {\icl}.

\textit{What is the relative representation capability of LLMs compared to other sequence models, or more specifically, what classes of languages are learnable by an LLM?} LLMs with a transformer architecture may have a different representation capability than other neural language models, such as LSTMs and RNNs.  We refer to a recent survey discussing the expressiveness of LLMs as a language recognizer~\cite{strobl2023transformers}. Towards comparing representation capability, \citet{shi2022learning} find that both LSTMs and transformer networks can simulate context-free languages with bounded recursion, suggesting similar representational power between the two. However, unlike transformers, LSTMs fail to decompose the latent representation space.~\citep{bhattamishra2020ability} observe a clear contrast between the performance of transformers and LSTMs on regular languages. They find that in comparison with LSTMs, transformers achieve limited performance on languages involving periodicity, modular counting, and even simpler star-free variants of Dyck-$1$ languages. \citet{deletang2022neural} explore how neural network models used for program induction relate to the idealized computational models defined by the Chomsky hierarchy~\cite{chomsky1956three}. They find that neural language models are hard to place on the standard Chomsky hierarchy. Several works criticize their setup, since they consider a language transduction task (mapping one language to another), which is  different from the language recognition task~\cite{icard2020calibrating}.~\citet{borenstein2024languages} consider learning strings from deterministic and probabilistic finite state automata. They empirically test the learnability as a function of various complexity parameters of the language and the hidden state size of the transformer and RNN. In a different line of work, ~\cite{akyurek2024context} evaluate neural networks' abilities to learn regular languages in {\icl}. Rather than learning one particular distribution from the training dataset, they infer the generating mechanism using {\icl}. Similar to~\citet{deletang2022neural}, they find that RNNs are better suited to modeling formal languages than transformers. \citet{kallini2024mission} construct a continuum of languages that differ in their hardness to learn and show that GPT-$2$ has difficulty in learning the carefully constructed impossible languages, compared to English.

While most of the works in this line capture the expressiveness of LLMs and their differing representation ability with other sequence models, we fundamentally criticize their evaluation metrics. As elaborated in Section~\ref{sec:language_proficiency}, they focus on testing how well an LLM learns the grammar rules or states of the automata, whereas our discriminative measure is rule/state agnostic and focuses on whether the LLMs can generate strings from the language better than strings outside the language -- LLMs may learn in a way different from specific rules/states, and it is non-trivial to measure this.

\textit{Can LLMs learn the underlying structure of formal languages, and if so, how?} Several studies utilize the controlled data generation of formal languages to study different NLP (natural language processing) aspects of the LLM. Formal languages, particularly those derived from context-free grammars, can imitate the rich recursive structure of natural languages. Therefore, many studies focus on teaching the LLM strings from a formal language and explaining how LLMs might learn them~\citep{allen2023physics,murty2022characterizing,liu2022transformers}. 
In another line,~\citet{jumelet2023transparency} study if causal and masked LLMs capture the true underlying patterns if trained on a true distribution. They find that causal LLMs approximate the theoretically optimal perplexity of the PCFG more closely than masked LLMs. Along this direction, several studies consider the known distribution to analyze the impact of topological features of a language ~\cite{cotterell2018all,mielke2019kind,ravfogel2019studying,mielke2019kind,papadimitriou2023injecting,white2021examining}. Several studies propose augmenting LLMs with additional components to enable them to learn certain classes of languages with ease. For example,~\citet{chi2023transformer} propose to add working memory, such as weight sharing, adaptive-Depth, and sliding-dilated attention to a GPT model to enable it to learn the parity function, which is hard for an LLM to learn~\cite{hahn2024sensitive}. In contrast to this line of work, our focus is to apply formal languages to study different modes of learning in LLMs: {\ft} and {\icl}, which, to the best of our knowledge, is novel.

\section{Extended Experimental Setup}
\label{app_sec:exp_setup}

All experiments are conducted in compute clusters with Python as the programming language (version $ 3.10 $), where we use $ 8 $x Nvidia H$ 100 $ $ 94 $GB NVL GPUs  and $ 2 $x AMD EPYC $ 9554 $ CPU @ $ 3.1 $ GHz, $ 2 $x$ 64 $ cores, and $ 24 $x$ 96 $GB RAM. {\ft} is performed with a batch size of $8$ and a linear learning rate scheduler with a warm-up ratio of $0.05$. We fix the learning rate to $5\times 10^{-5}$ for the Qwen, Gemma, and Llama-$ 3 $ families; $5\times 10^{-6}$ for the Mistral, Opt, and Llama-$ 2 $ families; and $10^{-5}$ for the Pythia family. During inference, we sidestep temperature sampling and instead record the loss of each target token given the preceding tokens in the string.

Below, we provide details of the formal languages used in our experiments, along with their formal definitions. Intuitively, we carefully design languages to show the robustness of our results by changing the grammar rules and token types.

\paragraph{Formal Languages and Grammars.}

\begin{figure*}[!t]
\begin{minipage}{0.5\textwidth}
\centering

\begin{align*}
	& \textcolor{red}{S}\;\textcolor{black}{\rightarrow}\;\textcolor{red}{A19}\;\textcolor{blue}{[1]}\\
	& \textcolor{red}{A19}\;\textcolor{black}{\rightarrow}\;\textcolor{red}{A18}\;\textcolor{red}{A16}\;\textcolor{blue}{[0.50]}\\
	& \textcolor{red}{A19}\;\textcolor{black}{\rightarrow}\;\textcolor{red}{A16}\;\textcolor{red}{A18}\;\textcolor{red}{A17}\;\textcolor{blue}{[0.50]}\\
	& \textcolor{red}{A18}\;\textcolor{black}{\rightarrow}\;\textcolor{red}{A15}\;\textcolor{red}{A14}\;\textcolor{red}{A13}\;\textcolor{blue}{[0.50]}\\
	& \textcolor{red}{A18}\;\textcolor{black}{\rightarrow}\;\textcolor{red}{A14}\;\textcolor{red}{A15}\;\textcolor{red}{A13}\;\textcolor{blue}{[0.50]}\\
	& \textcolor{red}{A17}\;\textcolor{black}{\rightarrow}\;\textcolor{red}{A14}\;\textcolor{red}{A13}\;\textcolor{red}{A15}\;\textcolor{blue}{[0.50]}\\
	& \textcolor{red}{A17}\;\textcolor{black}{\rightarrow}\;\textcolor{red}{A13}\;\textcolor{red}{A14}\;\textcolor{red}{A15}\;\textcolor{blue}{[0.50]}\\
	& \textcolor{red}{A16}\;\textcolor{black}{\rightarrow}\;\textcolor{red}{A14}\;\textcolor{red}{A15}\;\textcolor{blue}{[0.50]}\\
	& \textcolor{red}{A16}\;\textcolor{black}{\rightarrow}\;\textcolor{red}{A15}\;\textcolor{red}{A14}\;\textcolor{blue}{[0.50]}\\
	& \textcolor{red}{A15}\;\textcolor{black}{\rightarrow}\;\textcolor{red}{A11}\;\textcolor{red}{A12}\;\textcolor{red}{A10}\;\textcolor{blue}{[0.50]}\\
	& \textcolor{red}{A15}\;\textcolor{black}{\rightarrow}\;\textcolor{red}{A12}\;\textcolor{red}{A10}\;\textcolor{red}{A11}\;\textcolor{blue}{[0.50]}\\
	& \textcolor{red}{A14}\;\textcolor{black}{\rightarrow}\;\textcolor{red}{A11}\;\textcolor{red}{A10}\;\textcolor{red}{A12}\;\textcolor{blue}{[0.50]}\\
	& \textcolor{red}{A14}\;\textcolor{black}{\rightarrow}\;\textcolor{red}{A10}\;\textcolor{red}{A11}\;\textcolor{red}{A12}\;\textcolor{blue}{[0.50]}\\
	& \textcolor{red}{A13}\;\textcolor{black}{\rightarrow}\;\textcolor{red}{A10}\;\textcolor{red}{A12}\;\textcolor{red}{A11}\;\textcolor{blue}{[0.50]}\\
	& \textcolor{red}{A13}\;\textcolor{black}{\rightarrow}\;\textcolor{red}{A12}\;\textcolor{red}{A11}\;\textcolor{red}{A10}\;\textcolor{blue}{[0.50]}\\
	& \textcolor{red}{A12}\;\textcolor{black}{\rightarrow}\;\textcolor{teal}{9}\;\textcolor{teal}{8}\;\textcolor{teal}{7}\;\textcolor{blue}{[0.50]}\\
	& \textcolor{red}{A12}\;\textcolor{black}{\rightarrow}\;\textcolor{teal}{8}\;\textcolor{teal}{7}\;\textcolor{blue}{[0.50]}\\
	& \textcolor{red}{A11}\;\textcolor{black}{\rightarrow}\;\textcolor{teal}{6}\;\textcolor{teal}{5}\;\textcolor{blue}{[0.50]}\\
	& \textcolor{red}{A11}\;\textcolor{black}{\rightarrow}\;\textcolor{teal}{6}\;\textcolor{teal}{4}\;\textcolor{teal}{5}\;\textcolor{blue}{[0.50]}\\
	& \textcolor{red}{A10}\;\textcolor{black}{\rightarrow}\;\textcolor{teal}{3}\;\textcolor{teal}{1}\;\textcolor{blue}{[0.50]}\\
	& \textcolor{red}{A10}\;\textcolor{black}{\rightarrow}\;\textcolor{teal}{1}\;\textcolor{teal}{2}\;\textcolor{teal}{3}\;\textcolor{blue}{[0.50]}\\
\end{align*}

\end{minipage}%
\begin{minipage}{0.5\textwidth}
\centering

\begin{align*}
	& \textcolor{red}{S}\;\textcolor{black}{\rightarrow}\;\textcolor{red}{A19}\;\textcolor{blue}{[1]}\\
	& \textcolor{red}{A19}\;\textcolor{black}{\rightarrow}\;\textcolor{red}{A18}\;\textcolor{red}{A16}\;\textcolor{blue}{[0.50]}\\
	& \textcolor{red}{A19}\;\textcolor{black}{\rightarrow}\;\textcolor{red}{A16}\;\textcolor{red}{A18}\;\textcolor{red}{A17}\;\textcolor{blue}{[0.50]}\\
	& \textcolor{red}{A18}\;\textcolor{black}{\rightarrow}\;\textcolor{red}{A15}\;\textcolor{red}{A14}\;\textcolor{red}{A13}\;\textcolor{blue}{[0.50]}\\
	& \textcolor{red}{A18}\;\textcolor{black}{\rightarrow}\;\textcolor{red}{A14}\;\textcolor{red}{A15}\;\textcolor{red}{A13}\;\textcolor{blue}{[0.50]}\\
	& \textcolor{red}{A17}\;\textcolor{black}{\rightarrow}\;\textcolor{red}{A14}\;\textcolor{red}{A13}\;\textcolor{red}{A15}\;\textcolor{blue}{[0.50]}\\
	& \textcolor{red}{A17}\;\textcolor{black}{\rightarrow}\;\textcolor{red}{A13}\;\textcolor{red}{A14}\;\textcolor{red}{A15}\;\textcolor{blue}{[0.50]}\\
	& \textcolor{red}{A16}\;\textcolor{black}{\rightarrow}\;\textcolor{red}{A14}\;\textcolor{red}{A15}\;\textcolor{blue}{[0.50]}\\
	& \textcolor{red}{A16}\;\textcolor{black}{\rightarrow}\;\textcolor{red}{A15}\;\textcolor{red}{A14}\;\textcolor{blue}{[0.50]}\\
	& \textcolor{red}{A15}\;\textcolor{black}{\rightarrow}\;\textcolor{red}{A11}\;\textcolor{red}{A12}\;\textcolor{red}{A10}\;\textcolor{blue}{[0.50]}\\
	& \textcolor{red}{A15}\;\textcolor{black}{\rightarrow}\;\textcolor{red}{A12}\;\textcolor{red}{A10}\;\textcolor{red}{A11}\;\textcolor{blue}{[0.50]}\\
	& \textcolor{red}{A14}\;\textcolor{black}{\rightarrow}\;\textcolor{red}{A11}\;\textcolor{red}{A10}\;\textcolor{red}{A12}\;\textcolor{blue}{[0.50]}\\
	& \textcolor{red}{A14}\;\textcolor{black}{\rightarrow}\;\textcolor{red}{A10}\;\textcolor{red}{A11}\;\textcolor{red}{A12}\;\textcolor{blue}{[0.50]}\\
	& \textcolor{red}{A13}\;\textcolor{black}{\rightarrow}\;\textcolor{red}{A10}\;\textcolor{red}{A12}\;\textcolor{red}{A11}\;\textcolor{blue}{[0.50]}\\
	& \textcolor{red}{A13}\;\textcolor{black}{\rightarrow}\;\textcolor{red}{A12}\;\textcolor{red}{A11}\;\textcolor{red}{A10}\;\textcolor{blue}{[0.50]}\\
	& \textcolor{red}{A12}\;\textcolor{black}{\rightarrow}\;\textcolor{teal}{i}\;\textcolor{teal}{h}\;\textcolor{teal}{g}\;\textcolor{blue}{[0.50]}\\
	& \textcolor{red}{A12}\;\textcolor{black}{\rightarrow}\;\textcolor{teal}{h}\;\textcolor{teal}{g}\;\textcolor{blue}{[0.50]}\\
	& \textcolor{red}{A11}\;\textcolor{black}{\rightarrow}\;\textcolor{teal}{f}\;\textcolor{teal}{e}\;\textcolor{blue}{[0.50]}\\
	& \textcolor{red}{A11}\;\textcolor{black}{\rightarrow}\;\textcolor{teal}{f}\;\textcolor{teal}{d}\;\textcolor{teal}{e}\;\textcolor{blue}{[0.50]}\\
	& \textcolor{red}{A10}\;\textcolor{black}{\rightarrow}\;\textcolor{teal}{c}\;\textcolor{teal}{a}\;\textcolor{blue}{[0.50]}\\
	& \textcolor{red}{A10}\;\textcolor{black}{\rightarrow}\;\textcolor{teal}{a}\;\textcolor{teal}{b}\;\textcolor{teal}{c}\;\textcolor{blue}{[0.50]}\\
\end{align*}

\end{minipage}%

\caption{Production rules of ${G}_\alpha^{\text{Numerical}}$ (left) and ${G}_\alpha^{\text{Latin}}$ (right).}

\label{fig:grammar_g1_g2}
\end{figure*}

\begin{figure*}[!t]
\begin{minipage}{0.5\textwidth}
\centering

\begin{align*}
	& \textcolor{red}{S}\;\textcolor{black}{\rightarrow}\;\textcolor{red}{S5}\;\textcolor{blue}{[1]}\\
	& \textcolor{red}{S5}\;\textcolor{black}{\rightarrow}\;\textcolor{red}{B4}\;\textcolor{red}{C1_1}\;\textcolor{red}{E4}\;\textcolor{red}{T1_1}\;\textcolor{blue}{[0.25]}\\
	& \textcolor{red}{S5}\;\textcolor{black}{\rightarrow}\;\textcolor{red}{B4}\;\textcolor{red}{C1_2}\;\textcolor{red}{E4}\;\textcolor{red}{T1_2}\;\textcolor{blue}{[0.25]}\\
	& \textcolor{red}{S5}\;\textcolor{black}{\rightarrow}\;\textcolor{red}{B4}\;\textcolor{red}{C1_3}\;\textcolor{red}{E4}\;\textcolor{red}{T1_3}\;\textcolor{blue}{[0.25]}\\
	& \textcolor{red}{S5}\;\textcolor{black}{\rightarrow}\;\textcolor{red}{B4}\;\textcolor{red}{C1_4}\;\textcolor{red}{E4}\;\textcolor{red}{T1_4}\;\textcolor{blue}{[0.25]}\\
	& \textcolor{red}{B4}\;\textcolor{black}{\rightarrow}\;\textcolor{red}{B3}\;\textcolor{blue}{[0.3333]}\\
	& \textcolor{red}{B4}\;\textcolor{black}{\rightarrow}\;\textcolor{red}{B3}\;\textcolor{red}{B3}\;\textcolor{red}{B3}\;\textcolor{blue}{[0.3333]}\\
	& \textcolor{red}{B4}\;\textcolor{black}{\rightarrow}\;\textcolor{red}{B3}\;\textcolor{red}{B3}\;\textcolor{blue}{[0.3333]}\\
	& \textcolor{red}{B3}\;\textcolor{black}{\rightarrow}\;\textcolor{red}{B2}\;\textcolor{blue}{[0.3333]}\\
	& \textcolor{red}{B3}\;\textcolor{black}{\rightarrow}\;\textcolor{red}{B2}\;\textcolor{blue}{[0.3333]}\\
	& \textcolor{red}{B3}\;\textcolor{black}{\rightarrow}\;\textcolor{red}{B2}\;\textcolor{red}{B2}\;\textcolor{blue}{[0.3333]}\\
	& \textcolor{red}{B2}\;\textcolor{black}{\rightarrow}\;\textcolor{red}{B1}\;\textcolor{blue}{[0.3333]}\\
	& \textcolor{red}{B2}\;\textcolor{black}{\rightarrow}\;\textcolor{red}{B1}\;\textcolor{blue}{[0.3333]}\\
	& \textcolor{red}{B2}\;\textcolor{black}{\rightarrow}\;\textcolor{red}{B1}\;\textcolor{red}{B1}\;\textcolor{red}{B1}\;\textcolor{blue}{[0.3333]}\\
	& \textcolor{red}{B1}\;\textcolor{black}{\rightarrow}\;\textcolor{teal}{2}\;\textcolor{teal}{9}\;\textcolor{teal}{3}\;\textcolor{blue}{[0.3333]}\\
	& \textcolor{red}{B1}\;\textcolor{black}{\rightarrow}\;\textcolor{teal}{9}\;\textcolor{teal}{6}\;\textcolor{teal}{1}\;\textcolor{blue}{[0.3333]}\\
	& \textcolor{red}{B1}\;\textcolor{black}{\rightarrow}\;\textcolor{teal}{1}\;\textcolor{teal}{8}\;\textcolor{teal}{6}\;\textcolor{teal}{2}\;\textcolor{blue}{[0.3333]}\\
	& \textcolor{red}{E4}\;\textcolor{black}{\rightarrow}\;\textcolor{red}{E3}\;\textcolor{blue}{[0.3333]}\\
	& \textcolor{red}{E4}\;\textcolor{black}{\rightarrow}\;\textcolor{red}{E3}\;\textcolor{red}{E3}\;\textcolor{blue}{[0.3333]}\\
	& \textcolor{red}{E4}\;\textcolor{black}{\rightarrow}\;\textcolor{red}{E3}\;\textcolor{red}{E3}\;\textcolor{red}{E3}\;\textcolor{blue}{[0.3333]}\\
	& \textcolor{red}{E3}\;\textcolor{black}{\rightarrow}\;\textcolor{red}{E2}\;\textcolor{blue}{[0.3333]}\\
	& \textcolor{red}{E3}\;\textcolor{black}{\rightarrow}\;\textcolor{red}{E2}\;\textcolor{red}{E2}\;\textcolor{blue}{[0.3333]}\\
	& \textcolor{red}{E3}\;\textcolor{black}{\rightarrow}\;\textcolor{red}{E2}\;\textcolor{blue}{[0.3333]}\\
	& \textcolor{red}{E2}\;\textcolor{black}{\rightarrow}\;\textcolor{red}{E1}\;\textcolor{red}{E1}\;\textcolor{blue}{[0.3333]}\\
	& \textcolor{red}{E2}\;\textcolor{black}{\rightarrow}\;\textcolor{red}{E1}\;\textcolor{blue}{[0.3333]}\\
	& \textcolor{red}{E2}\;\textcolor{black}{\rightarrow}\;\textcolor{red}{E1}\;\textcolor{red}{E1}\;\textcolor{red}{E1}\;\textcolor{blue}{[0.3333]}\\
	& \textcolor{red}{E1}\;\textcolor{black}{\rightarrow}\;\textcolor{teal}{5}\;\textcolor{teal}{6}\;\textcolor{blue}{[0.3333]}\\
	& \textcolor{red}{E1}\;\textcolor{black}{\rightarrow}\;\textcolor{teal}{1}\;\textcolor{teal}{8}\;\textcolor{teal}{6}\;\textcolor{teal}{6}\;\textcolor{blue}{[0.3333]}\\
	& \textcolor{red}{E1}\;\textcolor{black}{\rightarrow}\;\textcolor{teal}{1}\;\textcolor{teal}{5}\;\textcolor{teal}{1}\;\textcolor{teal}{5}\;\textcolor{teal}{5}\;\textcolor{teal}{9}\;\textcolor{blue}{[0.3333]}\\
	& \textcolor{red}{T1_1}\;\textcolor{black}{\rightarrow}\;\textcolor{teal}{1}\;\textcolor{blue}{[1]}\\
	& \textcolor{red}{T1_2}\;\textcolor{black}{\rightarrow}\;\textcolor{teal}{2}\;\textcolor{blue}{[1]}\\
	& \textcolor{red}{T1_3}\;\textcolor{black}{\rightarrow}\;\textcolor{teal}{3}\;\textcolor{blue}{[1]}\\
	& \textcolor{red}{T1_4}\;\textcolor{black}{\rightarrow}\;\textcolor{teal}{4}\;\textcolor{blue}{[1]}\\
	& \textcolor{red}{C1_1}\;\textcolor{black}{\rightarrow}\;\textcolor{teal}{5}\;\textcolor{blue}{[1]}\\
	& \textcolor{red}{C1_2}\;\textcolor{black}{\rightarrow}\;\textcolor{teal}{6}\;\textcolor{blue}{[1]}\\
	& \textcolor{red}{C1_3}\;\textcolor{black}{\rightarrow}\;\textcolor{teal}{7}\;\textcolor{blue}{[1]}\\
	& \textcolor{red}{C1_4}\;\textcolor{black}{\rightarrow}\;\textcolor{teal}{8}\;\textcolor{blue}{[1]}\\
	& \textcolor{red}{C1_5}\;\textcolor{black}{\rightarrow}\;\textcolor{teal}{9}\;\textcolor{blue}{[1]}\\
\end{align*}

\end{minipage}%
\begin{minipage}{0.5\textwidth}
\centering

\begin{align*}
	& \textcolor{red}{S}\;\textcolor{black}{\rightarrow}\;\textcolor{red}{S5}\;\textcolor{blue}{[1]}\\
	& \textcolor{red}{S5}\;\textcolor{black}{\rightarrow}\;\textcolor{red}{B4}\;\textcolor{red}{C1_1}\;\textcolor{red}{E4}\;\textcolor{red}{T1_1}\;\textcolor{blue}{[0.25]}\\
	& \textcolor{red}{S5}\;\textcolor{black}{\rightarrow}\;\textcolor{red}{B4}\;\textcolor{red}{C1_2}\;\textcolor{red}{E4}\;\textcolor{red}{T1_2}\;\textcolor{blue}{[0.25]}\\
	& \textcolor{red}{S5}\;\textcolor{black}{\rightarrow}\;\textcolor{red}{B4}\;\textcolor{red}{C1_3}\;\textcolor{red}{E4}\;\textcolor{red}{T1_3}\;\textcolor{blue}{[0.25]}\\
	& \textcolor{red}{S5}\;\textcolor{black}{\rightarrow}\;\textcolor{red}{B4}\;\textcolor{red}{C1_4}\;\textcolor{red}{E4}\;\textcolor{red}{T1_4}\;\textcolor{blue}{[0.25]}\\
	& \textcolor{red}{B4}\;\textcolor{black}{\rightarrow}\;\textcolor{red}{B3}\;\textcolor{blue}{[0.3333]}\\
	& \textcolor{red}{B4}\;\textcolor{black}{\rightarrow}\;\textcolor{red}{B3}\;\textcolor{red}{B3}\;\textcolor{red}{B3}\;\textcolor{blue}{[0.3333]}\\
	& \textcolor{red}{B4}\;\textcolor{black}{\rightarrow}\;\textcolor{red}{B3}\;\textcolor{red}{B3}\;\textcolor{blue}{[0.3333]}\\
	& \textcolor{red}{B3}\;\textcolor{black}{\rightarrow}\;\textcolor{red}{B2}\;\textcolor{blue}{[0.3333]}\\
	& \textcolor{red}{B3}\;\textcolor{black}{\rightarrow}\;\textcolor{red}{B2}\;\textcolor{blue}{[0.3333]}\\
	& \textcolor{red}{B3}\;\textcolor{black}{\rightarrow}\;\textcolor{red}{B2}\;\textcolor{red}{B2}\;\textcolor{blue}{[0.3333]}\\
	& \textcolor{red}{B2}\;\textcolor{black}{\rightarrow}\;\textcolor{red}{B1}\;\textcolor{blue}{[0.3333]}\\
	& \textcolor{red}{B2}\;\textcolor{black}{\rightarrow}\;\textcolor{red}{B1}\;\textcolor{blue}{[0.3333]}\\
	& \textcolor{red}{B2}\;\textcolor{black}{\rightarrow}\;\textcolor{red}{B1}\;\textcolor{red}{B1}\;\textcolor{red}{B1}\;\textcolor{blue}{[0.3333]}\\
	& \textcolor{red}{B1}\;\textcolor{black}{\rightarrow}\;\textcolor{teal}{b}\;\textcolor{teal}{i}\;\textcolor{teal}{c}\;\textcolor{blue}{[0.3333]}\\
	& \textcolor{red}{B1}\;\textcolor{black}{\rightarrow}\;\textcolor{teal}{i}\;\textcolor{teal}{f}\;\textcolor{teal}{a}\;\textcolor{blue}{[0.3333]}\\
	& \textcolor{red}{B1}\;\textcolor{black}{\rightarrow}\;\textcolor{teal}{a}\;\textcolor{teal}{h}\;\textcolor{teal}{f}\;\textcolor{teal}{b}\;\textcolor{blue}{[0.3333]}\\
	& \textcolor{red}{E4}\;\textcolor{black}{\rightarrow}\;\textcolor{red}{E3}\;\textcolor{blue}{[0.3333]}\\
	& \textcolor{red}{E4}\;\textcolor{black}{\rightarrow}\;\textcolor{red}{E3}\;\textcolor{red}{E3}\;\textcolor{blue}{[0.3333]}\\
	& \textcolor{red}{E4}\;\textcolor{black}{\rightarrow}\;\textcolor{red}{E3}\;\textcolor{red}{E3}\;\textcolor{red}{E3}\;\textcolor{blue}{[0.3333]}\\
	& \textcolor{red}{E3}\;\textcolor{black}{\rightarrow}\;\textcolor{red}{E2}\;\textcolor{blue}{[0.3333]}\\
	& \textcolor{red}{E3}\;\textcolor{black}{\rightarrow}\;\textcolor{red}{E2}\;\textcolor{red}{E2}\;\textcolor{blue}{[0.3333]}\\
	& \textcolor{red}{E3}\;\textcolor{black}{\rightarrow}\;\textcolor{red}{E2}\;\textcolor{blue}{[0.3333]}\\
	& \textcolor{red}{E2}\;\textcolor{black}{\rightarrow}\;\textcolor{red}{E1}\;\textcolor{red}{E1}\;\textcolor{blue}{[0.3333]}\\
	& \textcolor{red}{E2}\;\textcolor{black}{\rightarrow}\;\textcolor{red}{E1}\;\textcolor{blue}{[0.3333]}\\
	& \textcolor{red}{E2}\;\textcolor{black}{\rightarrow}\;\textcolor{red}{E1}\;\textcolor{red}{E1}\;\textcolor{red}{E1}\;\textcolor{blue}{[0.3333]}\\
	& \textcolor{red}{E1}\;\textcolor{black}{\rightarrow}\;\textcolor{teal}{e}\;\textcolor{teal}{f}\;\textcolor{blue}{[0.3333]}\\
	& \textcolor{red}{E1}\;\textcolor{black}{\rightarrow}\;\textcolor{teal}{a}\;\textcolor{teal}{h}\;\textcolor{teal}{f}\;\textcolor{teal}{f}\;\textcolor{blue}{[0.3333]}\\
	& \textcolor{red}{E1}\;\textcolor{black}{\rightarrow}\;\textcolor{teal}{a}\;\textcolor{teal}{e}\;\textcolor{teal}{a}\;\textcolor{teal}{e}\;\textcolor{teal}{e}\;\textcolor{teal}{i}\;\textcolor{blue}{[0.3333]}\\
	& \textcolor{red}{T1_1}\;\textcolor{black}{\rightarrow}\;\textcolor{teal}{a}\;\textcolor{blue}{[1]}\\
	& \textcolor{red}{T1_2}\;\textcolor{black}{\rightarrow}\;\textcolor{teal}{b}\;\textcolor{blue}{[1]}\\
	& \textcolor{red}{T1_3}\;\textcolor{black}{\rightarrow}\;\textcolor{teal}{c}\;\textcolor{blue}{[1]}\\
	& \textcolor{red}{T1_4}\;\textcolor{black}{\rightarrow}\;\textcolor{teal}{d}\;\textcolor{blue}{[1]}\\
	& \textcolor{red}{C1_1}\;\textcolor{black}{\rightarrow}\;\textcolor{teal}{e}\;\textcolor{blue}{[1]}\\
	& \textcolor{red}{C1_2}\;\textcolor{black}{\rightarrow}\;\textcolor{teal}{f}\;\textcolor{blue}{[1]}\\
	& \textcolor{red}{C1_3}\;\textcolor{black}{\rightarrow}\;\textcolor{teal}{g}\;\textcolor{blue}{[1]}\\
	& \textcolor{red}{C1_4}\;\textcolor{black}{\rightarrow}\;\textcolor{teal}{h}\;\textcolor{blue}{[1]}\\
	& \textcolor{red}{C1_5}\;\textcolor{black}{\rightarrow}\;\textcolor{teal}{i}\;\textcolor{blue}{[1]}\\
\end{align*}

\end{minipage}%

\caption{Production rules of ${G}_{\beta}^{\text{Numerical}}$ (left) and ${G}_{\beta}^{\text{Latin}}$ (right).}

\label{fig:grammar_g4_g5}
\end{figure*}

Throughout our experiments, we provide the LLM with strings sampled from a probabilistic formal language. Formally, a probabilistic formal language is represented by a \emph{probabilistic formal grammar}, or simply \emph{grammars}~\cite{collins2013probabilistic}.
A grammar consists of two sets of symbols called the \emph{non-terminals} and \emph{terminals}, a set of production rules for rewriting strings that contain at least one nonterminal, and a probability distribution over the production rules. More precisely, a probabilistic formal grammar is defined as a quintuple.  
\begin{align*}
    G \triangleq (\textcolor{red}{\mathbf{N}}, \textcolor{teal}{\mathbf{T}}, \textcolor{black}{\mathbf{R}}, \textcolor{red}{S}, \textcolor{blue}{\mathbf{P}})
\end{align*}

where $\textcolor{red}{\mathbf{N}}$ is the set of non-terminals, $\textcolor{teal}{\mathbf{T}}$ is the set of terminals (equivalently, tokens), $\textcolor{black}{\mathbf{R}}$ is the set of production rules, $\textcolor{red}{S} \in \textcolor{red}{\mathbf{N}} $ is the start non-terminal, and $\textcolor{blue}{\mathbf{P}}$ is the set of probabilities on production rules.

Formal languages are divided into well-known classes based on the \emph{complexity} of the language membership problem, i.e., the \emph{complexity} of the grammars needed to generate them~\cite{chomsky1956three}. In this paper, we use one class of grammars, namely, hierarchical probabilistic context-free grammars (HPCFGs)~\cite{allen2023physics}. Specifically, our experiments are based on teaching LLMs languages represented by HPCFGs, which are syntactically simple and can represent languages that are structurally similar to natural languages~\cite{allen2023physics,shi2022learning}.

\paragraph{Description of Grammars and Identified Languages.} In our experiments, we consider two generic structures for the considered grammars, one adapted from~\citet{allen2023physics}, namely $ G_\alpha $, and another proposed by us, namely $ G_\beta $. We propose variants of these grammars by considering different alphabet sets.

In Figure~\ref{fig:grammar_g1_g2}, in the first generic structure $ G_\alpha $, each grammar has $ \mathbf{N} = \{S, A10, A11, \dots, A19\}$ and $\mathbf{T} = \{1, 2, 3, \dots, 9\}$. The grammar has four levels of hierarchy: the non-terminals from top to bottom levels are $\{A19\}$, $\{A16, A17, A18\}$, $\{A13, A14, A15\}$, and $\{A10, A11, A12\}$, followed by terminals $\{1, 2, 3, \dots, 9\}$. Since the terminals are derived from numerical characters, we call this grammar $ G_{\alpha}^{\text{Numerical}} $; and if the terminals are derived from Latin characters, we call this grammar $ G_{\alpha}^{\text{Latin}} $. Each non-terminal (except the start non-terminal) has two expansion rules, consisting of non-terminals from the immediate lower level. Further, the expansion rules are probabilistic, where the sum of probabilities of all expansion rules from a given non-terminal is $1$.

In Figure~\ref{fig:grammar_g4_g5}, the second generic structure $ G_\beta $ is inspired by bridging two HPCFGs together, and simulating long-range dependencies within the generated strings. Specifically, the sub-grammar rooted at $ B4 $ and the sub-grammar rooted at $ E4 $ are connected by non-terminal $C1_i$; and $E4$ ends with $T1_j$. Long-range dependencies are communicated through $C1_i$ and $T1_j$, by enforcing $ i = j $ at each expansion of $ S5 $.

\begin{table*}
	\centering
	\caption{Notations of grammars and identified languages.}
	\label{tab:grammar_notations}
	\begin{tabular}{ll}
		\toprule
		Grammar & Identified Language\\
		\midrule
		$ G_{\alpha}^{\text{Numerical}} $ & $ \lang_1 $\\
		$ G_{\alpha}^{\text{Latin}} $ & $ \lang_2 $\\
		$ G_{\alpha}^{\text{Under-trained-tokens}} $ & $ \lang_3 $\\
		$ G_{\beta}^{\text{Numerical}} $ & $ \lang_4 $\\
		$ G_{\beta}^{\text{Latin}} $ & $ \lang_5 $\\
		$ G_{\beta}^{\text{Under-trained-tokens}} $ & $ \lang_6 $\\
		\bottomrule
	\end{tabular}
\end{table*}

\begin{figure*}
    \centering

    \subfloat[$\lang_1$ (also $\lang_2$, $ \lang_3 $)]{
    \includegraphics[scale=0.6]{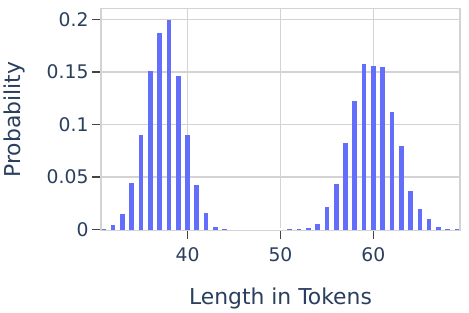}
    }
    \subfloat[$\lang_4$ (also $\lang_5$, $ \lang_6 $)]{
    \includegraphics[scale=0.6]{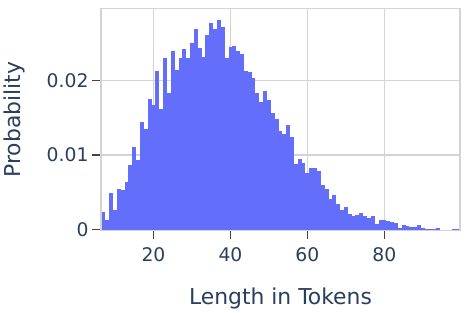}
    }

    \caption{Length distribution of considered probabilistic languages, based on $10000$ sampled strings per language.}
    \label{fig:length_distribution}
\end{figure*}

\begin{figure*}
    \centering

    \subfloat[Language $ \lang_1 $  (Grammar $ G_{\alpha}^{\text{Numerical}}) $]{
    \includegraphics[trim={0 0 0 2.5cm},clip,scale=0.6]{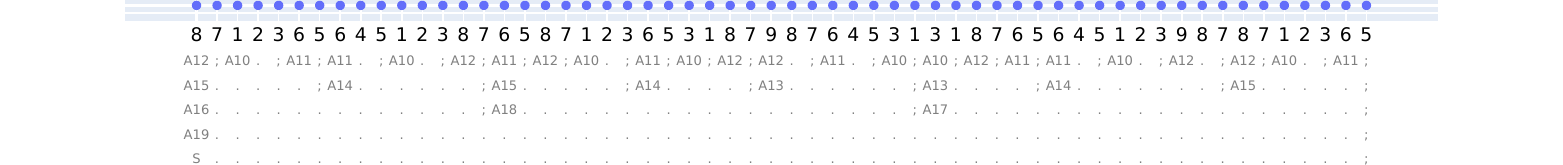}
    }

    \subfloat[Language $ \lang_2 $  (Grammar $ G_{\alpha}^{\text{Latin}}) $]{
    \includegraphics[trim={0 0 0 2.5cm},clip,scale=0.6]{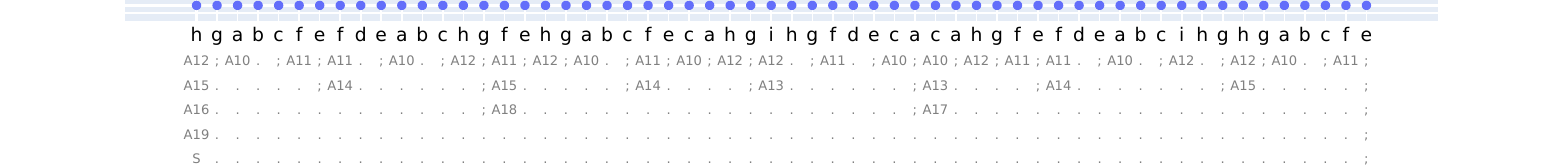}
    }

	\subfloat[Language $ \lang_4 $  (Grammar $ G_{\beta}^{\text{Numerical}}) $]{
    \includegraphics[trim={0 0 0 2.5cm},clip,scale=0.6]{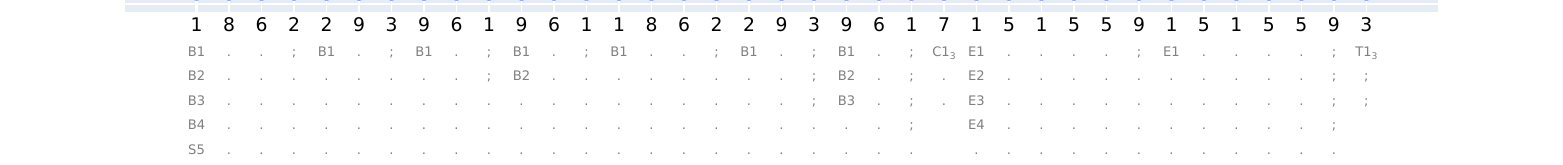}
    }

	\subfloat[Language $ \lang_5 $  (Grammar $ G_{\beta}^{\text{Latin}}) $]{
	\includegraphics[trim={0 0 0 2.5cm},clip,scale=0.6]{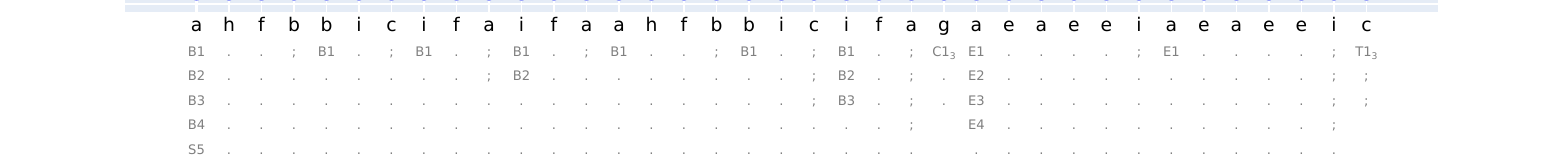}
	}

    \caption{Representative strings from different languages, annotated with non-terminals applied in different positions by the respective hierarchical grammar.}

    \label{fig:representaive_string}
    
\end{figure*}

Table~\ref{tab:grammar_notations} shows the mapping of notations between grammars and identified languages. Figure~\ref{fig:length_distribution} shows the length distribution of generated strings from different languages. Figure~\ref{fig:representaive_string} demonstrates how hierarchical non-terminals are applied in different positions in the representative strings.

{
\color{black}

\paragraph{Sampling Strings from a Formal Language.}
Given a language $\lang$ generated by an HPCFG, our objective is to sample a set of i.i.d.\  strings from the language.
To \emph{sample a string from the language},
we start from a special string in the grammar containing a single, distinguished nonterminal called the ``start'' or ``root'' symbol, and apply the production rules to rewrite the string repeatedly. 
If several rules can be used to rewrite the string at any stage, we sample one such rule from the probability distribution over the rules and apply it. 
We stop when we obtain a string containing terminals only. This string is a sample drawn from the language.
We can repeat this process to draw any number of i.i.d.\ samples from the language.
 
}

In our experiments, we aim to split the sampled strings into disjoint training and test sets that have a similar probability distribution over string occurrences. To realize this goal, we first sample a finite number of strings from the language. We then perform a stratified split: we iterate over unique strings in random order, alternately assigning all occurrences of each string to the training or test set. This process repeats until the initial finite set is exhausted.

\clearpage
\paragraph{Distance Between Languages.} In probabilistic languages, a well-known approach to compute language distance is to compare the distribution of strings generated by both languages~\cite{de2014computation}. In our implementation, we choose a simplified distance metric based on $L_2$-norm.

\begin{equation}
\mathtt{dist}_{L_2}(\lang_1, \lang_2) =   \sqrt{\Sigma_{s\in \mathbf{T}^*}(P_{\lang_1}(s) - P_{\lang_2}(s))^2}
\label{eq:language_distance}
\end{equation}

While distance metrics have their nuances, our goal is to systematically modify the original language, more specifically the underlying grammar, such that we can intuitively interpret language distance, irrespective of the distance metric used.

For simulating out-of-distribution generalization of learning modes, we modify the base grammar $ G_{\alpha}^{\text{Numerical}} $ (abbreviated as $ G $) in the following way: We construct five grammars $\{\grammar^{(1)},\dots, \grammar^{(5)}\}$ by perturbing $\ell \in \{1,2,3,4,5\}$ production rules of $\grammar$, such that $\grammar^{(\ell)}$ contains all perturbed production rules in  $\grammar^{(\ell-1)}$. The order in which rule-perturbation is applied is the following:

\begin{align*}
    & \textcolor{red}{A10^{(1)}}\;\textcolor{black}{\rightarrow}\;\textcolor{teal}{1}\;\textcolor{teal}{3}\;\textcolor{teal}{2}\;\textcolor{blue}{[0.50]}\\
	& \textcolor{red}{A11^{(2)}}\;\textcolor{black}{\rightarrow}\;\textcolor{teal}{5}\;\textcolor{teal}{6}\;\textcolor{blue}{[0.50]}\\
 & \textcolor{red}{A12^{(3)}}\;\textcolor{black}{\rightarrow}\;\textcolor{teal}{8}\;\textcolor{teal}{7}\;\textcolor{teal}{9}\;\textcolor{blue}{[0.50]}\\
 & \textcolor{red}{A10^{(4)}}\;\textcolor{black}{\rightarrow}\;\textcolor{teal}{3}\;\textcolor{teal}{1}\;\textcolor{blue}{[0.50]}\\
	& \textcolor{red}{A12^{(5)}}\;\textcolor{black}{\rightarrow}\;\textcolor{teal}{8}\;\textcolor{teal}{7}\;\textcolor{blue}{[0.50]}\\
\end{align*}

Intuitively, $\grammar^{(1)}$ contains perturbed rule $\{A10^{(1)}\}$, $\grammar^{(2)}$ contains perturbed rule $\{A10^{(1)}, A11^{(2)}\}$,  and so on. Finally, each grammar $\grammar^{(\ell)}$ identifies a language $\lang^{(\ell)}$ in Figure~\ref{fig:ood_generalization}.

\clearpage
\section{Additional Experimental Results}
\label{app_sec:additional_results}

In the following, we outline additional experimental results.

\begin{figure*}[!t]
\centering
        \subfloat[Language $\lang_1$]{
	\includegraphics[scale=0.5]{figures/fine-tuning/fine-tuning_exp_all_models_pcfg_cfg3b_disjoint_terminals_non_grammatical_test_sequences_edit_distance_1_auc_edit_distance.pdf}
	}
	\subfloat[Language $\lang_2$]{
	\includegraphics[scale=0.5]{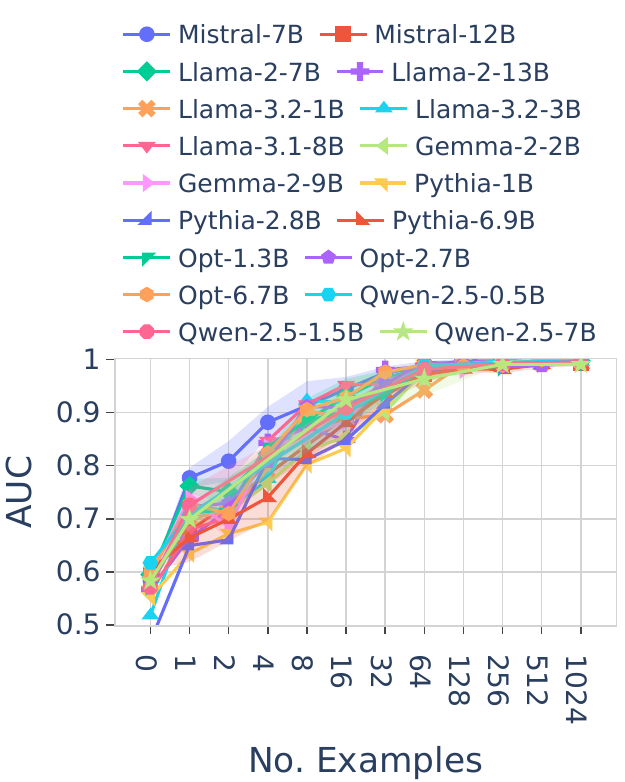}
	}

	\subfloat[Language $\lang_4$]{
	\includegraphics[scale=0.5]{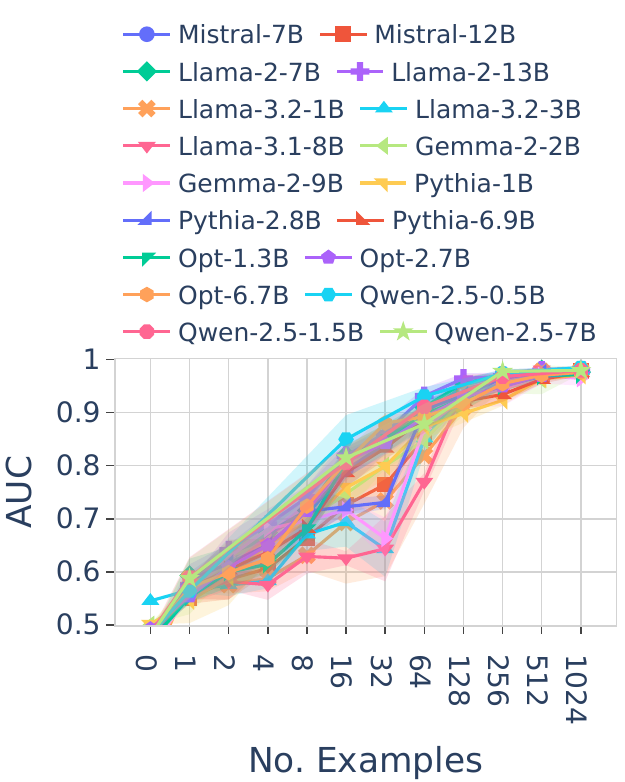}
	}
	\subfloat[Language $\lang_5$]{
	\includegraphics[scale=0.5]{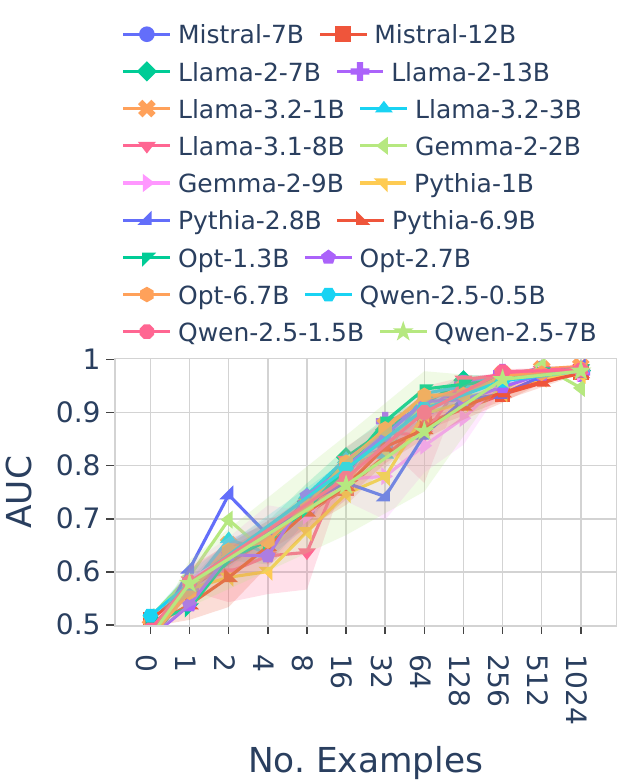}
	}

	\caption{Optimal fine-tuning performance in all models across different languages.}

    \label{fig:optimal_fine_tuning_all_grammars}
\end{figure*}

\begin{figure*}[!t]
	\centering
	\subfloat[Language $\lang_1$]{
	\includegraphics[scale=0.5]{figures/incontext_experiments/incontext_exp_all_models_pcfg_cfg3b_disjoint_terminals_non_grammatical_test_sequences_edit_distance_1_auc_edit_distance.pdf}
	}
	\subfloat[Language $\lang_2$]{
	\includegraphics[scale=0.5]{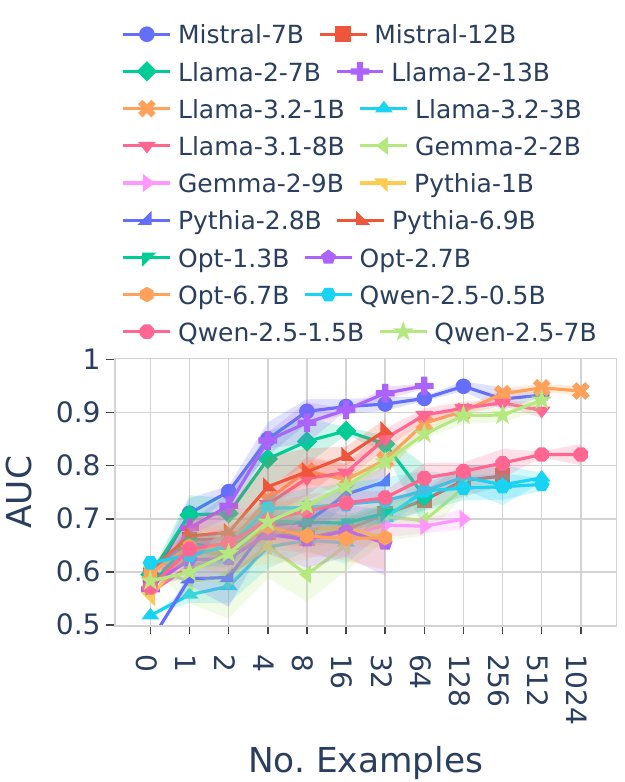}
	}

	\subfloat[Language $\lang_4$]{
	\includegraphics[scale=0.5]{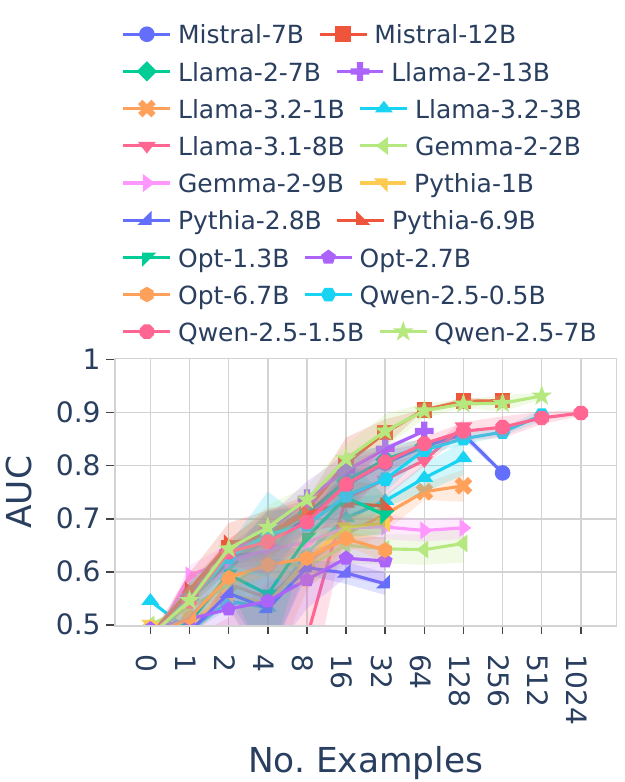}
	}
	\subfloat[Language $\lang_5$]{
	\includegraphics[scale=0.5]{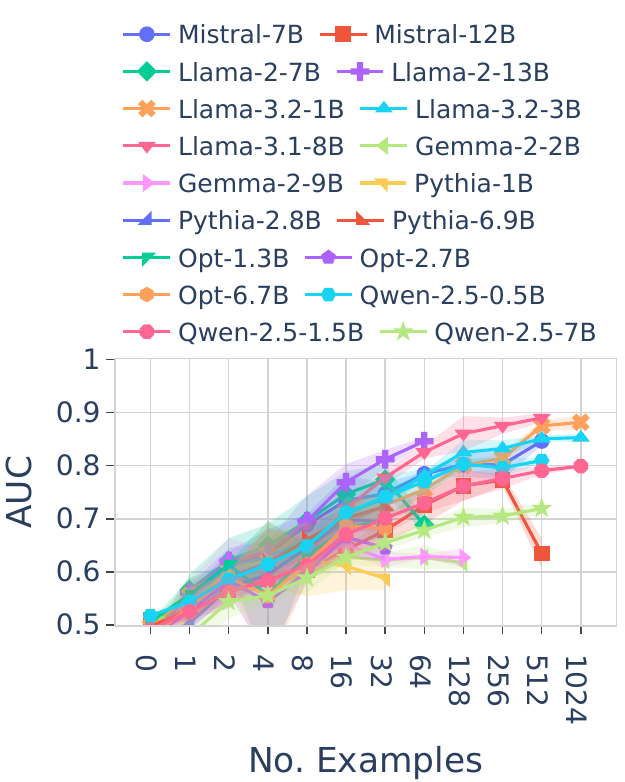}
	}

	\caption{In-context learning performance of  all models across different languages}

    \label{fig:in_context_learning_all_grammars}
\end{figure*}

\begin{figure*}
	\centering
        \captionsetup[subfigure]{labelformat=empty}

        \subfloat[Qwen, Language $\lang_1$]{
        \includegraphics[scale=0.35]{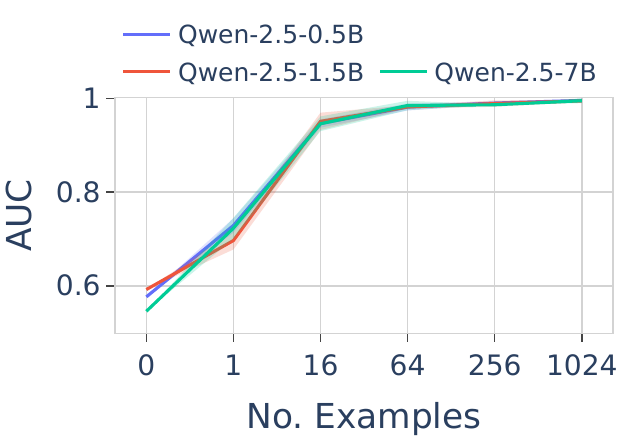}
        }
        \subfloat[Qwen, Language $\lang_2$]{
        \includegraphics[scale=0.35]{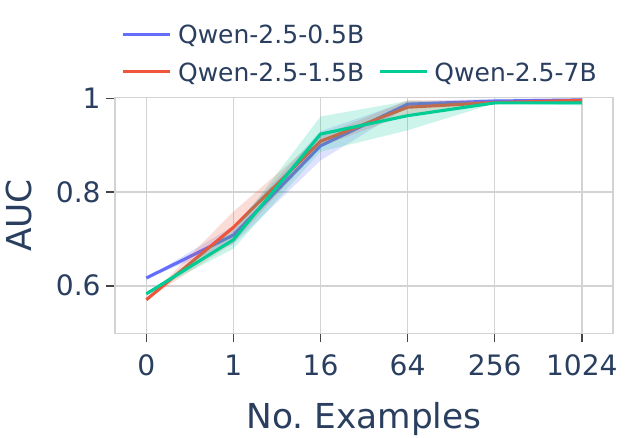}
        }
        \subfloat[Qwen, Language $\lang_4$]{
        \includegraphics[scale=0.35]{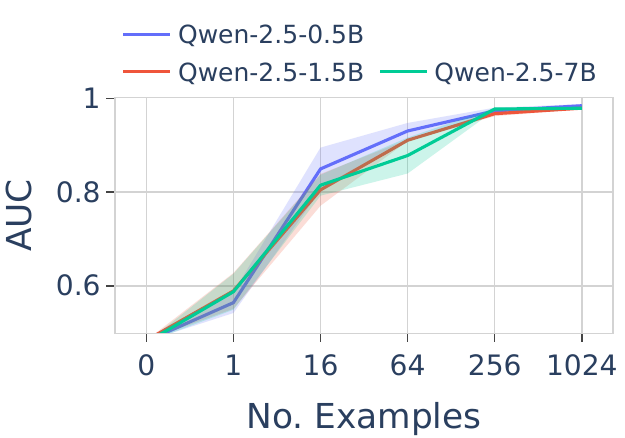}
        }
        \subfloat[Qwen, Language $\lang_5$]{
        \includegraphics[scale=0.35]{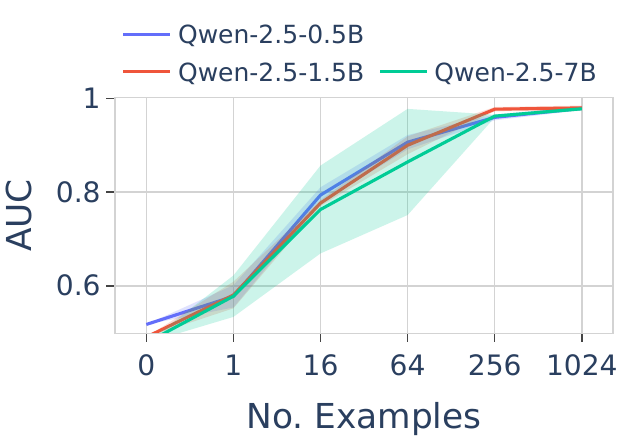}
        }

        \subfloat[Mistral, Language $\lang_1$]{
        \includegraphics[scale=0.35]{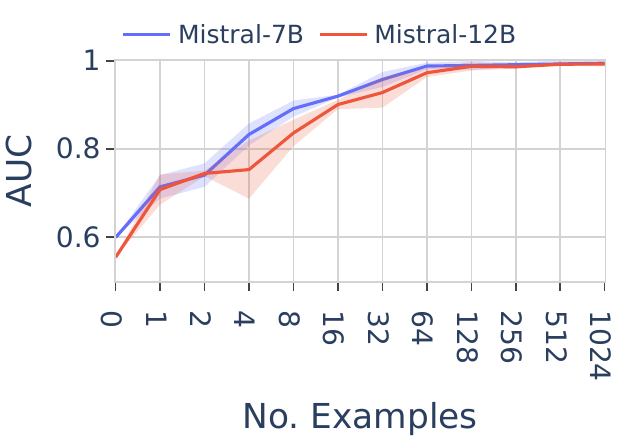}
        }
        \subfloat[Mistral, Language $\lang_2$]{
        \includegraphics[scale=0.35]{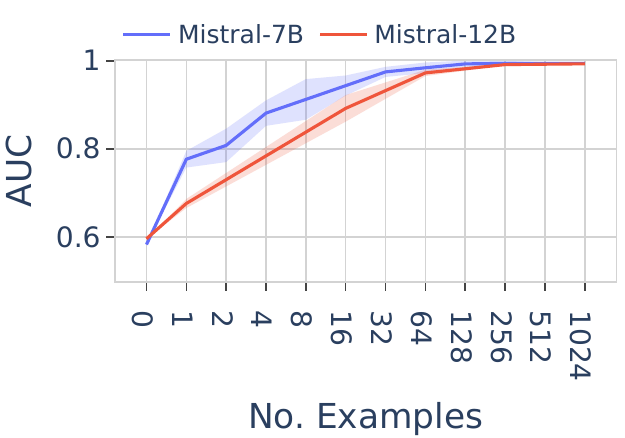}
        }
        \subfloat[Mistral, Language $\lang_4$]{
        \includegraphics[scale=0.35]{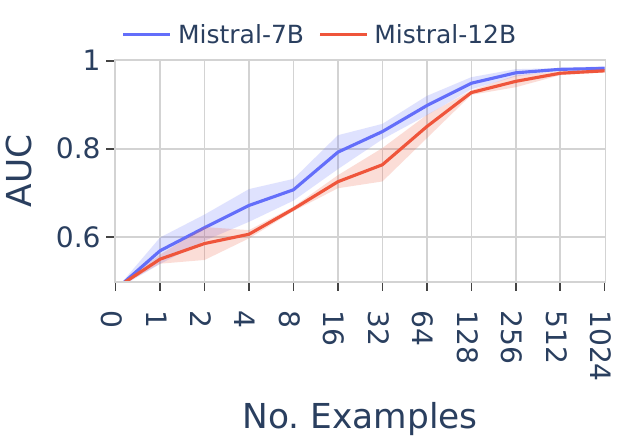}
        }
        \subfloat[Mistral, Language $\lang_5$]{
        \includegraphics[scale=0.35]{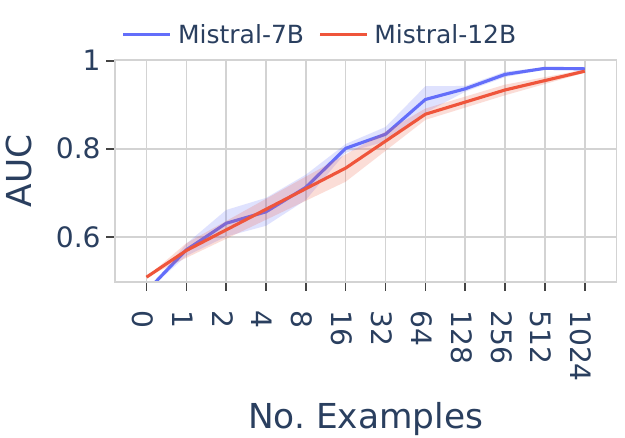}
        }

        \subfloat[Llama-$2$, Language $\lang_1$]{
        \includegraphics[scale=0.35]{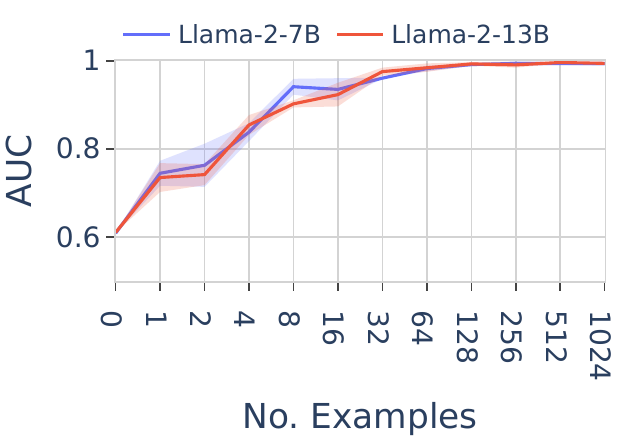}
        }
        \subfloat[Llama-$2$, Language $\lang_2$]{
        \includegraphics[scale=0.35]{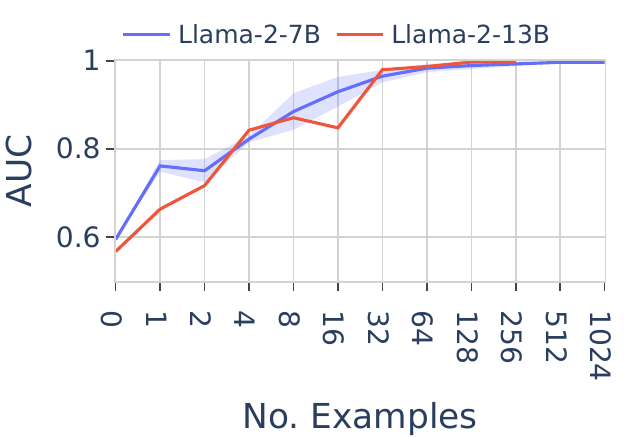}
        }
        \subfloat[Llama-$2$, Language $\lang_4$]{
        \includegraphics[scale=0.35]{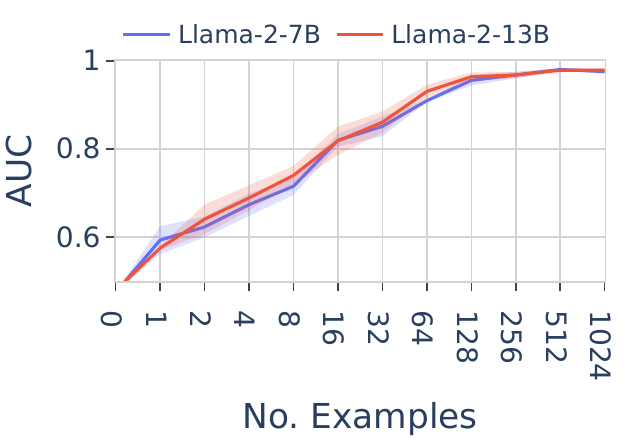}
        }
        \subfloat[Llama-$2$, Language $\lang_5$]{
        \includegraphics[scale=0.35]{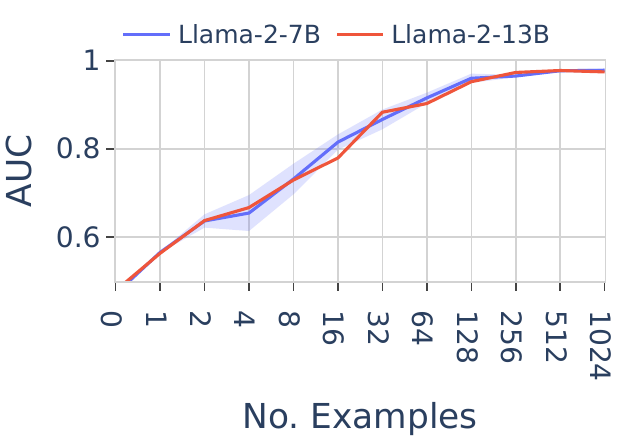}
        }

        \subfloat[Llama-$3$, Language $\lang_1$]{
        \includegraphics[scale=0.35]{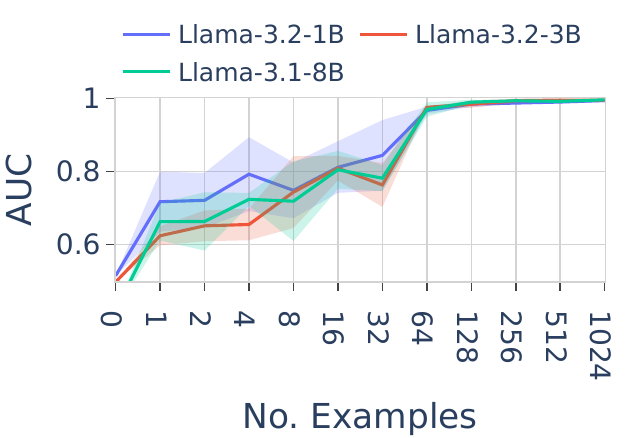}
        }
        \subfloat[Llama-$3$, Language $\lang_2$]{
        \includegraphics[scale=0.35]{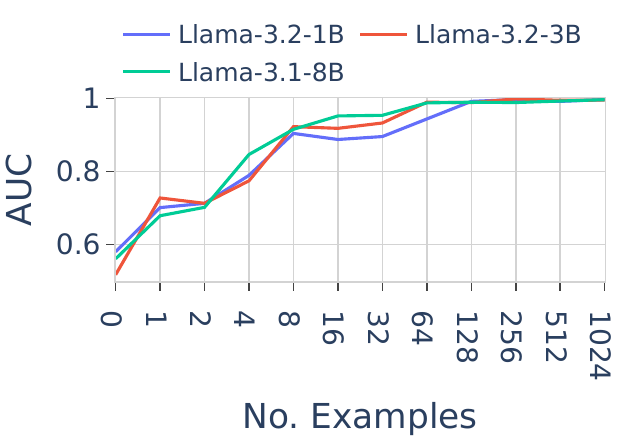}
        }
        \subfloat[Llama-$3$, Language $\lang_4$]{
        \includegraphics[scale=0.35]{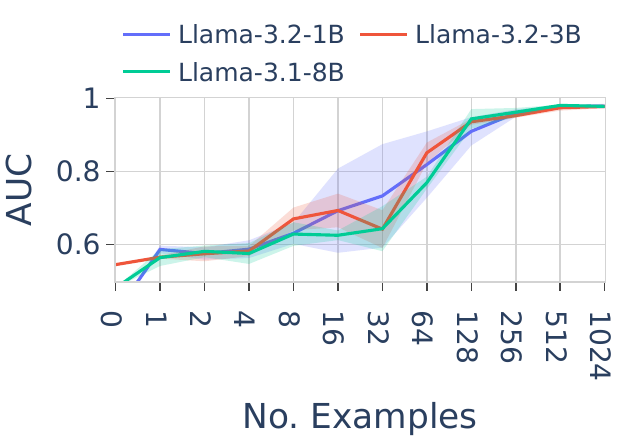}
        }
        \subfloat[Llama-$3$, Language $\lang_5$]{
        \includegraphics[scale=0.35]{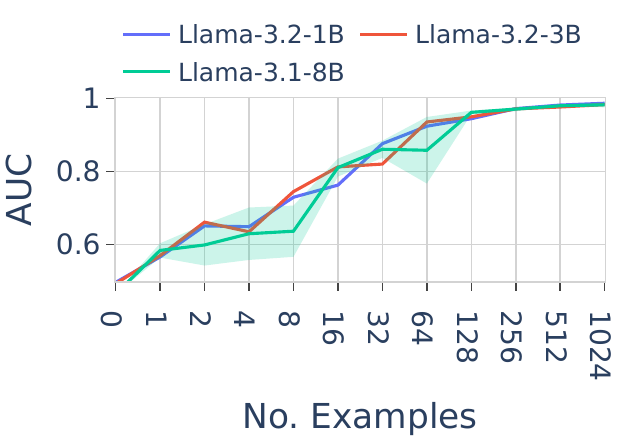}
        }

        \subfloat[Gemma-, Language $\lang_1$]{
        \includegraphics[scale=0.35]{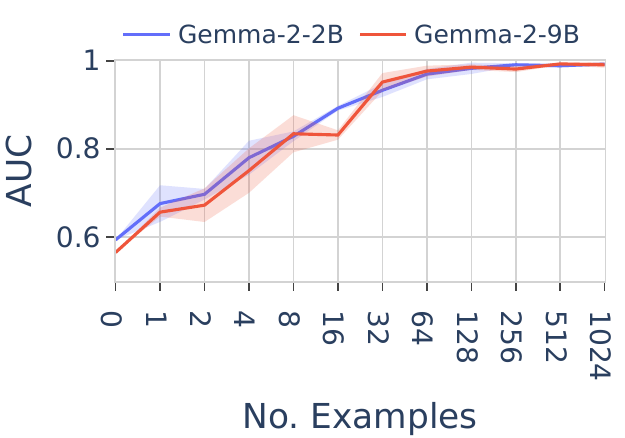}
        }
        \subfloat[Gemma, Language $\lang_2$]{
        \includegraphics[scale=0.35]{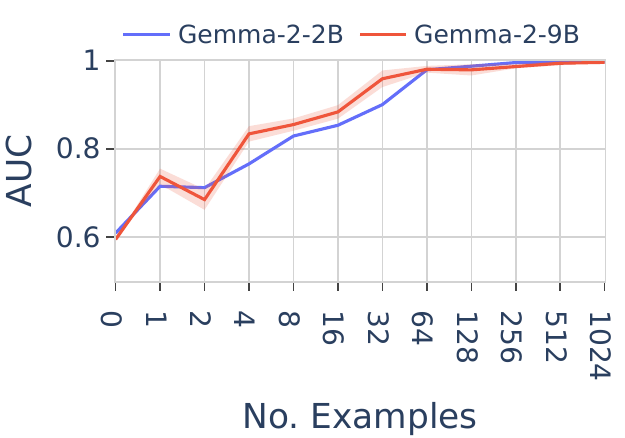}
        }
        \subfloat[Gemma, Language $\lang_4$]{
        \includegraphics[scale=0.35]{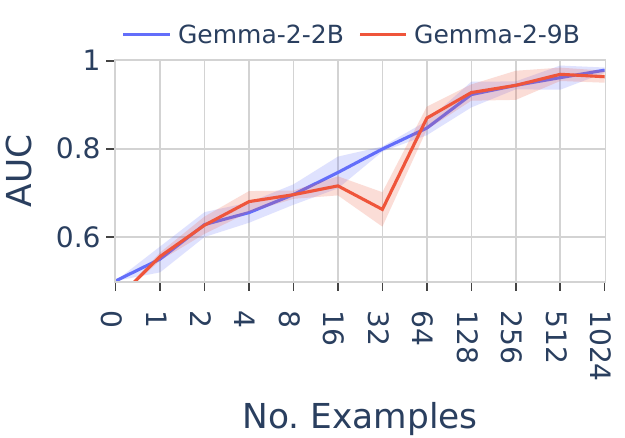}
        }
        \subfloat[Gemma, Language $\lang_5$]{
        \includegraphics[scale=0.35]{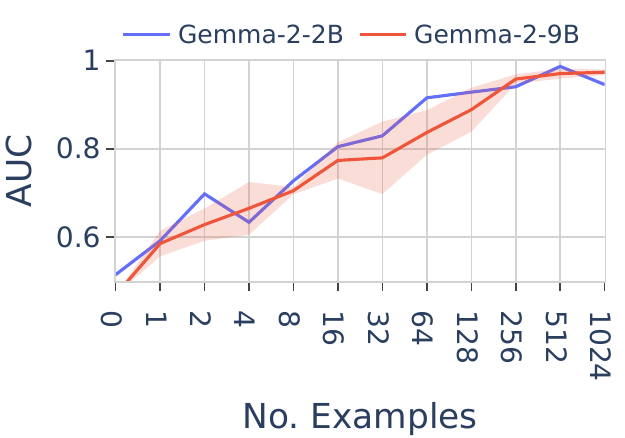}
        }

        \subfloat[Pythia, Language $\lang_1$]{
        \includegraphics[scale=0.35]{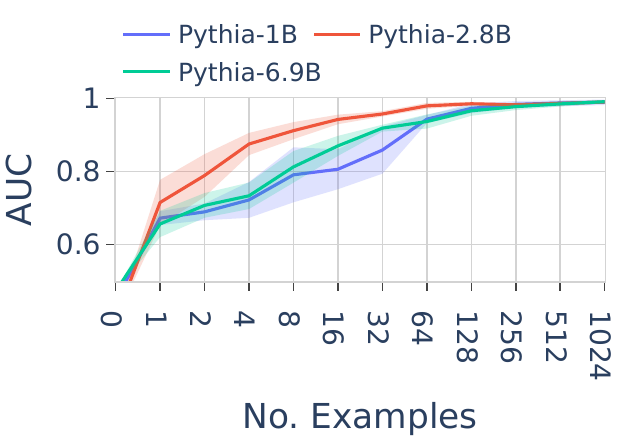}
        }
        \subfloat[Pythia, Language $\lang_2$]{
        \includegraphics[scale=0.35]{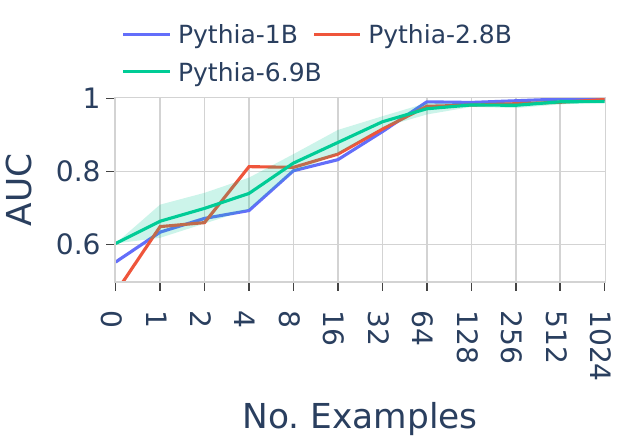}
        }
        \subfloat[Pythia, Language $\lang_4$]{
        \includegraphics[scale=0.35]{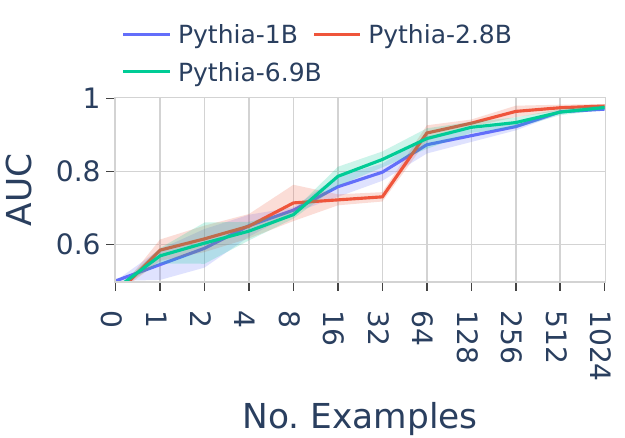}
        }
        \subfloat[Pythia, Language $\lang_5$]{
        \includegraphics[scale=0.35]{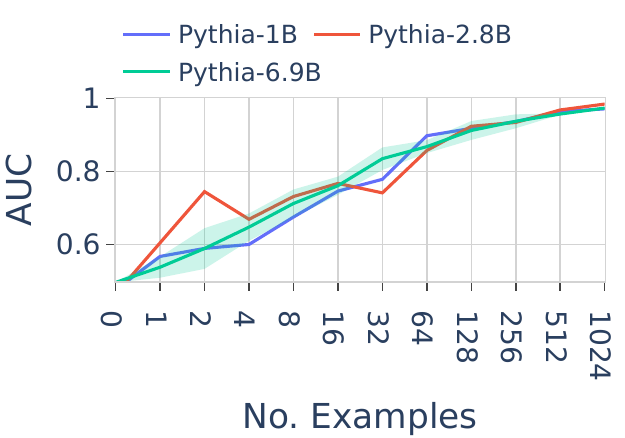}
        }

        \subfloat[Opt, Language $\lang_1$]{
        \includegraphics[scale=0.35]{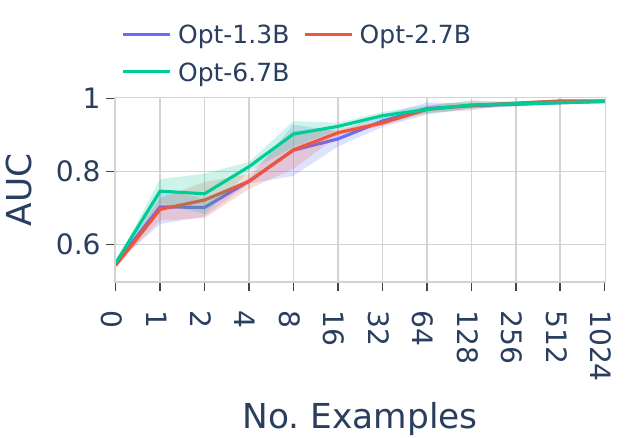}
        }
        \subfloat[Opt, Language $\lang_2$]{
        \includegraphics[scale=0.35]{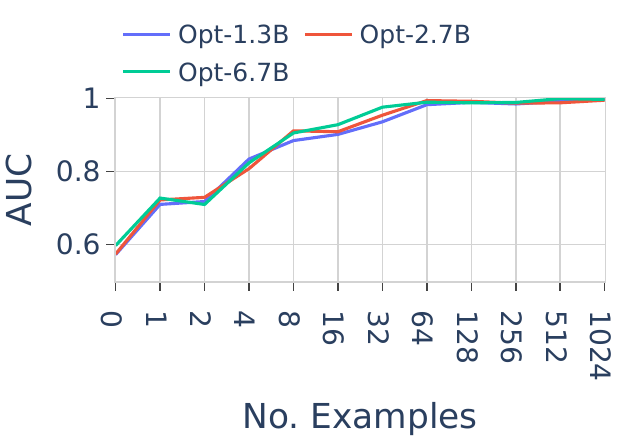}
        }
        \subfloat[Opt, Language $\lang_4$]{
        \includegraphics[scale=0.35]{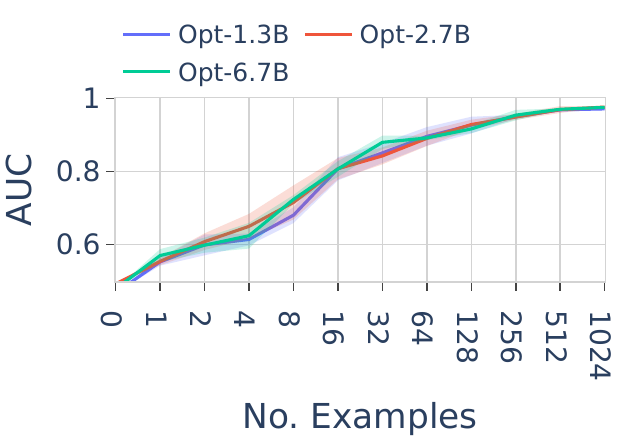}
        }
        \subfloat[Opt, Language $\lang_5$]{
        \includegraphics[scale=0.35]{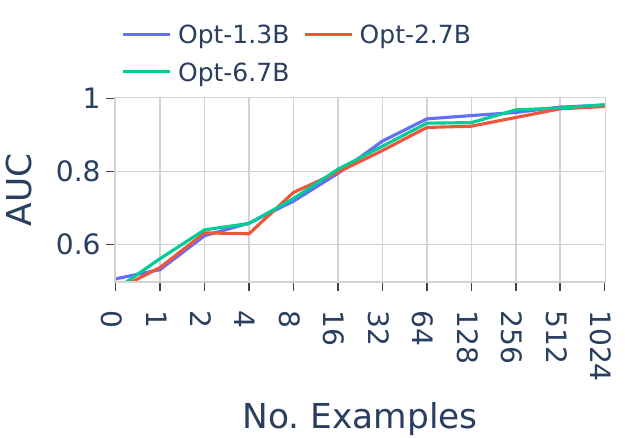}
        }

        \caption{Intra-family {\ft} performance. }

        \label{fig:intra-model_family_fine-tuning}

\end{figure*}

\begin{figure*}
	\centering
        \captionsetup[subfigure]{labelformat=empty}

        \subfloat[Qwen, Language $\lang_1$]{
        \includegraphics[scale=0.35]{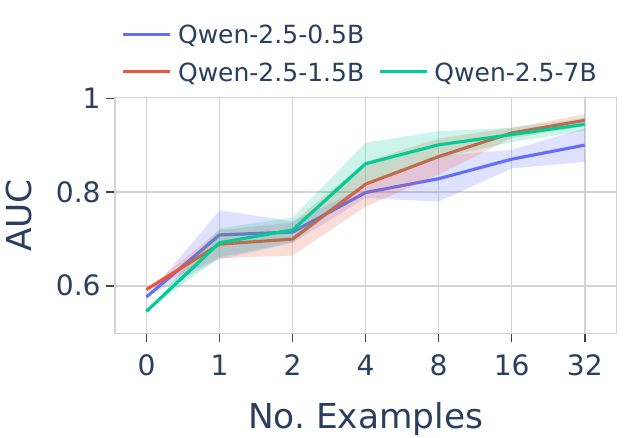}
        }
        \subfloat[Qwen, Language $\lang_2$]{
        \includegraphics[scale=0.35]{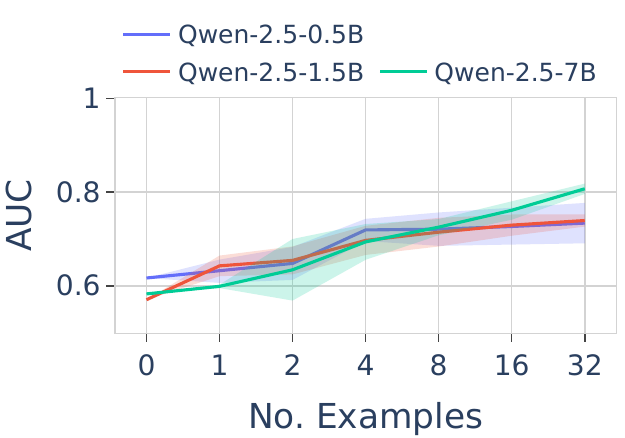}
        }
        \subfloat[Qwen, Language $\lang_4$]{
        \includegraphics[scale=0.35]{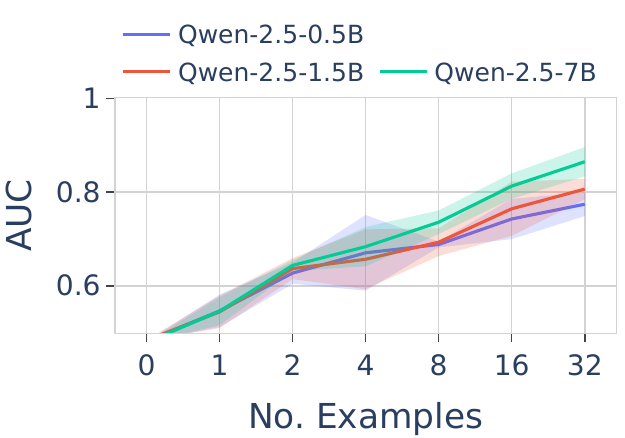}
        }
        \subfloat[Qwen, Language $\lang_5$]{
        \includegraphics[scale=0.35]{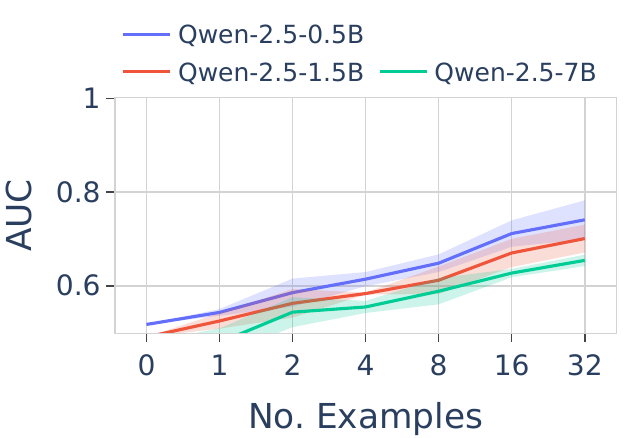}
        }

        \subfloat[Mistral, Language $\lang_1$]{
        \includegraphics[scale=0.35]{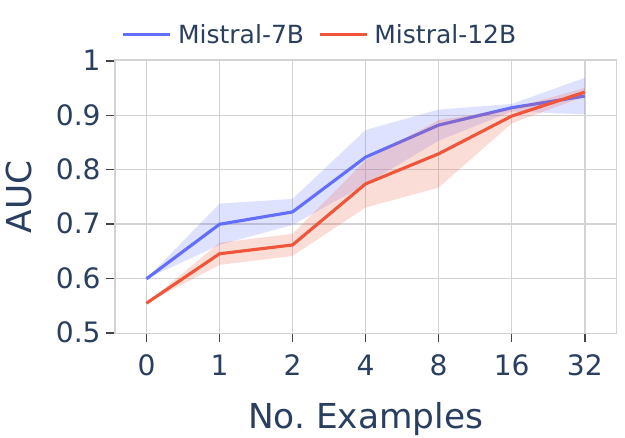}
        }
        \subfloat[Mistral, Language $\lang_2$]{
        \includegraphics[scale=0.35]{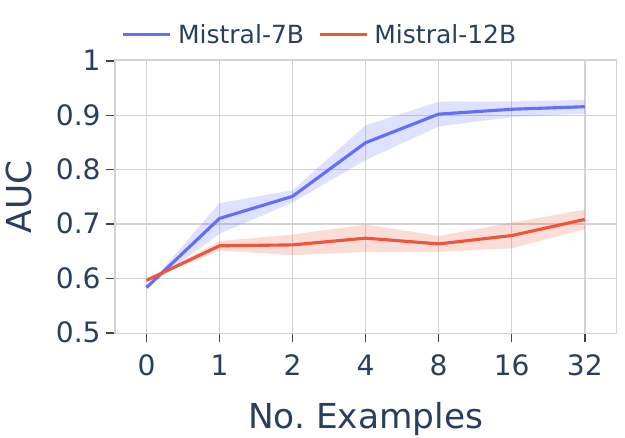}
        }
        \subfloat[Mistral, Language $\lang_4$]{
        \includegraphics[scale=0.35]{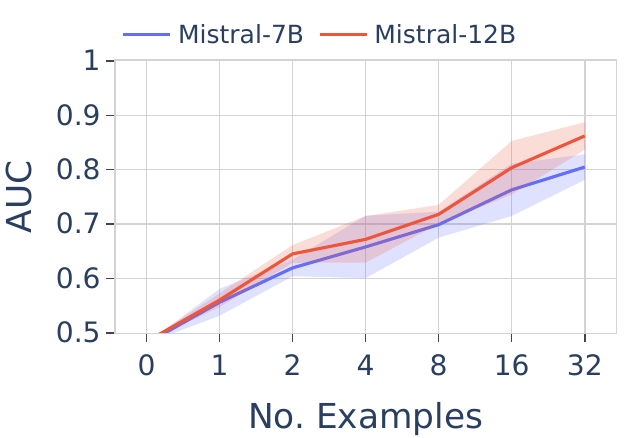}
        }
        \subfloat[Mistral, Language $\lang_5$]{
        \includegraphics[scale=0.35]{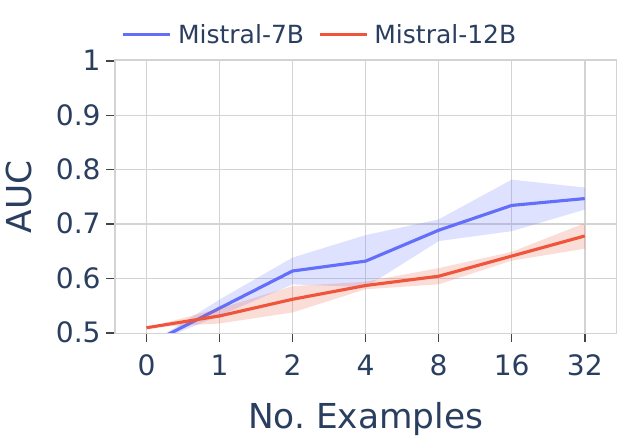}
        }

        \subfloat[Llama-$2$, Language $\lang_1$]{
        \includegraphics[scale=0.35]{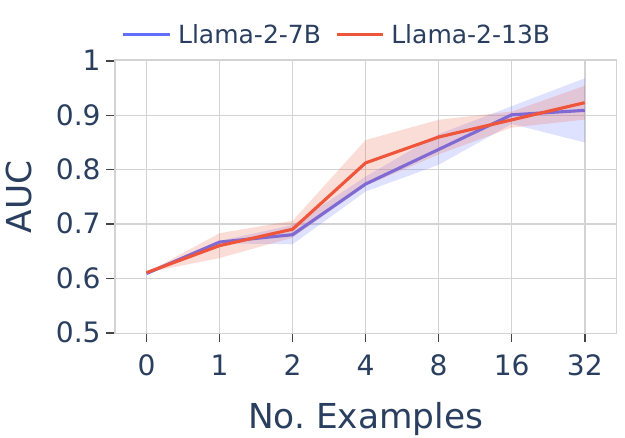}
        }
        \subfloat[Llama-$2$, Language $\lang_2$]{
        \includegraphics[scale=0.35]{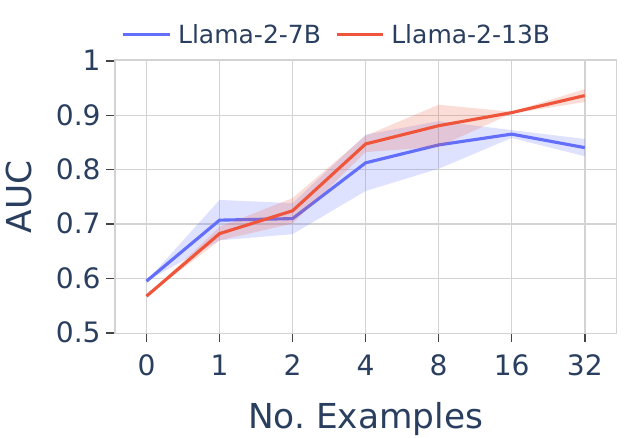}
        }
        \subfloat[Llama-$2$, Language $\lang_4$]{
        \includegraphics[scale=0.35]{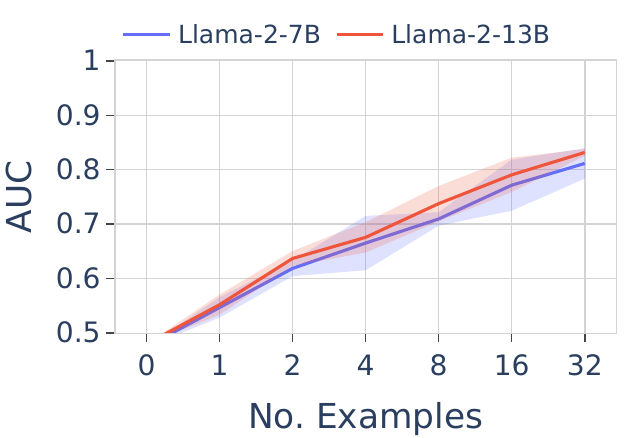}
        }
        \subfloat[Llama-$2$, Language $\lang_5$]{
        \includegraphics[scale=0.35]{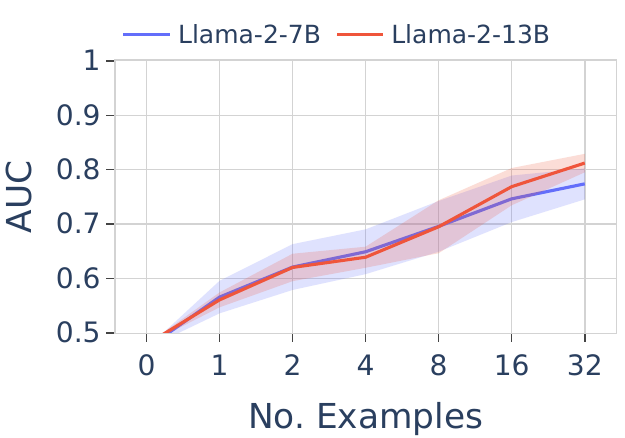}
        }

        \subfloat[Llama-$3$, Language $\lang_1$]{
        \includegraphics[scale=0.35]{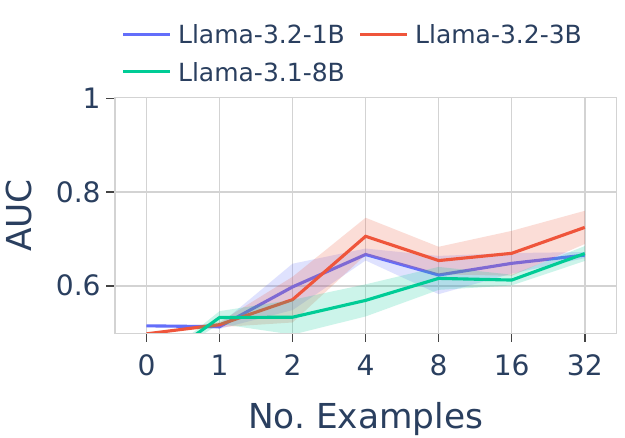}
        }
        \subfloat[Llama-$3$, Language $\lang_2$]{
        \includegraphics[scale=0.35]{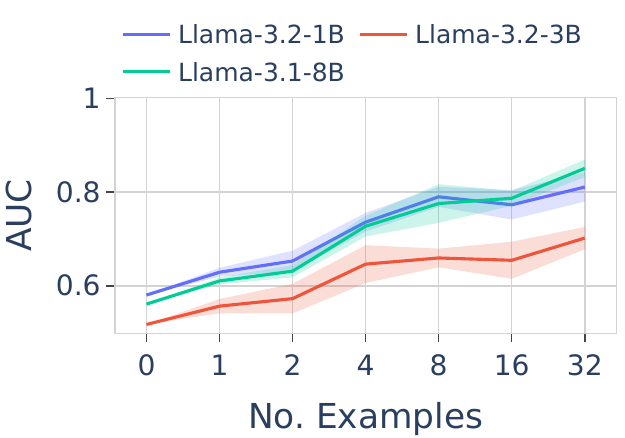}
        }
        \subfloat[Llama-$3$, Language $\lang_4$]{
        \includegraphics[scale=0.35]{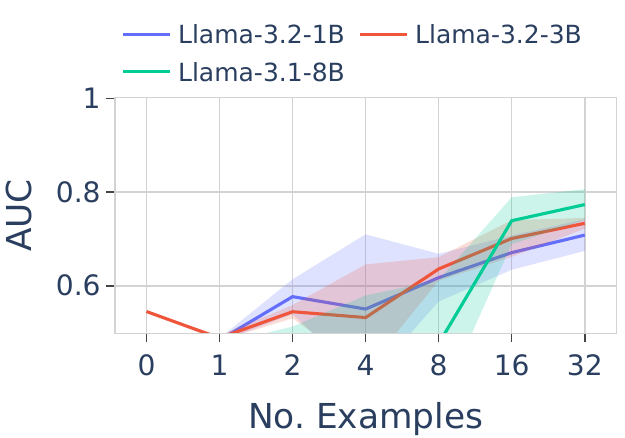}
        }
        \subfloat[Llama-$3$, Language $\lang_5$]{
        \includegraphics[scale=0.35]{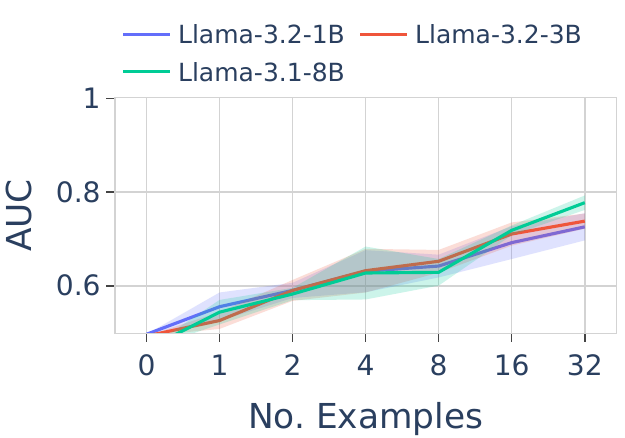}
        }

        \subfloat[Gemma-, Language $\lang_1$]{
        \includegraphics[scale=0.35]{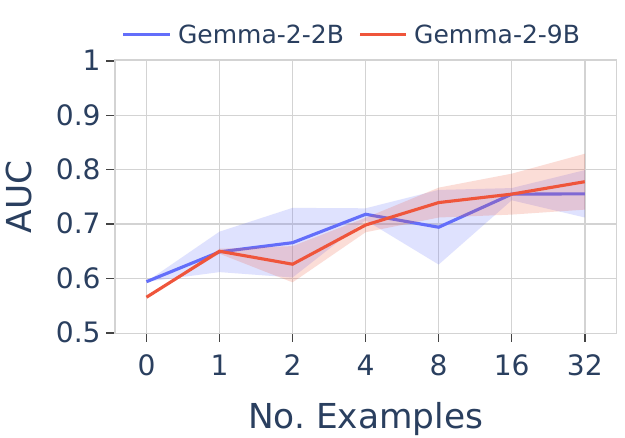}
        }
        \subfloat[Gemma, Language $\lang_2$]{
        \includegraphics[scale=0.35]{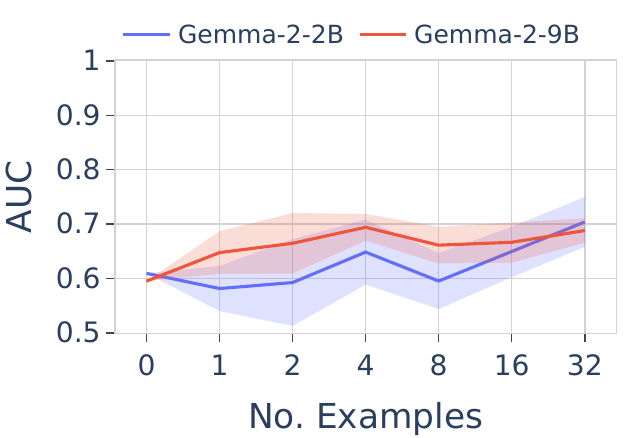}
        }
        \subfloat[Gemma, Language $\lang_4$]{
        \includegraphics[scale=0.35]{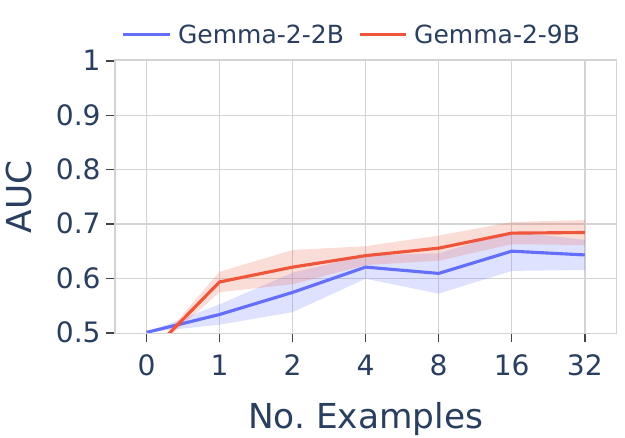}
        }
        \subfloat[Gemma, Language $\lang_5$]{
        \includegraphics[scale=0.35]{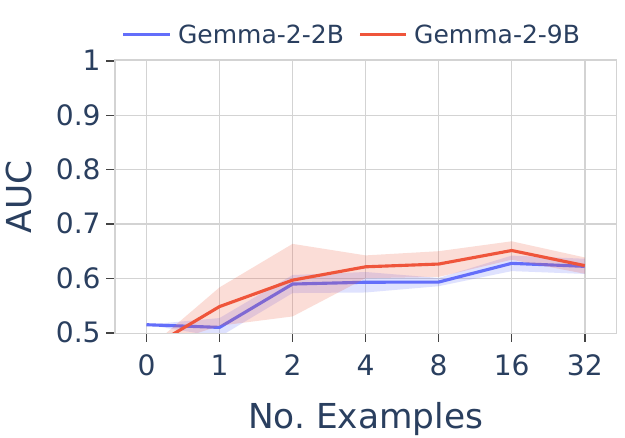}
        }

        \subfloat[Pythia, Language $\lang_1$]{
        \includegraphics[scale=0.35]{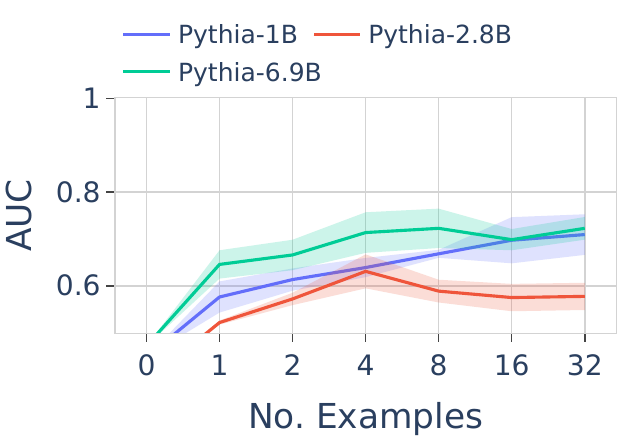}
        }
        \subfloat[Pythia, Language $\lang_2$]{
        \includegraphics[scale=0.35]{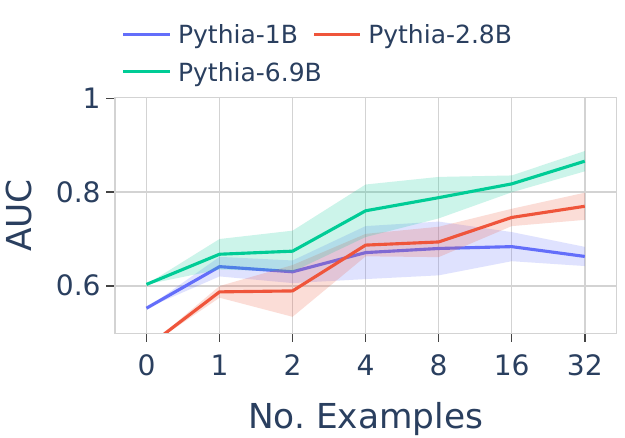}
        }
        \subfloat[Pythia, Language $\lang_4$]{
        \includegraphics[scale=0.35]{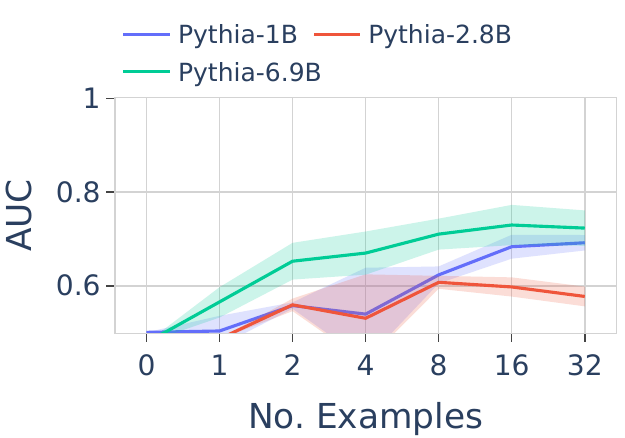}
        }
        \subfloat[Pythia, Language $\lang_5$]{
        \includegraphics[scale=0.35]{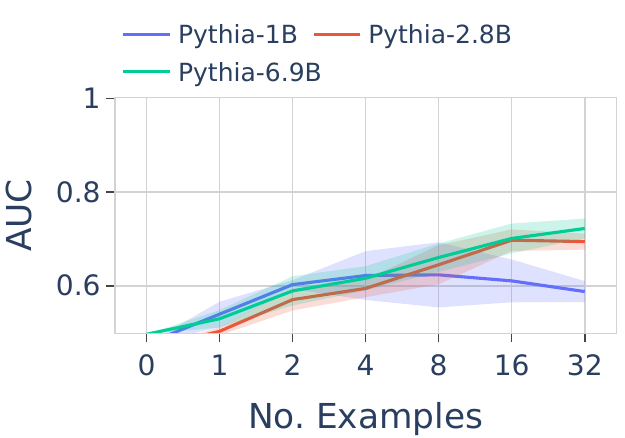}
        }

        \subfloat[Opt, Language $\lang_1$]{
        \includegraphics[scale=0.35]{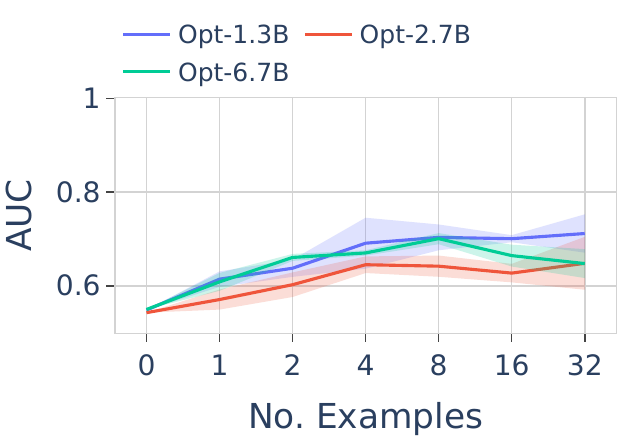}
        }
        \subfloat[Opt, Language $\lang_2$]{
        \includegraphics[scale=0.35]{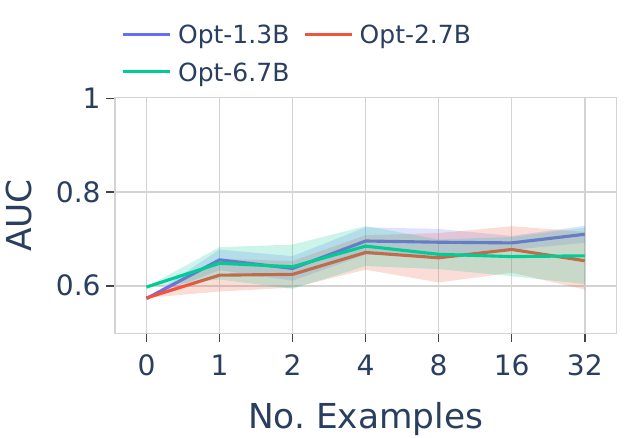}
        }
        \subfloat[Opt, Language $\lang_4$]{
        \includegraphics[scale=0.35]{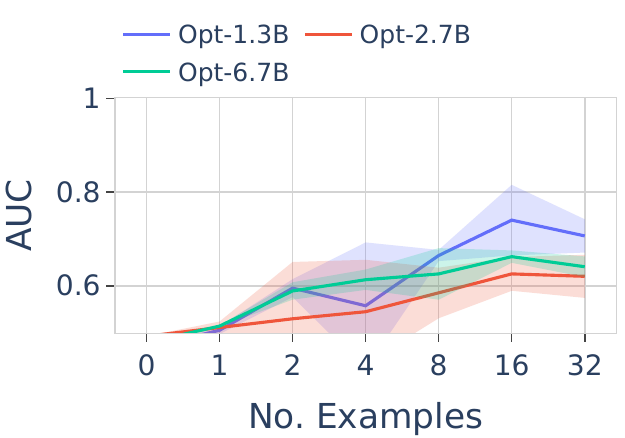}
        }
        \subfloat[Opt, Language $\lang_5$]{
        \includegraphics[scale=0.35]{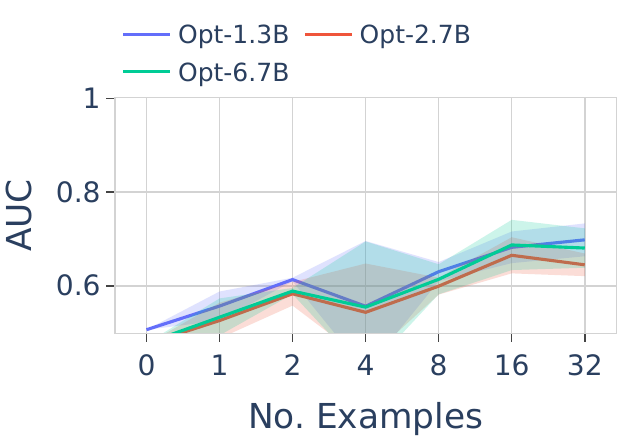}
        }

        \caption{Intra-family {\icl} performance. }

        \label{fig:intra-model_family_incontext_learning}

\end{figure*}

\begin{figure*}
        \centering

        \subfloat[Language $ \lang_1 $\\($ G_\alpha^{\text{Numerical}} $)]{
                \includegraphics[scale=0.4]{figures/incontext_experiments/comparison_incontext_vs_finetuning_qwen-2.5-7b_pcfg_cfg3b_disjoint_terminals_auc_edit_distance.pdf}
	}
        \subfloat[Language $ \lang_2 $\\($ G_\alpha^{\text{Latin}} $)]{
                \includegraphics[scale=0.4]{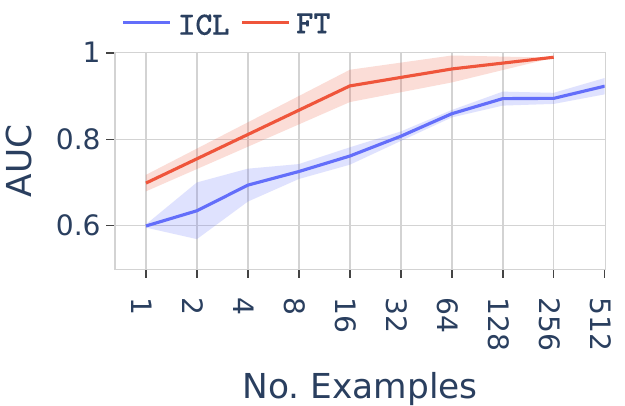}
	}
        \subfloat[Language $ \lang_3 $\\($ G_\alpha^{\text{Under-trained}} $)]{
                \includegraphics[scale=0.4]{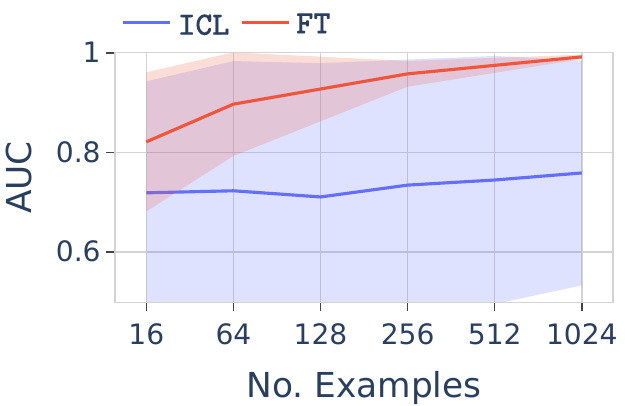}
	}

        \subfloat[Language $ \lang_4 $\\($ G_\beta^{\text{Numerical}} $)]{
		\includegraphics[scale=0.4]{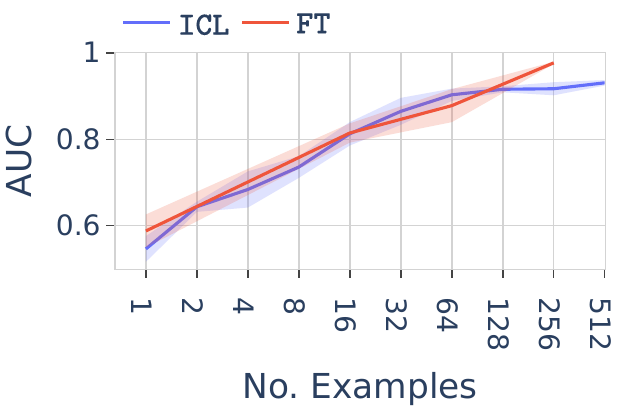}
	}
        \subfloat[Language $ \lang_5 $\\($ G_\beta^{\text{Latin}} $)]{
		\includegraphics[scale=0.4]{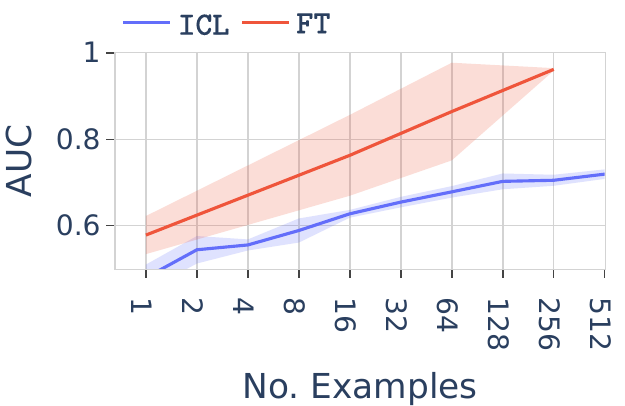}
	}
        \subfloat[Language $ \lang_6 $\\($ G_\beta^{\text{Under-trained}} $)]{
		\includegraphics[scale=0.4]{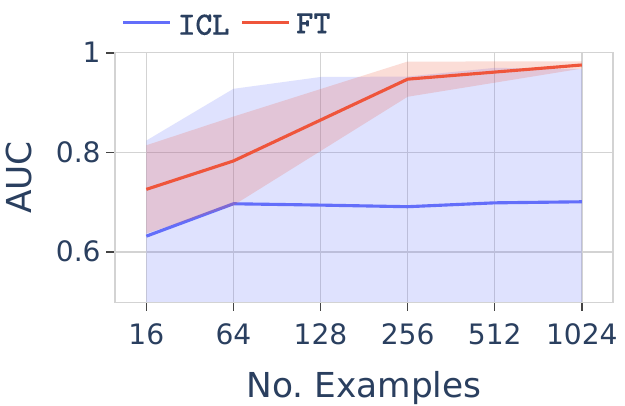}
	}

        \caption{Qwen-$ 2.5 $-$ 7 $B: comparison between fine-tuning and in-context learning across different languages}

        \label{fig:fine_tuning_vs_incontext_learning_qwen_2.5_7b}
\end{figure*}

\begin{figure*}
        \centering

        \subfloat[Language $ \lang_1 $\\($ G_\alpha^{\text{Numerical}} $)]{
                \includegraphics[scale=0.4]{figures/incontext_experiments/comparison_incontext_vs_finetuning_mistral-7b_pcfg_cfg3b_disjoint_terminals_auc_edit_distance.pdf}
	}
        \subfloat[Language $ \lang_2 $\\($ G_\alpha^{\text{Latin}} $)]{
                \includegraphics[scale=0.4]{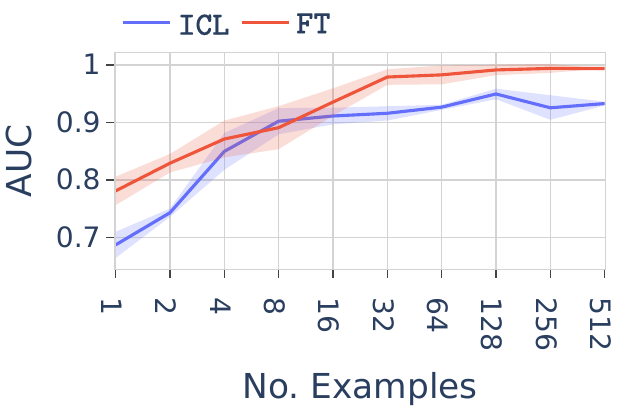}
	}
        \subfloat[Language $ \lang_3 $\\($ G_\alpha^{\text{Under-trained}} $)]{
                \includegraphics[scale=0.4]{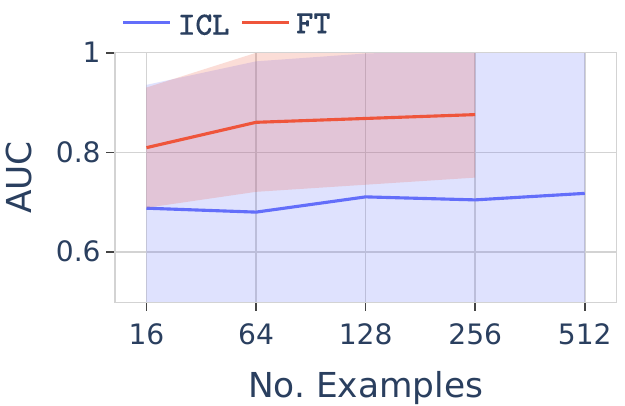}
	}

        \subfloat[Language $ \lang_4 $\\($ G_\beta^{\text{Numerical}} $)]{
		\includegraphics[scale=0.4]{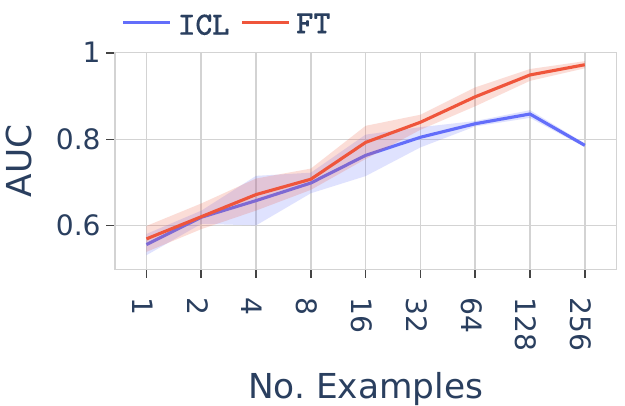}
	}
        \subfloat[Language $ \lang_5 $\\($ G_\beta^{\text{Latin}} $)]{
		\includegraphics[scale=0.4]{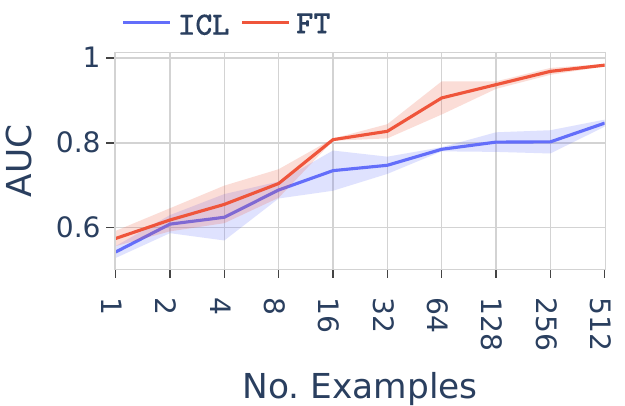}
	}
        \subfloat[Language $ \lang_6 $\\($ G_\beta^{\text{Under-trained}} $)]{
		\includegraphics[scale=0.4]{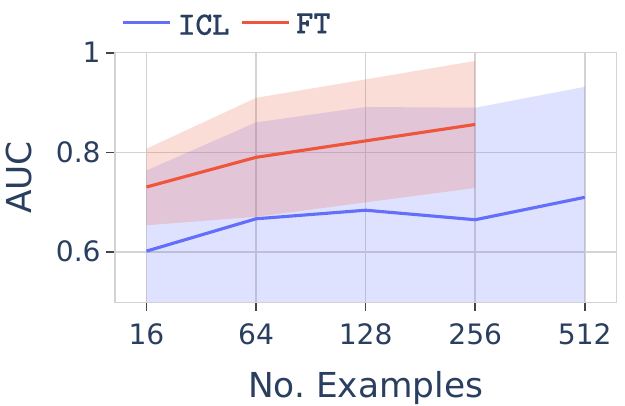}
	}

        \caption{Mistral-$ 7 $B: comparison between fine-tuning and in-context learning across different languages}

        \label{fig:fine_tuning_vs_incontext_learning_mistral_7b}
\end{figure*}

\begin{figure*}[!t]
        \captionsetup[subfigure]{justification=centering}
	\centering
        
        \subfloat[Language $ \lang_1 $\\($ G_\alpha^{\text{Numerical}} $)]{
                \includegraphics[scale=0.4]{figures/incontext_experiments/comparison_incontext_vs_finetuning_llama-2-7b_pcfg_cfg3b_disjoint_terminals_auc_edit_distance.pdf}
	}
        \subfloat[Language $ \lang_3 $\\($ G_\alpha^{\text{Under-trained}} $)]{
                \includegraphics[scale=0.4]{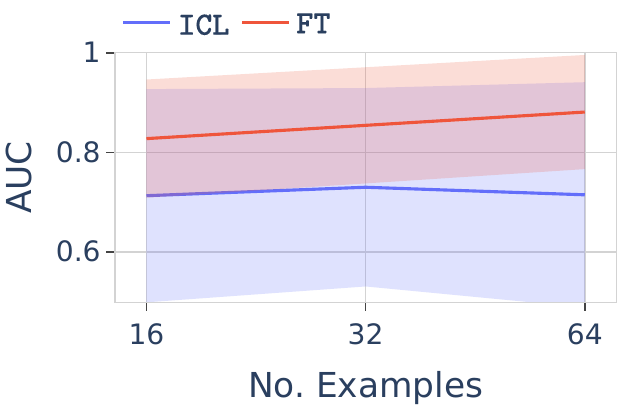}
	}

        \subfloat[Language $ \lang_4 $\\($ G_\beta^{\text{Numerical}} $)]{
		\includegraphics[scale=0.4]{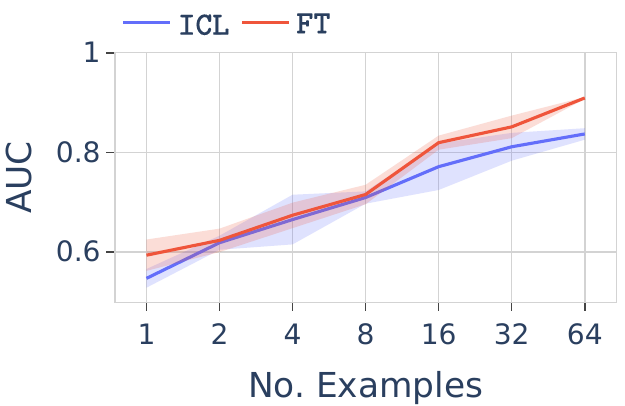}
	}
        \subfloat[Language $ \lang_6 $\\($ G_\beta^{\text{Under-trained}} $)]{
		\includegraphics[scale=0.4]{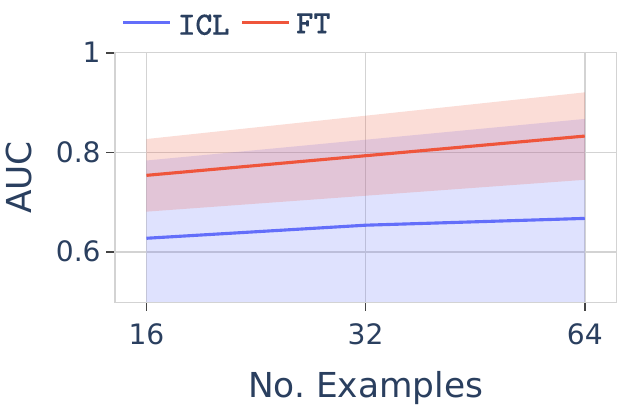}
	}
        
        \caption{Llama-$ 2 $-$7$B: comparison between fine-tuning and in-context learning across different languages.}

        \label{fig:fine_tuning_vs_incontext_learning_llama_2_7b}

\end{figure*}

\begin{figure*}
        \centering

        \subfloat[Language $ \lang_1 $\\($ G_\alpha^{\text{Numerical}} $)]{
                \includegraphics[scale=0.4]{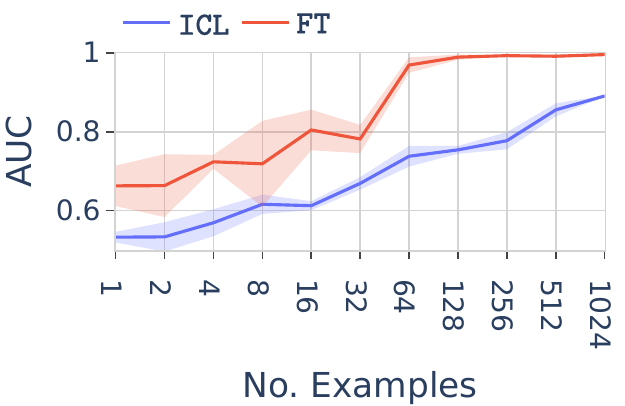}
	}
        \subfloat[Language $ \lang_3 $\\($ G_\alpha^{\text{Under-trained}} $)]{
                \includegraphics[scale=0.4]{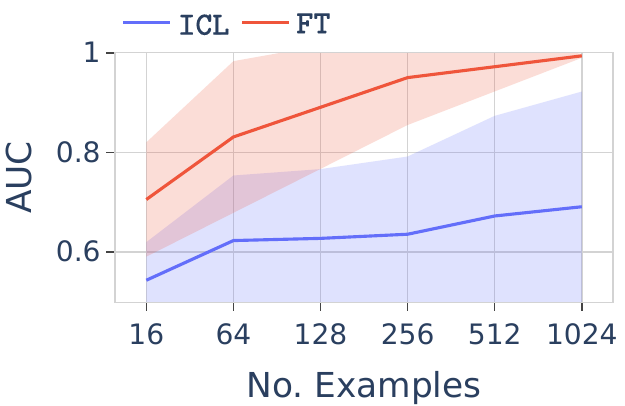}
	}

        \subfloat[Language $ \lang_4 $\\($ G_\beta^{\text{Numerical}} $)]{
		\includegraphics[scale=0.4]{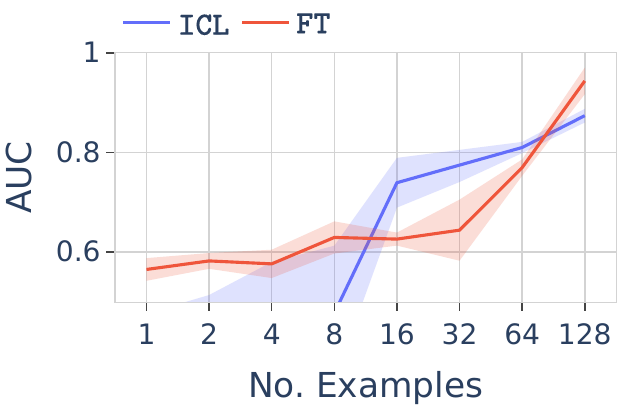}
	}
        \subfloat[Language $ \lang_6 $\\($ G_\beta^{\text{Under-trained}} $)]{
		\includegraphics[scale=0.4]{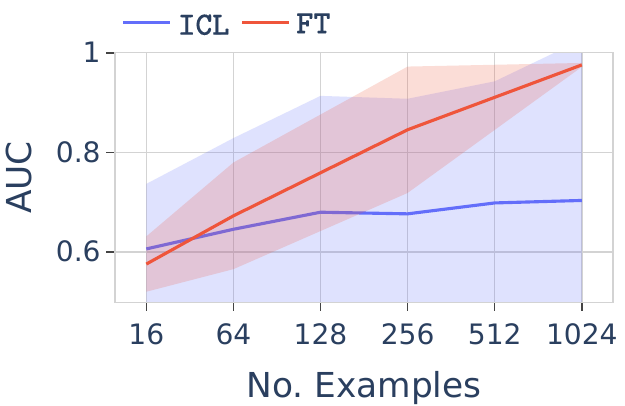}
	}

        \caption{Llama-$ 3.1 $-$ 8 $B: comparison between fine-tuning and in-context learning across different languages}

        \label{fig:fine_tuning_vs_incontext_learning_llama_3.1_8b}
\end{figure*}

\begin{figure*}
    \centering
    \captionsetup[subfigure]{justification=centering}

    \subfloat[Qwen-$2.5$-$0.5$B]{
        \includegraphics[scale=0.35]{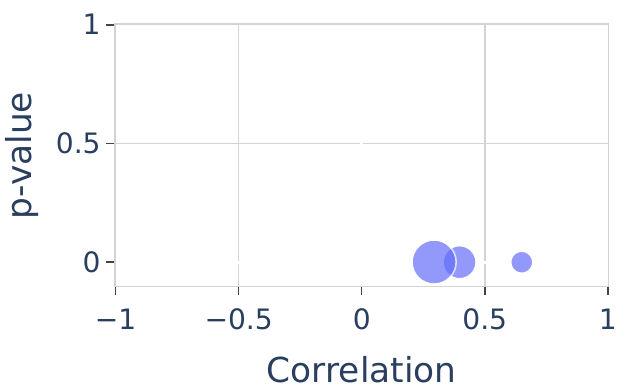}
    }
    \subfloat[Qwen-$2.5$-$1.5$B]{
        \includegraphics[scale=0.35]{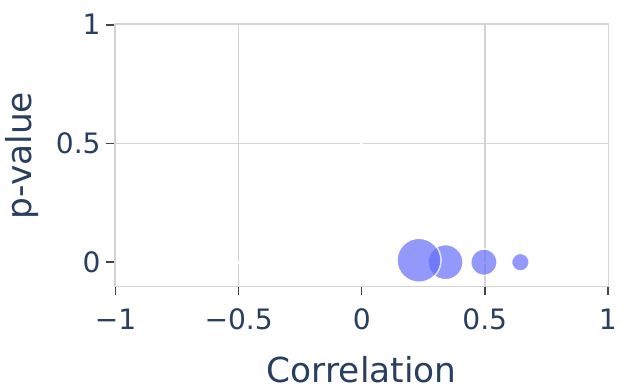}
    }
    \subfloat[Qwen-$2.5$-$7$B]{
        \includegraphics[scale=0.35]{figures/inductive_bias/inductive_bias_pcfg_cfg3b_disjoint_terminals_qwen-2.5-7b_0_edit_distance.pdf}
    }

    \subfloat[Mistral-$7$B]{
        \includegraphics[scale=0.35]{figures/inductive_bias/inductive_bias_pcfg_cfg3b_disjoint_terminals_mistral-7b_0_edit_distance.pdf}
    }
    \subfloat[Mistral-$12$B]{
        \includegraphics[scale=0.35]{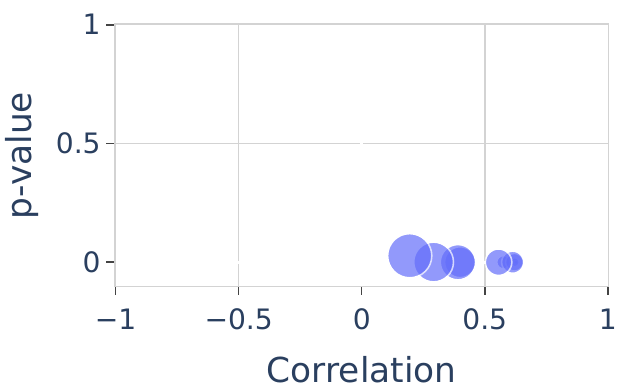}
    }
    \subfloat[Llama-$2$-$7$B]{
        \includegraphics[scale=0.35]{figures/inductive_bias/inductive_bias_pcfg_cfg3b_disjoint_terminals_llama-2-7b_0_edit_distance.pdf}
    }
    \subfloat[Llama-$2$-$13$B]{
        \includegraphics[scale=0.35]{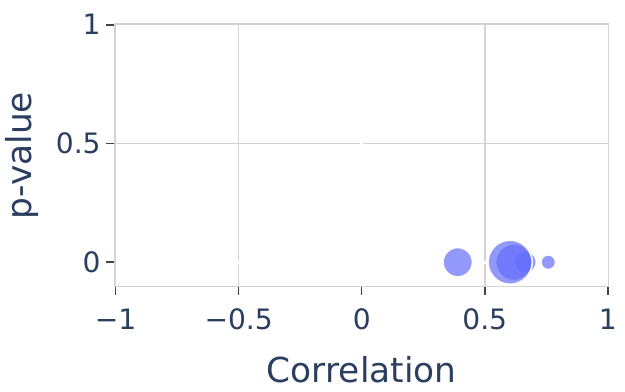}
    }

    \subfloat[Llama-$3.2$-$1$B]{
        \includegraphics[scale=0.35]{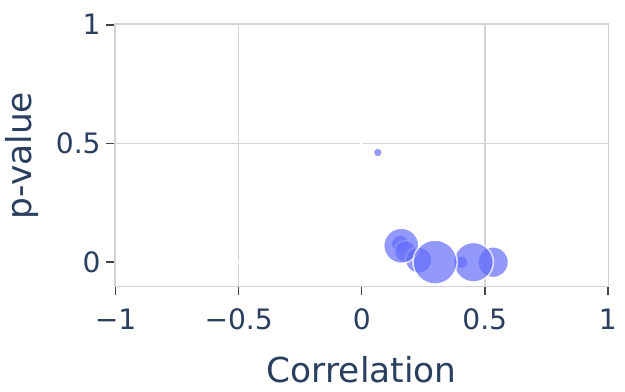}
    }
    \subfloat[Llama-$3.2$-$3$B]{
        \includegraphics[scale=0.35]{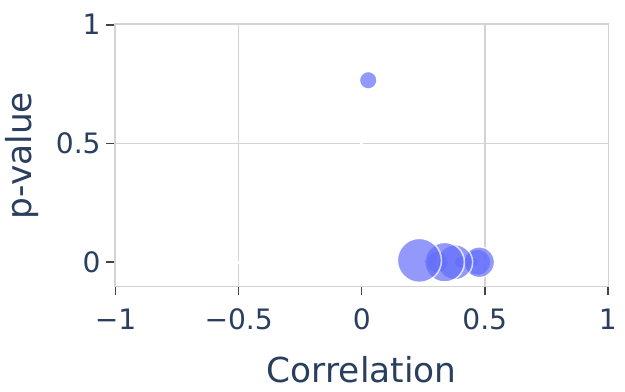}
    }
    \subfloat[Llama-$3.1$-$8$B]{
        \includegraphics[scale=0.35]{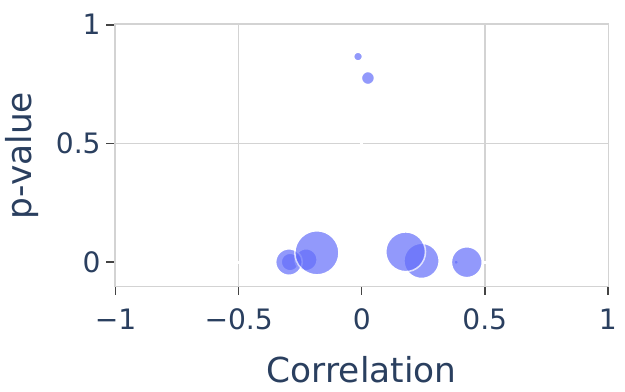}
    }

    \subfloat[Gemma-$2$-$2$B]{
        \includegraphics[scale=0.35]{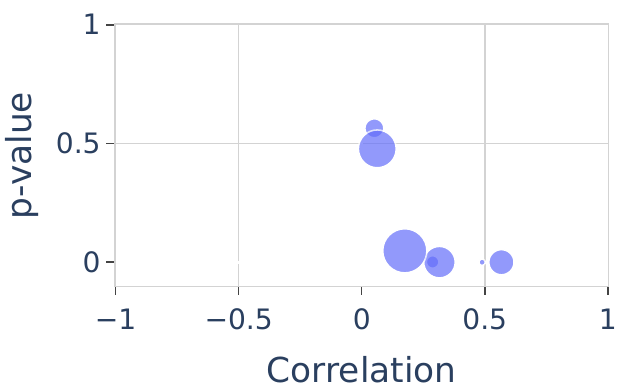}
    }
    \subfloat[Gemma-$2$-$9$B]{
        \includegraphics[scale=0.35]{figures/inductive_bias/inductive_bias_pcfg_cfg3b_disjoint_terminals_gemma-2-9b_0_edit_distance.pdf}
    }

    \subfloat[Pythia-$1$B]{
        \includegraphics[scale=0.35]{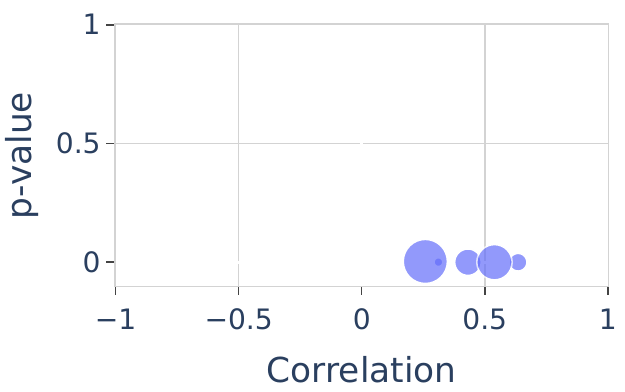}
    }
    \subfloat[Pythia-$2.8$B]{
        \includegraphics[scale=0.35]{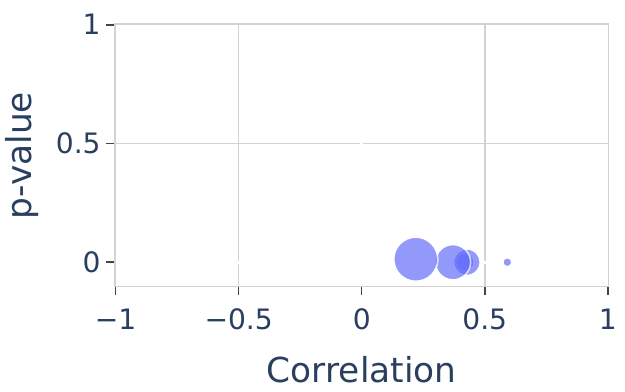}
    }
    \subfloat[Pythia-$6.9$B]{
        \includegraphics[scale=0.35]{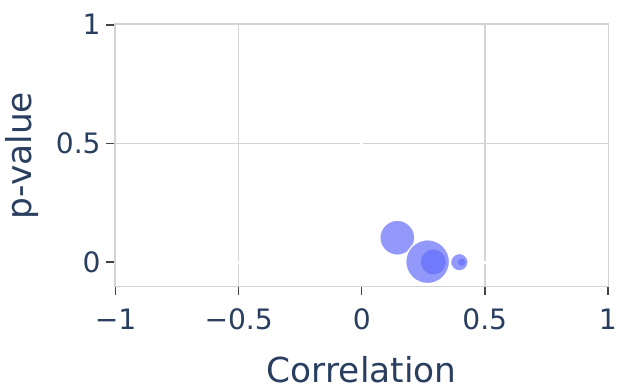}
    }

    \subfloat[Opt-$1.3$B]{
        \includegraphics[scale=0.35]{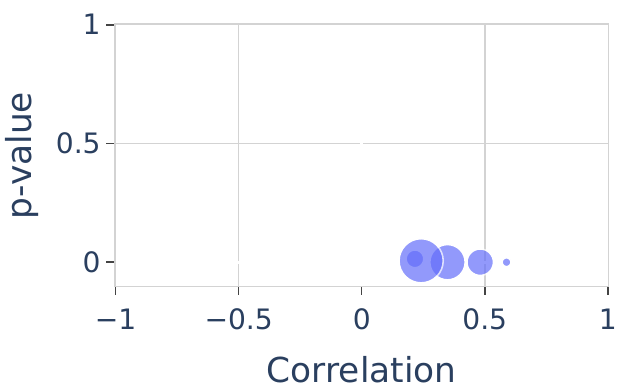}
    }
    \subfloat[Opt-$2.7$B]{
        \includegraphics[scale=0.35]{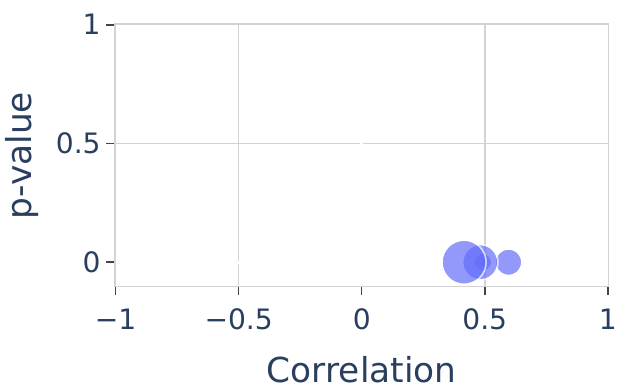}
    }
    \subfloat[Opt-$6.7$B]{
        \includegraphics[scale=0.35]{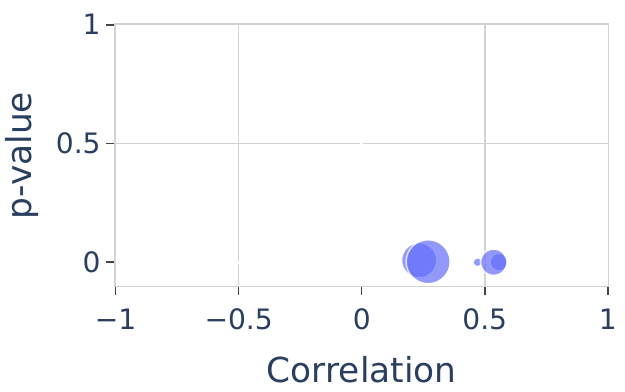}
    }
    
    \caption{Inductive bias of {\icl} and {\ft} on language $ \lang_1 $, computed as the Pearson correlation of generation loss of {\ft} and {\icl} on identical test strings. Correlation, despite being positive, tends to decrease with higher examples (larger markers).}
    \label{fig:inductive_bias_l1}
    
\end{figure*}

\begin{figure*}
    \centering
    \captionsetup[subfigure]{justification=centering}

    \subfloat[Qwen-$2.5$-$0.5$B]{
        \includegraphics[scale=0.35]{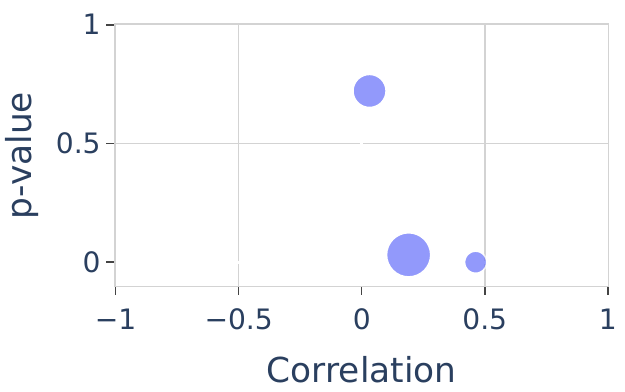}
    }
    \subfloat[Qwen-$2.5$-$1.5$B]{
        \includegraphics[scale=0.35]{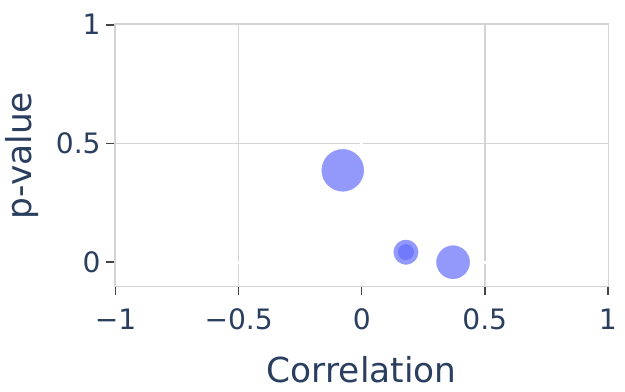}
    }
    \subfloat[Qwen-$2.5$-$7$B]{
        \includegraphics[scale=0.35]{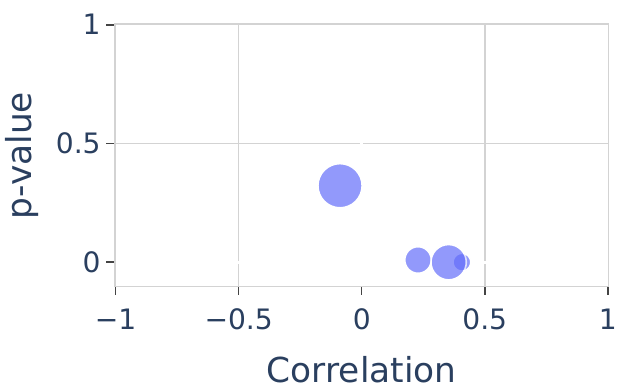}
    }

    \subfloat[Mistral-$7$B]{
        \includegraphics[scale=0.35]{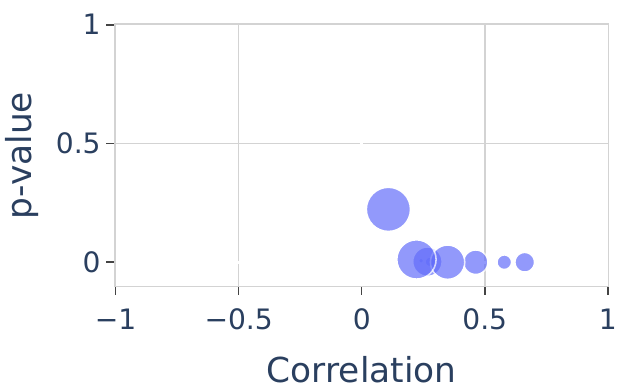}
    }
    \subfloat[Mistral-$12$B]{
        \includegraphics[scale=0.35]{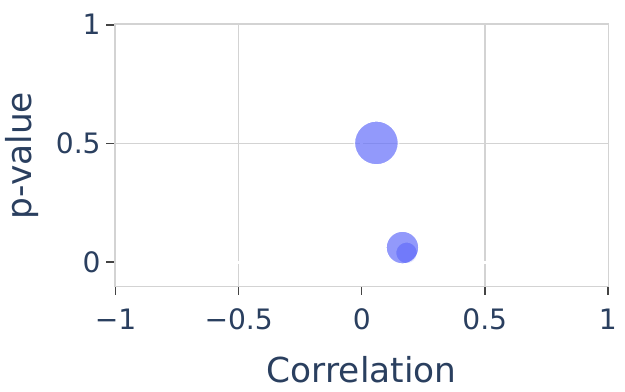}
    }
    \subfloat[Llama-$2$-$7$B]{
        \includegraphics[scale=0.35]{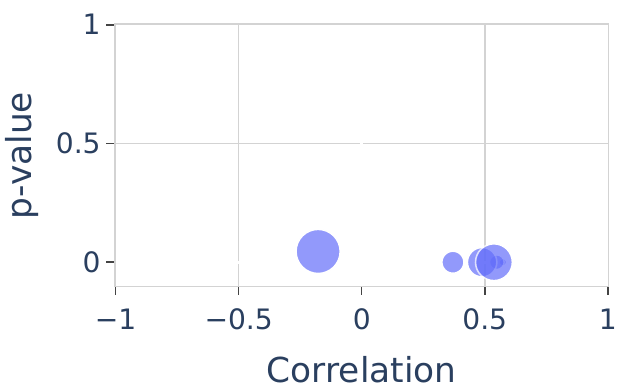}
    }
    \subfloat[Llama-$2$-$13$B]{
        \includegraphics[scale=0.35]{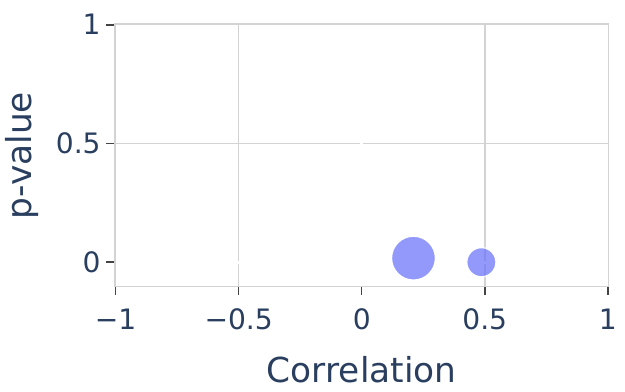}
    }

    \subfloat[Llama-$3.2$-$1$B]{
        \includegraphics[scale=0.35]{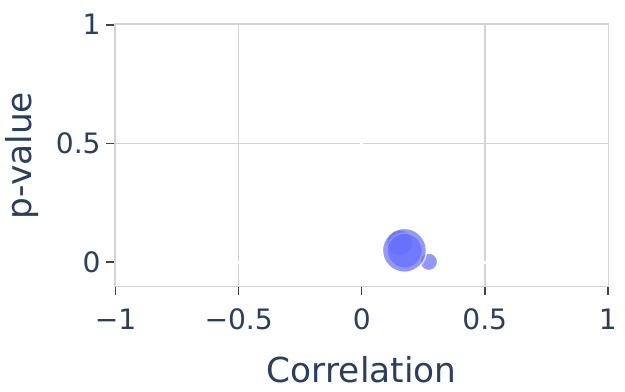}
    }
    \subfloat[Llama-$3.2$-$3$B]{
        \includegraphics[scale=0.35]{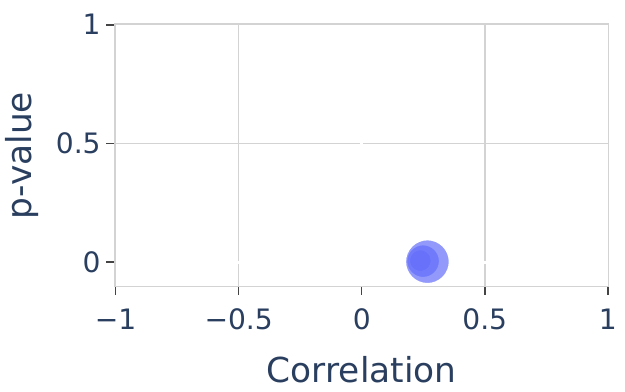}
    }
    \subfloat[Llama-$3.1$-$8$B]{
        \includegraphics[scale=0.35]{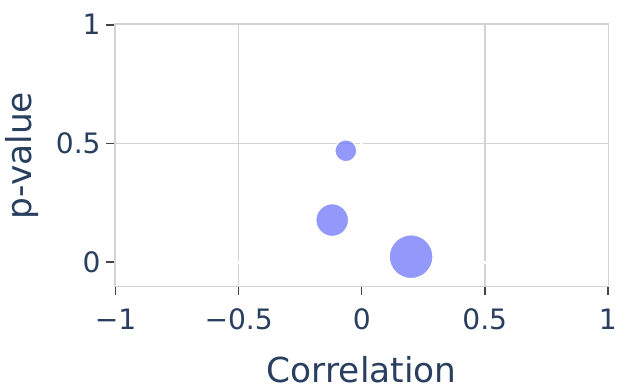}
    }

    \subfloat[Gemma-$2$-$2$B]{
        \includegraphics[scale=0.35]{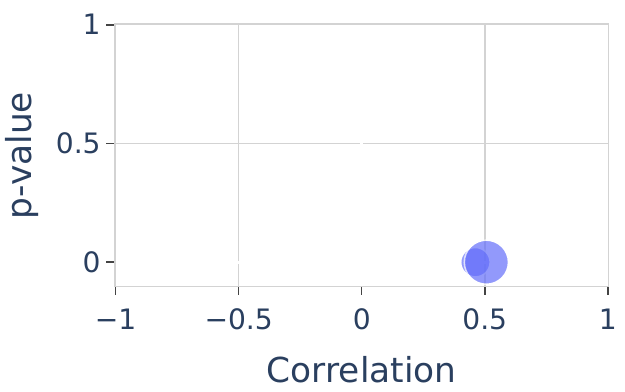}
    }
    \subfloat[Gemma-$2$-$9$B]{
        \includegraphics[scale=0.35]{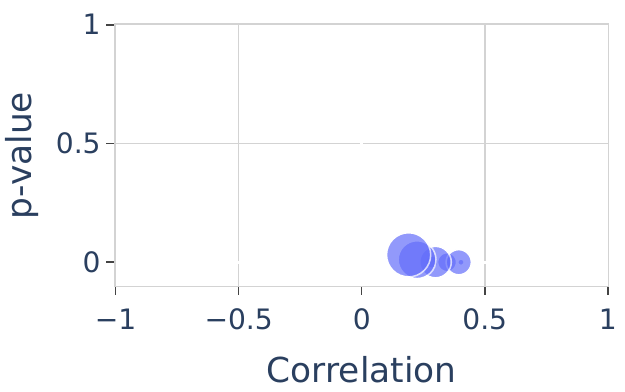}
    }

    \subfloat[Pythia-$1$B]{
        \includegraphics[scale=0.35]{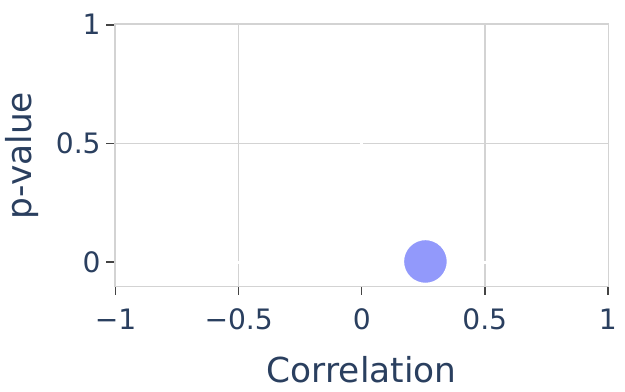}
    }
    \subfloat[Pythia-$2.8$B]{
        \includegraphics[scale=0.35]{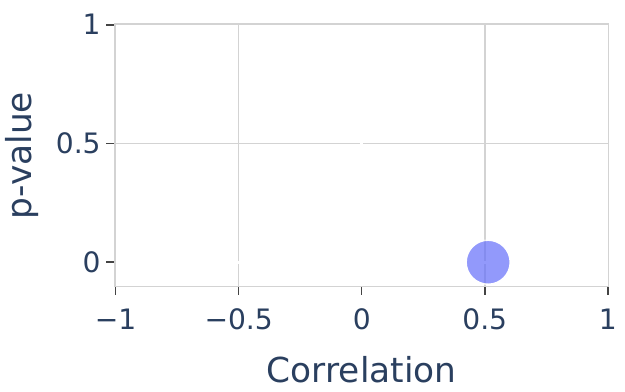}
    }
    \subfloat[Pythia-$6.9$B]{
        \includegraphics[scale=0.35]{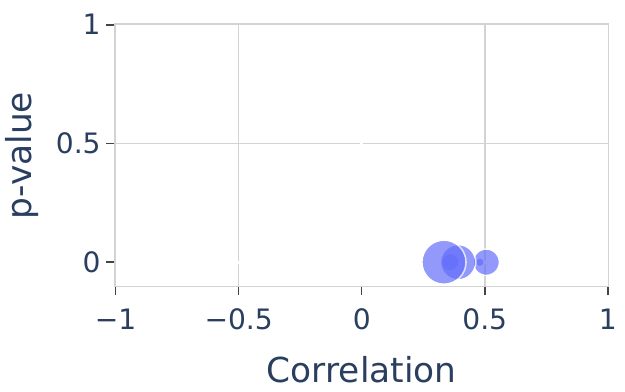}
    }

    \subfloat[Opt-$1.3$B]{
        \includegraphics[scale=0.35]{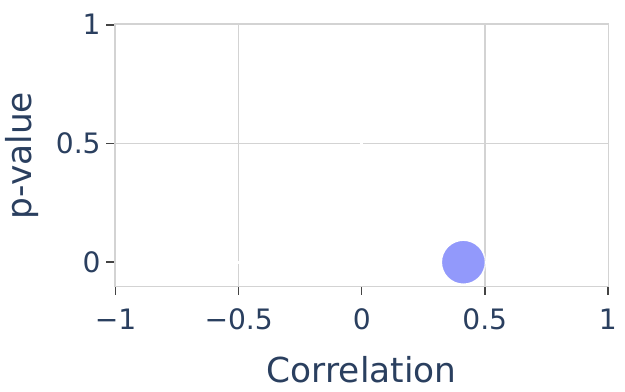}
    }
    \subfloat[Opt-$2.7$B]{
        \includegraphics[scale=0.35]{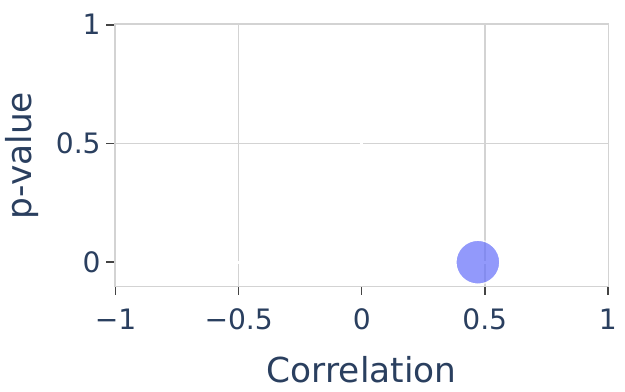}
    }
    \subfloat[Opt-$6.7$B]{
        \includegraphics[scale=0.35]{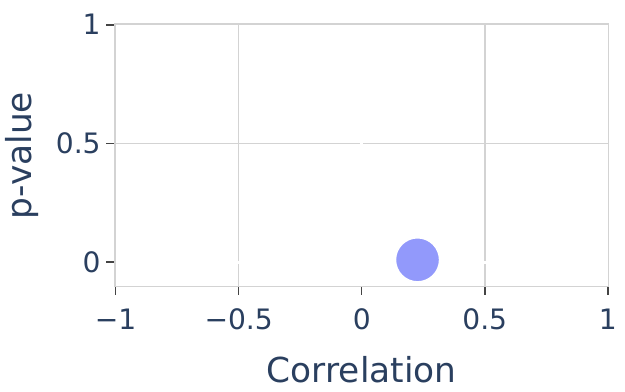}
    }
    
    \caption{Inductive bias of {\icl} and {\ft} on language $ \lang_2 $, computed as the Pearson correlation of generation loss of {\ft} and {\icl} on identical test strings. Correlation, despite being positive, tends to decrease with higher examples (larger markers).}
    \label{fig:inductive_bias_l2}
    
\end{figure*}

\begin{figure*}
    \centering
    \captionsetup[subfigure]{justification=centering}

    \subfloat[Qwen-$2.5$-$0.5$B]{
        \includegraphics[scale=0.35]{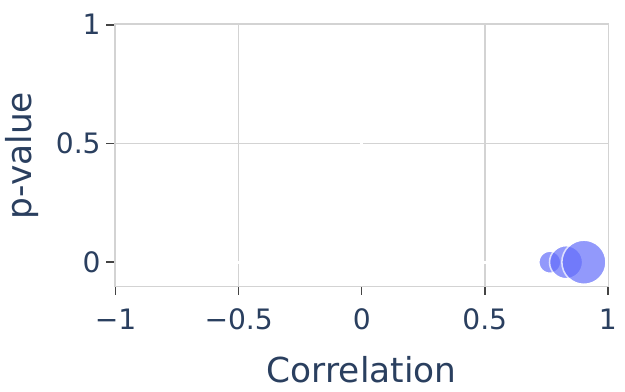}
    }
    \subfloat[Qwen-$2.5$-$1.5$B]{
        \includegraphics[scale=0.35]{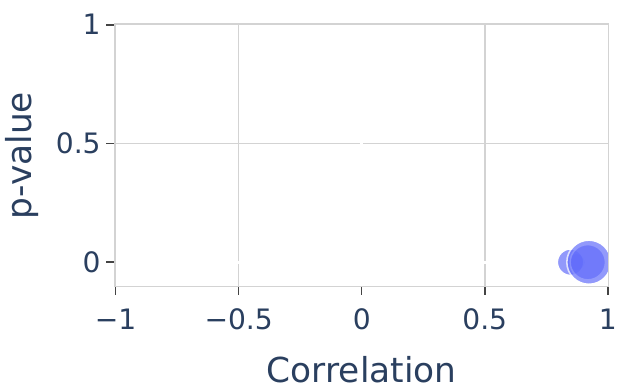}
    }
    \subfloat[Qwen-$2.5$-$7$B]{
        \includegraphics[scale=0.35]{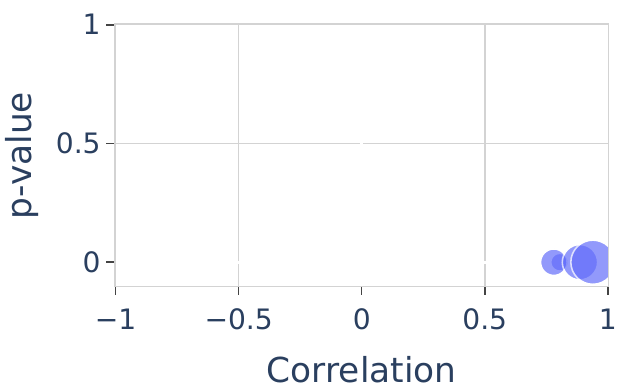}
    }

    \subfloat[Mistral-$7$B]{
        \includegraphics[scale=0.35]{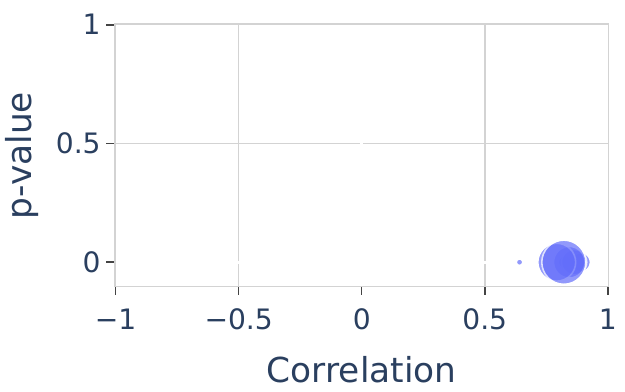}
    }
    \subfloat[Mistral-$12$B]{
        \includegraphics[scale=0.35]{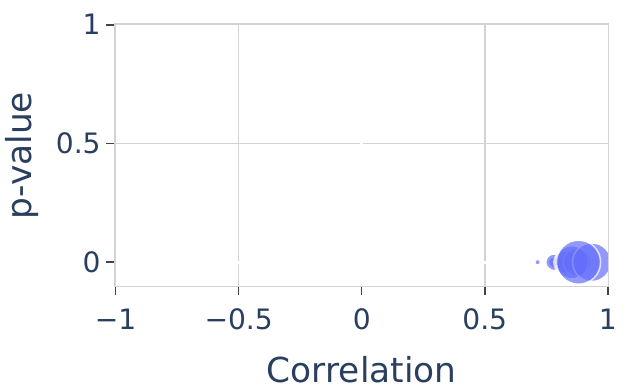}
    }
    \subfloat[Llama-$2$-$7$B]{
        \includegraphics[scale=0.35]{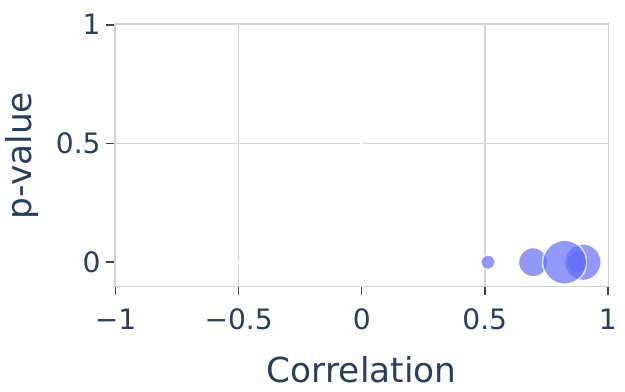}
    }
    \subfloat[Llama-$2$-$13$B]{
        \includegraphics[scale=0.35]{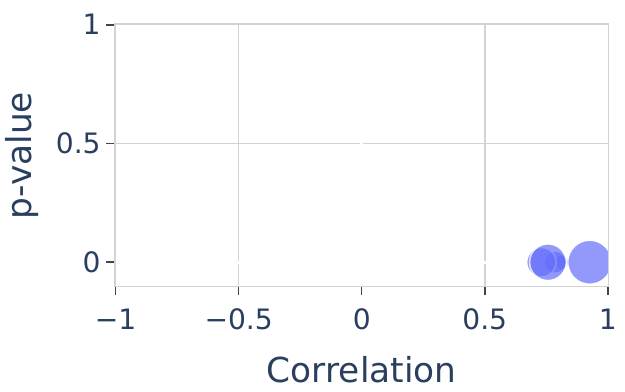}
    }

    \subfloat[Llama-$3.2$-$1$B]{
        \includegraphics[scale=0.35]{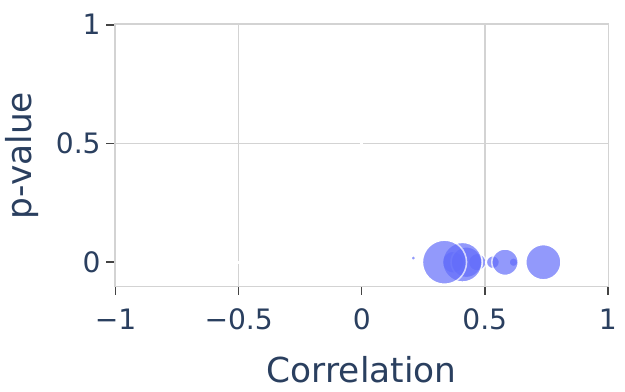}
    }
    \subfloat[Llama-$3.2$-$3$B]{
        \includegraphics[scale=0.35]{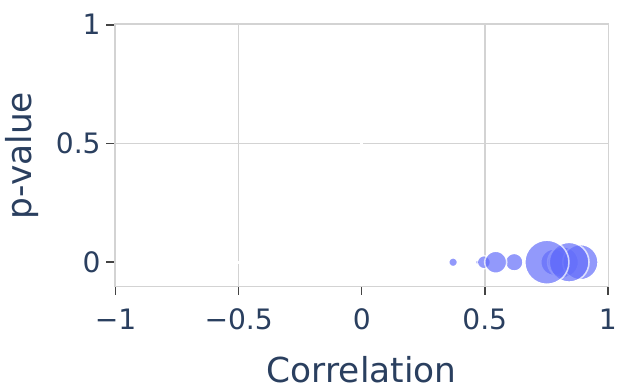}
    }
    \subfloat[Llama-$3.1$-$8$B]{
        \includegraphics[scale=0.35]{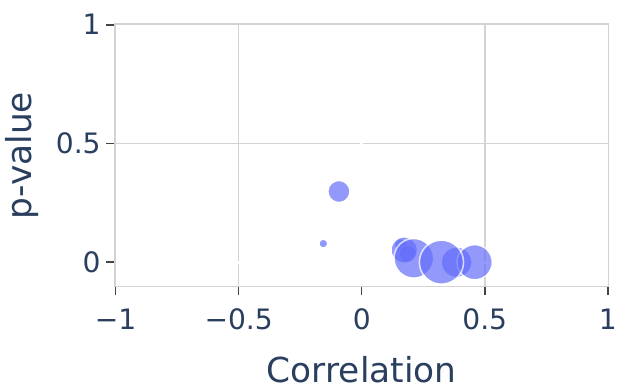}
    }

    \subfloat[Gemma-$2$-$2$B]{
        \includegraphics[scale=0.35]{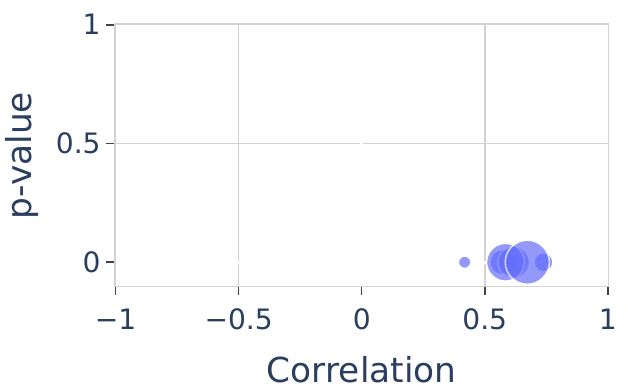}
    }
    \subfloat[Gemma-$2$-$9$B]{
        \includegraphics[scale=0.35]{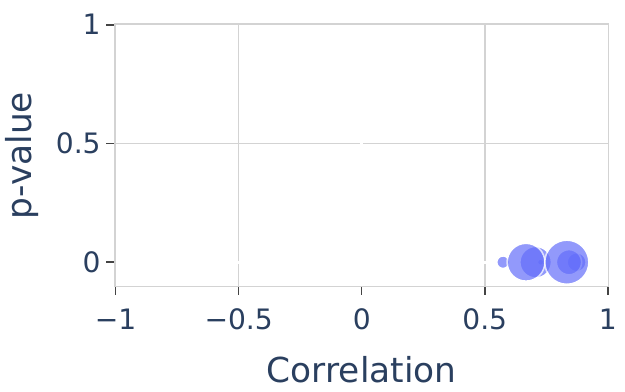}
    }

    \subfloat[Pythia-$1$B]{
        \includegraphics[scale=0.35]{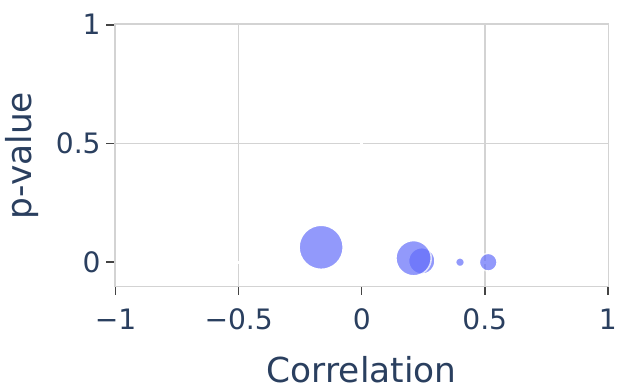}
    }
    \subfloat[Pythia-$2.8$B]{
        \includegraphics[scale=0.35]{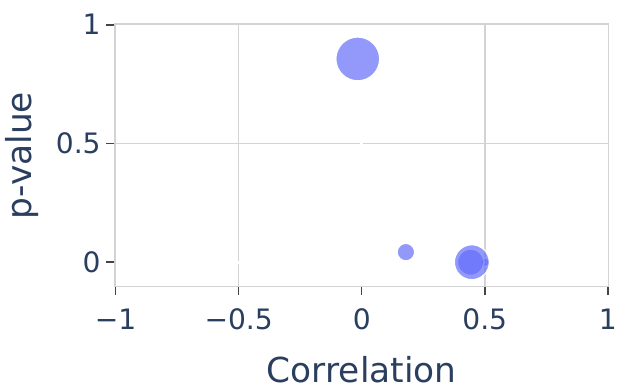}
    }
    \subfloat[Pythia-$6.9$B]{
        \includegraphics[scale=0.35]{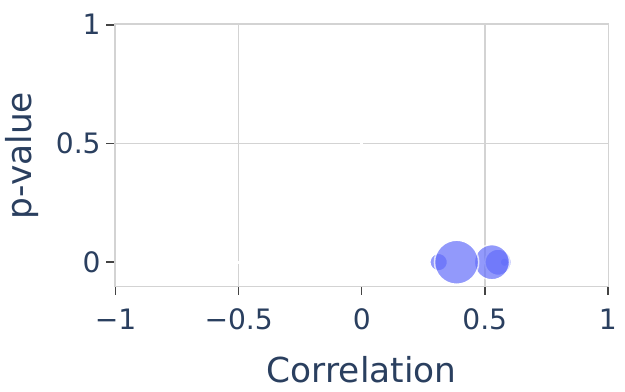}
    }

    \subfloat[Opt-$1.3$B]{
        \includegraphics[scale=0.35]{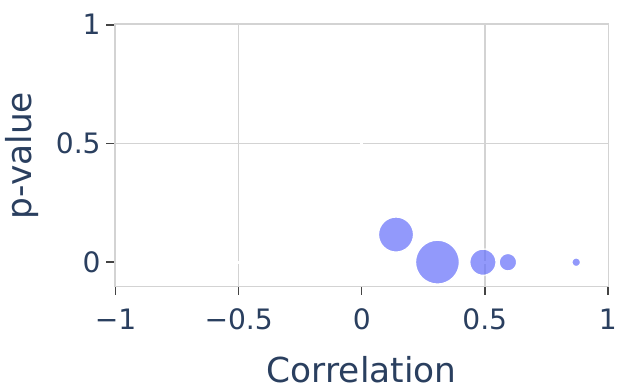}
    }
    \subfloat[Opt-$2.7$B]{
        \includegraphics[scale=0.35]{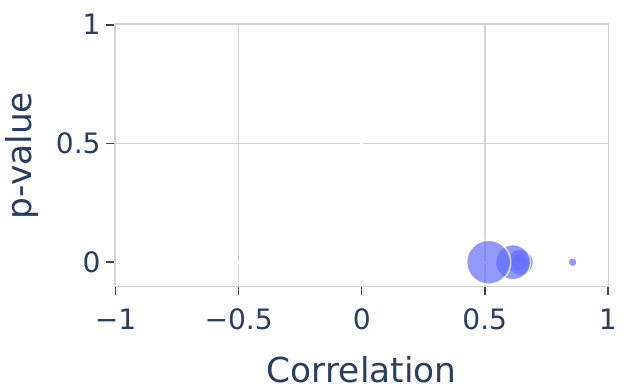}
    }
    \subfloat[Opt-$6.7$B]{
        \includegraphics[scale=0.35]{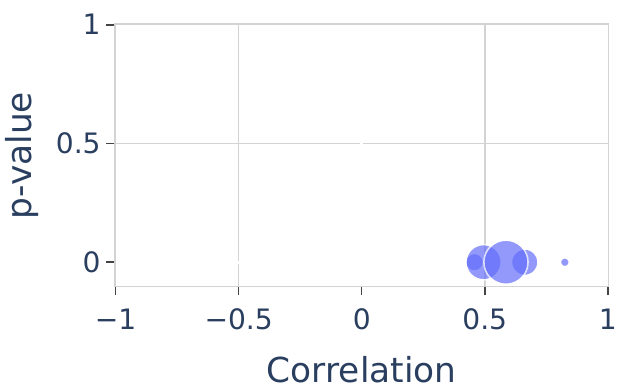}
    }
    
    \caption{Inductive bias of {\icl} and {\ft} on language $ \lang_4 $, computed as the Pearson correlation of generation loss of {\ft} and {\icl} on identical test strings. Correlation, despite being positive, tends to decrease with higher examples (larger markers).}
    \label{fig:inductive_bias_l4}
    
\end{figure*}

\begin{figure*}
    \centering
    \captionsetup[subfigure]{justification=centering}

    \subfloat[Qwen-$2.5$-$0.5$B]{
        \includegraphics[scale=0.35]{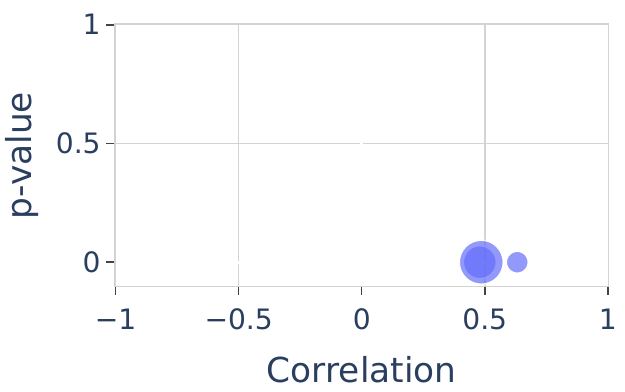}
    }
    \subfloat[Qwen-$2.5$-$1.5$B]{
        \includegraphics[scale=0.35]{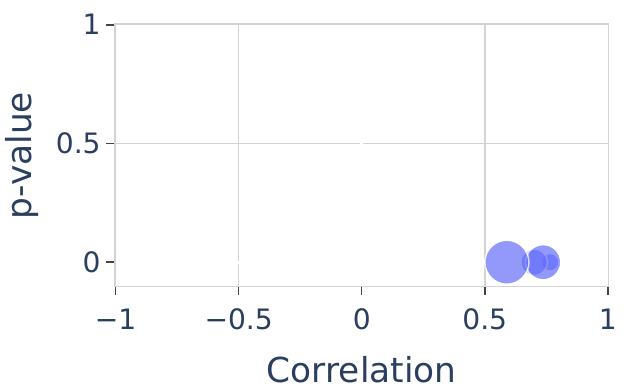}
    }
    \subfloat[Qwen-$2.5$-$7$B]{
        \includegraphics[scale=0.35]{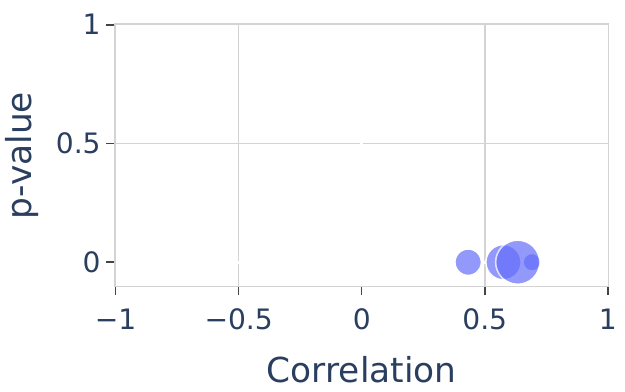}
    }

    \subfloat[Mistral-$7$B]{
        \includegraphics[scale=0.35]{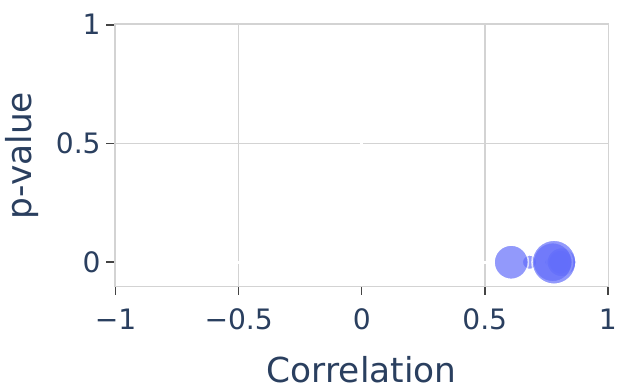}
    }
    \subfloat[Mistral-$12$B]{
        \includegraphics[scale=0.35]{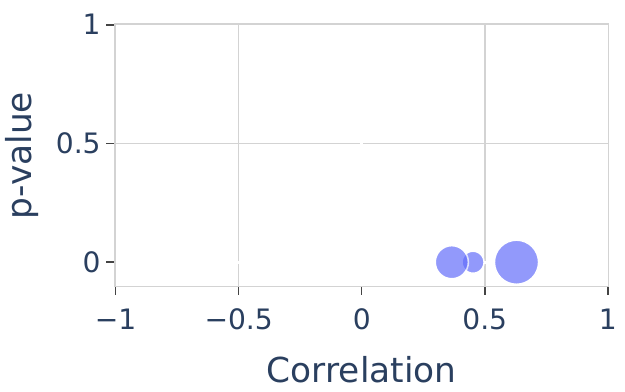}
    }
    \subfloat[Llama-$2$-$7$B]{
        \includegraphics[scale=0.35]{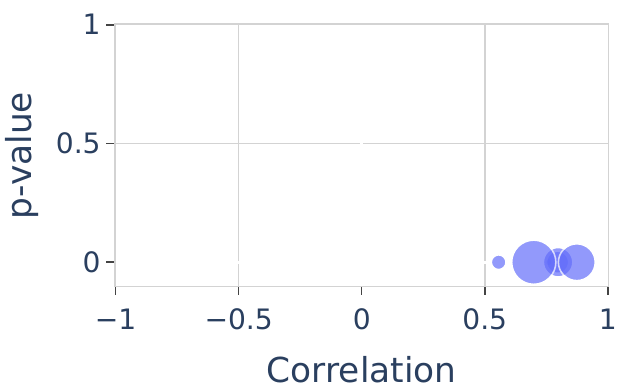}
    }
    \subfloat[Llama-$2$-$13$B]{
        \includegraphics[scale=0.35]{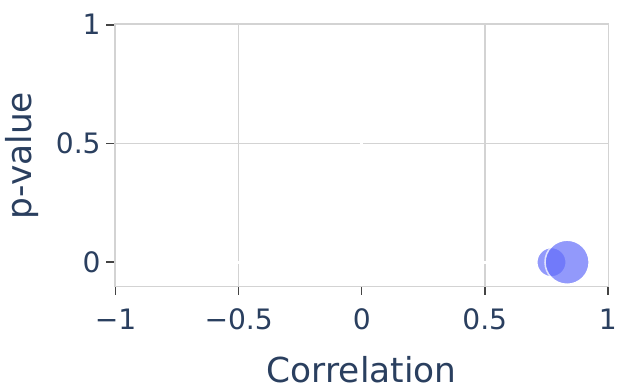}
    }

    \subfloat[Llama-$3.2$-$1$B]{
        \includegraphics[scale=0.35]{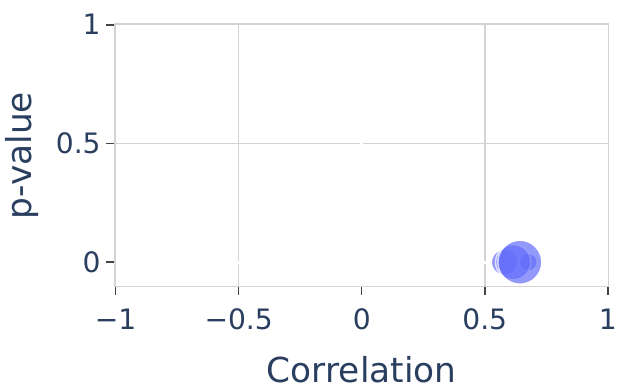}
    }
    \subfloat[Llama-$3.2$-$3$B]{
        \includegraphics[scale=0.35]{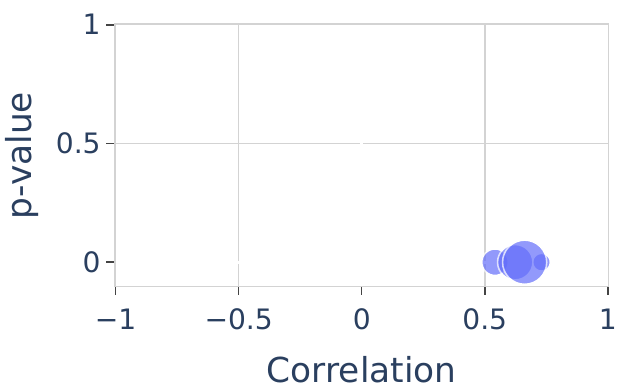}
    }
    \subfloat[Llama-$3.1$-$8$B]{
        \includegraphics[scale=0.35]{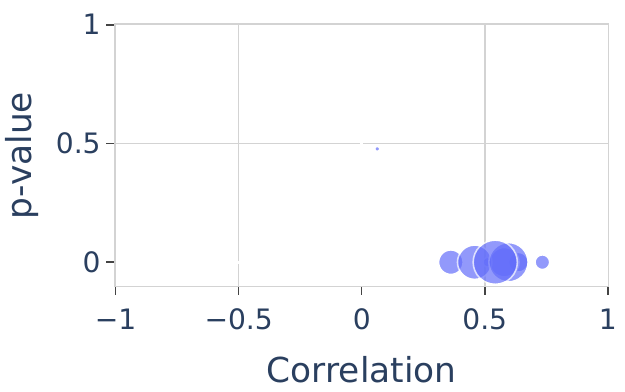}
    }

    \subfloat[Gemma-$2$-$2$B]{
        \includegraphics[scale=0.35]{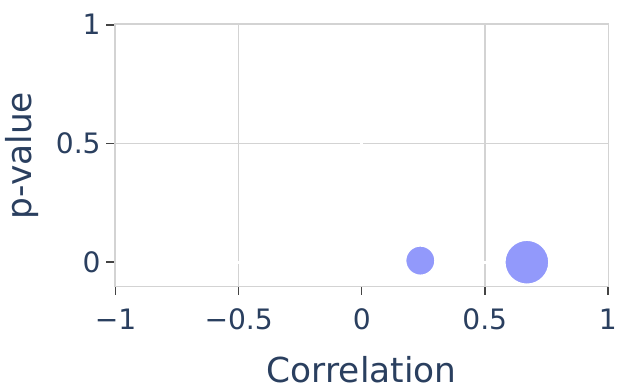}
    }
    \subfloat[Gemma-$2$-$9$B]{
        \includegraphics[scale=0.35]{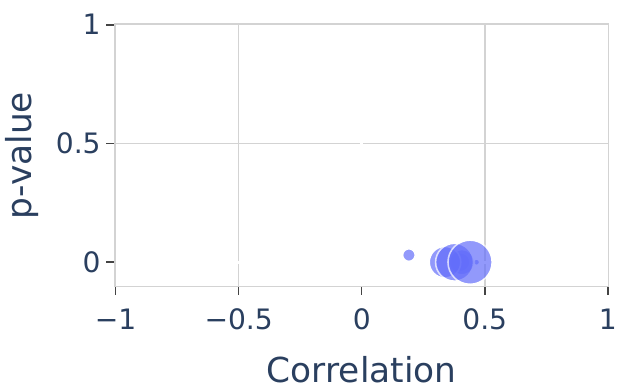}
    }

    \subfloat[Pythia-$1$B]{
        \includegraphics[scale=0.35]{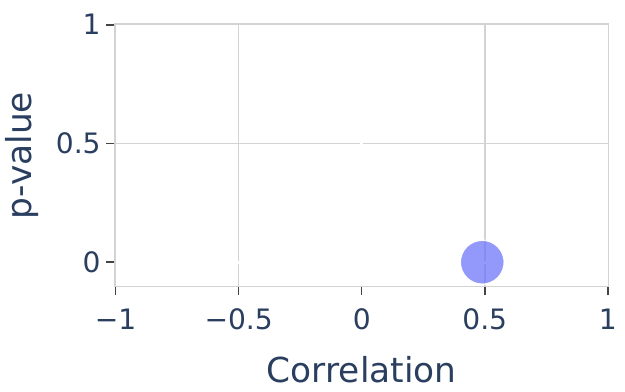}
    }
    \subfloat[Pythia-$2.8$B]{
        \includegraphics[scale=0.35]{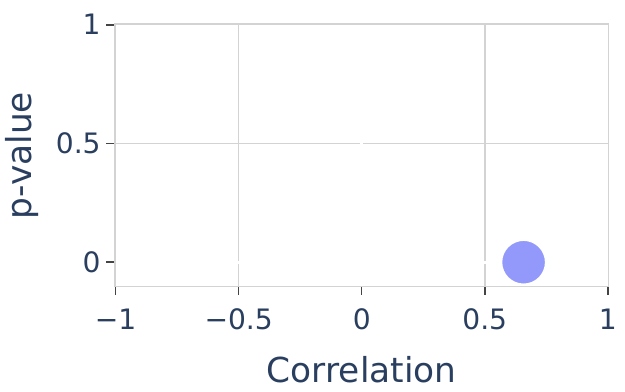}
    }
    \subfloat[Pythia-$6.9$B]{
        \includegraphics[scale=0.35]{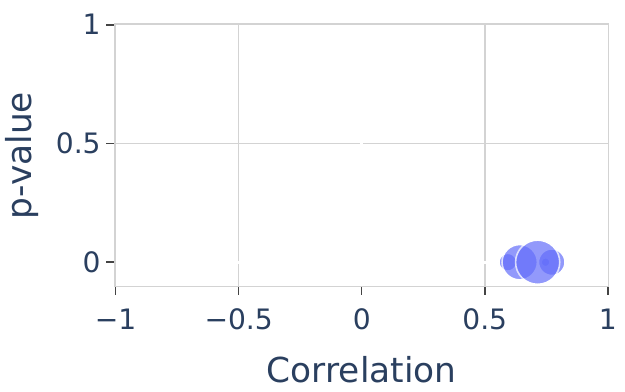}
    }

    \subfloat[Opt-$1.3$B]{
        \includegraphics[scale=0.35]{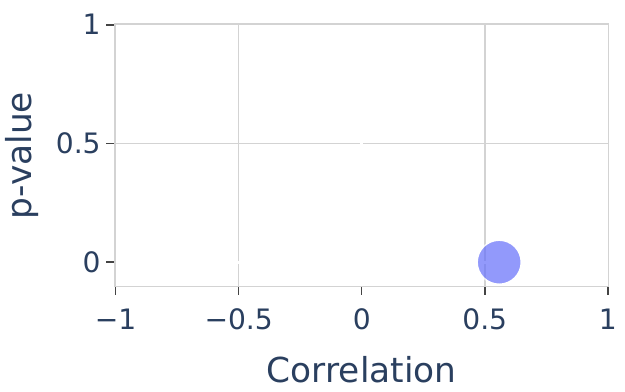}
    }
    \subfloat[Opt-$2.7$B]{
        \includegraphics[scale=0.35]{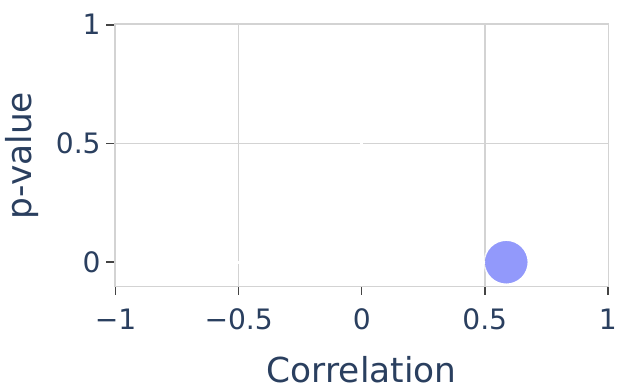}
    }
    \subfloat[Opt-$6.7$B]{
        \includegraphics[scale=0.35]{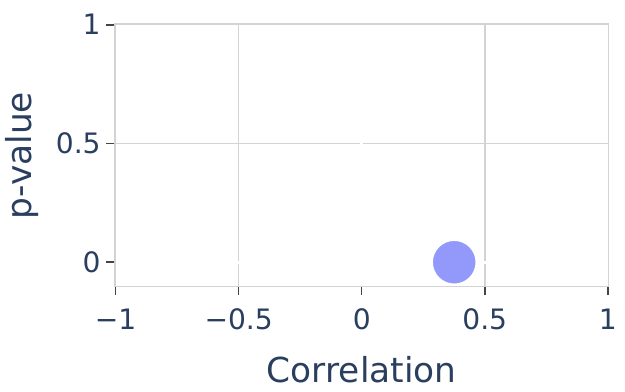}
    }
    
    \caption{Inductive bias of {\icl} and {\ft} on language $ \lang_5 $, computed as the Pearson correlation of generation loss of {\ft} and {\icl} on identical test strings. Correlation, despite being positive, tends to decrease with higher examples (larger markers).}
    \label{fig:inductive_bias_l5}
    
\end{figure*}

\begin{figure*}[!t]
    \captionsetup[subfigure]{justification=centering}
    \centering

    \subfloat[Mistral-$7$B]{
    \includegraphics[scale=0.35]{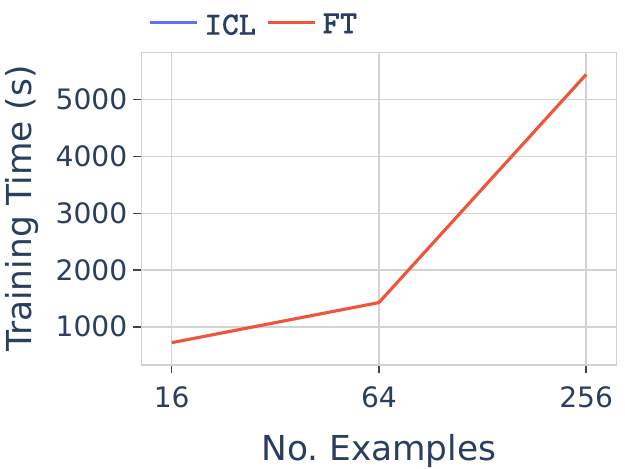}
    }
    \subfloat[Mistral-$7$B]{
    \includegraphics[scale=0.35]{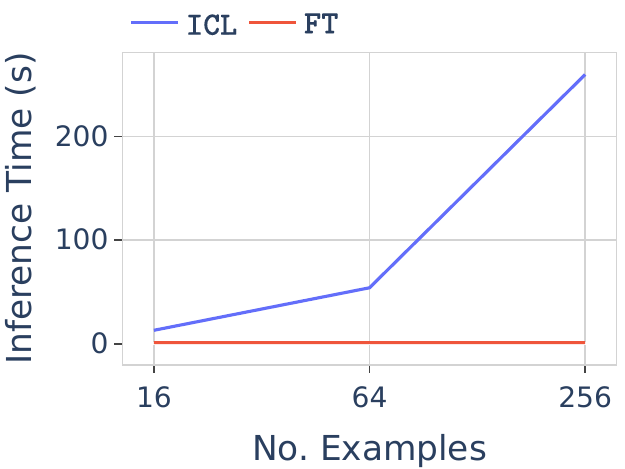}
    }
    \subfloat[Mistral-$7$B]{
    \includegraphics[scale=0.35]{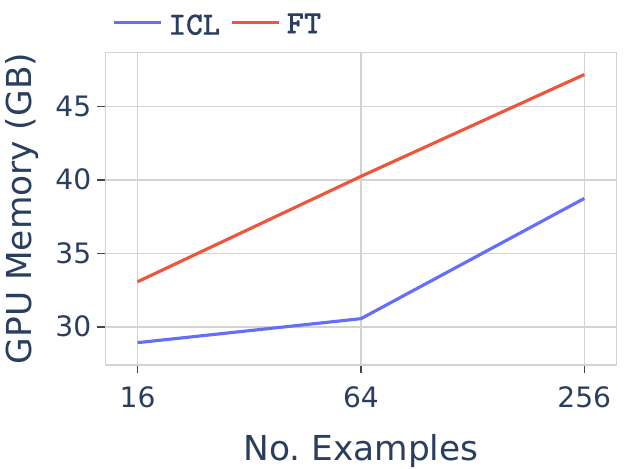}
    }

     \subfloat[Mistral-$12$B]{
    \includegraphics[scale=0.35]{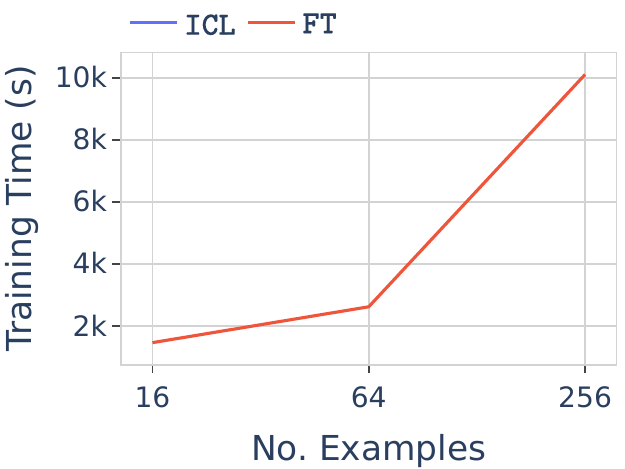}
    }
    \subfloat[Mistral-$12$B]{
    \includegraphics[scale=0.35]{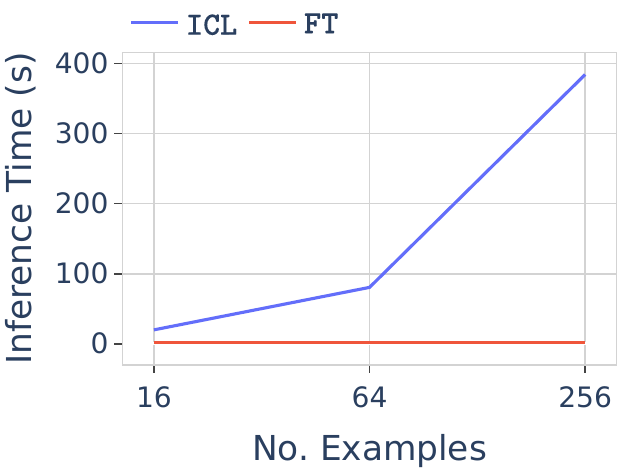}
    }
    \subfloat[Mistral-$12$B]{
    \includegraphics[scale=0.35]{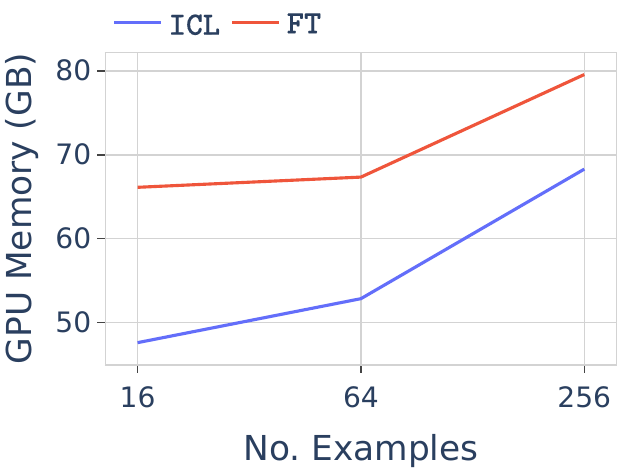}
    }

    \caption{Comparing {\ft} and {\icl} across compute cost, such as training cost (column 1), inference cost (column 2), and memory cost (column 3), recorded for language $ \lang_1 $. The results show that {\ft} and {\icl} are expensive in different phases of computation: {\ft} incurs training cost, which does not apply to {\icl}. In contrast, {\icl} has significantly higher inference cost, despite requiring less memory. Our paper therefore compares both learning modes on equal data budget, using the same number of examples for training and inference.}
    \label{fig:other_dimension}

\end{figure*}

\begin{itemize}
	\item Independent evaluation of {\ft} and {\icl} on different languages across datasets in Figure~\ref{fig:optimal_fine_tuning_all_grammars},~\ref{fig:in_context_learning_all_grammars}.
	
	\item Intra-family {\ft} and {\icl} performance in  Figure~\ref{fig:intra-model_family_fine-tuning},~\ref{fig:intra-model_family_incontext_learning}.
	
	\item Robustness of {\ft} and {\icl} of individual models across languages in Figure~\ref{fig:fine_tuning_vs_incontext_learning_qwen_2.5_7b},~\ref{fig:fine_tuning_vs_incontext_learning_mistral_7b}~\ref{fig:fine_tuning_vs_incontext_learning_llama_2_7b},~\ref{fig:fine_tuning_vs_incontext_learning_llama_3.1_8b}.
	
	\item Inductive bias of LLMs across languages in  Figure~\ref{fig:inductive_bias_l1},~\ref{fig:inductive_bias_l2},~\ref{fig:inductive_bias_l4},~\ref{fig:inductive_bias_l5}.
	
	\item Out-of-distribution generalization on languages of different distances in Figure~\ref{fig:ood_generalization_loss_auc}.

	\item {\ft} vs.\ {\icl} on natural language datasets in Appendix~\ref{app_sec:nlp_dataset}.

	\item Evaluating the utilization of the full context for {\icl} in Appendix~\ref{app_sec:icl_limit}.

	\item Generative vs.\ discriminative tests for determining language proficiency in Appendix~\ref{app_sec:discriminative_test}.

	\item Comparison of different learning modes across compute cost in Figure~\ref{fig:other_dimension}.

\end{itemize}

\begin{figure*}[!t]
    \captionsetup[subfigure]{justification=centering}
    \centering

    \subfloat[{\ft},  Mistral-$7$B]{
        \includegraphics[scale=0.35]{figures/fine-tuning/generalization_finetuning_different_mistral-7b_auc_edit_distance_truncated.pdf
        }
    }
    \subfloat[{\icl},  Mistral-$7$B]{
        \includegraphics[scale=0.35]{figures/incontext_experiments/generalization_incontext_learning_different_mistral-7b_auc_edit_distance_truncated.pdf
        }
    }

    \subfloat[{\ft},  Llama-$2$-$7$B]{
        \includegraphics[scale=0.35]{figures/fine-tuning/generalization_finetuning_different_llama-2-7b_auc_edit_distance_truncated.pdf
        }
    }
    \subfloat[{\icl},  Llama-$2$-$7$B]{
        \includegraphics[scale=0.35]{figures/incontext_experiments/generalization_incontext_learning_different_llama-2-7b_auc_edit_distance_truncated.pdf
        }
    }

    \subfloat[{\ft}, Pythia-$6.9$B]{
        \includegraphics[scale=0.35]{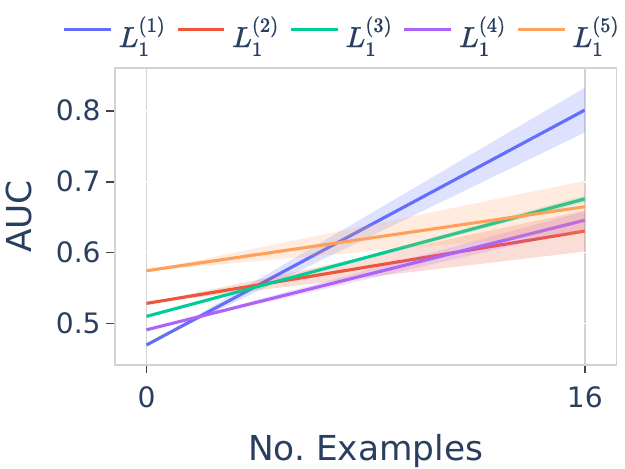
        }
    }
    \subfloat[{\icl},  Pythia-$6.9$B]{
        \includegraphics[scale=0.35]{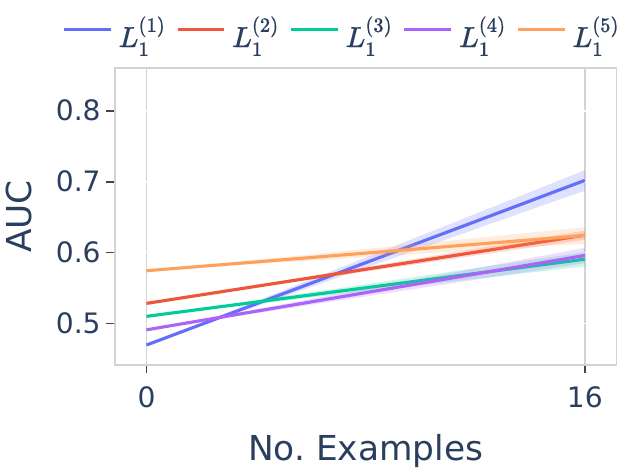
        }
    }

    \subfloat[{\ft},  Gemma-$2$-$2$B]{
        \includegraphics[scale=0.35]{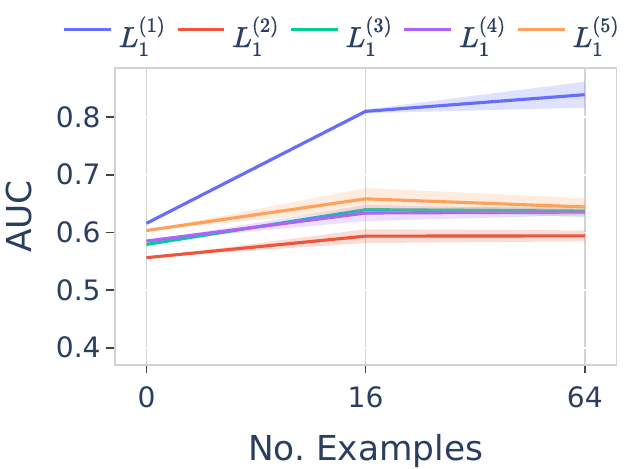
        }
    }
    \subfloat[{\icl},  Gemma-$2$-$2$B]{
        \includegraphics[scale=0.35]{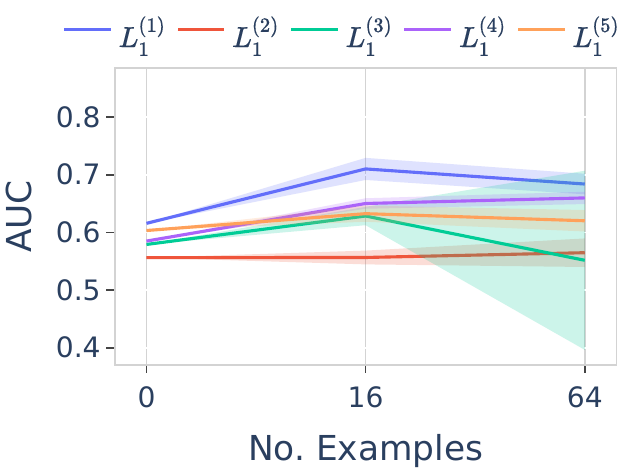
        }
    }

    \subfloat[{\ft},  Llama-$3.2$-$1$B]{
        \includegraphics[scale=0.35]{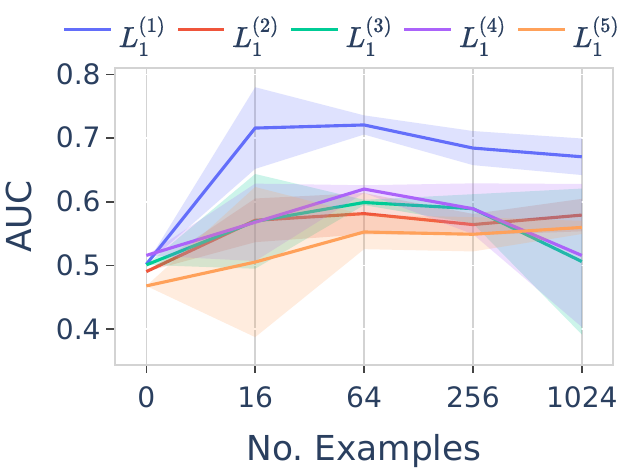
        }
    }
    \subfloat[{\icl},  Llama-$3.2$-$1$B]{
        \includegraphics[scale=0.35]{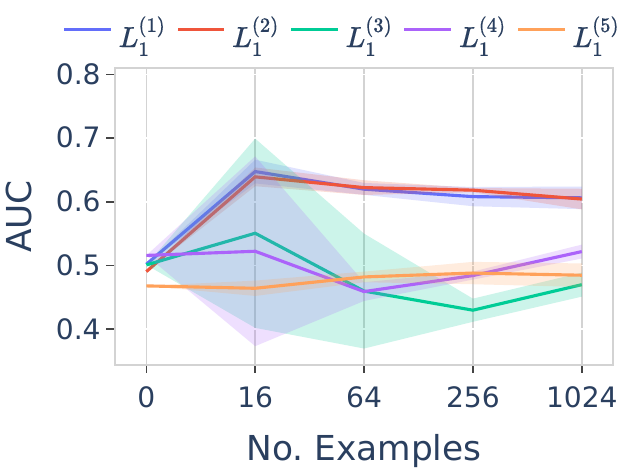
        }
    }

    \caption{Out-of-distribution generalization to languages of increasing distance using {\ft} and {\icl}. We consider $\lang_1$ as the base language. We create languages of higher distance, denoted by $\lang_1^{(\ell)}$, by changing $\ell$ production rules in the {grammar} of $\lang_1$. $\lang_1^{(\ell)}$ contains all changed rules in $\lang_1^{(\ell-1)}$. Hence, $\mathtt{dist}(\lang_1, \lang_1^{(1)}) \leq \cdots \leq \mathtt{dist}(\lang_1, \lang_1^{(5)})$ (see Eq.~\eqref{eq:language_distance})}

    \label{fig:ood_generalization_loss_auc}
\end{figure*}

\cleardoublepage

\begin{table*}
    \caption{Construction of in-language and out-language strings for the MNLI dataset~\cite{N18-1101}, where the out-language string differs from the in-language string only in the sentiment label. The discriminative test is successful, if the generation loss of the correct label in the in-language string is lower than that of the incorrect label in the out-language string. The prompt instruction is shown in the table below.}
    \label{tab:mnli_example}
    
    \centering
    \begin{tabular}{p{0.45\textwidth}p{0.45\textwidth}}
        \toprule
        In-language string & Out-language string (\textcolor{red}{edit} at label) \\
        \midrule

        \textbf{Premise}: One of our number will carry out your instructions minutely.\newline
        \textbf{Hypothesis}: A member of my team will execute your orders with immense precision.\newline
        \textbf{Label}: entailment
        &

        \textbf{Premise}: One of our number will carry out your instructions minutely.\newline
        \textbf{Hypothesis}: A member of my team will execute your orders with immense precision.\newline
        \textbf{Label}: \textcolor{red}{neutral}
        \\

        \midrule

        \textbf{Premise}: Fun for adults and children.\newline
        \textbf{Hypothesis}: Fun for only children.\newline
        \textbf{Label}: contradiction
        &

        \textbf{Premise}: Fun for adults and children.\newline
        \textbf{Hypothesis}: Fun for only children.\newline
        \textbf{Label}: \textcolor{red}{entailment}
        \\

        \bottomrule

    \end{tabular}
    
    \vspace{3em}

    \begin{tabular}{p{0.9\textwidth}}
        \toprule
        Prompt Instruction (beginning of the prompt)\\
        \midrule
        Provide a classification label for the pair, indicating the relationship between the premise and hypothesis:\newline
        - entailment : The hypothesis logically follows from the premise.\newline
        - neutral : The hypothesis is neither entailed nor contradicted by the premise.\newline
        - contradiction : The hypothesis contradicts the premise.\\
        \bottomrule
    \end{tabular}
\end{table*}

\begin{figure*}
    \centering
    \captionsetup[subfigure]{justification=centering}

    \subfloat[Qwen-$2.5$-$7$B]{
        \includegraphics[scale=0.35]{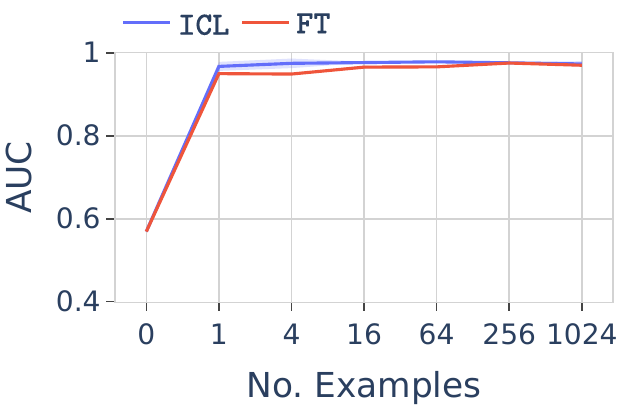}
    }
    \subfloat[Mistral-$7$B]{
        \includegraphics[scale=0.35]{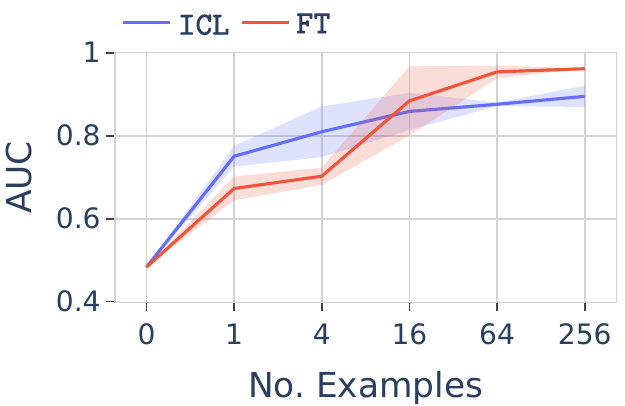}
    }
    \subfloat[Mistral-$ 12 $B]{
        \includegraphics[scale=0.35]{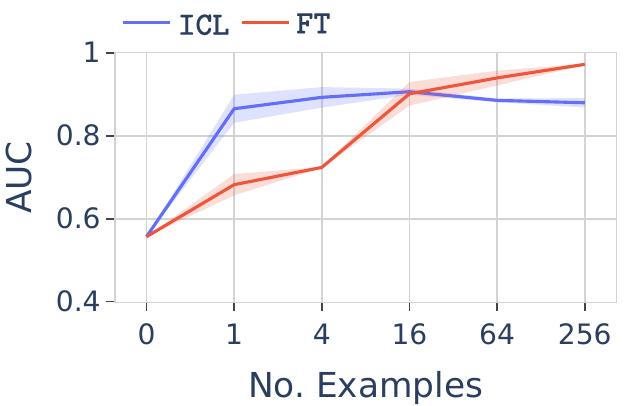}
    }
    \subfloat[Llama-$ 3.1 $-$ 8 $B]{
        \includegraphics[scale=0.35]{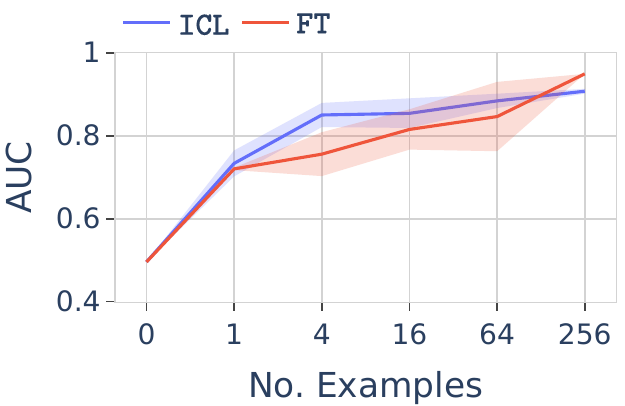}
    }

    \subfloat[Llama-$ 2 $-$ 13 $B]{
        \includegraphics[scale=0.35]{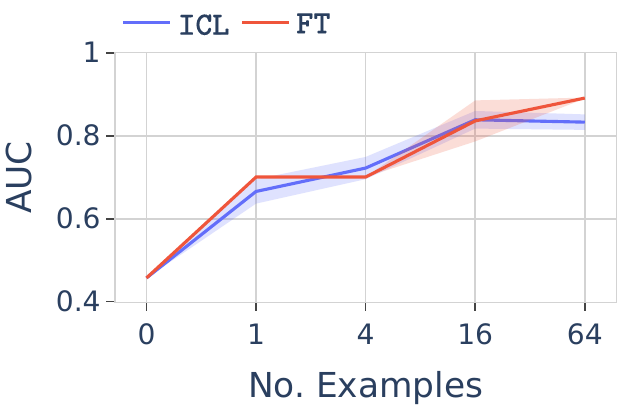}
    }
    \subfloat[Opt-$6.7$B]{
        \includegraphics[scale=0.35]{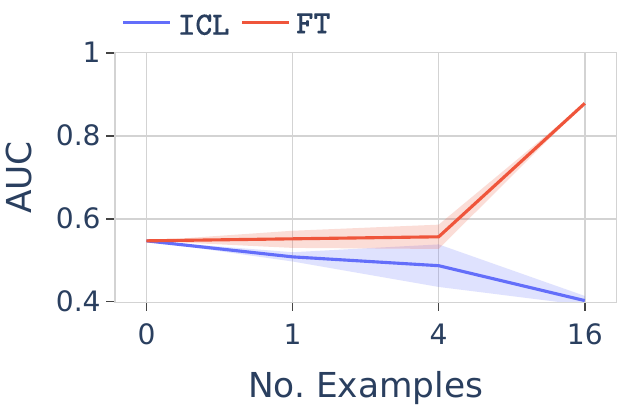}
    }
   
    \caption{In-distribution generalization of {\ft} and {\icl} on the MNLI dataset, where the learning task is to perform natural language inference by generating the sentiment label \{entailment, neutral, contradiction\} given premise and hypothesis. At a high level, {\ft} is better than {\icl} with more examples, consistent with results on formal languages. In a detailed analysis, we observe that different LLMs perform differently given the same problem, indicating the possibility of data contamination in some well-performing LLMs, such as Qwen-$2.5$-$7$B.}
    \label{fig:in_distribution_generalization_mnli}
\end{figure*}

\section{Fine-tuning vs. In-context Learning on Natural Language Datasets}
\label{app_sec:nlp_dataset}

We conduct a comparison of {\ft} and {\icl} on a natural language dataset to observe whether our findings on formal languages generalize to natural language datasets. We consider a natural language inference task on the MNLI dataset~\citep{N18-1101}, as studied in the related work by~\citet{mosbach2023few}. The learning objective is to generate the sentiment label given the premise and hypothesis (see Table~\ref{tab:mnli_example}).

\paragraph{Issues of Data Contamination.} In Figure~\ref{fig:in_distribution_generalization_mnli},  {\ft} surpasses {\icl} on the MNLI dataset with increasing examples, \textit{which is consistent with our findings on formal languages} in Figure~\ref{fig:fine_tuning_vs_few_shot_direct_comparison}. However, Qwen-$ 2.5 $-$ 7 $B model performs much better than other models in both learning modes, suggesting the \textit{possibility} of data contamination. As evidence, the MNLI dataset was proposed in 2018, which is earlier than the release of Qwen-$ 2.5 $-$ 7 $B model in 2024. Therefore, it is difficult to fairly compare different models or their learning modes on publicly available datasets, if a subset of models is possibly trained on the testing dataset~\cite{dominguez2024training}. This further strengthens our case that \textit{synthetic formal languages should be adopted widely to critically evaluate the performance of LLMs, where the risk of data contamination is minimal}.

\begin{figure*}
    \centering
    \captionsetup[subfigure]{justification=centering}

    \subfloat[Qwen-$2.5$-$7$B,\\In-distribution]{
        \includegraphics[scale=0.35]{figures/incontext_experiments/comparison_incontext_vs_finetuning_qwen-2.5-7b_mnli_dataset_in_distribution_auc_nlp.pdf}
    }
    \subfloat[Qwen-$2.5$-$7$B,\\Out-of-distribution]{
        \includegraphics[scale=0.35]{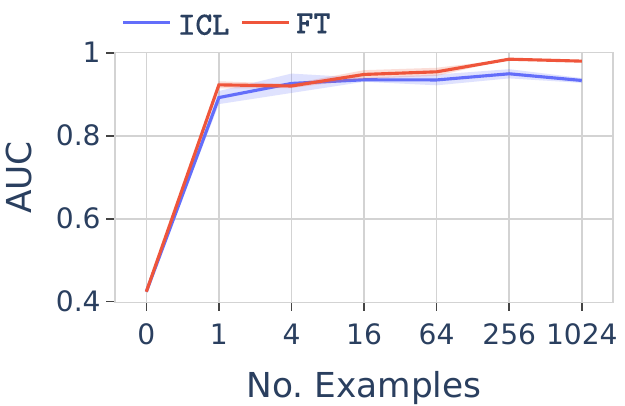}
    }
    \subfloat[Mistral-$7$B,\\In-distribution]{
        \includegraphics[scale=0.35]{figures/incontext_experiments/comparison_incontext_vs_finetuning_mistral-7b_mnli_dataset_in_distribution_auc_nlp.pdf}
    }
    \subfloat[Mistral-$7$B,\\Out-of-distribution]{
        \includegraphics[scale=0.35]{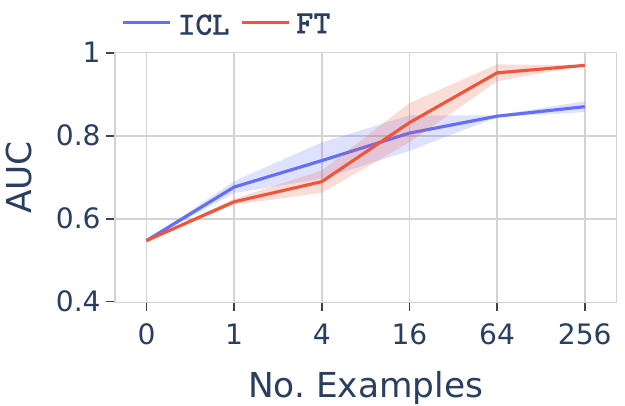}
    }

    \subfloat[Mistral-$ 12 $B,\\In-distribution]{
        \includegraphics[scale=0.35]{figures/incontext_experiments/comparison_incontext_vs_finetuning_mistral-12b_mnli_dataset_in_distribution_auc_nlp.pdf}
    }
    \subfloat[Mistral-$ 12 $B,\\Out-of-distribution]{
        \includegraphics[scale=0.35]{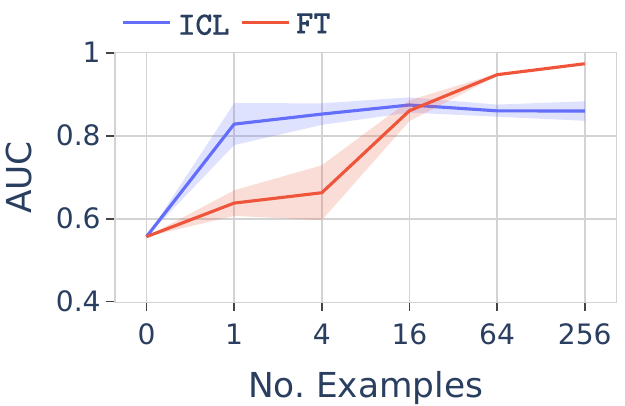}
    }
    \subfloat[Llama-$ 3.1 $-$ 8 $B,\\In-distribution]{
        \includegraphics[scale=0.35]{figures/incontext_experiments/comparison_incontext_vs_finetuning_llama-3.1-8b_mnli_dataset_in_distribution_auc_nlp.pdf}
    }
    \subfloat[Llama-$ 3.1 $-$ 8 $B,\\Out-of-distribution]{
        \includegraphics[scale=0.35]{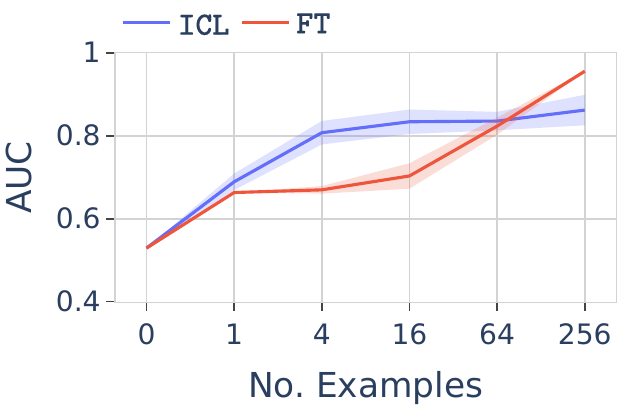}
    }

    \subfloat[Llama-$ 2 $-$ 13 $B,\\In-distribution]{
        \includegraphics[scale=0.35]{figures/incontext_experiments/comparison_incontext_vs_finetuning_llama-2-13b_mnli_dataset_in_distribution_auc_nlp.pdf}
    }
    \subfloat[Llama-$ 2 $-$ 13 $B,\\Out-of-distribution]{
        \includegraphics[scale=0.35]{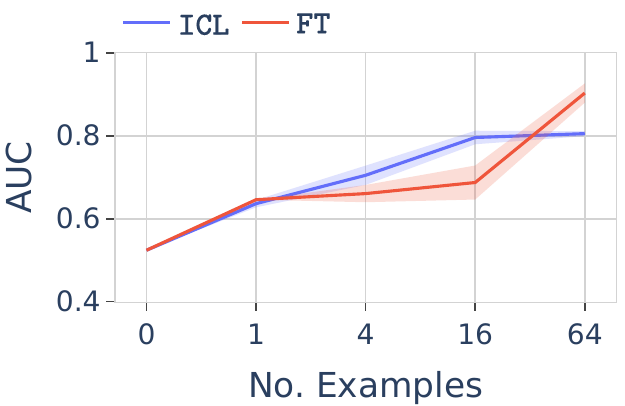}
    }
    \subfloat[Opt-$6.7$B,\\In-distribution]{
        \includegraphics[scale=0.35]{figures/incontext_experiments/comparison_incontext_vs_finetuning_opt-6.7b_mnli_dataset_in_distribution_auc_nlp.pdf}
    }
    \subfloat[Opt-$6.7$B,\\Out-of-distribution]{
        \includegraphics[scale=0.35]{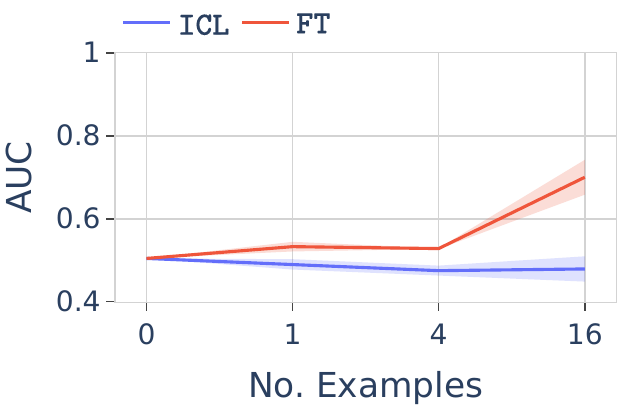}
    }

    \caption{MNLI dataset: In-distribution (inference within the same genre, Column $ 1 $ and $ 3 $) vs.\ out-of-distribution (inference across genres, Column $ 2 $ and $ 4 $) generalization performance of {\ft} and {\icl}, where there is no substantial difference across tasks. This is a fundamental problem in natural language datasets, where the identification of tasks can be ambiguous, and LLMs may not distinguish them. Overall, {\ft} is better than {\icl}, which contradicts our results on formal languages where {\ft} is better than {\icl} in in-distribution generalization only, but both learning modes perform equally well in out-of-distribution generalization.}
    \label{fig:in_out_distribution_generalization_mnli}
\end{figure*}

\paragraph{Difficulty in Identifying In-distribution vs.\ Out-of-distribution Tasks.} In Figure~\ref{fig:in_out_distribution_generalization_mnli}, we demonstrate in-distribution and out-of-distribution performance side-by-side on the MNLI dataset for both learning modes. The differentiation of tasks is determined by the genre of (premise, hypothesis) pairs. If the genre of the testing pair matches with training pairs, then the task is in-distribution. Otherwise, the task is out-of-distribution. However, we do not observe any difference in the comparison of {\ft} vs.\ {\icl} based on tasks -- {\ft} is better than {\icl} in both tasks. This contradicts our findings in formal languages in Figure~\ref{fig:ood_generalization}, where the distance between tasks is well-defined, and  both {\ft} and {\icl} perform equally well in the out-of-distribution task. This experiment highlights the ambiguity of specifying learning tasks in natural language datasets, the core theme in desideratum \textbf{D1} in Section~\ref{sec:intro}. \textit{Therefore, for an objective comparison, it is important to carefully define in-distribution and out-of-distribution tasks, which is easier in formal languages than natural languages.}

\cleardoublepage
\section{Testing the Limit of In-context Learning}
\label{app_sec:icl_limit}

To find the limit of {\icl} ability of an LLM, we rely on the convergence of training and test loss in {\icl} as examples are added. Intuitively, training loss provides a practical \textit{lower bound} on test loss in {\icl}. An LLM can no longer improve in {\icl} when both losses converge. To obtain training loss, we first provide {\icl} examples from the training set and later compute the loss of generating each training example already present in the context.

\begin{figure*}[!t]
    \centering
    
    \subfloat[Qwen-$ 2.5 $-$ 0.5$B]{
        \includegraphics[scale=0.35]{
            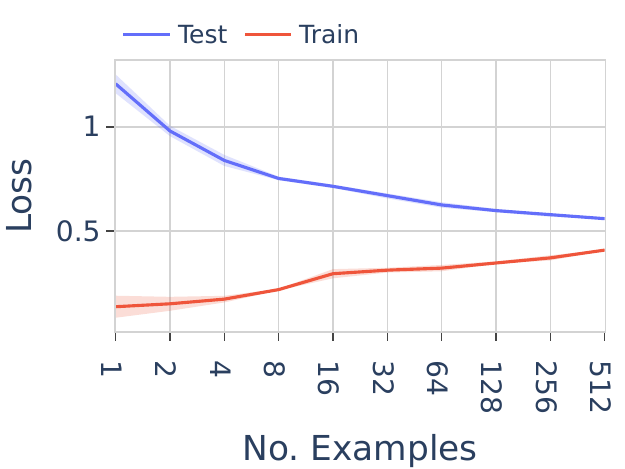
        }
    }
    \subfloat[Qwen-$ 2.5 $-$ 1.5$B]{
        \includegraphics[scale=0.35]{
            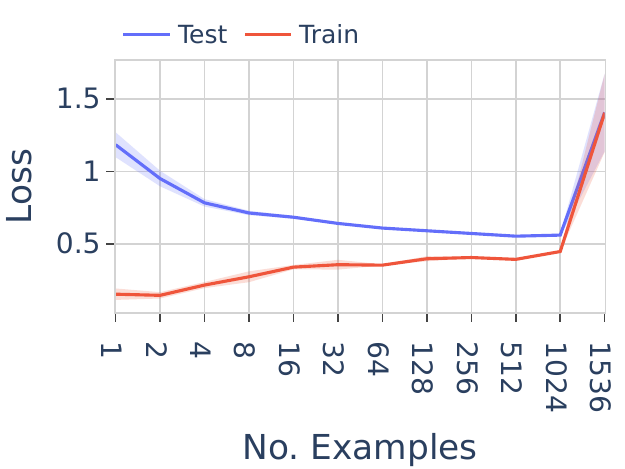
        }
    }
    \subfloat[Qwen-$ 2.5 $-$ 7$B]{
        \includegraphics[scale=0.35]{
            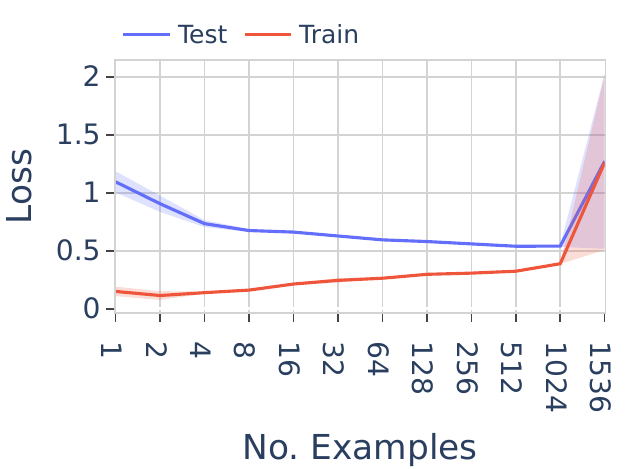
        }
    }
    
    \subfloat[Mistral-$7$B]{
        \includegraphics[scale=0.35]{
            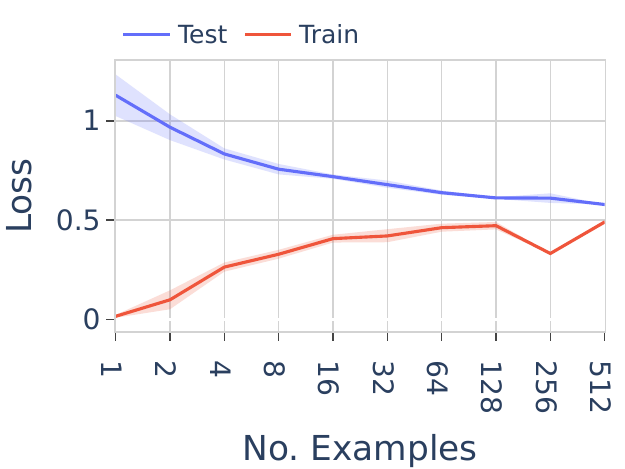
        }
    }
    \subfloat[Mistral-$12$B]{
        \includegraphics[scale=0.35]{
            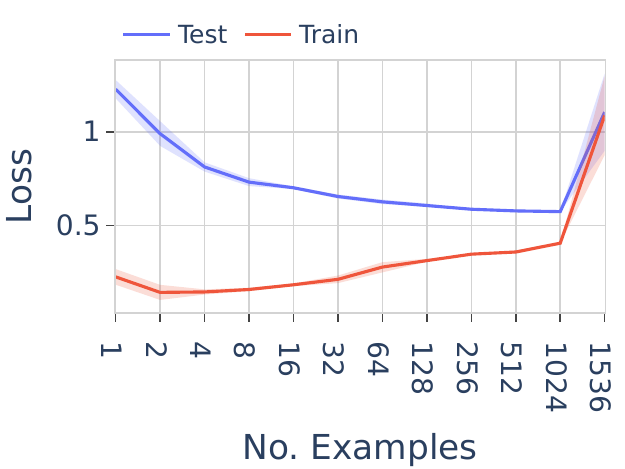
        }
    }
    \subfloat[Llama-$2$-$7$B]{
        \includegraphics[scale=0.35]{
            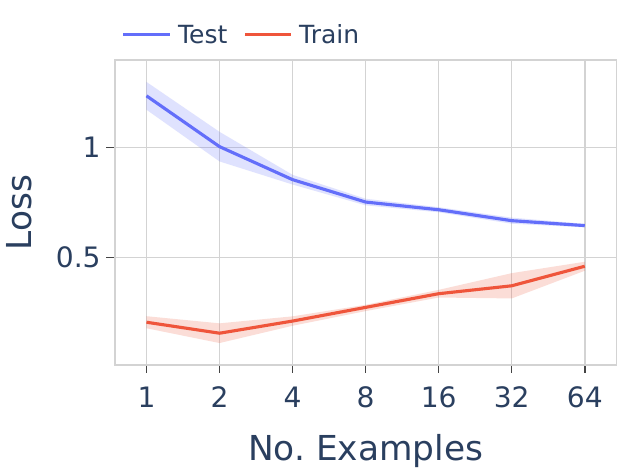
        }
    }
    \subfloat[Llama-$2$-$13$B]{
        \includegraphics[scale=0.35]{
            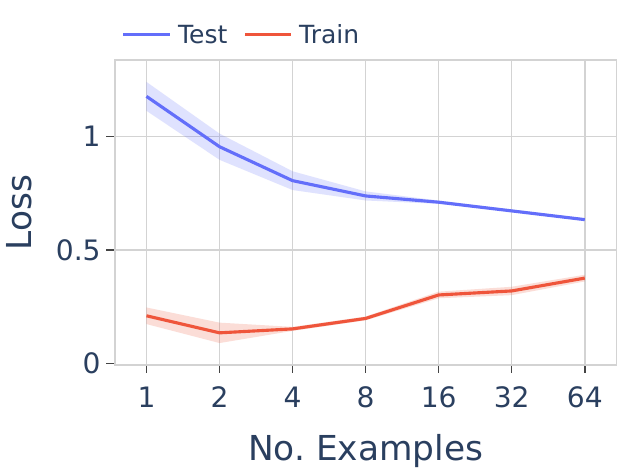
        }
    }

    \subfloat[Llama-$3.2$-$1$B]{
        \includegraphics[scale=0.35]{
            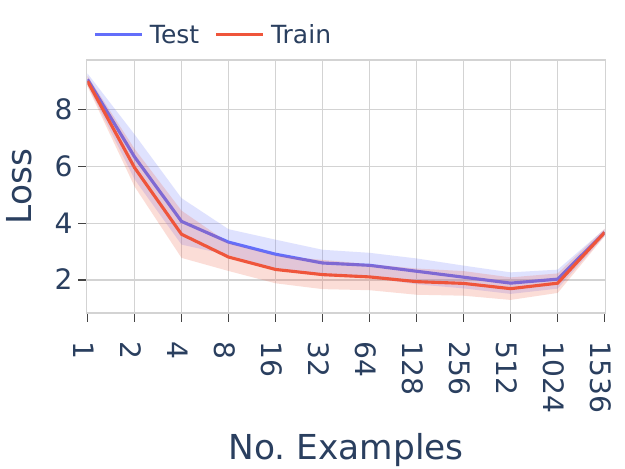
        }
    }
    \subfloat[Llama-$3.2$-$3$B]{
        \includegraphics[scale=0.35]{
            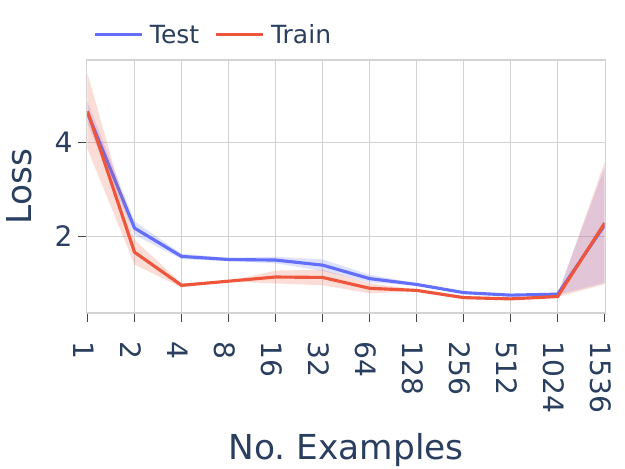
        }
    }
    \subfloat[Llama-$3.1$-$8$B]{
        \includegraphics[scale=0.35]{
            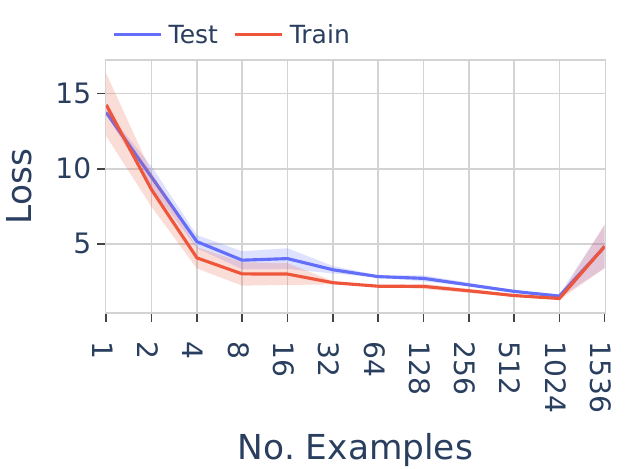
        }
    }

    \subfloat[Gemma-$2$-$2$B]{
        \includegraphics[scale=0.35]{
            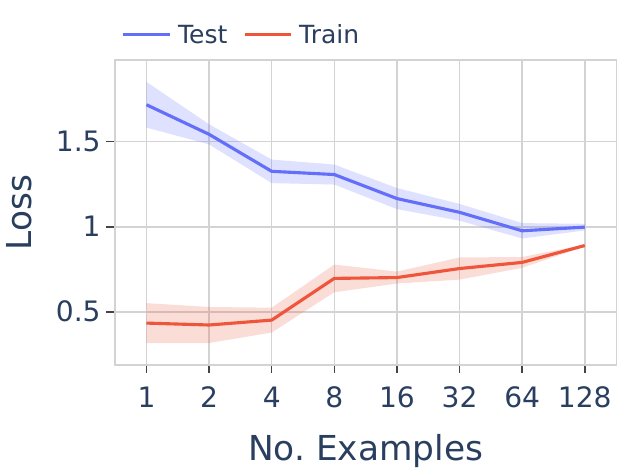
        }
    }\subfloat[Gemma-$2$-$9$B]{
        \includegraphics[scale=0.35]{
            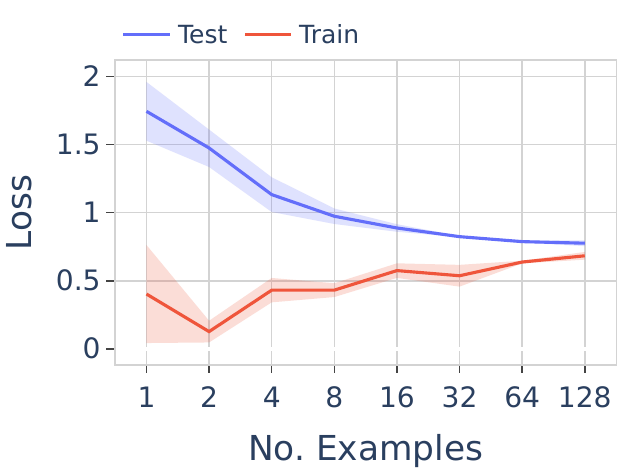
        }
    }

    \subfloat[Pythia-$1$B]{
        \includegraphics[scale=0.35]{
            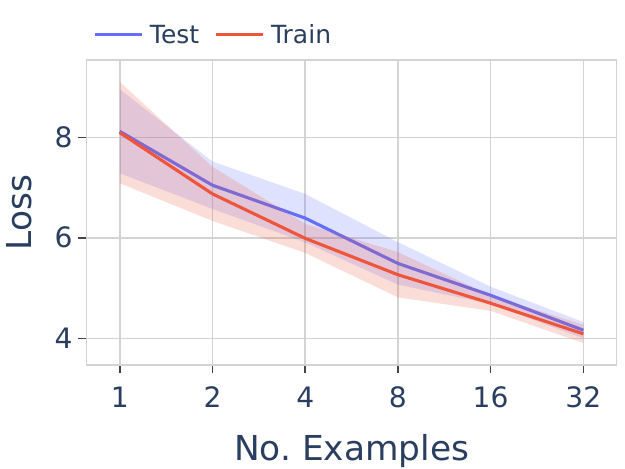
        }
    }
    \subfloat[Pythia-$2.8$B]{
        \includegraphics[scale=0.35]{
            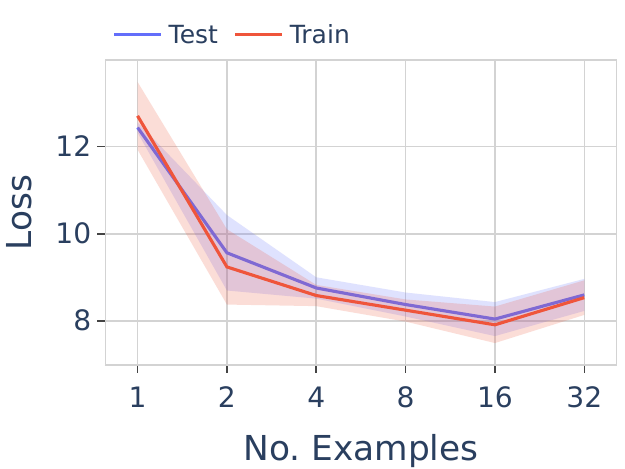
        }
    }
    \subfloat[Pythia-$6.9$B]{
        \includegraphics[scale=0.35]{
            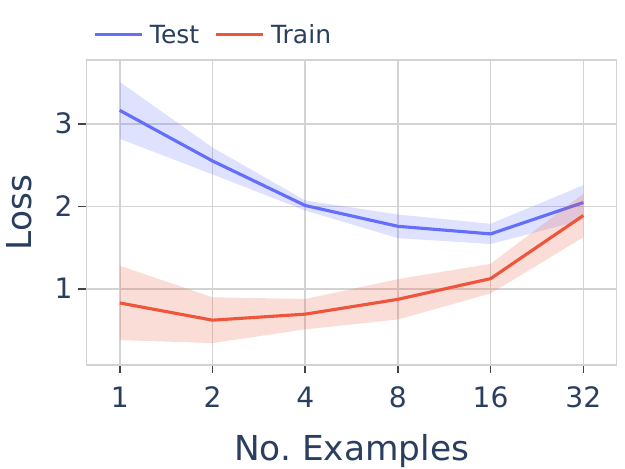
        }
    }

    \subfloat[Opt-$1.3$B]{
        \includegraphics[scale=0.35]{
            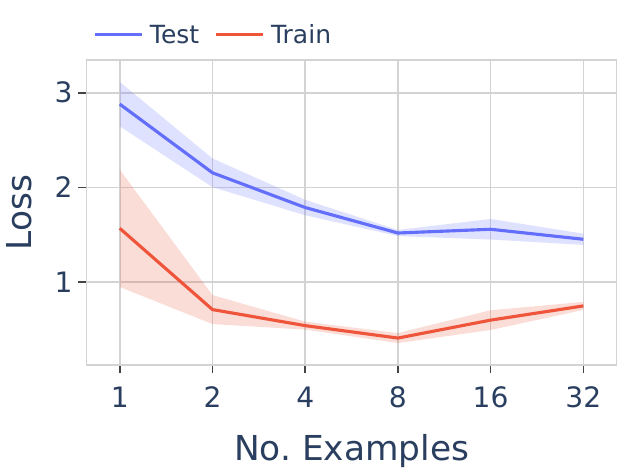
        }
    }
    \subfloat[Opt-$2.7$B]{
        \includegraphics[scale=0.35]{
            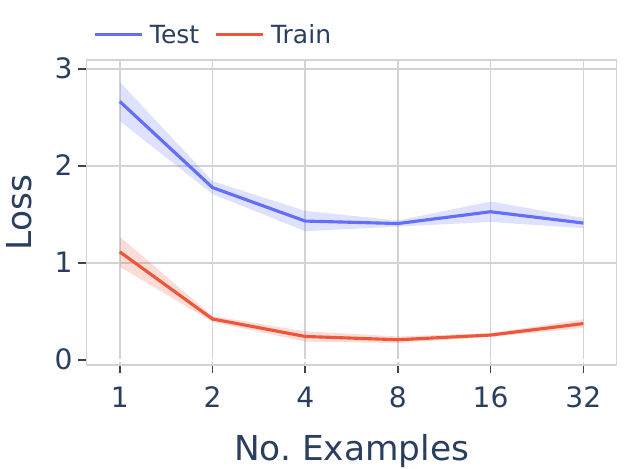
        }
    }
    \subfloat[Opt-$6.7$B]{
        \includegraphics[scale=0.35]{
            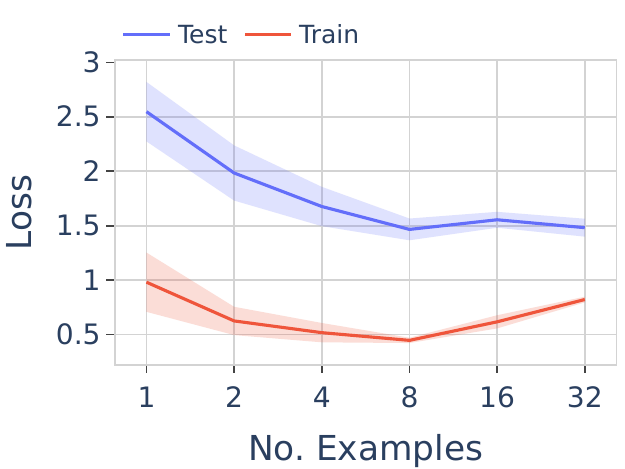
        }
    }

    \caption{Testing the limit of utilizing {\icl} context ($ 1536 $ examples $ \approx $ $ 77 $K tokens) on language $ \lang_1 $. Training loss provides a lower bound of test loss in {\icl}. Long context LLMs cannot further improve from additional examples.
    }
    \label{fig:utilizing_long_context_lang_1}

\end{figure*}

Empirically, across all languages, test loss converges to train loss, i.e.,  \textit{{\icl} limit is reached} in the majority of LLMs, except in the Llama-2 and Opt families. These two families have limited context ($ 4 $K and $ 2 $K tokens, respectively), and there is a gap between losses even upon exhausting their context length. Moreover, long-context LLMs, such as Qwen-$ 2.5 $-$ 7 $B and Llama-$ 3.1 $-$ 8 $B with $ 128 $K context length, cannot further improve from additional examples as both losses converge and later increase near the limit (see Figure~\ref{fig:utilizing_long_context_lang_1}). \textit{Therefore, formal language learning enables us to categorize LLMs into two: (a) LLMs that cannot reach the {\icl} limit, and (b) LLMs that reach their {\icl} limit and do not improve with additional examples towards their context length limit.}

\begin{figure*}[!t]
    \centering
    
    \subfloat[Qwen-$ 2.5 $-$ 0.5$B]{
        \includegraphics[scale=0.35]{
            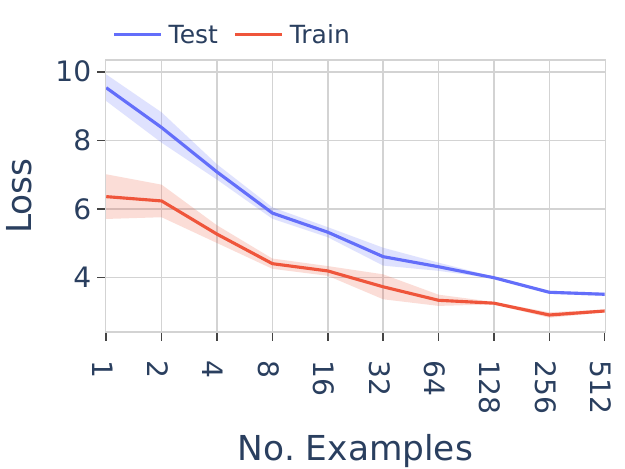
        }
    }
    \subfloat[Qwen-$ 2.5 $-$ 1.5$B]{
        \includegraphics[scale=0.35]{
            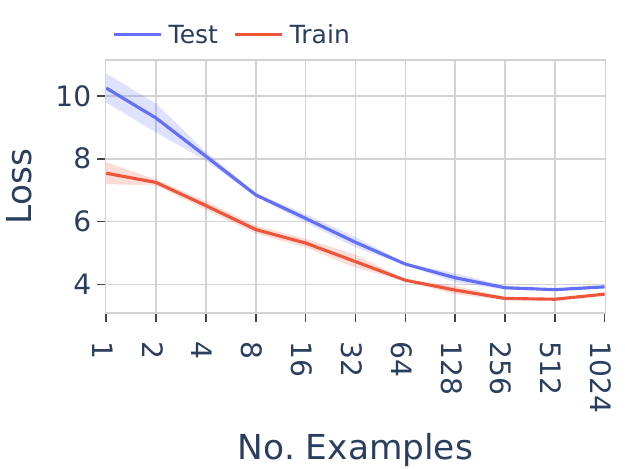
        }
    }
    \subfloat[Qwen-$ 2.5 $-$ 7$B]{
        \includegraphics[scale=0.35]{
            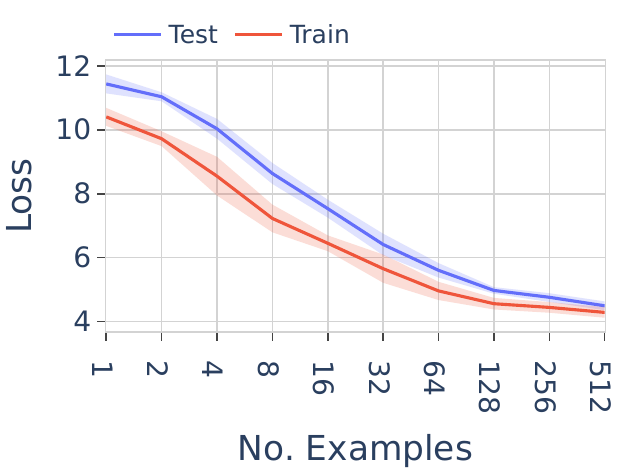
        }
    }

    \subfloat[Mistral-$7$B]{
        \includegraphics[scale=0.35]{
            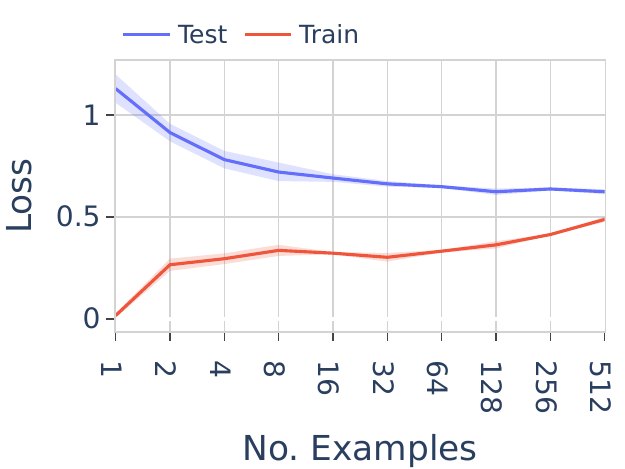
        }
    }
    \subfloat[Mistral-$12$B]{
        \includegraphics[scale=0.35]{
            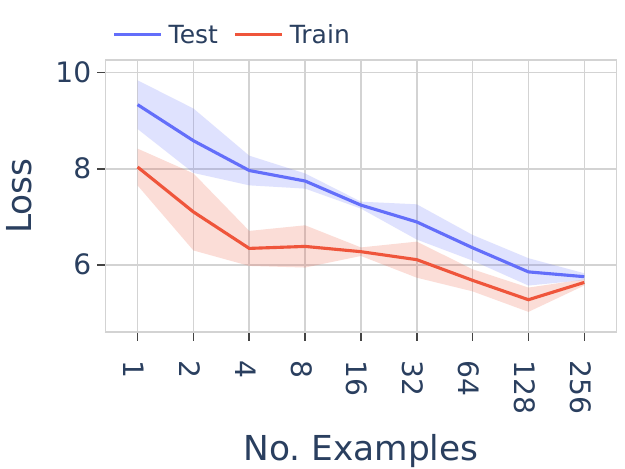
        }
    }
    \subfloat[Llama-$2$-$7$B]{
        \includegraphics[scale=0.35]{
            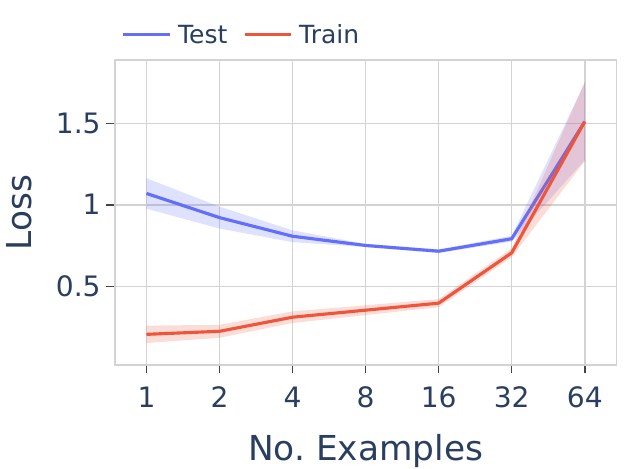
        }
    }
    \subfloat[Llama-$2$-$13$B]{
        \includegraphics[scale=0.35]{
            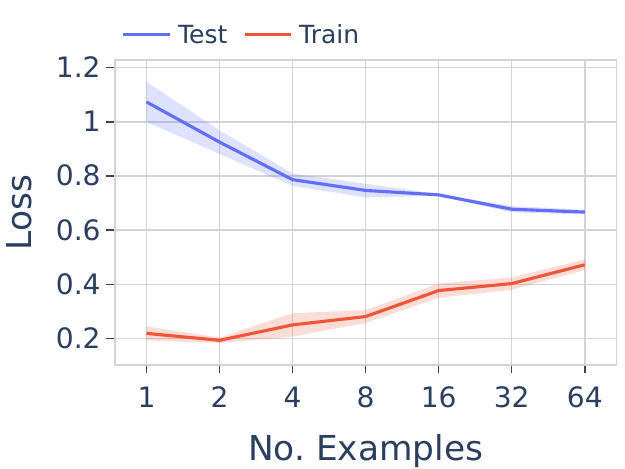
        }
    } 
    
    \subfloat[Llama-$3.2$-$1$B]{
        \includegraphics[scale=0.35]{
            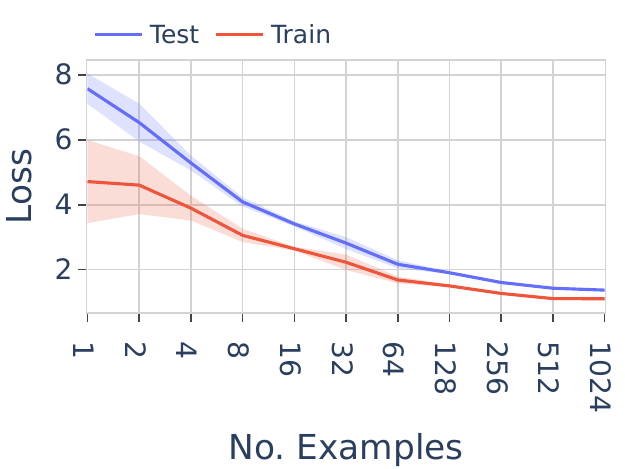
        }
    }
    \subfloat[Llama-$3.2$-$3$B]{
        \includegraphics[scale=0.35]{
            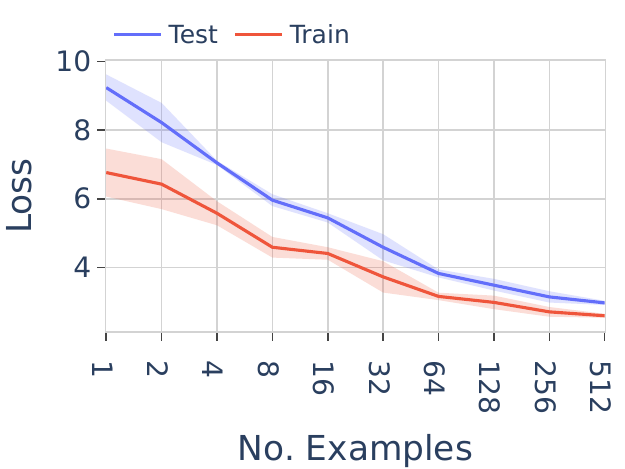
        }
    }
    \subfloat[Llama-$3.1$-$8$B]{
        \includegraphics[scale=0.35]{
            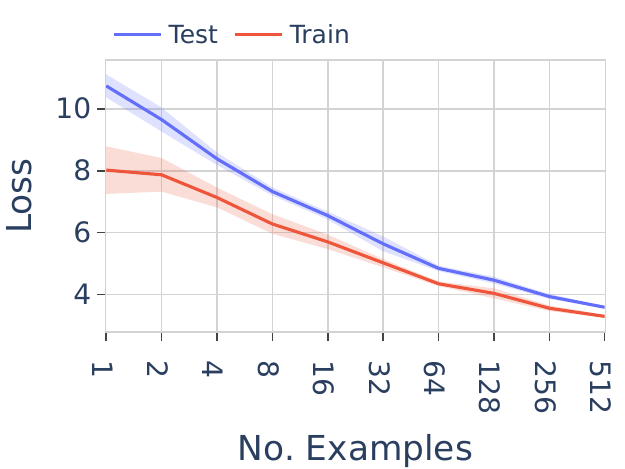
        }
    }

    \subfloat[Gemma-$2$-$2$B]{
        \includegraphics[scale=0.35]{
            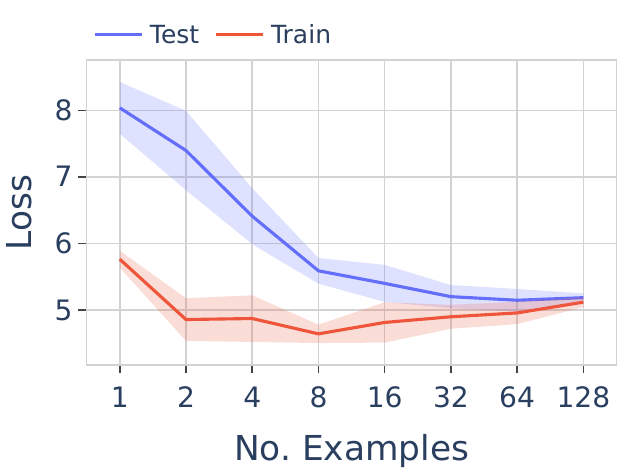
        }
    }\subfloat[Gemma-$2$-$9$B]{
        \includegraphics[scale=0.35]{
            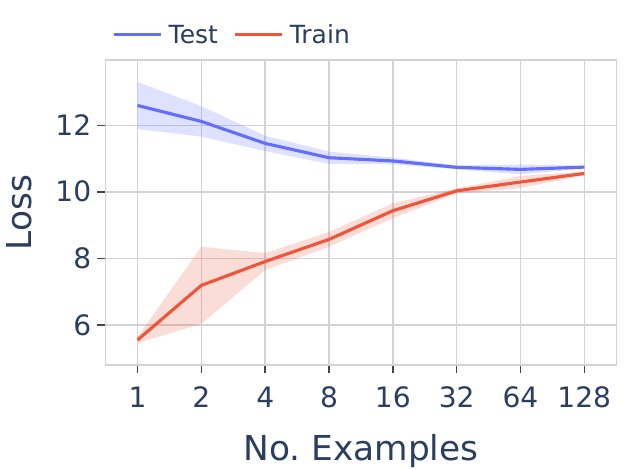
        }
    }

    \subfloat[Pythia-$1$B]{
        \includegraphics[scale=0.35]{
            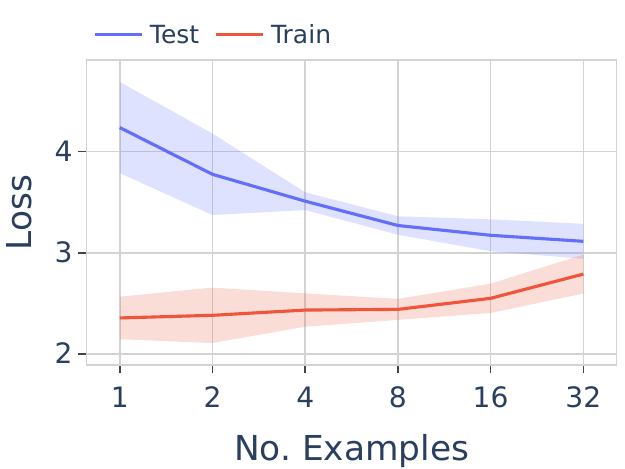
        }
    }
    \subfloat[Pythia-$2.8$B]{
        \includegraphics[scale=0.35]{
            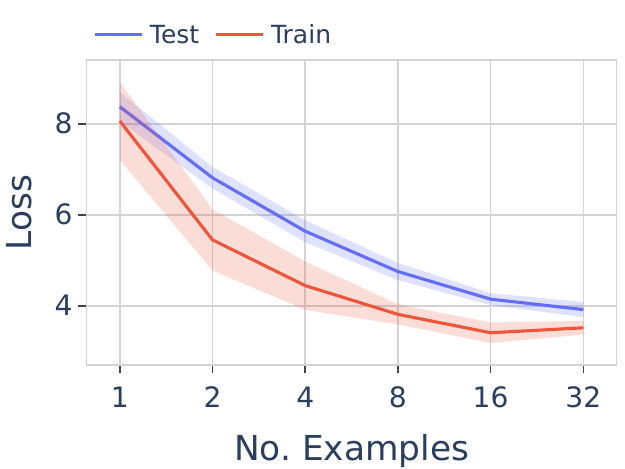
        }
    }
    \subfloat[Pythia-$6.9$B]{
        \includegraphics[scale=0.35]{
            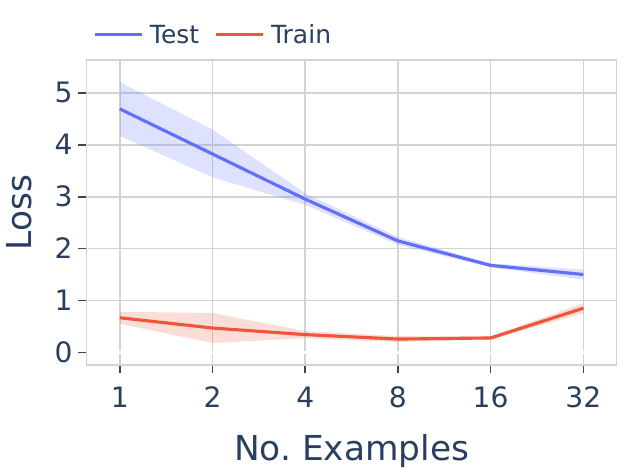
        }
    }
    
    \subfloat[Opt-$1.3$B]{
        \includegraphics[scale=0.35]{
            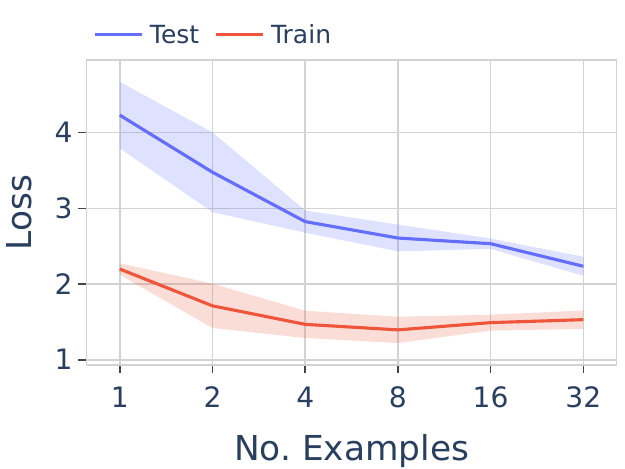
        }
    }
    \subfloat[Opt-$2.7$B]{
        \includegraphics[scale=0.35]{
            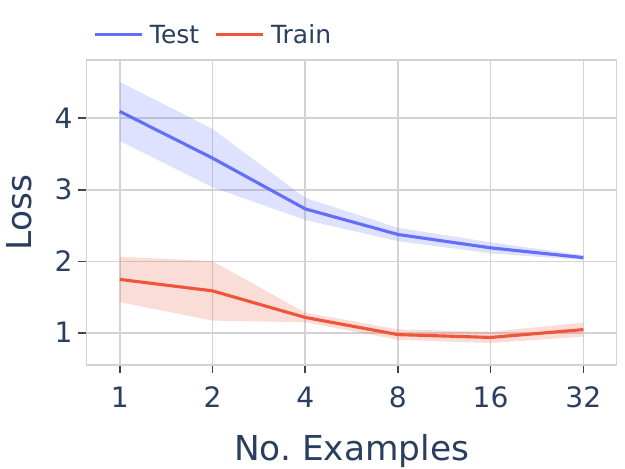
        }
    }
    \subfloat[Opt-$6.7$B]{
        \includegraphics[scale=0.35]{
            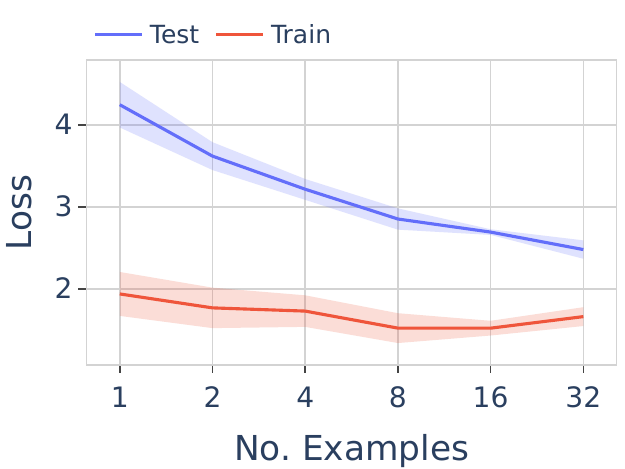
        }
    }

    \caption{Testing the limit of utilizing {\icl} context ($ 1536 $ examples $ \approx $ $ 77 $K tokens) on language $ \lang_2 $. Training loss provides a lower bound of test loss in {\icl}. Long context LLMs cannot further improve from additional examples.}
    \label{fig:utilizing_long_context_lang_2}

\end{figure*}

\begin{figure*}[!t]
    \centering
    
    \subfloat[Qwen-$ 2.5 $-$ 0.5$B]{
        \includegraphics[scale=0.35]{
            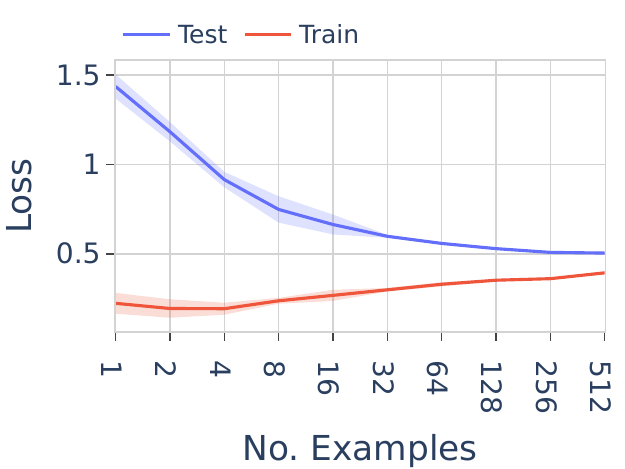
        }
    }
    \subfloat[Qwen-$ 2.5 $-$ 1.5$B]{
        \includegraphics[scale=0.35]{
            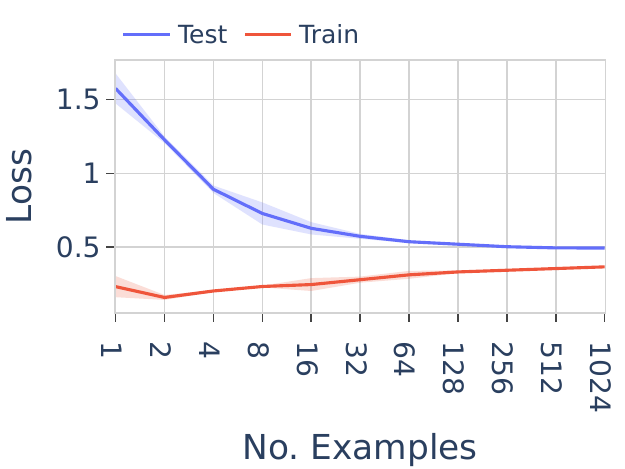
        }
    }
    \subfloat[Qwen-$ 2.5 $-$ 7$B]{
        \includegraphics[scale=0.35]{
            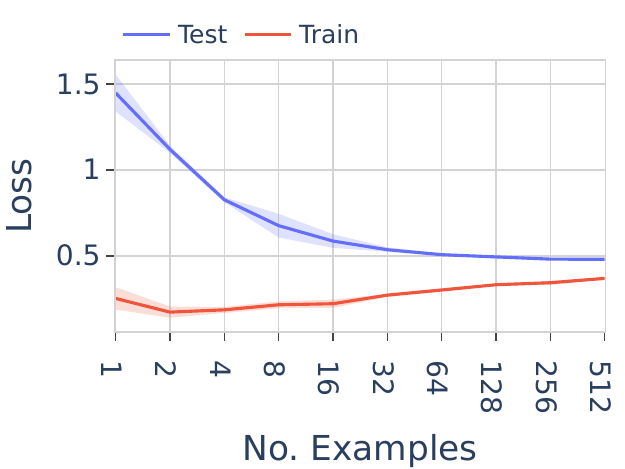
        }
    }

    \subfloat[Mistral-$7$B]{
        \includegraphics[scale=0.35]{
            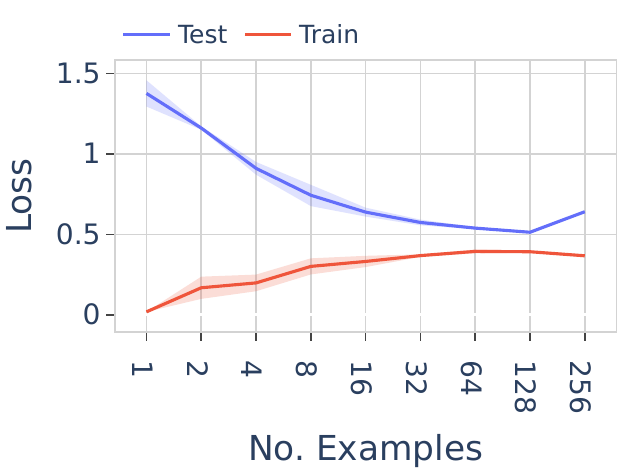
        }
    }
    \subfloat[Mistral-$12$B]{
        \includegraphics[scale=0.35]{
            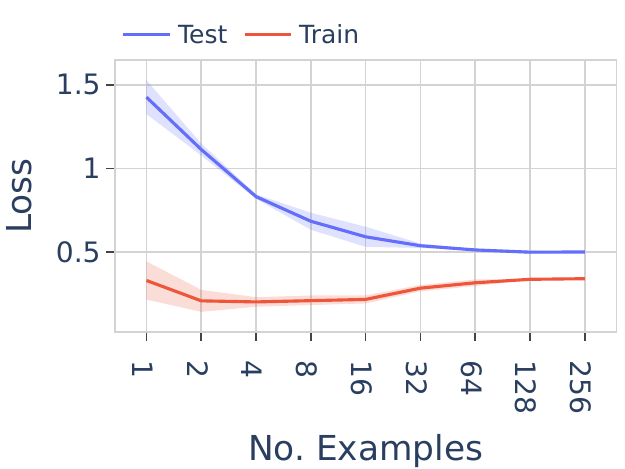
        }
    }
    \subfloat[Llama-$2$-$7$B]{
        \includegraphics[scale=0.35]{
            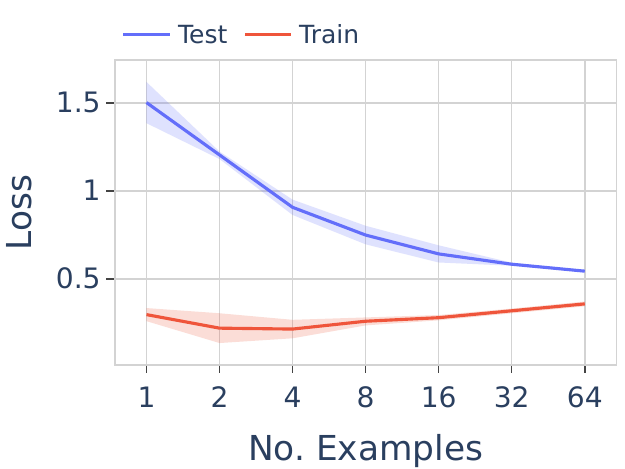
        }
    }
    \subfloat[Llama-$2$-$13$B]{
        \includegraphics[scale=0.35]{
            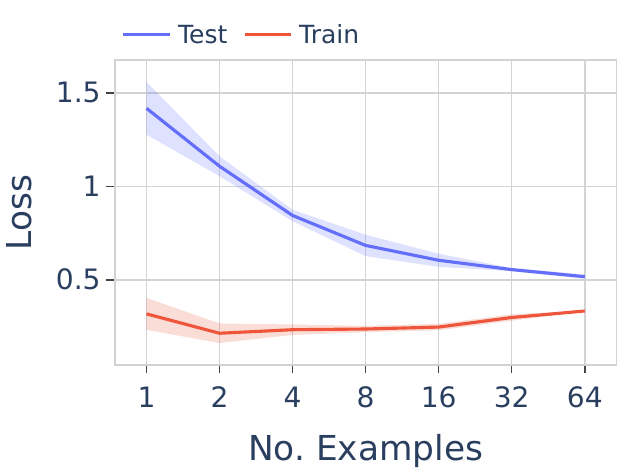
        }
    }

    \subfloat[Llama-$3.2$-$1$B]{
        \includegraphics[scale=0.35]{
            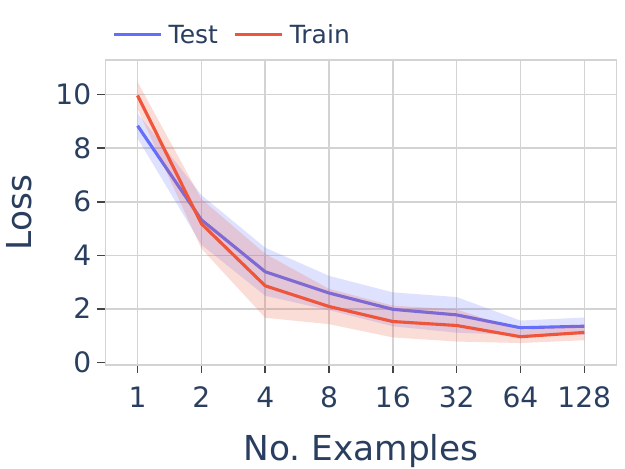
        }
    }
    \subfloat[Llama-$3.2$-$3$B]{
        \includegraphics[scale=0.35]{
            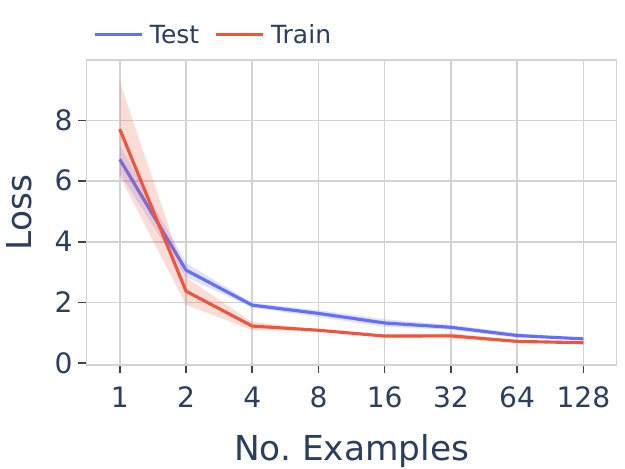
        }
    }
    \subfloat[Llama-$3.1$-$8$B]{
        \includegraphics[scale=0.35]{
            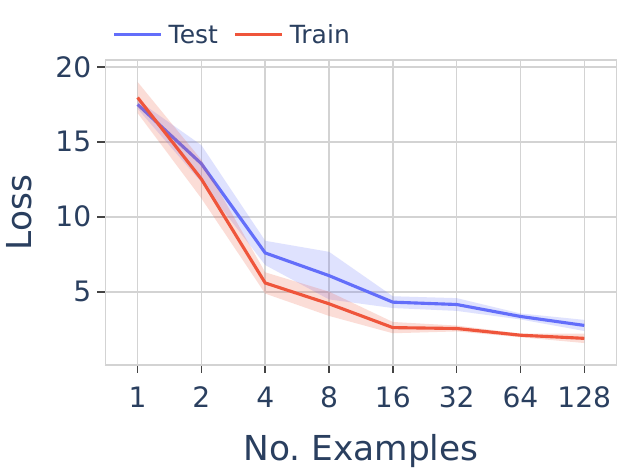
        }
    }

    \subfloat[Gemma-$2$-$2$B]{
        \includegraphics[scale=0.35]{
            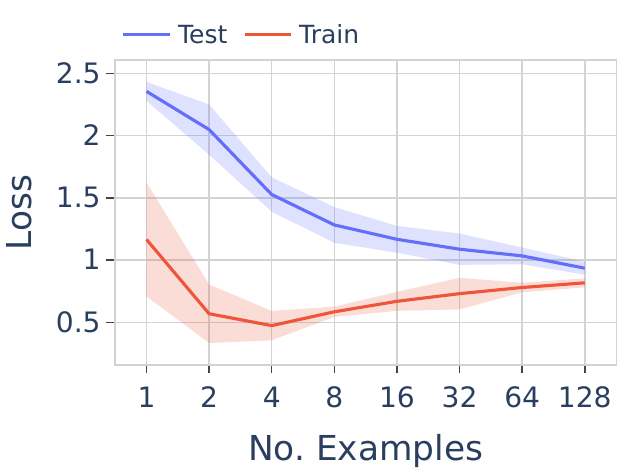
        }
    }\subfloat[Gemma-$2$-$9$B]{
        \includegraphics[scale=0.35]{
            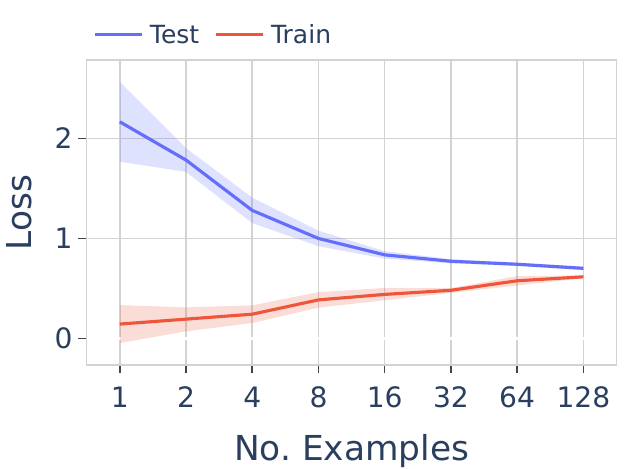
        }
    }

    \subfloat[Pythia-$1$B]{
        \includegraphics[scale=0.35]{
            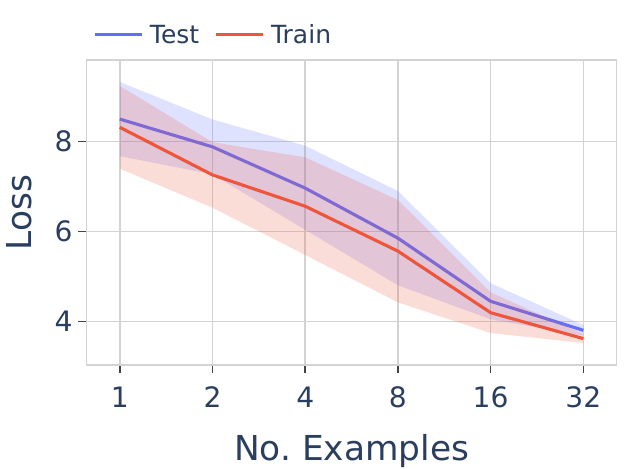
        }
    }
    \subfloat[Pythia-$2.8$B]{
        \includegraphics[scale=0.35]{
            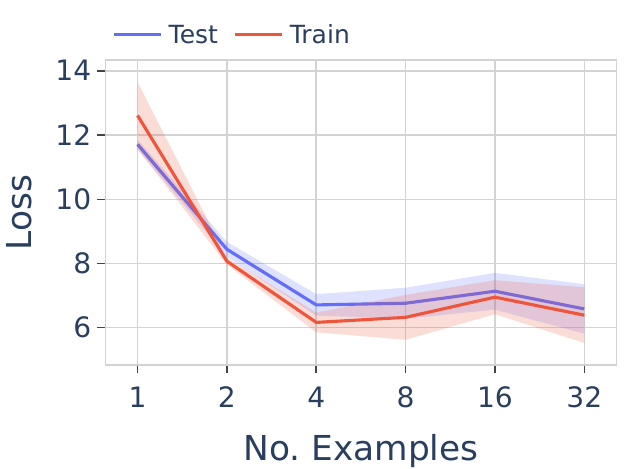
        }
    }
    \subfloat[Pythia-$6.9$B]{
        \includegraphics[scale=0.35]{
            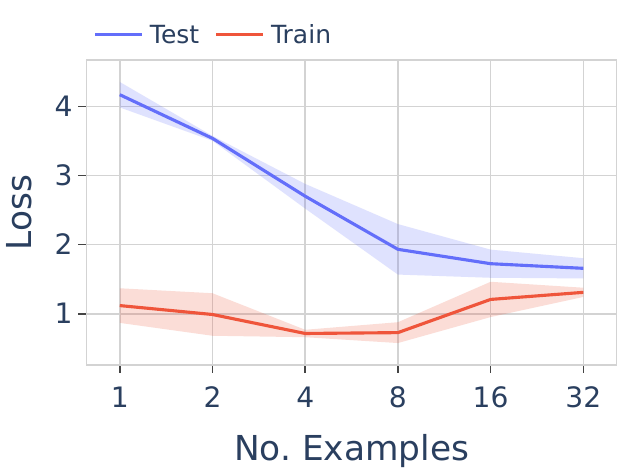
        }
    }

    \subfloat[Opt-$1.3$B]{
        \includegraphics[scale=0.35]{
            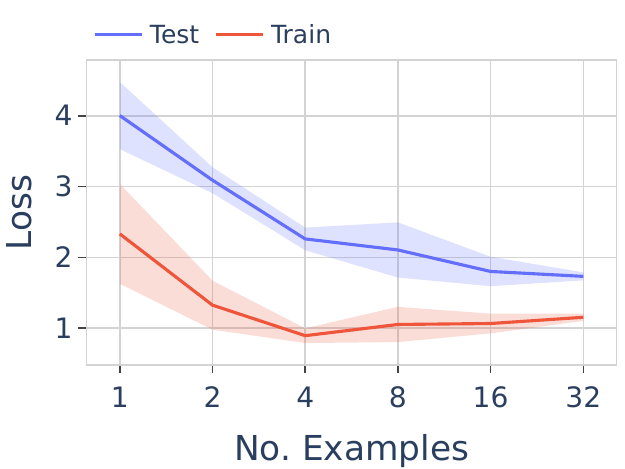
        }
    }
    \subfloat[Opt-$2.7$B]{
        \includegraphics[scale=0.35]{
            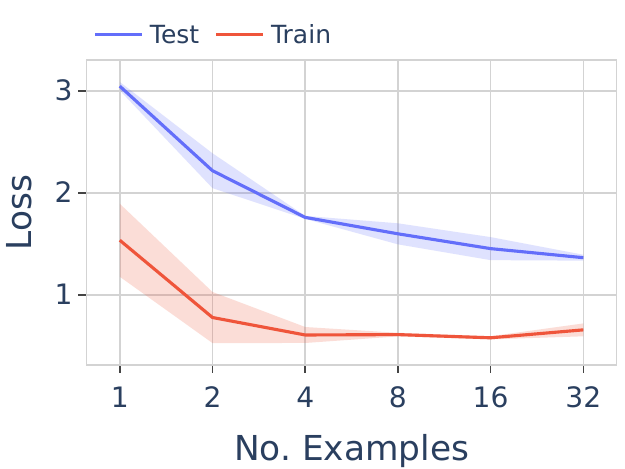
        }
    }
    \subfloat[Opt-$6.7$B]{
        \includegraphics[scale=0.35]{
            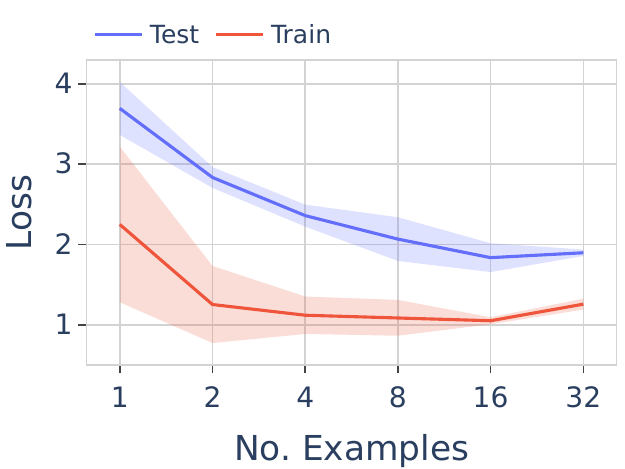
        }
    }

    \caption{Testing the limit of utilizing {\icl} context on language $ \lang_4 $. Training loss provides a lower bound of test loss in {\icl}. Long context LLMs cannot further improve from additional examples.}
    \label{fig:utilizing_long_context_lang_4}

\end{figure*}

\begin{figure*}[!t]
    \centering
    
    \subfloat[Qwen-$ 2.5 $-$ 0.5$B]{
        \includegraphics[scale=0.35]{
            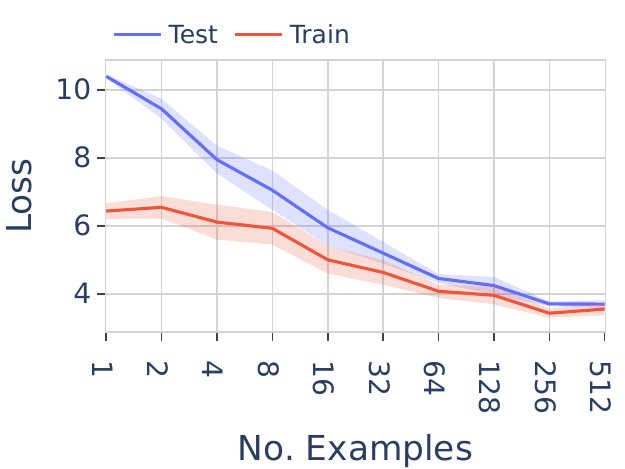
        }
    }
    \subfloat[Qwen-$ 2.5 $-$ 1.5$B]{
        \includegraphics[scale=0.35]{
            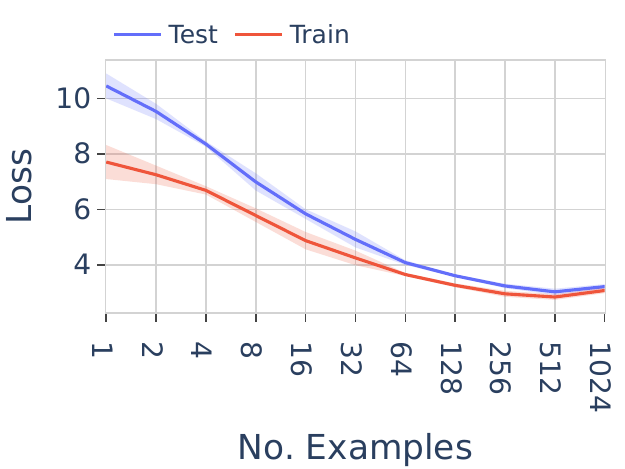
        }
    }
    \subfloat[Qwen-$ 2.5 $-$ 7$B]{
        \includegraphics[scale=0.35]{
            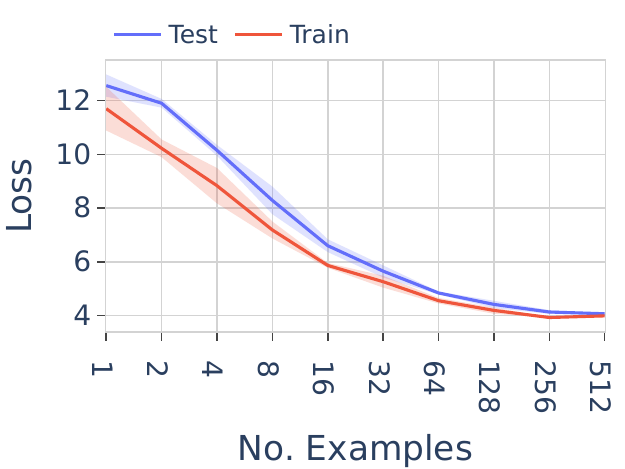
        }
    }

    \subfloat[Mistral-$7$B]{
        \includegraphics[scale=0.35]{
            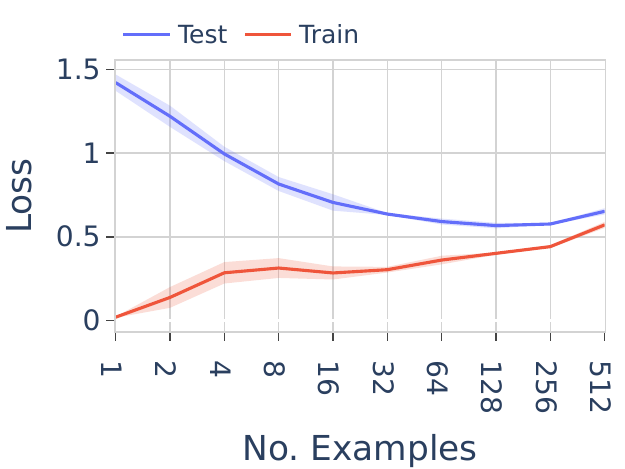
        }
    }
    \subfloat[Mistral-$12$B]{
        \includegraphics[scale=0.35]{
            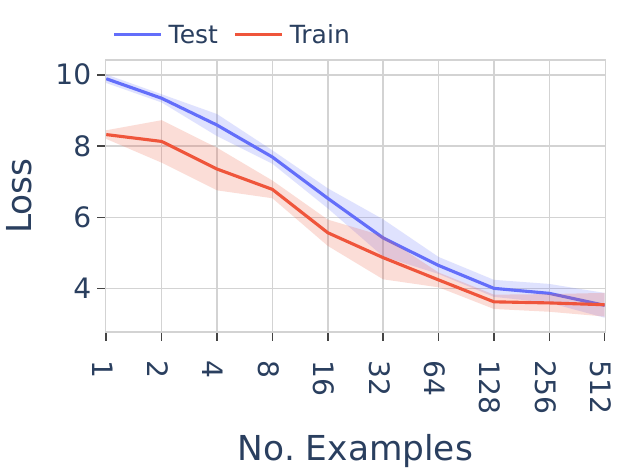
        }
    }
    \subfloat[Llama-$2$-$7$B]{
        \includegraphics[scale=0.35]{
            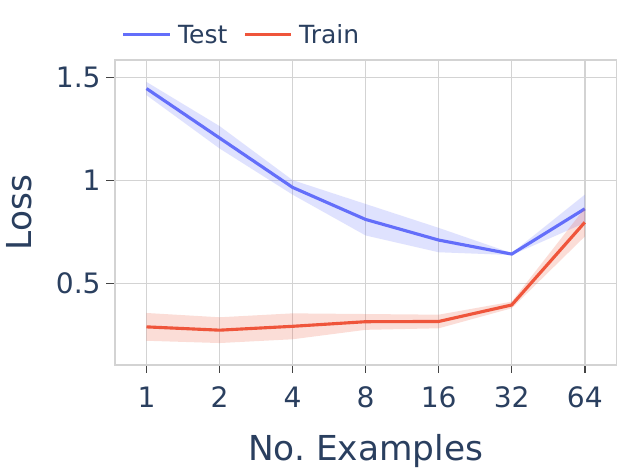
        }
    }
    \subfloat[Llama-$2$-$13$B]{
        \includegraphics[scale=0.35]{
            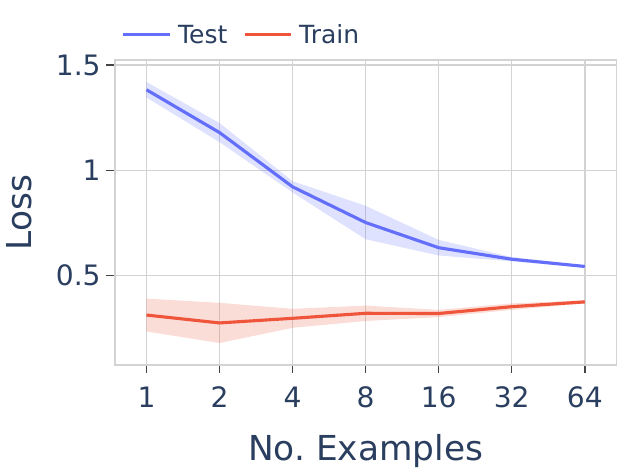
        }
    }

    \subfloat[Llama-$3.2$-$1$B]{
        \includegraphics[scale=0.35]{
            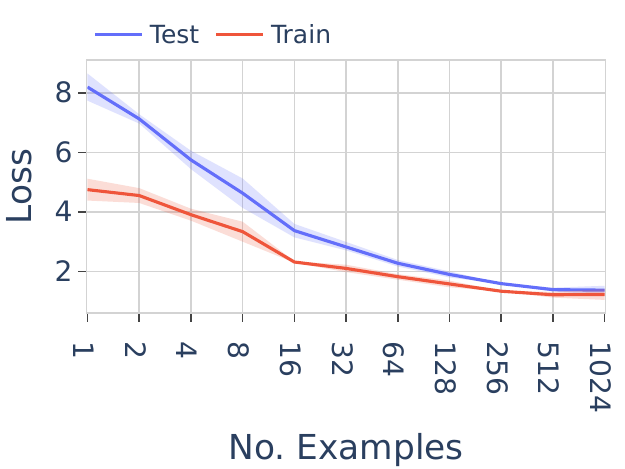
        }
    }
    \subfloat[Llama-$3.2$-$3$B]{
        \includegraphics[scale=0.35]{
            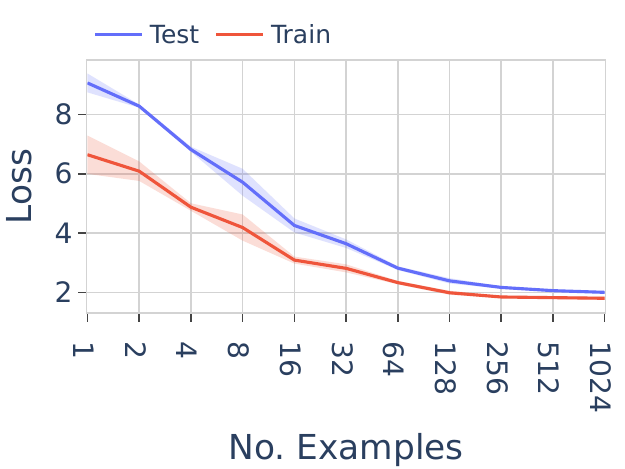
        }
    }
    \subfloat[Llama-$3.1$-$8$B]{
        \includegraphics[scale=0.35]{
            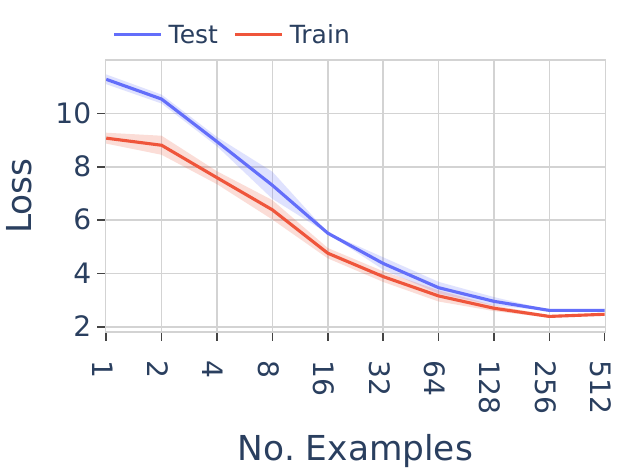
        }
    }

    \subfloat[Gemma-$2$-$2$B]{
        \includegraphics[scale=0.35]{
            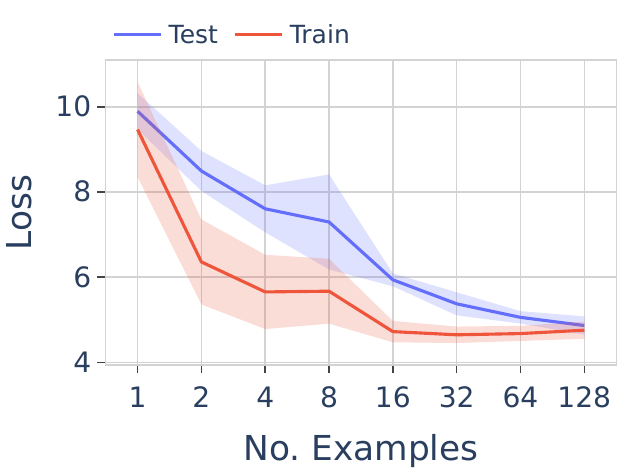
        }
    }\subfloat[Gemma-$2$-$9$B]{
        \includegraphics[scale=0.35]{
            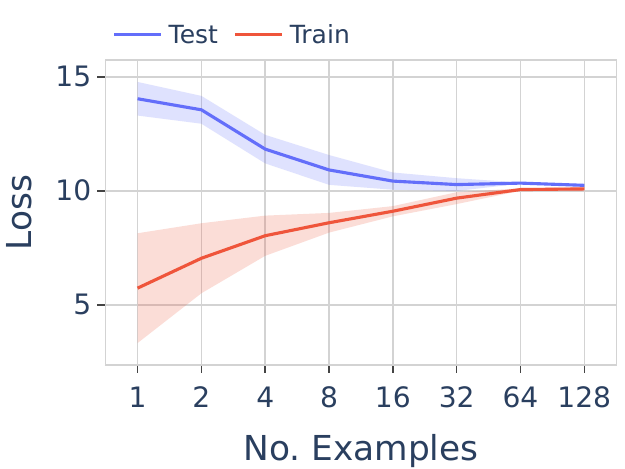
        }
    }

    \subfloat[Pythia-$1$B]{
        \includegraphics[scale=0.35]{
            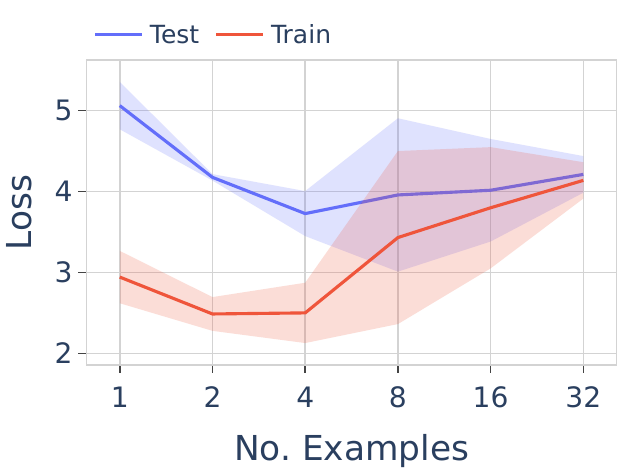
        }
    }
    \subfloat[Pythia-$2.8$B]{
        \includegraphics[scale=0.35]{
            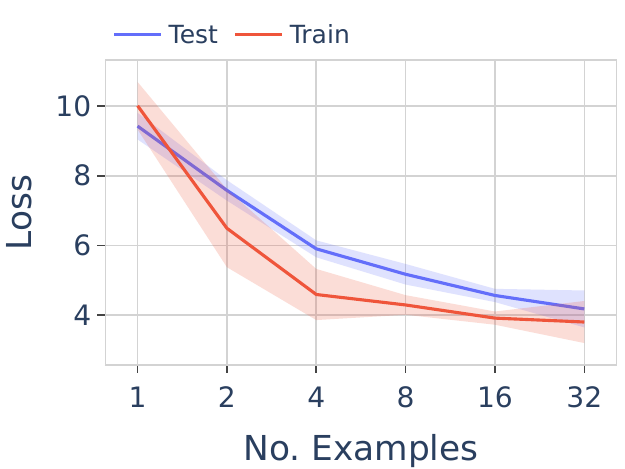
        }
    }
    \subfloat[Pythia-$6.9$B]{
        \includegraphics[scale=0.35]{
            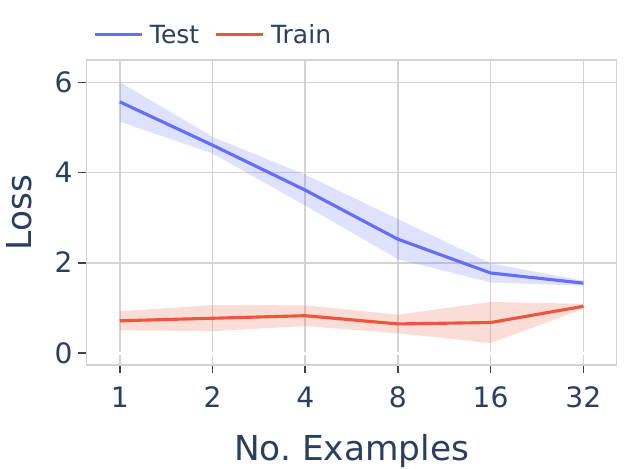
        }
    }

    \subfloat[Opt-$1.3$B]{
        \includegraphics[scale=0.35]{
            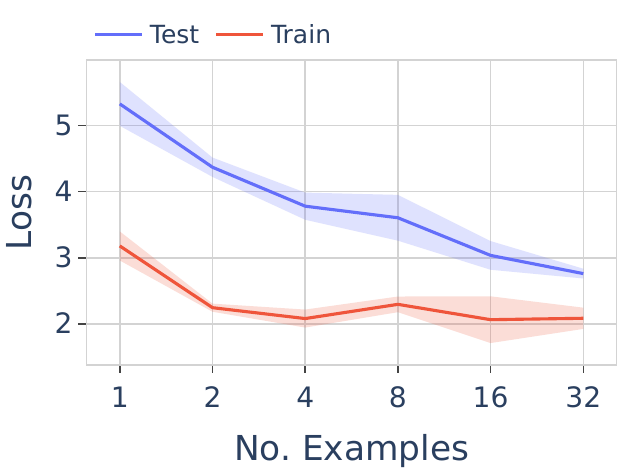
        }
    }
    \subfloat[Opt-$2.7$B]{
        \includegraphics[scale=0.35]{
            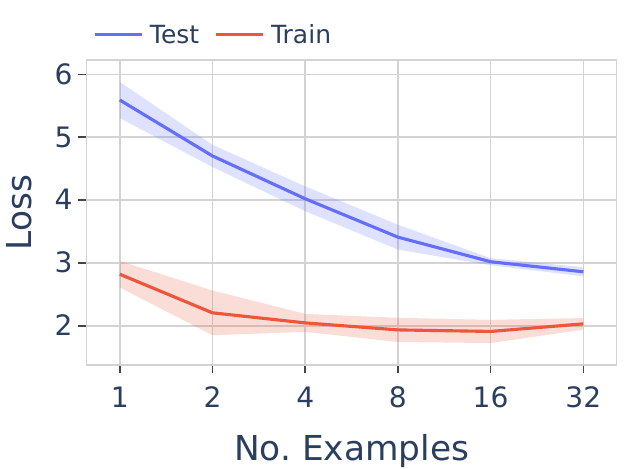
        }
    }
    \subfloat[Opt-$6.7$B]{
        \includegraphics[scale=0.35]{
            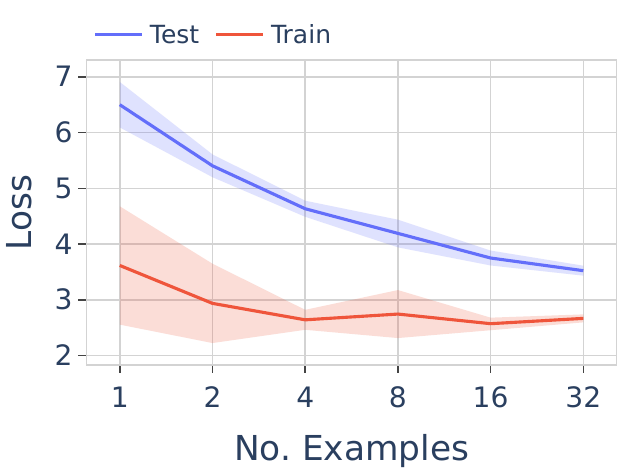
        }
    }

    \caption{Testing the limit of utilizing {\icl} context on language $ \lang_5 $. Training loss provides a lower bound of test loss in {\icl}. Long context LLMs cannot further improve from additional examples.}
    \label{fig:utilizing_long_context_lang_5}

\end{figure*}

\clearpage
\section{Discriminative Test}
\label{app_sec:discriminative_test}

\begin{repclaim}{claim:discriminative_test}
    For a given language, the discriminative test yields a numerically comparable score between two learning modes of an LLM and across LLMs, unlike the generative test.
\end{repclaim}

The discriminative test computes a classification score to determine how well strings in a language are discriminated from strings outside the language, using generation loss. Crucially, within each learning mode, both in-language and out-language strings undergo the same input formatting and are evaluated under the same parameters and hyperparameters of the LLM. For example, in {\icl}, the same concatenated prefix is applied to all strings; in {\ft}, all strings use a null prefix (see Figure~\ref{fig:intro}). As a result, any mode-specific or model-specific bias in generation loss is cancelled out in the relative comparison, making the classification score numerically comparable across learning modes and LLMs. The generative test, however, compares absolute generation loss on in-language strings only. Since LLMs with different pretraining priors assign different baseline generation loss to the same strings, generative scores are not numerically comparable across LLMs.

\begin{figure*}[!t]
    \captionsetup[subfigure]{justification=centering}
    \centering

    \subfloat[{\ft}, Language $\lang_1$, Generative performance]{
    \includegraphics[scale=0.35]{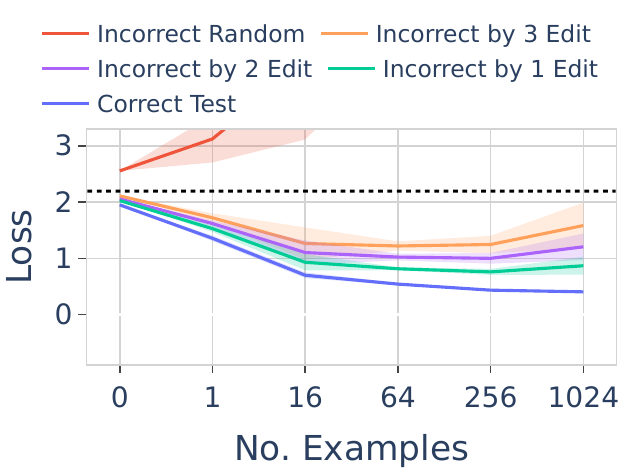}
    }
    \subfloat[{\icl}, Language $\lang_1$, Generative performance]{
    \includegraphics[scale=0.35]{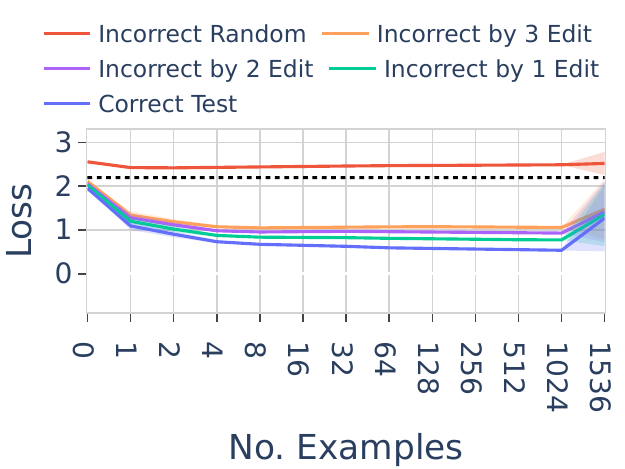}
    }
    \subfloat[{\ft}, Language $\lang_4$, Generative performance]{
    \includegraphics[scale=0.35]{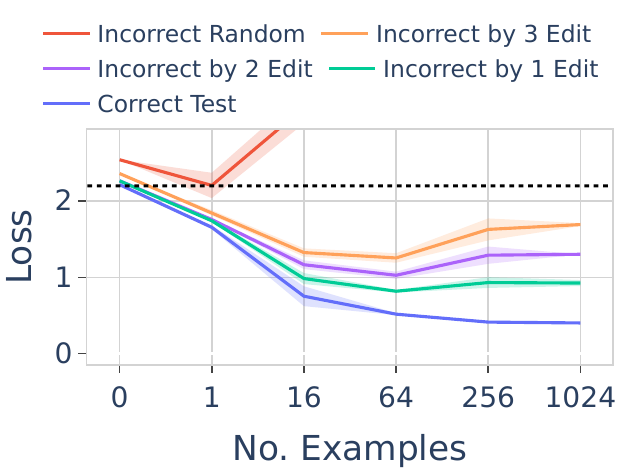}
    }
    \subfloat[{\icl}, Language $\lang_4$, Generative performance]{
    \includegraphics[scale=0.35]{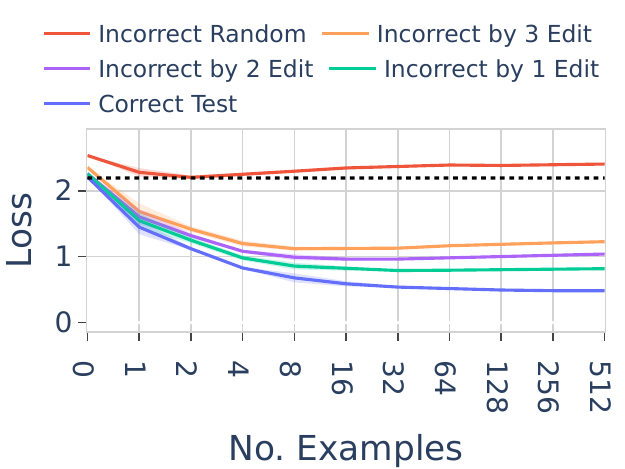}
    }

    \subfloat[{\ft}, Language $\lang_1$, Discriminative performance]{
    \includegraphics[scale=0.35]{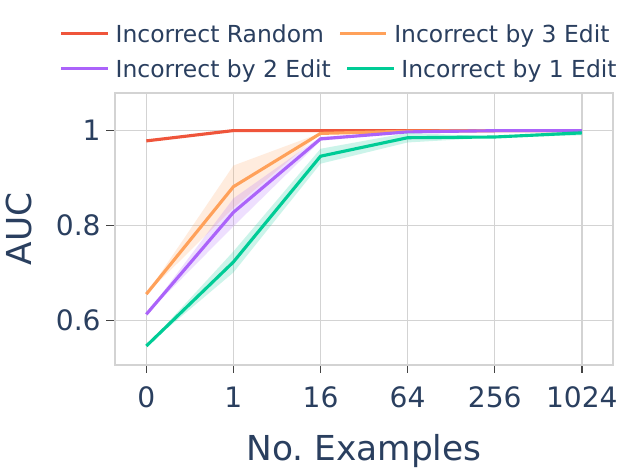}
    }
    \subfloat[{\icl}, Language $\lang_1$, Discriminative performance]{
    \includegraphics[scale=0.35]{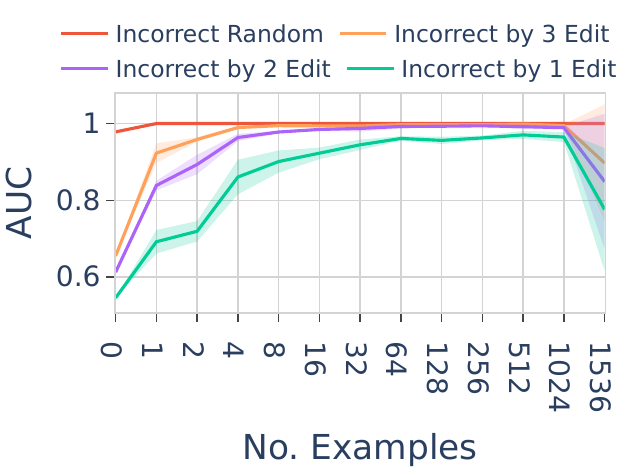}
    }
    \subfloat[{\ft}, Language $\lang_4$, Discriminative performance]{
    \includegraphics[scale=0.35]{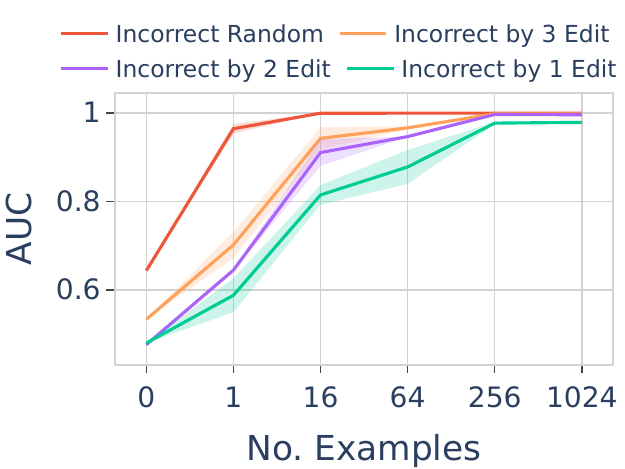}
    }
    \subfloat[{\icl}, Language $\lang_4$, Discriminative performance]{
    \includegraphics[scale=0.35]{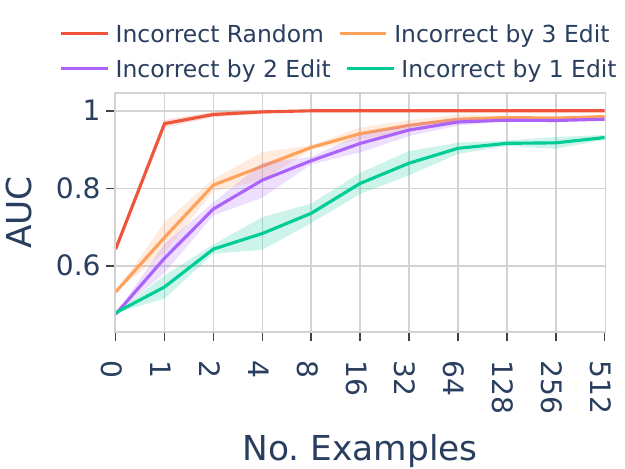}
    }

    \caption{Qwen-$ 2.5 $-$7$B: Language proficiency according to generative (first row) and discriminative (second row) tests. First two columns are for language $\lang_1$, and the last two columns are for language $\lang_4$.}
    \label{fig:gen_disc_qwen-2.5}

\end{figure*}

\begin{figure*}[!t]
    \captionsetup[subfigure]{justification=centering}
    \centering

    \subfloat[{\ft}, Language $\lang_1$, Generative performance]{
    \includegraphics[scale=0.35]{figures/fine-tuning/fine-tuning_eval_dataset_mistral-7b_pcfg_cfg3b_disjoint_terminals_loss_edit_distance_truncated.pdf}
    }
    \subfloat[{\icl}, Language $\lang_1$, Generative performance]{
    \includegraphics[scale=0.35]{figures/incontext_experiments/incontext_exp_eval_dataset_mistral-7b_pcfg_cfg3b_disjoint_terminals_loss_edit_distance_truncated.pdf}
    }
    \subfloat[{\ft}, Language $\lang_4$, Generative performance]{
    \includegraphics[scale=0.35]{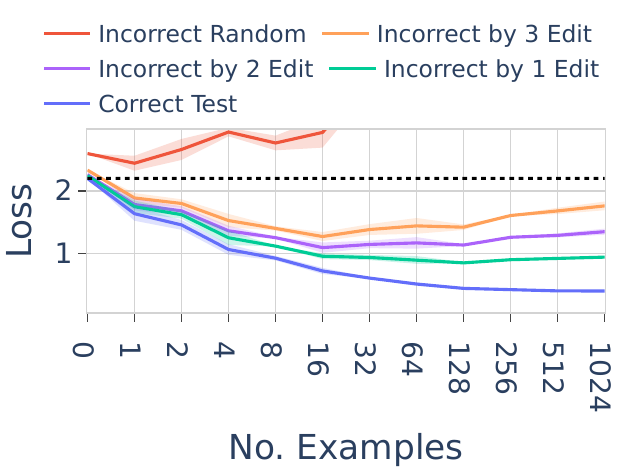}
    }
    \subfloat[{\icl}, Language $\lang_4$, Generative performance]{
    \includegraphics[scale=0.35]{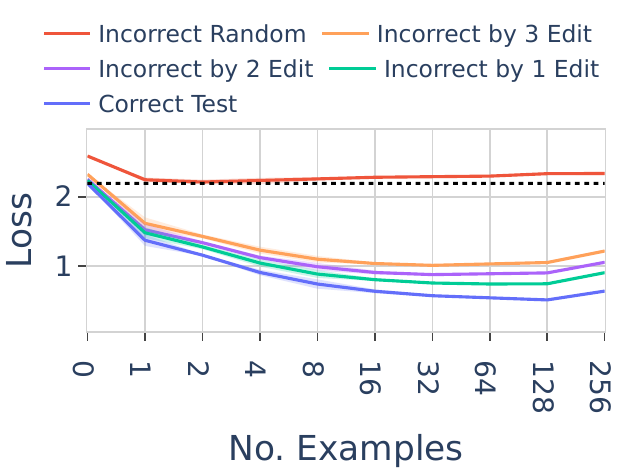}
    }

    \subfloat[{\ft}, Language $\lang_1$, Discriminative performance]{
    \includegraphics[scale=0.35]{figures/fine-tuning/fine-tuning_eval_dataset_mistral-7b_pcfg_cfg3b_disjoint_terminals_auc_edit_distance_truncated.pdf}
    }
    \subfloat[{\icl}, Language $\lang_1$, Discriminative performance]{
    \includegraphics[scale=0.35]{figures/incontext_experiments/incontext_exp_eval_dataset_mistral-7b_pcfg_cfg3b_disjoint_terminals_auc_edit_distance_truncated.pdf}
    }
    \subfloat[{\ft}, Language $\lang_4$, Discriminative performance]{
    \includegraphics[scale=0.35]{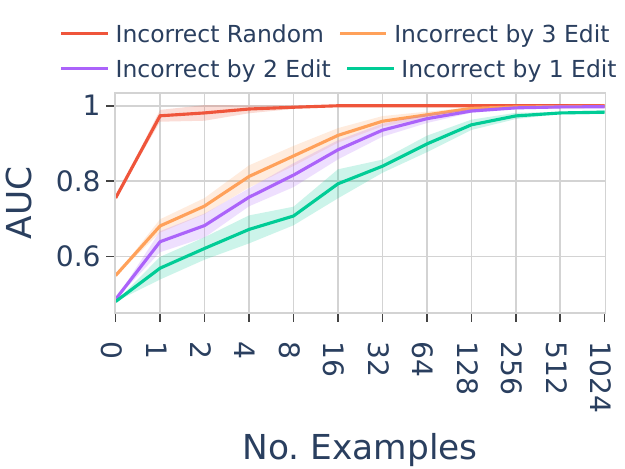}
    }
    \subfloat[{\icl}, Language $\lang_4$, Discriminative performance]{
    \includegraphics[scale=0.35]{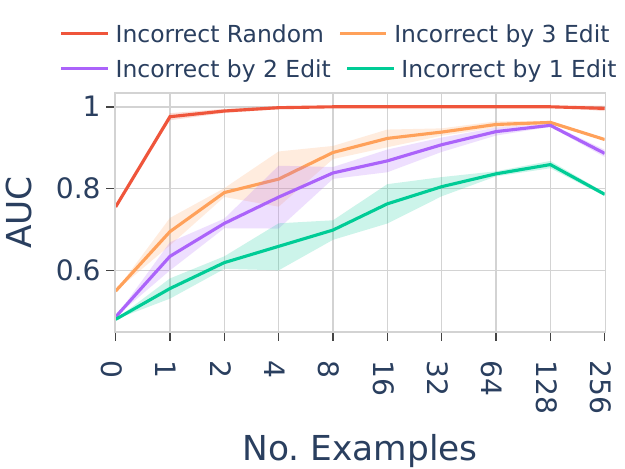}
    }

    \caption{Mistral-$7$B: Language proficiency according to generative (first row) and discriminative (second row) tests. First two columns are for language $\lang_1$, and the last two columns are for language $\lang_4$.}
    \label{fig:gen_disc_mistral}

\end{figure*}

\begin{figure*}[!t]
    \captionsetup[subfigure]{justification=centering}
    \centering

    \subfloat[{\ft}, Language $\lang_1$, Generative performance]{
    \includegraphics[scale=0.35]{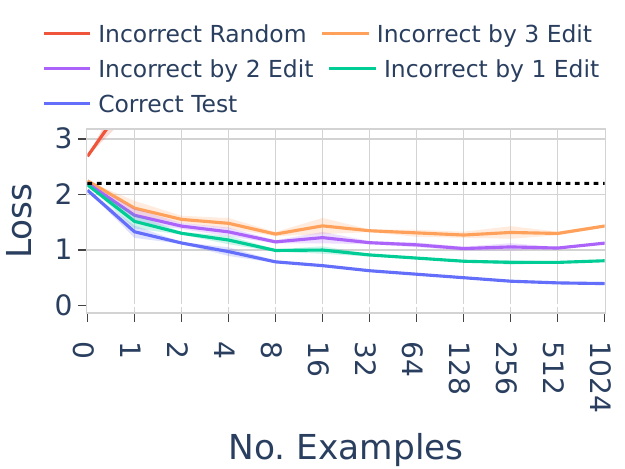}
    }
    \subfloat[{\icl}, Language $\lang_1$, Generative performance]{
    \includegraphics[scale=0.35]{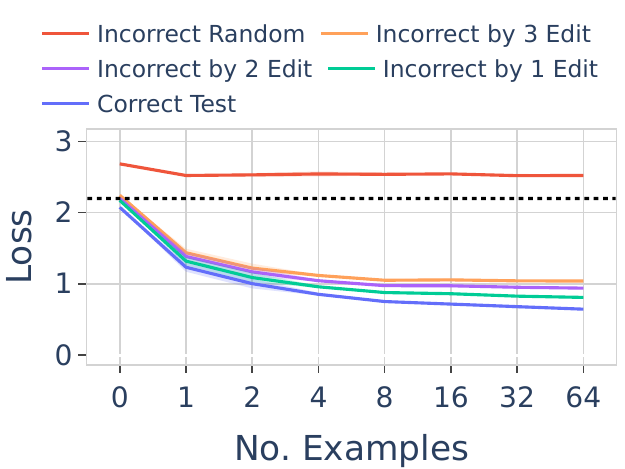}
    }
    \subfloat[{\ft}, Language $\lang_4$, Generative performance]{
    \includegraphics[scale=0.35]{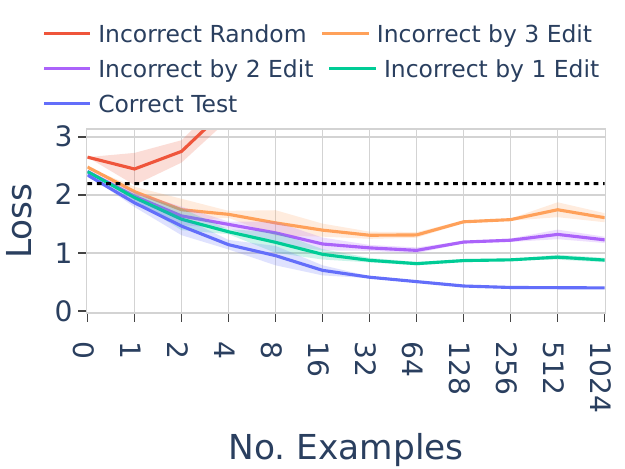}
    }
    \subfloat[{\icl}, Language $\lang_4$, Generative performance]{
    \includegraphics[scale=0.35]{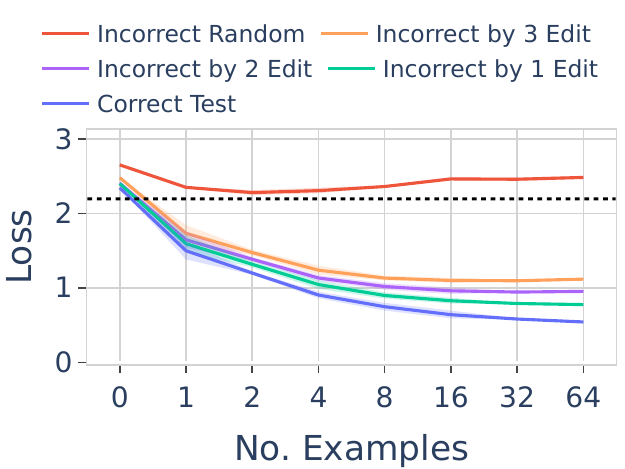}
    }

    \subfloat[{\ft}, Language $\lang_1$, Discriminative performance]{
    \includegraphics[scale=0.35]{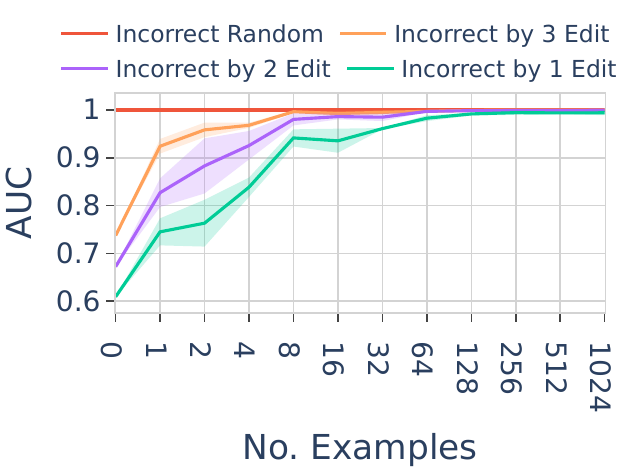}
    }
    \subfloat[{\icl}, Language $\lang_1$, Discriminative performance]{
    \includegraphics[scale=0.35]{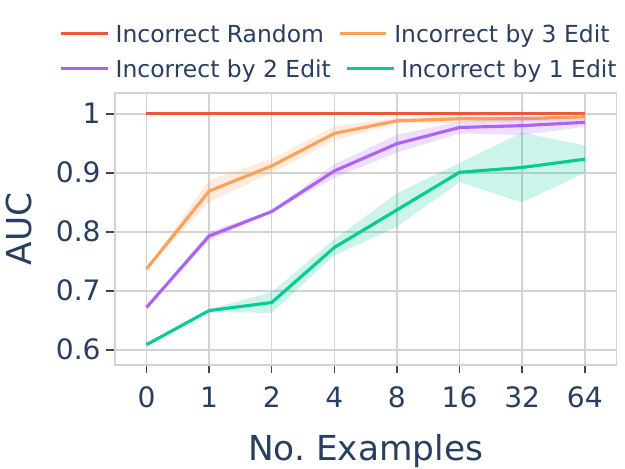}
    }
    \subfloat[{\ft}, Language $\lang_4$, Discriminative performance]{
    \includegraphics[scale=0.35]{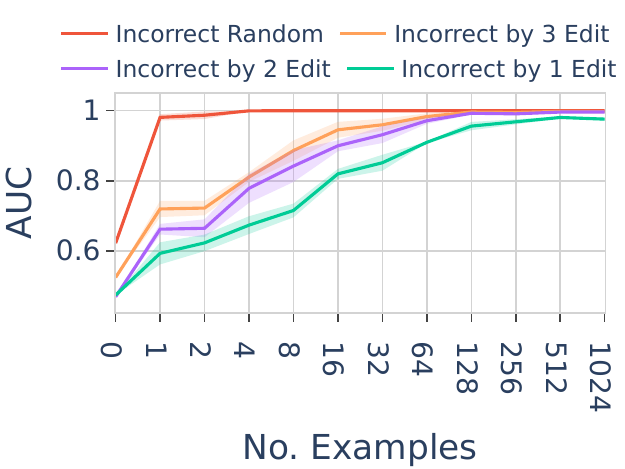}
    }
    \subfloat[{\icl}, Language $\lang_4$, Discriminative performance]{
    \includegraphics[scale=0.35]{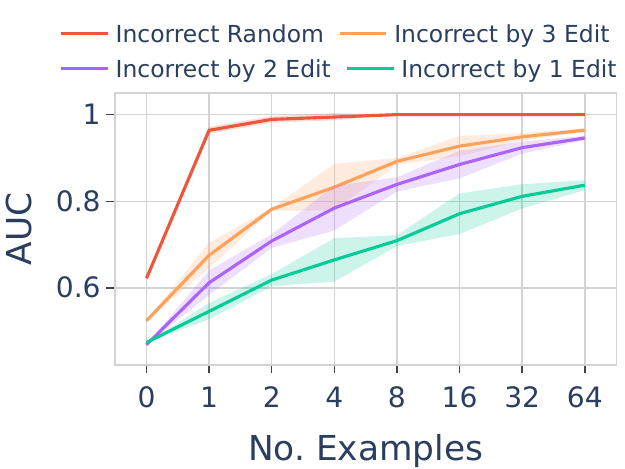}
    }

    \caption{Llama-$2$-$7$B:  Language proficiency according to generative (first row) and discriminative (second row) tests. First two columns are for language $\lang_1$, and the last two columns are for language $\lang_4$.}
    \label{fig:gen_disc_llama2}

\end{figure*}

\begin{figure*}[!t]
    \captionsetup[subfigure]{justification=centering}
    \centering

    \subfloat[{\ft}, Language $\lang_1$, Generative performance]{
    \includegraphics[scale=0.35]{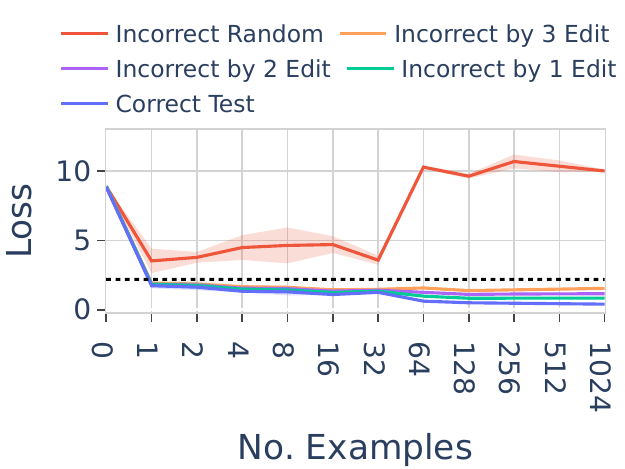}
    }
    \subfloat[{\icl}, Language $\lang_1$, Generative performance]{
    \includegraphics[scale=0.35]{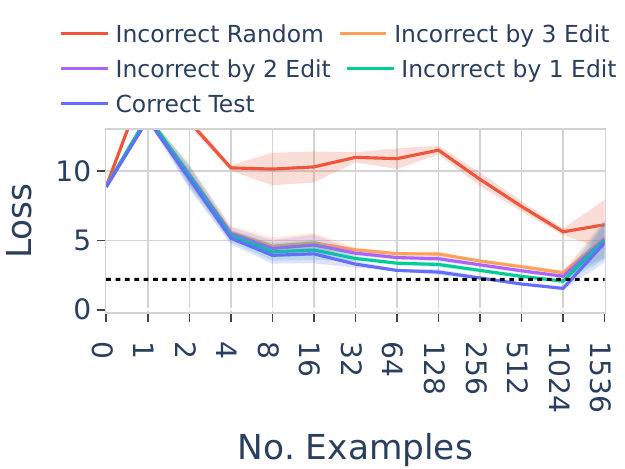}
    }
    \subfloat[{\ft}, Language $\lang_4$, Generative performance]{
    \includegraphics[scale=0.35]{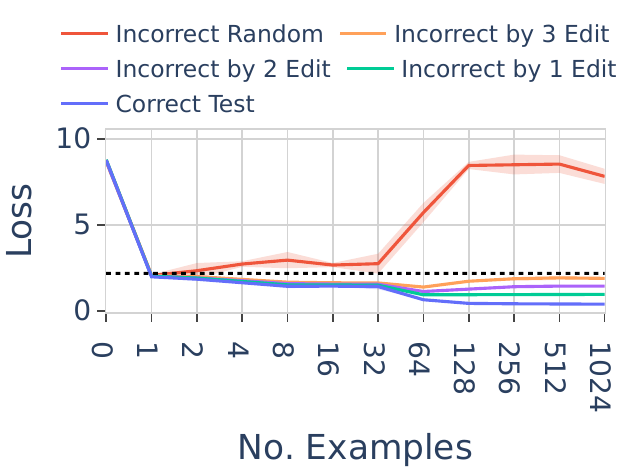}
    }
    \subfloat[{\icl}, Language $\lang_4$, Generative performance]{
    \includegraphics[scale=0.35]{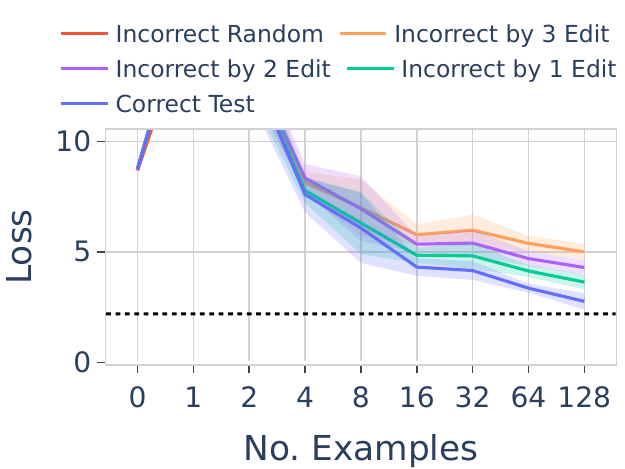}
    }

    \subfloat[{\ft}, Language $\lang_1$, Discriminative performance]{
    \includegraphics[scale=0.35]{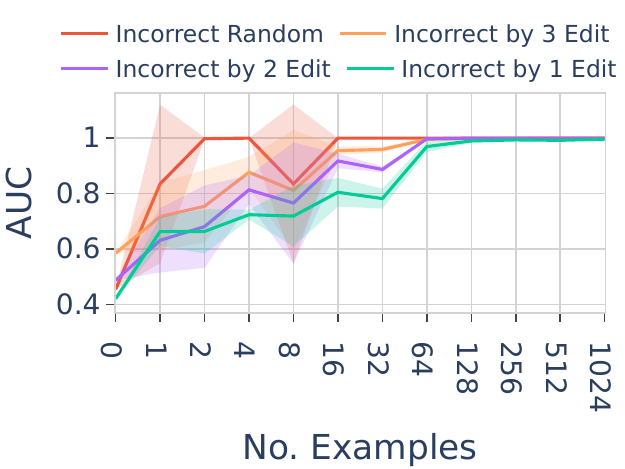}
    }
    \subfloat[{\icl}, Language $\lang_1$, Discriminative performance]{
    \includegraphics[scale=0.35]{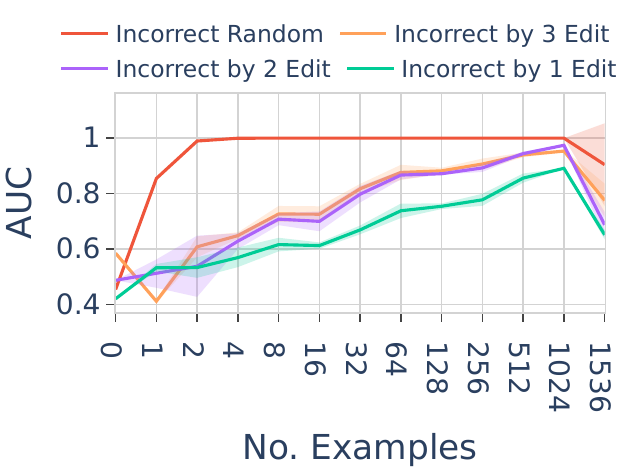}
    }
    \subfloat[{\ft}, Language $\lang_4$, Discriminative performance]{
    \includegraphics[scale=0.35]{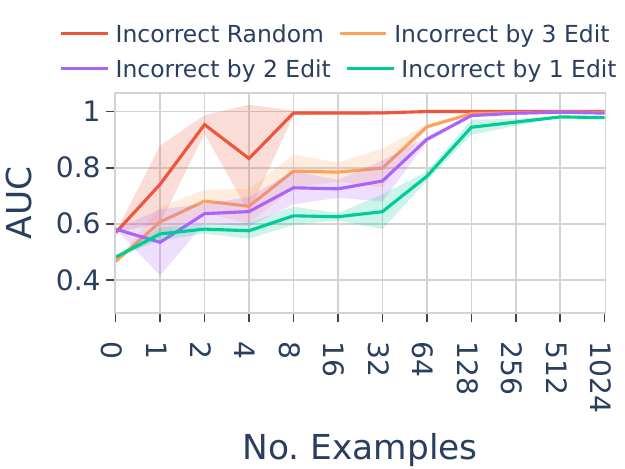}
    }
    \subfloat[{\icl}, Language $\lang_4$, Discriminative performance]{
    \includegraphics[scale=0.35]{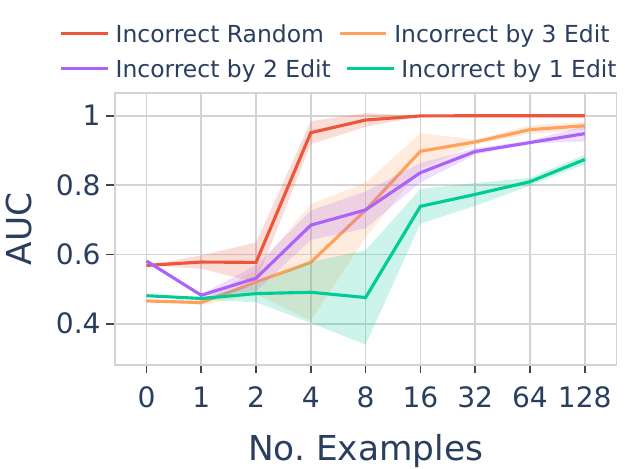}
    }

    \caption{Llama-$3.1$-$8$B:  Language proficiency according to generative (first row) and discriminative (second row) tests. First two columns are for language $\lang_1$, and the last two columns are for language $\lang_4$.}
    \label{fig:gen_disc_llama3}

\end{figure*}

\begin{figure*}[!t]
    \captionsetup[subfigure]{justification=centering}
    \centering

    \subfloat[{\ft}, Language $\lang_1$, Generative performance]{
    \includegraphics[scale=0.35]{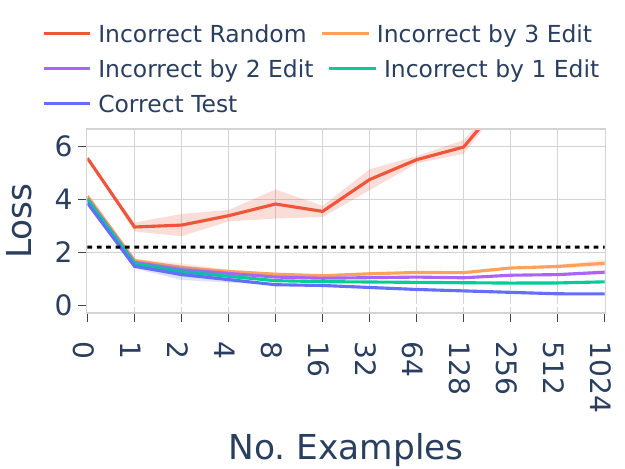}
    }
    \subfloat[{\icl}, Language $\lang_1$, Generative performance]{
    \includegraphics[scale=0.35]{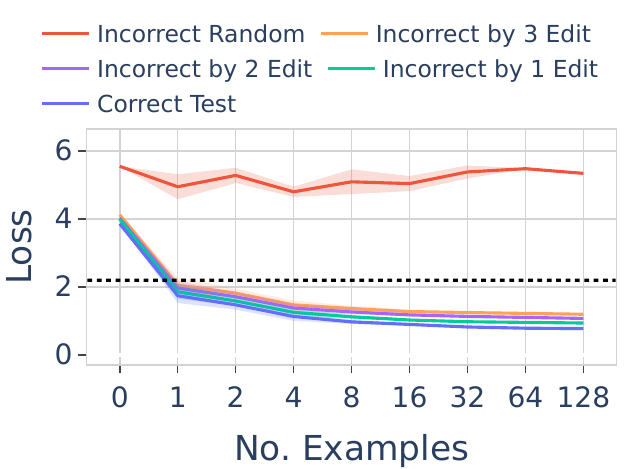}
    }
    \subfloat[{\ft}, Language $\lang_4$, Generative performance]{
    \includegraphics[scale=0.35]{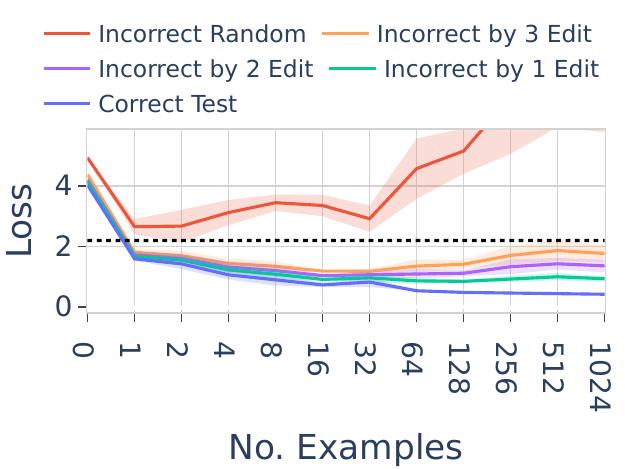}
    }
    \subfloat[{\icl}, Language $\lang_4$, Generative performance]{
    \includegraphics[scale=0.35]{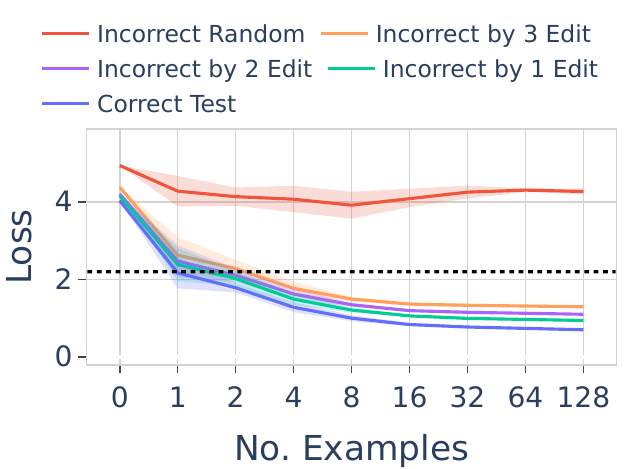}
    }

    \subfloat[{\ft}, Language $\lang_1$, Discriminative performance]{
    \includegraphics[scale=0.35]{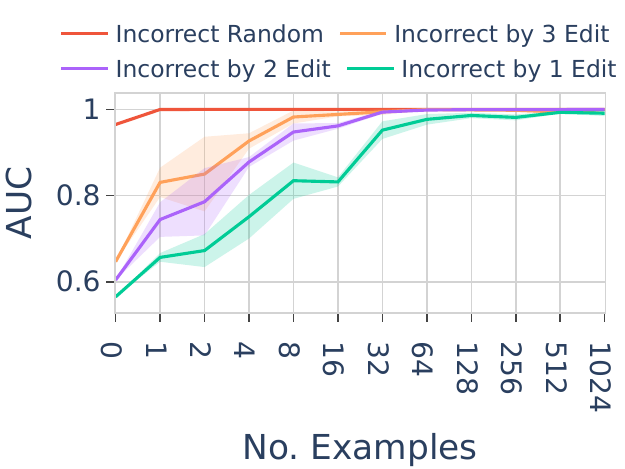}
    }
    \subfloat[{\icl}, Language $\lang_1$, Discriminative performance]{
    \includegraphics[scale=0.35]{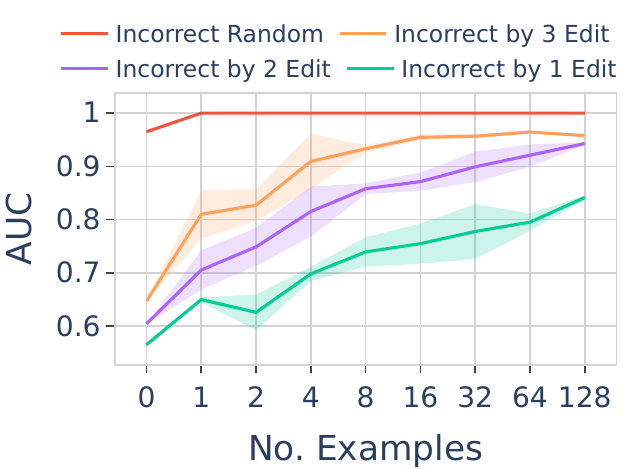}
    }
    \subfloat[{\ft}, Language $\lang_4$, Discriminative performance]{
    \includegraphics[scale=0.35]{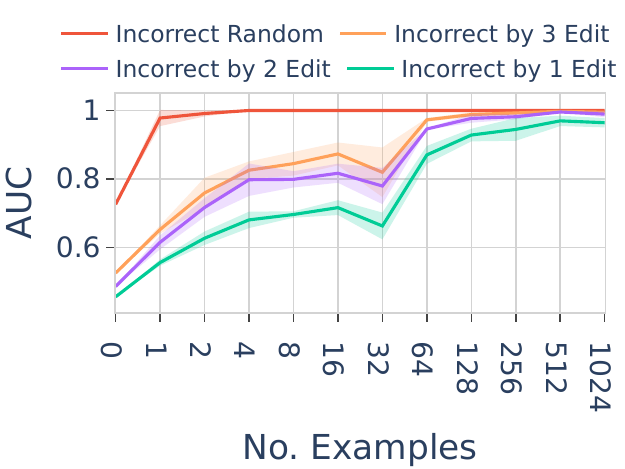}
    }
    \subfloat[{\icl}, Language $\lang_4$, Discriminative performance]{
    \includegraphics[scale=0.35]{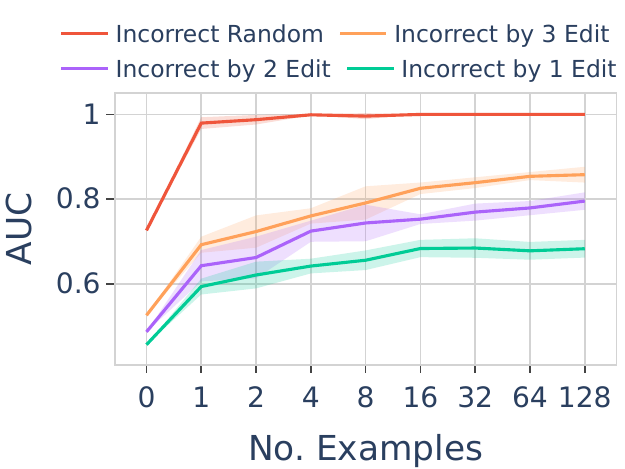}
    }

    \caption{Gemma-$2$-$9$B:  Language proficiency according to generative (first row) and discriminative (second row) tests. First two columns are for language $\lang_1$, and the last two columns are for language $\lang_4$.}
    \label{fig:gen_disc_gemma}

\end{figure*}

\begin{figure*}[!t]
    \captionsetup[subfigure]{justification=centering}
    \centering
    
    \subfloat[{\ft}, Language $\lang_1$, Generative performance]{
    \includegraphics[scale=0.35]{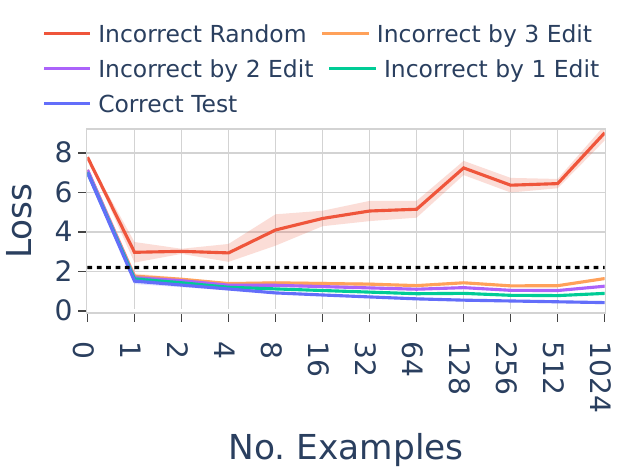}
    }
    \subfloat[{\icl}, Language $\lang_1$, Generative performance]{
    \includegraphics[scale=0.35]{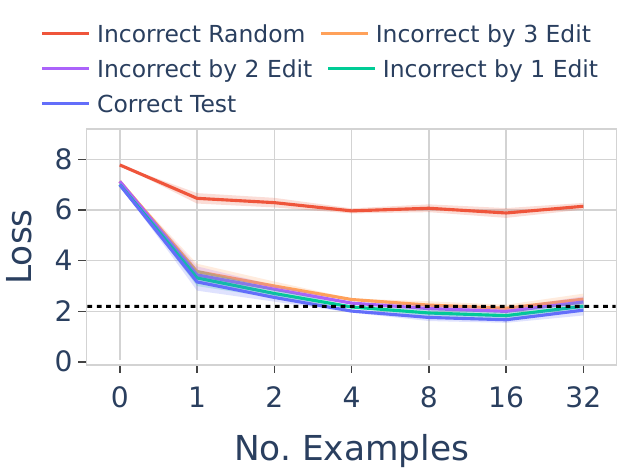}
    }
    \subfloat[{\ft}, Language $\lang_4$, Generative performance]{
    \includegraphics[scale=0.35]{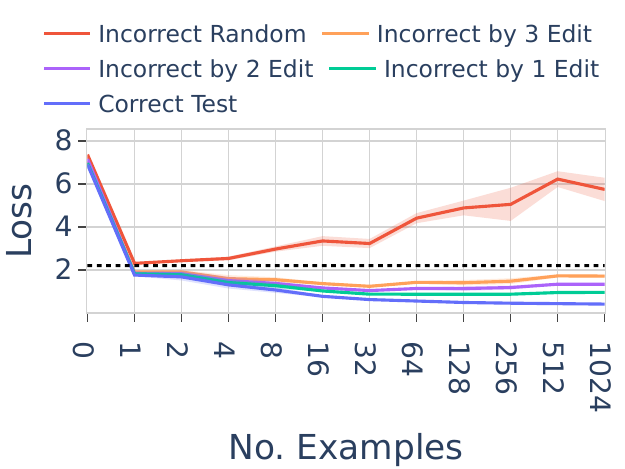}
    }
    \subfloat[{\icl}, Language $\lang_4$, Generative performance]{
    \includegraphics[scale=0.35]{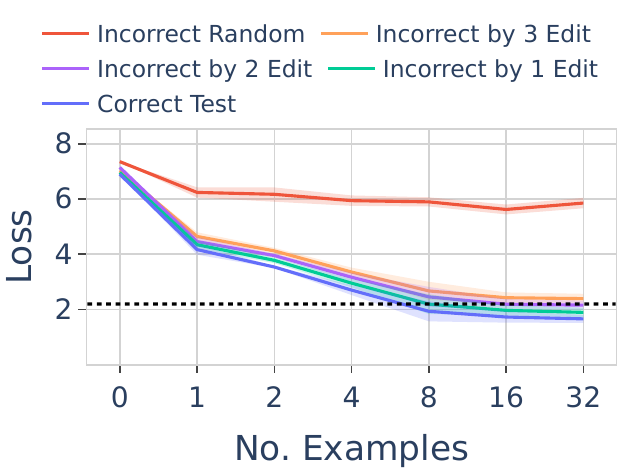}
    }

    \subfloat[{\ft}, Language $\lang_1$, Discriminative performance]{
    \includegraphics[scale=0.35]{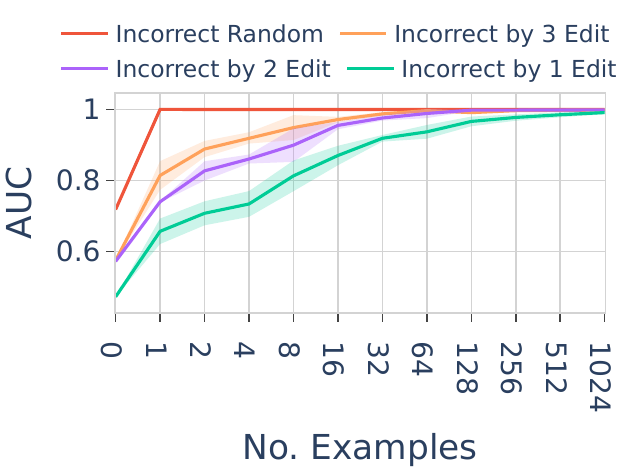}
    }
    \subfloat[{\icl}, Language $\lang_1$, Discriminative performance]{
    \includegraphics[scale=0.35]{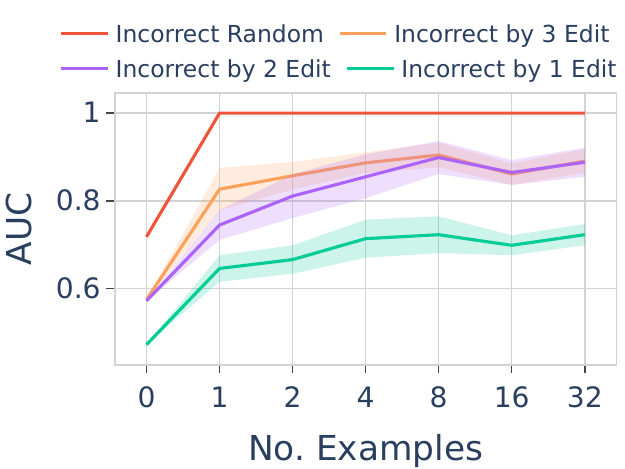}
    }
    \subfloat[{\ft}, Language $\lang_4$, Discriminative performance]{
    \includegraphics[scale=0.35]{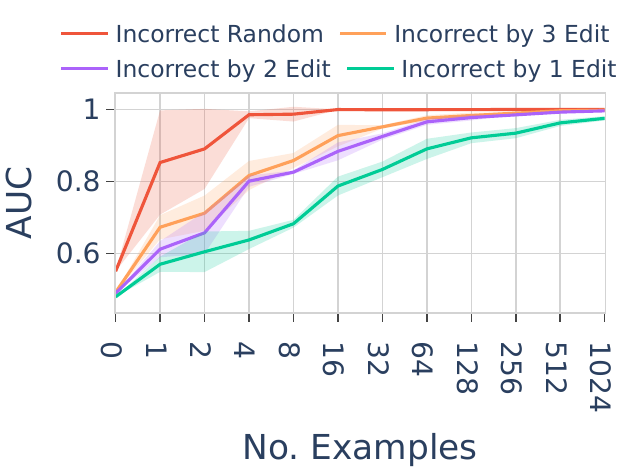}
    }
    \subfloat[{\icl}, Language $\lang_4$, Discriminative performance]{
    \includegraphics[scale=0.35]{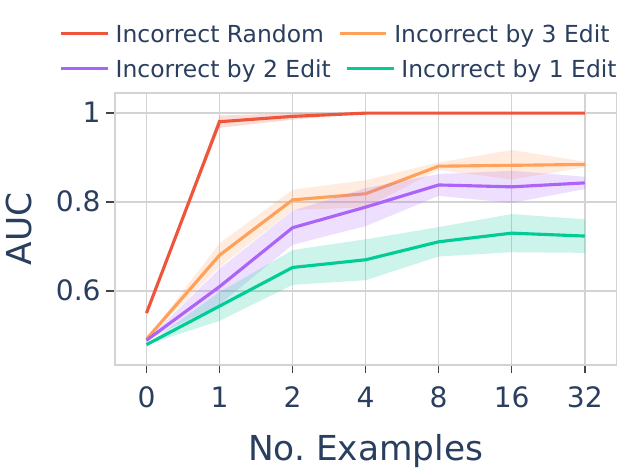}
    }

    \caption{Pythia-$6.9$B:  Language proficiency according to generative (first row) and discriminative (second row) tests. First two columns are for language $\lang_1$, and the last two columns are for language $\lang_4$.}
    \label{fig:gen_disc_pythia}

\end{figure*}

\begin{figure*}[!t]
    \captionsetup[subfigure]{justification=centering}
    \centering

    \subfloat[{\ft}, Language $\lang_1$, Generative performance]{
    \includegraphics[scale=0.35]{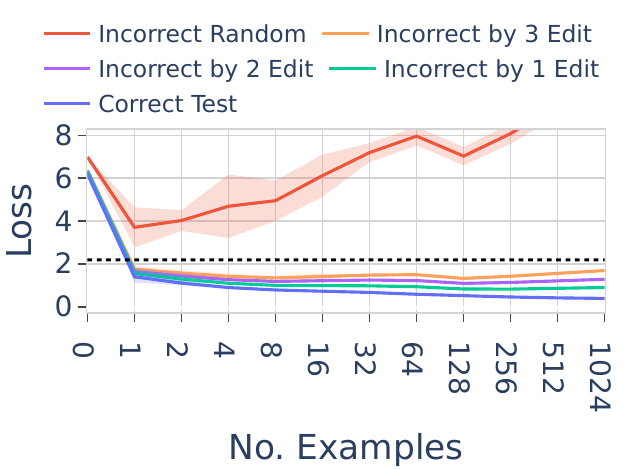}
    }
    \subfloat[{\icl}, Language $\lang_1$, Generative performance]{
    \includegraphics[scale=0.35]{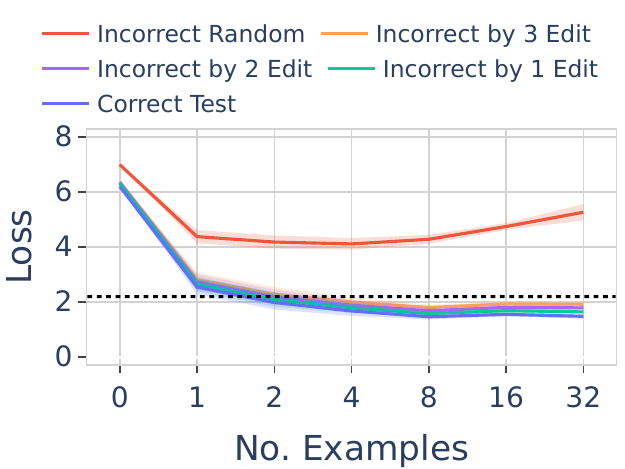}
    }
    \subfloat[{\ft}, Language $\lang_4$, Generative performance]{
    \includegraphics[scale=0.35]{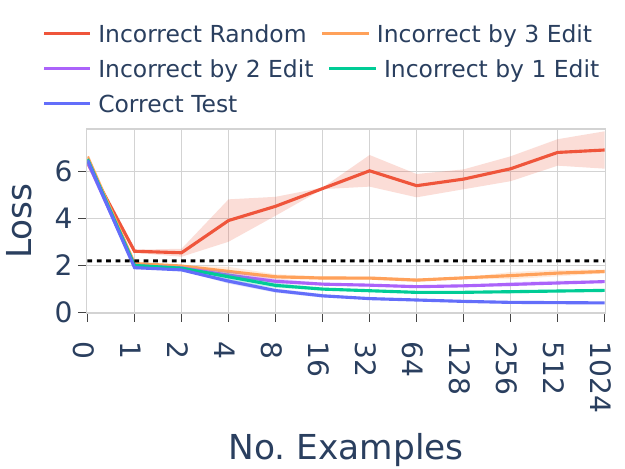}
    }
    \subfloat[{\icl}, Language $\lang_4$, Generative performance]{
    \includegraphics[scale=0.35]{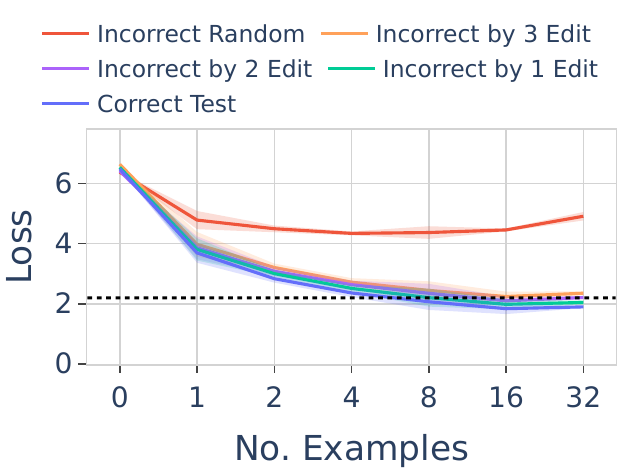}
    }

    \subfloat[{\ft}, Language $\lang_1$, Discriminative performance]{
    \includegraphics[scale=0.35]{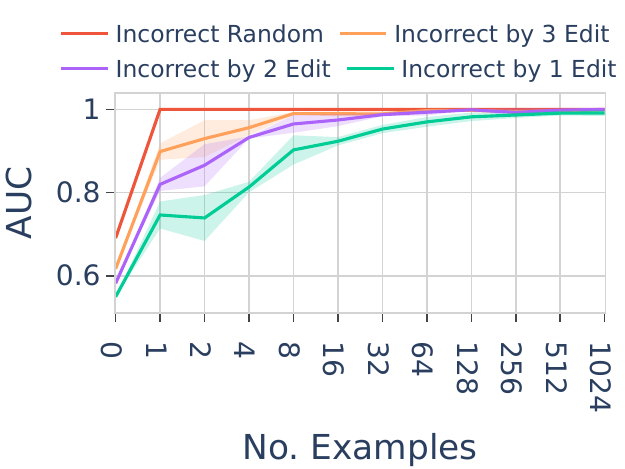}
    }
    \subfloat[{\icl}, Language $\lang_1$, Discriminative performance]{
    \includegraphics[scale=0.35]{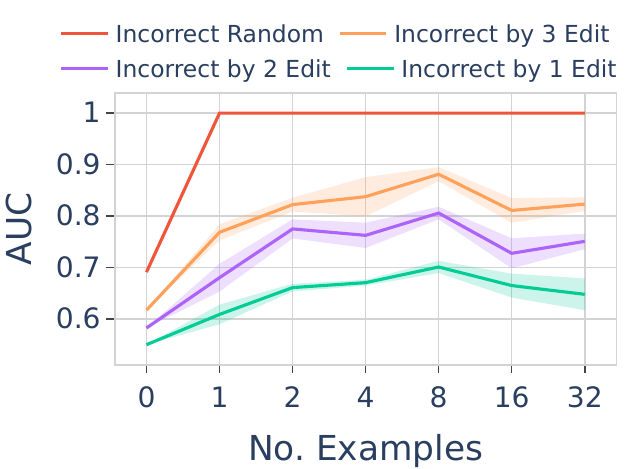}
    }
    \subfloat[{\ft}, Language $\lang_4$, Discriminative performance]{
    \includegraphics[scale=0.35]{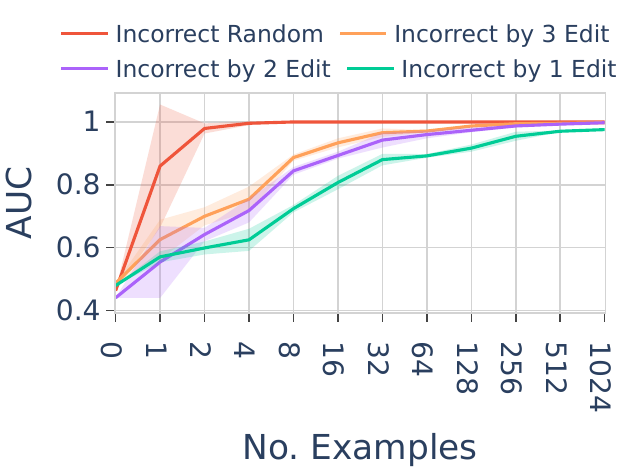}
    }
    \subfloat[{\icl}, Language $\lang_4$, Discriminative performance]{
    \includegraphics[scale=0.35]{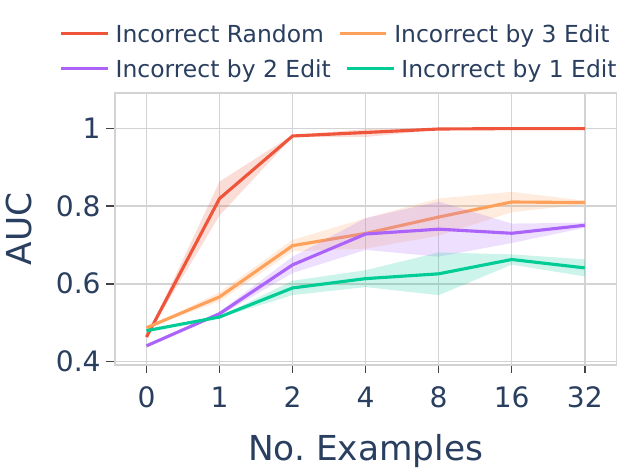} 
    }

    \caption{Opt-$6.7$B:  Language proficiency according to generative (first row) and discriminative (second row) tests. First two columns are for language $\lang_1$, and the last two columns are for language $\lang_4$.}
    \label{fig:gen_disc_opt}

\end{figure*}

\clearpage
\section{Implications of the Study}
\label{app_sec:implications}

We elaborate on the implications of our findings across four research questions in Section~\ref{sec:comparison}. We provide our hypothesis for each finding, which may inspire future research.

\begin{itemize}
    \item In \textbf{RQ1}, when learning a language, {\ft} performance converges across LLMs but {\icl} performance is variable. Our hypothesis is that {\ft} is a direct form of learning, where parameters are explicitly updated toward the task. Since we evaluate {\ft} at its optimal performance across epochs, and the considered language is simple with a hierarchical recursive structure, all LLMs reach a similar performance ceiling, leading to convergence across LLMs.

    {\icl}, however, is an indirect form of learning, where the model infers patterns from the context without any parameter update. As a result, {\icl} performance retains model-specific pre-training biases, which differ across LLMs of different sizes and families, leading to variable performance. A more subtle analysis is given below in \textbf{RQ4}.

    \item In \textbf{RQ2}, {\ft} outperforms {\icl} in in-distribution generalization, where the training and test languages are the same. In formal languages, however, both modes perform equally in out-of-distribution generalization, generalizing only to languages closer to the training language. Therefore, if the test language is different, {\ft} is no longer the better mode, and explicit parameter update in {\ft} does not help. In this case, {\icl} is a more natural choice, since its parameters remain unchanged and no specialization toward the training language occurs, unlike {\ft} where parameter updates specialize the model toward the training language, without improving out-of-distribution generalization.

    \item In \textbf{RQ3}, the inductive bias of {\ft} and {\icl} is similar, but this similarity often decreases with more training examples, where similarity is measured by the Pearson correlation of generation losses on identical test strings. Thus, when learning is stronger, each mode learns the language differently, leading to different inductive biases.
    
    \item In \textbf{RQ4}, {\icl} is less robust than {\ft} across languages. Since {\icl} relies on pre-training, its performance depends on how well the pre-training corpus covers the specific tokens of the language, making it sensitive to token variation. {\ft} avoids this problem by directly updating parameters toward the language, leading to more robust performance.
    
    \item We emphasize the adoption of the discriminative test for evaluating language proficiency in LLMs, across both formal and natural languages. The discriminative test ensures that in-language strings are generated with higher probability than, and are even separable from, out-of-language strings -- a stronger condition than the generative test on in-language strings only.

    For future work on the adoption of the discriminative test, one needs to systematically generate strings outside the language, which we have shown for formal languages in Section~\ref{sec:preliminary}, and for natural language, with sentiment classification as one instance, in Appendix~\ref{app_sec:nlp_dataset}. Since natural language is less well-defined than formal language, the boundary between in-language and out-of-language strings may be harder to define precisely in natural language, warranting careful study.
\end{itemize}

\section{Use of AI Assistants}

We use AI assistants for the following purposes:
\begin{itemize}[leftmargin=*]
    \item Paper writing: We use Claude to correct grammatical mistakes, and paraphrase sentences to improve the quality and flow of the writing.
    \item Coding: We use Claude and Windsurf as coding assistants.
\end{itemize}
Nevertheless, we take full responsibility for the content of the paper and code.

\end{document}